\documentclass[12pt]{article}

\RequirePackage[T1]{fontenc}
\usepackage{natbib}
\RequirePackage[colorlinks,citecolor=blue,urlcolor=blue]{hyperref}
\usepackage[english]{babel}
\usepackage{amsthm}
\usepackage{amsmath}
\usepackage{amsfonts}
\usepackage{amssymb}
\usepackage{graphicx}
\usepackage{enumerate}
\usepackage{color}
\usepackage{array}
\usepackage{bbm}
\usepackage{multirow}
\usepackage{rotating}
\usepackage{xr}
\usepackage{setspace}

\newcommand\nocell[1]{\multicolumn{#1}{c|}{}}

\begin{document}
\thispagestyle{empty}
\baselineskip=28pt
\vskip 5mm
\begin{center} {\Large{\bf Regression modelling of spatiotemporal extreme U.S. wildfires via partially-interpretable neural networks}}
\end{center}

\baselineskip=12pt
\vskip 5mm

\begin{center}
\large
Jordan Richards$^{1*}$ and Rapha\"el Huser$^2$
\end{center}

\footnotetext[1]{
\baselineskip=10pt School of Mathematics, University of Edinburgh, UK. $^*$E-mail: jordan.richards@ed.ac.uk}
\footnotetext[2]{
\baselineskip=10pt Statistics Program, Computer, Electrical and Mathematical Sciences and Engineering (CEMSE) Division, King Abdullah University of Science and Technology (KAUST), Thuwal 23955-6900, Saudi Arabia.}

\baselineskip=17pt
\vskip 4mm
\centerline{\today}
\vskip 6mm

\begin{center}
{\large{\bf Abstract}}
\end{center}

Risk management in many environmental settings requires an understanding of the mechanisms that drive extreme events. Useful metrics for quantifying such risk are extreme quantiles of response variables conditioned on predictor variables that describe, e.g., climate, biosphere and environmental states. Typically these quantiles lie outside the range of observable data and so, for estimation, require specification of parametric extreme value models within a regression framework. Classical approaches in this context utilise linear or additive relationships between predictor and response variables and suffer in either their predictive capabilities or computational efficiency; moreover, their simplicity is unlikely to capture the truly complex structures that lead to the creation of extreme wildfires. In this paper, we propose a new methodological framework for performing extreme quantile regression using artificial neutral networks, which are able to capture complex non-linear relationships and scale well to high-dimensional data. The ``black box" nature of neural networks means that they lack the desirable trait of interpretability often favoured by practitioners; thus, we unify  linear, and additive, regression methodology with deep learning to create partially-interpretable neural networks that can be used for statistical inference but retain high prediction accuracy. To complement this methodology, we further propose a novel point process model for extreme values which overcomes the finite lower-endpoint problem associated with the generalised extreme value class of distributions. Efficacy of our unified framework is illustrated on U.S. wildfire data with a high-dimensional predictor set and we illustrate vast improvements in predictive performance over linear and spline-based regression techniques. Our model is used to quantify the spatially-varying effect of temperature and drought on wildfire extremes and occurrences across the U.S., as well as identify high risk regions.
\baselineskip=16pt

\par\vfill\noindent

\pagenumbering{arabic}
\baselineskip=24pt

\newpage

\section{Introduction}
\subsection{Context and motivation}
\label{intro_sec}
Uncontrolled wildfires are a significant cause of death and property damage across the world. Climate change is expected to increase both the severity and occurrence rate of wildfires with worrying trends predicted for the western United States (U.S.) \citep{smith2020sciencebrief}, due to an increase in the frequency and intensity of extreme meteorological events, including high temperatures, droughts and sufficiently fast windspeeds. Some of the most devastating wildfires in U.S. history have occurred most recently in California, with fires occurring between 2017 and 2020 being responsible for hundreds of deaths and an excess of one million acres of burnt land \citep{keeley2021large}. Wildfires contribute to the accelerating rate of climate change through the emission of greenhouses gases; the year 2021 saw the release of an estimated 1760 megatonnes of carbon into the atmosphere as a direct consequence of wildfires \citep{2021copernicus}, with a large proportion being attributed to wildfires in the northern U.S. Clearly the mitigation of risk and prevention of extreme wildfires, particularly events that lead to vast consumption of fuel, and hence, high carbon emissions, is of the utmost importance from both an economical and environmental perspective. In this paper, we develop new statistical methodology to quantify the effects of extreme wildfires through measures of burnt area, which are a useful proxy for both fuel consumption and emissions \citep{koh2021spatiotemporal}. \par
Estimation of extreme quantiles for spatiotemporal processes is important for risk management in a number of environmental applications, e.g., extreme precipitation \citep{huser2014space,opitz2018inla}, heatwaves \citep{winter2016modelling, zhong2020modeling}, and wind gusts \citep{castro2019spliced,youngman2019generalized}. For processes observed over complex or large space-time domains, it is highly likely that the process will exhibit marginal non-stationarity, which can be handled by allowing the marginal distribution of the process to vary with both space and time. This can be achieved through a regression framework, i.e., with quantiles represented as functions of predictors. Let $\{Y(s,t):s \in \mathcal{S},t \in \mathcal{T}\}$ be a spatiotemporal process indexed by a spatial set $\mathcal{S}\subset\mathbb{R}^2$ and temporal domain $\mathcal{T} \subset \mathbb{R}_+$, and let $\{\mathbf{X}(s,t):s\in\mathcal{S},t\in\mathcal{T}\}$ denote a $d$-dimensional space-time process of predictor variables where $\mathbf{X}(s,t)=(X_1(s,t),\dots,X_d(s,t))$ for all $(s,t)\in \mathcal{S}\times\mathcal{T}$ and for $d\in\mathbb{N}$. We denote observations of $\mathbf{X}(s,t)$ by $\mathbf{x}({s,t})\in \mathbb{R}^d$ and note that the process need not be smooth spatially, nor temporally. Our interest lies in the upper-tail behaviour of $Y(s,t)\mid\{\mathbf{X}(s,t)=\mathbf{x}({s,t}):(s,t)\in\mathcal{S}\times\mathcal{T}\}$, which can be characterised through its quantile function; this can be estimated either non-parametrically or parametrically, with the latter approach typically relying on parametric extreme value distributions \citep[see, e.g.,][]{chavez2005generalized}. Although non-parametric approaches avoid making restrictive assumptions about the behaviour of $Y(s,t)$, they cannot reliably be used to make inference beyond the range of observations, i.e., estimation of quantiles larger than those previously observed; to that end, we turn to asymptotically-justified extreme value models. 
\subsection{Classical extreme-value regression modelling}
\label{intro_evmodel_sec}
 \cite{coles2001introduction} details three main approaches to modelling univariate extreme values of a sequence of independent random variables $Y_1,\dots,Y_n$ with common distribution function $F$; see also the review by \cite{davison2015statistics}. The first concerns block-maxima $M_n=\max\{Y_1,\dots,Y_n\}$; if there exist sequences $\{a_n >0\}$ and $\{b_n\}$ such that 
 \begin{equation}
 \label{max_lim}
 \Pr\{(M_n-b_n)/a_n \leq z \}\rightarrow G(z)\;\; \text{as}\;\; n\rightarrow \infty,
 \end{equation} 
 for non-degenerate $G$, then $F$ is in the max-domain of attraction (MDA) of $G$, the generalised extreme value GEV$(\mu,\sigma,\xi)$ distribution function, where
 \begin{equation}
 G(z\mid\mu,\sigma,\xi)= \begin{cases}\exp\left[-\left\{1+\xi\left(\frac{z-\mu}{\sigma}\right)\right\}_+^{-1/\xi}\right], \;\;&\xi\neq 0,\\
 \exp\left\{-\exp\left(-\frac{z-\mu}{\sigma}\right)\right\},\;\;&\xi=0
 ,\end{cases}
 \label{GEVcdf}
 \end{equation}
 with $\{y\}_+=\max\{0,y\}$, location, scale, and shape parameters $\mu \in \mathbb{R},\sigma >0$, and $\xi \in \mathbb{R}$, respectively, and support $\{z \in \mathbb{R} : 1+\xi(z-\mu)/\sigma > 0\}$. Another approach, the so-called peaks-over-threshold method, utilises the generalised Pareto distribution (GPD), which can be used to model exceedances of $Y$ above some high threshold $u$. Given that $F$ is in the MDA of a GEV$(\mu,\sigma,\xi)$ distribution, then $(Y-u)\mid Y>u$ can be approximated by a GPD$(\sigma_u,\xi)$ random variable with distribution function $H(z)=1-(1+\xi z/\sigma_u)^{-1/\xi}$ and support $z\geq 0$ for $\xi \geq 0$ and $0 \leq z \leq -\sigma_u/\xi$ for $\xi < 0$, and where $\sigma_u=\sigma+\xi(u-\mu)>0$. For both the GEV and GPD models, the shape $\xi$ controls the lower and upper bounds of $H$ and $G$. If $\xi <0$, then $H$ and $G$ are bounded above; for $\xi >  0$, then $G$ is bounded below. Modelling with distributions that have finite bounds dependent on model parameters can lead to unstable inference \citep{smith1985maximum}, and so we constrain $\xi >0$ throughout; the assumption that wildfire burnt areas are heavy-tailed has been shown to hold in a number of studies (see references in \cite{pereira2019statistical}). \cite{castro2021practical} overcome the finite lower-bound problem for $G$ when $\xi > 0$, which we discuss and extend in Section~\ref{bGEV_sec}. {To capture non-stationarity in extremes, we can represent the GEV and GPD parameters as functions of $\mathbf{X}(s,t)$, and denote these distributions by GEV$(\mu(s,t),\sigma(s,t),\xi(s,t))$ and GPD$(\sigma_u(s,t),\xi(s,t))$, respectively, where $\sigma_u(s,t)$ is dependent on a spatiotemporal threshold $u(s,t)$. We term such an approach as an extreme value regression model.}\par
Use of the GEV and GPD distributions for spatiotemporal regression is not practical in all applications. The GEV distribution is applicable to block-maxima and so transformation of covariates may be necessary to make modelling feasible, e.g., aggregating blocks of predictors. Whilst the GPD does not suffer from this issue, it requires two separate models for full inference on the distribution of $Y(s,t)$, i.e., models for $Y(s,t)\mid Y(s,t) > u(s,t)$ and $\Pr\{Y(s,t) > u(s,t)\}$; moreover, the scale parameter $\sigma_u(s,t)$ is dependent on the threshold $u(s,t)$, which makes interpretation of its drivers difficult. \cite{smith1989} details a third method for modelling extremes using point processes (PPs), which circumvents the issues associated with the GEV and GPD models. Assume limit \eqref{max_lim} holds and $G$ has lower- and upper-endpoints $z_-$ and $z_+$, respectively. Then for any $u>z_-$, the sequence of point processes $N_n=\{(i/(n+1),(Y_i-b_n)/a_n): i =1,\dots,n\}$ converges on regions $(0,1)\times(u,\infty)$ as $n\rightarrow \infty$ to a Poisson point process with intensity measure $\Lambda$ of the form $\Lambda(A)=-(t_2-t_1)\log G(z)$, where $A=[t_1,t_2]\times [z,z_+)$ for $ 0\leq t_1 \leq t_2 \leq 1$; with an abuse of notation, we write $Y\sim\mbox{PP}(\mu,\sigma,\xi; u)$ if $Y$ satisfies the limiting Poisson process conditions. Details for modelling using this approach and an extension thereof are given in Section~\ref{bgev_PP_sec}.\par
{As with the GEV and GPD models, extreme value regression models can be built using PPs by letting the corresponding GEV parameters vary with covariates. Consider now a generic parameter set $\{\boldsymbol{\theta}(s,t):s \in \mathcal{S}, t \in \mathcal{T}\}$ which collects the parameter of an extreme value regression model and which is dependent on the predictors, i.e., for all $i=1,\dots,p$ where $p$ is the number of model parameters in $\boldsymbol{\theta}(s,t)=(\theta_1(s,t),\dots,\theta_p(s,t))^T$, we have $\theta_i(s,t)=m_i(\mathbf{x}(s,t))$ for functions $m_i:\mathbb{R}^d\rightarrow \mathbb{R}$.} Similar extreme value regression models using linear functions for $m_i$ has been exploited in a number of spatial studies \citep{mannshardt2010downscaling,davison2012statistical,eastoe2019nonstationarity}. However, such models are incapable of capturing complex, non-linear relationships between the predictors and the response. Bayesian hierarchical modelling can be used to capture behaviour not explainable by linear models by assuming that the marginal parameters come from a latent process \citep{casson1999spatial,cooley2007bayesian,sang2010continuous,opitz2018inla, hrafnkelsson2021max}; these methods often require parametric assumptions for inference and scale poorly for large $d$ or for many observations. Semi-parametric regression has also been considered for spatial extremes modelling, where distribution parameters are represented as smooth, additive functions of predictors using splines or piece-wise linear functions \citep{chavez2005generalized,jonathan2014return,youngman2019generalized, zanini2020flexible}. For example, \cite{youngman2019generalized} estimate extreme spatiotemporal quantiles by representing the parameters of the GPD and exceedance threshold using generalised additive models (GAMs). Whilst semi-parametric approaches provide a much more flexible class of models compared to linear models, they can be computationally expensive to fit, particularly if $d$ or the number of observations are large. 
\subsection{Machine learning for extreme-value analysis}
Recent literature has seen the development of machine learning approaches for extreme value analyses. These algorithms often lack interpretability, but can be used to produce accurate and computationally fast predictions of extreme values for high-dimensional datasets; moreover, they are capable of capturing much more complex structures in data than linear models or GAMs. Such approaches applied in a regression context include the fitting of GPD models with trees \citep{farkas2020cyber}, random forests \citep{gnecco2022extremal} and gradient boosting \citep{velthoen2021gradient}. Other machine learning algorithms have also been adapted for modelling extreme values, e.g., anomaly detection and classification \citep{clifton2014extending,rudd2017extreme,vignotto2020extreme},  sparse learning \citep[see the review by][]{engelke2021sparse}, principal component analysis \citep{cooley2019decompositions,drees2021principal}, and clustering \citep{chautru2015dimension,janssen2020k}. \par
A somewhat untapped class of tools for modelling spatiotemporal extreme values are deep learning methods or neural networks (NNs). In the extreme value literature, \cite{cannon2018non} perform multiple-level extreme quantile regression for precipitation using a monotone quantile regression network. \cite{cannon2010flexible, cannon2011gevcdn}, \cite{vasiliades2015nonstationary}, \cite{bennett2015historical} and \cite{shrestha2017projecting} use NNs to estimate GEV parameters and \cite{rietsch2013network} apply a similar approach to fit GPD models; we refer to neural networks that are specifically used to fit parametric regression models as conditional density estimation networks (CDNs). \cite{carreau2007hybrid} and \cite{carreau2011stochastic} use CDNs to fit mixture models, with the upper-tails characterised using the GPD. A spatial extremes analysis is conducted by \cite{ceresetti2012evaluation}, who produce spatial maps of precipitation return levels by estimating non-stationary GEV and GPD parameters using a simple one-layer feed-foward network; this, and the above methods, all utilise multi-layered perceptron (MLP) neural networks. Here we use MLP to describe the class of feed-forward neural networks where all layers are densely-connected  \citep[see Appendix~\ref{dense_sec} of ][]{papersup} and which are not particularly well suited for identifying spatial or temporal structures within predictors; hence we consider more complex networks better suited to modelling spatial data. A comprehensive review of MLPs and their extensions, to be discussed, is given by \cite{ketkar2017deep}. In the context of extreme value analysis, extensions of MLPs are utilised by \cite{allouche2021approximation}, who construct a generative adversial network (GAN) for estimating quantiles of heavy-tailed distributions, and \cite{boulaguiem2021modelling}, who use a GAN for modelling extreme spatial dependence and performing stochastic weather generation.\par For spatial problems, a better alternative to an MLP are those that include layers with convolutional filters, often referred to as convolutional NNs \citep[CNNs;][]{lecun1995convolutional}. CNNs were first advocated for image classification due to their ability to capture spatial structures within images; \cite{medina2017comparison} and \cite{driss2017comparison} find that CNNs perform much better than MLPs for spatial classification problems. CNNs have been used in a number of environmental applications: classification \citep{gebrehiwot2019deep,zhang2019forest}, regression \citep{rodrigues2020beyond,yu2020probabilistic}, and downscaling \citep{harilal2021augmented}. \cite{lenzi2021neural} and \cite{sainsbury2023likelihood} use CNNs to fit max-stable processes for modelling extreme spatial dependence. 
\subsection{Towards interpretable deep learning for extremes}
\label{intro-towards-sec}
A potential drawback of using NNs for statistical modelling is their lack of interpretability, as they often contain a large number of estimable parameters; a trait shared by many other machine learning tools. Recent work has been done on interpreting the outputs of machine learning algorithms, see, e.g., \cite{bau2017network}, \cite{zhang2018visual}, and \cite{samek2021explaining}, but not in the context of performing regression. However, \cite{zhong2021neural} propose a framework for non-parametric quantile regression using neural networks which are partially linear, with the linear part of the model considered to be interpretable. For a univariate response variable $Y$, they represent its quantile at a pre-specified level by $\mathbf{x}^T\boldsymbol{\eta}+ m(\mathbf{z})+\epsilon$, where $(\mathbf{x},\mathbf{z}) \in \mathbb{R}^l \times \mathbb{R}^{d-l}$ are covariates, $\boldsymbol{\eta} \in \mathbb{R}^l$ is a set of parameters and $m:\mathbb{R}^{d-l}\rightarrow\mathbb{R}$ is an unknown function, with heteroscedastic errors $\epsilon$. Here $\mathbf{x}$ and $\mathbf{z}$ are both covariates that affect the distribution of $Y$, but only the contribution from $\mathbf{x}$ is of interest. Inference on the effect of $\mathbf{x}$ on $Y$ is conducted by considering the estimated values of the regression coefficients $\boldsymbol{\eta}$, whilst the covariates $\mathbf{z}$ feed an MLP which is used to estimate the unknown function $m$; inference on $m$ is not necessary as their interest lies in the effect of $\mathbf{x}$ on $Y$ only. {\cite{ruegamer2024} extend this framework by constructing semi-structured additive regression models. They allow the distributional parameters of $Y\mid\mathbf{X}$ to be partially-linear and further incorporate an extra additive component alongside the linear one, which can be modelled using splines. By replacing the usual linear functions in generalised linear models with an additive representation of parametric/semi-parametric functions and neural networks, their approach balances interpretability with the high predictive accuracy of deep learning methods. In this paper, we extend their framework by constructing such models that are tailored to regression for spatiotemporal extremes; we refer to these as partially-interpretable neural networks (PINNs) for extreme values \cite[PINNEVs; see][]{pinnEV}. In particular, we construct a novel extreme value point process PINN model for U.S. wildfires, which exploits convolutional neural networks}.\par
{Consider a parameter set $\boldsymbol{\theta}(s,t)=(\theta_1(s,t),\dots,\theta_p(s,t))^T$ for an arbitrary spatiotemporal regression model, i.e., where we assume $Y(s,t)\sim\mathcal{F}(\boldsymbol{\theta}(s,t))$ for some well-defined statistical model $\mathcal{F}$. Consider a general component $\theta(s,t)$ of $\boldsymbol{\theta}(s,t)$; then, for all space-time locations $(s,t)\in\mathcal{S}\times\mathcal{T}$, we divide the predictor process $\mathbf{X}(s,t)$ into two complementary components. For $I \in \mathbb{N}$ with $I \leq d$, let $\mathbf{X}_{\mathcal{I}}(s,t)\in\mathbb{R}^{I}$ and $\mathbf{X}_{\mathcal{N}}(s,t)\in\mathbb{R}^{d-I}$ be distinct sub-vectors of $\mathbf{X}(s,t)$, with observations of each component denoted $\mathbf{x}_{\mathcal{I}}(s,t)$ and $\mathbf{x}_{\mathcal{N}}(s,t)$, respectively. Note that the indices mapping $\mathbf{x}(s,t)$ to $\mathbf{x}_{\mathcal{I}}(s,t)$ are consistent for all $(s,t)$, and similarly for $\mathbf{x}_{\mathcal{N}}(s,t)$. Then, we represent the general parameter $\theta(s,t)$ through the unified model 
\begin{align}
\theta(s,t)=h[\eta_{0}+m_{\mathcal{I}}\{\mathbf{x}_\mathcal{I}(s,t)\}+m_\mathcal{N}\{\mathbf{x}_\mathcal{N}(s,t)\}]
\label{full_model}
\end{align}
for constant intercept $\eta_0 \in \mathbb{R}$, an ``interpretable'' function $m_{\mathcal{I}}:\mathbb{R}^{I}\rightarrow\mathbb{R},$ a (potentially) non-linear function $m_\mathcal{N}:\mathbb{R}^{d-I}\rightarrow\mathbb{R},$ and a link function $h:\mathbb{R}\rightarrow\mathbb{R}$ controlling the range of $\theta(s,t)$. The function $m_\mathcal{N}$ is approximated via a neural network with a suitable architecture, whilst $m_\mathcal{I}$ must be specified so that its estimates are interpretable, that is, practitioners can easily quantify the impact of ``interpreted'' predictors $\mathbf{X}_{\mathcal{I}}(s,t)$ on $\theta(s,t)$. Thus, we choose $\mathbf{x}_{\mathcal{I}}(s,t)$ as predictors of interest, for which we wish to infer their relationship on $\theta(s,t)$; all other predictors go into $\mathbf{x}_{\mathcal{N}}(s,t)$. 

Examples for $m_{\mathcal{I}}$ include parametric functions (e.g., linear or polynomial), semi-parametric functions (e.g., additive splines), or combinations thereof \citep[as in Appendix~\ref{sim_study}, ][]{papersup}. Note that $m_{\mathcal{I}}, m_{\mathcal{N}},$ and the indices mapping $\mathbf{x}(s,t)$ to $\mathbf{x}_{\mathcal{I}}(s,t)$ can differ amongst components of the parameter set $\boldsymbol{\theta}(s,t)$. We consider two cases below: i) $m_{\mathcal{I}}(\cdot)$ is linear, with $m_{\mathcal{I}}=\mathbf{x}^T\boldsymbol{\eta}$ for $\boldsymbol{\eta}\in\mathbb{R}^{I},$ and ii) $m_{\mathcal{I}}(\cdot)$ is modelled using splines. In both cases, for simplicity, we choose not to model interactions between covariates in $\mathbf{X}_\mathcal{I}(s,t)$. However, we note that these can be incorporated into the PINN modelling framework in the usual fashion as for classical regression methods, i.e., with product terms in a linear model or with a tensor product of splines in a GAM. We focus on the case where  $\mathcal{F}$ is an asymptotically-justified extreme value point process model (see Section~\ref{bgev_PP_sec}); hence, the strength and novelty of our proposed framework is that it unifies, for the first time, the {theoretical guarantees of parametric extreme value models for extrapolation into the tails}, the interpretability of generalised additive models, and the predictive power of artificial neural networks in a single model. Whilst our focus is on estimating an extreme value point process model, it is trivial to see how the PINN framework can be adapted to estimate other popular parametric statistical distributions, as well as perform more classical applications of NNs, e.g., classification and mean, median, or quantile regression; we perform logistic regression using a PINN model in Section~\ref{app_sec}.} \par
\subsection{Paper outline}
The rest of the paper is outlined as follows. Section~\ref{Wildfire_Data_sec} describes the data used in our analyses and the scientific questions we address. In Appendices~\ref{GAM_sec} and \ref{NN_sec} of the Supplementary Material \citep{papersup}, we describe the additive functions and neural networks used to model $m_{\mathcal{I}}$ and $m_{\mathcal{N}}$, respectively.
 The use of neural networks in our unified framework creates new challenges for fitting extreme value models where distributional lower bounds are dependent on model parameters; a discussion of these issues is provided in Appendix~\ref{NN_train_sec}, with details of a new approach for modelling extreme values using point processes that solves these problems given in Section~\ref{bgev_PP_sec}.  To illustrate the efficacy of our modelling framework, we apply our approach to simulated data in Appendix~\ref{sim_study} (with a concise summary provided in Section~\ref{sim_mainpaper_sec}) and we perform an in-depth study of U.S. wildfires data in Section~\ref{app_sec}; many different candidate models are proposed and their individual predictive performances are thoroughly assessed.
 \section{Wildfire data}
\label{Wildfire_Data_sec}
\begin{figure*}[t]
\begin{minipage}{0.495\linewidth}
\flushleft
\hspace{-.2cm}
\includegraphics[width=0.8\linewidth]{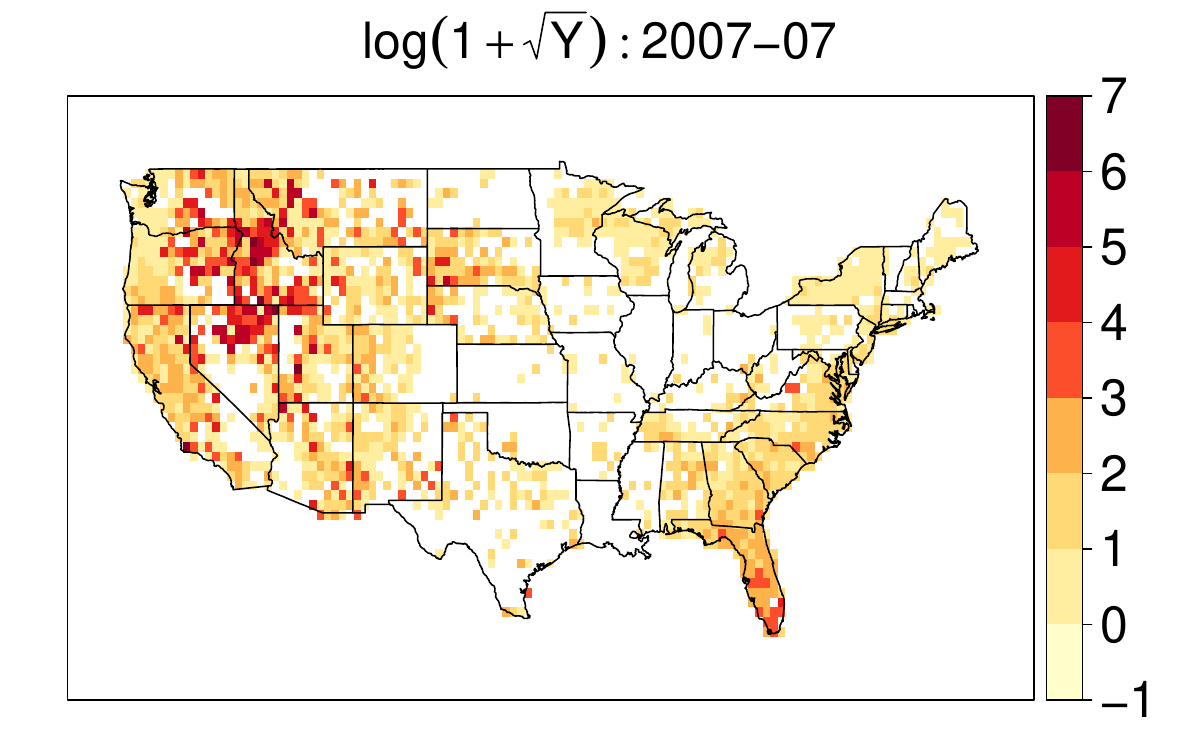} 
\end{minipage}
\vspace{-0.2cm}
\begin{minipage}{0.495\linewidth}
\flushleft
\includegraphics[width=0.8\linewidth]{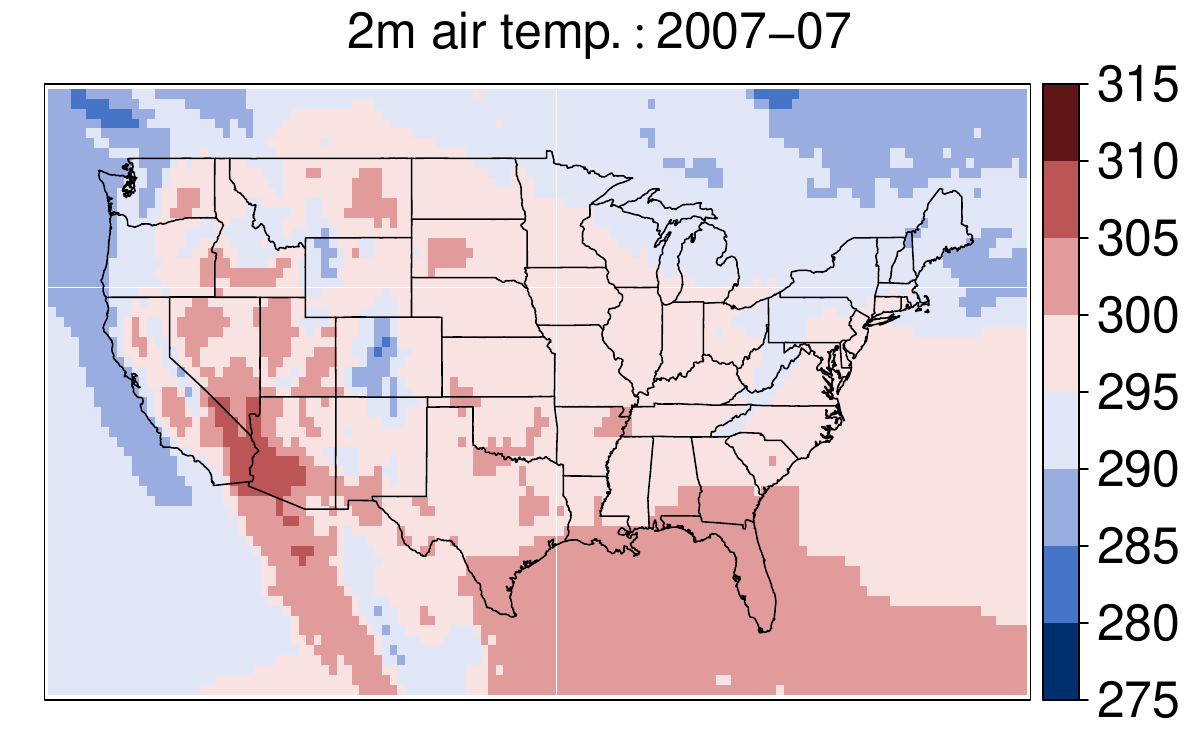} 
\end{minipage}
\vspace{-0.4cm}
\begin{minipage}{0.495\linewidth}
\flushleft
\includegraphics[width=0.8\linewidth]{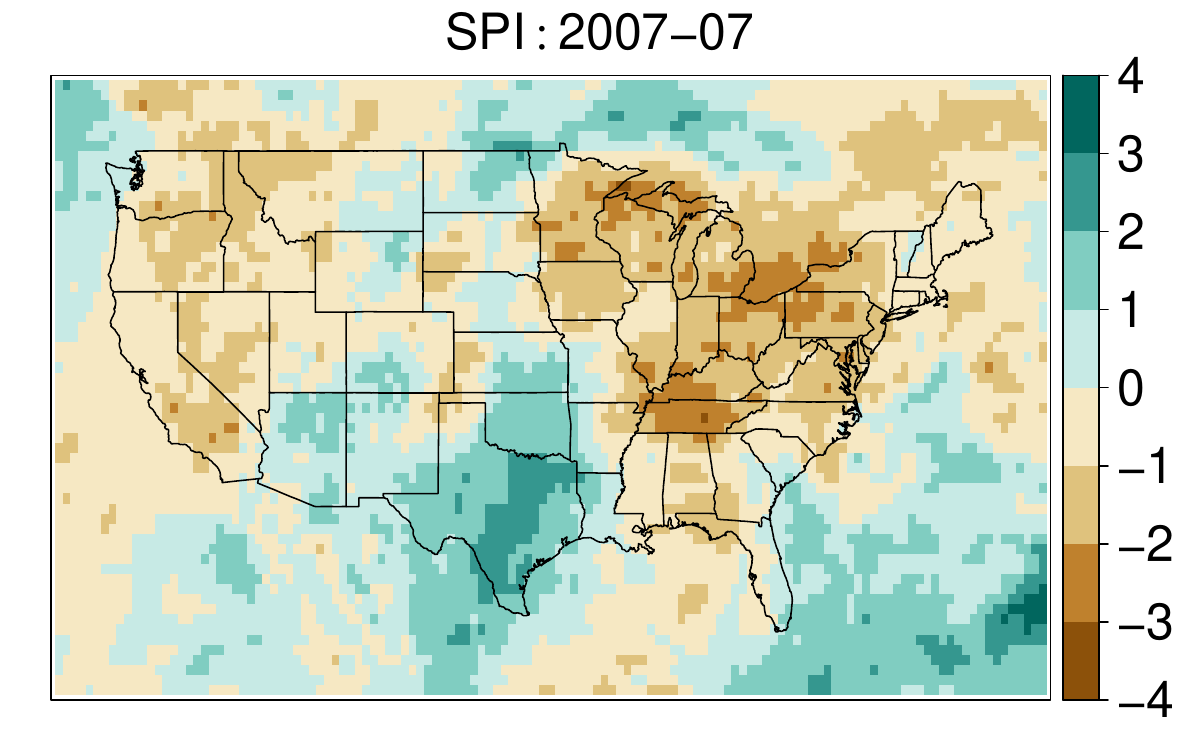} 
\end{minipage}
\begin{minipage}{0.495\linewidth}
\flushleft
\includegraphics[width=0.8\linewidth]{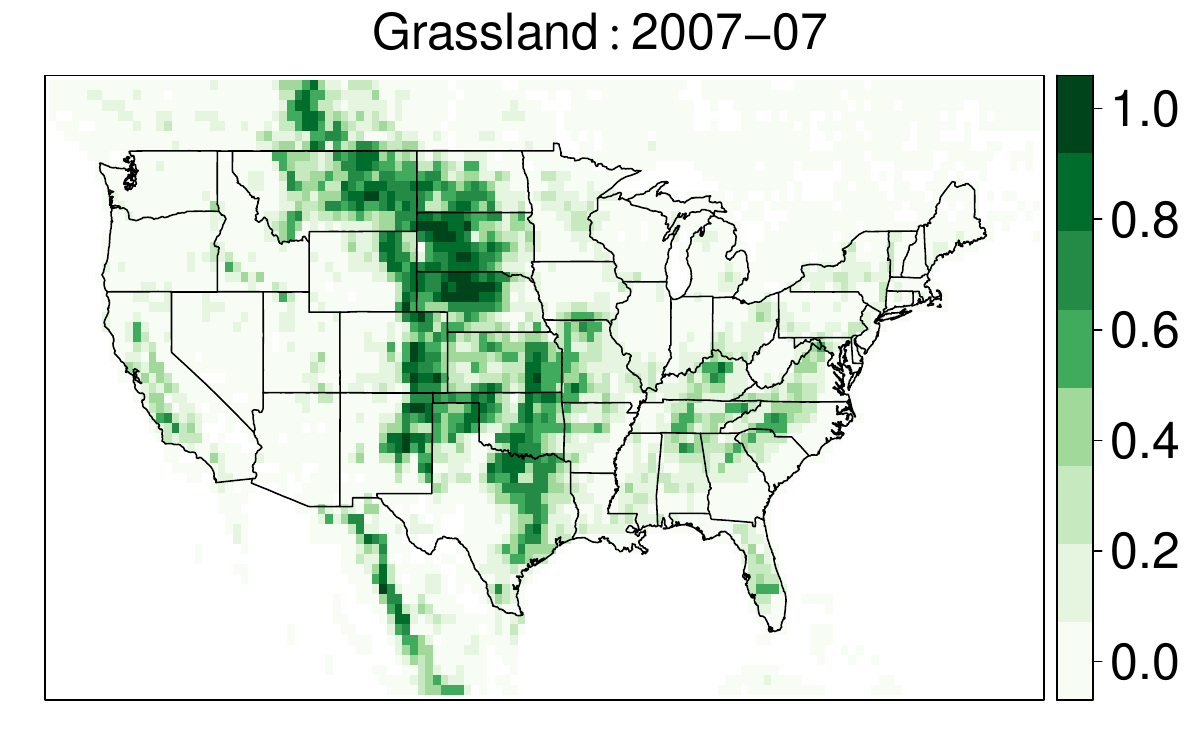} 
\end{minipage}
\caption{Maps of observed $\log\{1+\sqrt{Y}(s,t)\}$ (log-$\sqrt{\mbox{acres}}$; top-left), 2m air temperature (K; top-right), 3-month SPI (unitless; bottom-left), and proportion of grassland coverage (unitless; bottom-right) for July, 2007. }
\label{obs_maps}
\end{figure*}
The response data $Y(s,t)$ we use in our analyses are observations of monthly aggregated burnt area (acres) of 3503 spatial grid-cells located across the contiguous United States\footnote{The states of Alaska and Hawaii are excluded.}; the spatial domain $\mathcal{S}$ is illustrated in Figure~\ref{obs_maps}.
The observation period covers 1993 to 2015, using only months between March and September, inclusive, leaving 161 observed spatial fields. Grid-cells are arranged on a regular latitude/longitude grid with spatial resolution $0.5^{\circ} \times 0.5^{\circ}$. Observations are provided by the Fire Program Analysis fire-occurrence database \citep{short2017spatial} which collates U.S. wildfire records from the reporting systems of federal, state and local organisations. A subset of these data are described by \cite{opitz2022editorial} and have been previously studied as part of a data prediction challenge for the Extreme Value Analysis (EVA) 2021 conference, see, e.g., \cite{cisneros2021combined}, \cite{d2021flexible}, \cite{koh2021gradient}, and \cite{zhang2022joint}, with \cite{ivek2022reconstruction} using deep learning methods.  Whilst the focus of this challenge was on predicting extreme quantiles for missing spatiotemporal locations, our focus is instead on inference of relationships between predictors and extreme quantiles of burnt area; hence, we utilise all of the available data for modelling.\par
{We have $d=42$ predictors} of three classes: land coverage, orographical, and meteorological. The land cover variables that are considered describe the proportion of a grid-cell which is covered by one of {25 different types}, e.g., urban, grassland (illustrated in Figure~\ref{obs_maps}), water \citep[for full details, see][]{bontemps2015multi}. Land cover predictors are derived using a gridded land cover map, of spatial resolution $300$m and temporal resolution one year, produced by Copernicus and available through their Climate Data Service. For each $0.5^{\circ} \times 0.5^{\circ}$ grid-cell, the proportion of a cell consisting of a specific land cover type is derived from the high-resolution product. 
{In addition, 13 meteorological variables are considered} and given as monthly means. These variables are provided by the ERA5-reanalysis on single levels, also available through the Copernicus Climate Data Service, which is given on a $0.1^{\circ} \times 0.1^{\circ}$ grid; the values are then aggregated to a $0.5^{\circ} \times 0.5^{\circ}$ resolution. {The variables are: eastern and northern components of wind velocity at 10m above ground level (m/s), temperature at 2m above ground level (K; see Figure~\ref{obs_maps}), potential evaporation (m), evaporation (m of water equivalent), surface pressure (Pa), surface net solar and thermal radiation (J/m$^2$), snowfall, snow-melt, and snow evaporation (m of water equivalent), surface run-off (m), and a three-month standardised precipitation index\footnote{ The 3-month SPI is derived under the assumption that observations of monthly total precipitation follow a gamma distribution.} (SPI, unitless; see Figure~\ref{obs_maps}).  The reanalysis also provides three variables relating to sub-gridscale orography: aspect (radians), anisotropy (unitless), slope (unitless), as well as the standard deviation of orography (unitless). }
Note that, for us to utilise CNN layers in our model, the spatial domain over which predictors are observed is slightly larger than that associated with the response, see Figure~\ref{obs_maps} and Section~\ref{sec:pre_process}. \par
In Section~\ref{sec:interp_res}, we focus on quantifying the drivers of wildfire occurrence and the extremal behaviour of wildfire spread. {In particular, we interpret the effect of 2m air temperature and SPI on wildfire occurrence and spread, as these drivers are meteorological variables that will see their own extremal behaviour influenced by climate change. Note that, to reduce interactions (i.e., confounding) between the interpreted predictors and those in $\mathbf{x}_\mathcal{N}(s,t)$, we a priori estimated all pairwise correlations and removed any predictors in $\mathbf{x}_\mathcal{N}(s,t)$ that were highly correlated (correlation values $>0.5$ or $< -0.5$) with air temperature or SPI. The $d=42$ predictors described above are those that remained.} We also use our proposed methodology to produce maps of monthly occurrence risk and extreme quantiles for wildfire spread; the former can be used to identify areas of high susceptibility to wildfire occurrence, whilst the latter illustrates locations where it is of particular pertinence to prevent wildfires.
\section{Extreme point process model}
\label{bgev_PP_sec}
\subsection{Modelling extremes with point processes}
To model non-stationary extreme values with the PP approach, we first assume that limit $\eqref{max_lim}$ holds at each space-time location $(s,t)\in\mathcal{S}\times\mathcal{T}$. That is, for independent realisations $Y_i(s,t)$ of $Y(s,t)$ for $i=1,\dots,n$, a suitable normalisation of the $n_y$-block maxima, denoted $\max_{i=1,\dots,n_y}\{Y_i(s,t)\}$, follows a $\mbox{GEV}(\boldsymbol{\theta}(s,t))$ distribution with density $g$ and lower-endpoint $z_-(s,t)$; {note that, as we assume $\xi(s,t)>0$ for all $(s,t)$, the upper-endpoint of $Y(s,t)$, for all $(s,t)$, is infinite}. Then define a {1-dimensional} point process on the space $A_{s,t}=(u(s,t),\infty)$ for $u(s,t)>z_-(s,t)$ with corresponding intensity measure and density $\Lambda_{s,t}([z,\infty))=-(n/n_y) \log G(z)$ and $\lambda_{s,t}(z)=-\mathrm{d} \Lambda_{s,t}([z,\infty))/\mathrm{d}z \propto g(y)/G(z)$, respectively, for $z > u(s,t)$. Common practice is to take $n_y$ to be the number of observations in a year, e.g., for monthly observations $n_y=12$, so that the GEV parameters correspond to those for the associated annual maxima distribution. {We take $n_y=12$, but note that we have defined a 1-dimensional point process for $Y(s,t)$, instead of the customary 2-dimensional point process (see Section~\ref{intro_evmodel_sec}), as the model parameters vary with each space-time location $(s,t)$. Hence, it is misleading to refer to the annual maxima distribution, as we observe only a single realisation of the process $\{Y(s,t): s \in \mathcal{S}, t \in \mathcal{T}\}$, which is non-stationary in both space and time. It follows that $n=1$ }and the negative log-likelihood for a non-stationary point process model is 
\begin{align}
\label{PP_nll}
l(\boldsymbol{\theta}(s,t))&=\sum_{t \in \mathcal{T}}\sum_{s \in \mathcal{S}}\bigg[(1/n_y)\Lambda_{s,t}([u(s,t),\infty))-\mathbbm{1}\{y(s,t)> u(s,t)\}\log\lambda_{s,t}(y(s,t))\bigg],
\end{align} 
where $\mathbbm{1}\{\cdot\}$ denotes the indicator function \citep[see, Ch. 7.,][]{coles2001introduction}. {In this way, $Y(s,t)\sim \mbox {PP}(\boldsymbol{\theta}(s,t))$, which implies that $\{F_{s,t}(y)\}^{n_y}=G(y \mid  \boldsymbol{\theta}(s,t))$, where $F_{s,t}$ denotes the distribution function of $Y(s,t)$ and $G$ in \eqref{GEVcdf} is the GEV distribution function.}\par
As with GPD models, the threshold $u(s,t)$ must be estimated and fixed a priori. However, here the threshold is used in inference and the resulting parameters are not dependent on $u(s,t)$. If $y(s,t) < z_-(s,t)$, then \eqref{PP_nll} cannot be evaluated; for the reasons detailed in Appendix~\ref{NN_train_sec} of \cite{papersup}, this makes neural networks an unsuitable tool for fitting the classical point process model. To this end, we replace the GEV distribution $G$ with the blended-GEV (bGEV) distribution $G_b$, which has the same bulk and upper-tail but carries an infinite lower-bound, leading to a new extreme value point process model, described next, that can be fitted using deep learning methods.
\subsection{Blended GEV point process model}
\label{bGEV_sec}
Following \cite{castro2021practical}, we first re-parameterise the GEV distribution in terms of a new location parameter $q_\alpha \in \mathbb{R}$, the $\alpha$-quantile ($0<\alpha<1$) of the GEV model, and a spread parameter $s_\beta >0$, defined as $s_\beta=q_{1-\beta/2}-q_{\beta/2}$ for $0<\beta<1$; we denote the newly parameterised GEV distribution by GEV($q_\alpha,s_\beta,\xi$). There exists a one-to-one mapping between ($\mu,\sigma)$ and ($q_\alpha,s_\beta$); for $\xi>0$, we have
$\mu=q_\alpha-s_\beta( l_{\alpha,\xi}-1)/(l_{1-\beta/2,\xi}-l_{\beta/2,\xi})$ and $\sigma=\xi s_\beta/(l_{1-\beta/2,\xi}-l_{\beta/2,\xi})$, where $l_{x,\xi}=(-\log(x))^{-\xi}$; if $\xi=0$, we instead have $\mu=q_\alpha+s_\beta l_{\alpha}/(l_{\beta/2}-l_{1-\beta/2})$ and $\sigma= s_\beta/(l_{\beta/2}-l_{1-\beta/2})$, where $l_{x}= \log(-\log x)$.\par
The bGEV($q_\alpha,s_\beta,\xi$) distribution combines the lower-tail of the Gumbel distribution, with distribution function $G_{G}= \mbox{GEV}(\tilde{q}_\alpha,\tilde{s}_\beta,0)$, with the bulk and upper-tail of the Fr\'echet distribution ($\xi>0$) by mixing the two distributions in a small interval $[b_1,b_2]$. Precisely, the bGEV distribution function is defined by
\begin{align}
\label{bGEV_cdf}
G_{b}(z\mid q_\alpha,s_\beta,\xi,b_1,b_2)&=G_{G}(z\mid\tilde{q}_\alpha,\tilde{s}_\beta)^{1-p(z;b_1,b_2)}\times G(z\mid q_\alpha,s_\beta,\xi)^{p(z;b_1,b_2)},
\end{align}
where the weight function $p$ is defined by $p(z;b_1,b_2)=F_{beta}((z-b_1)/(b_2-b_1)\mid c_1,c_2)$, for $F_{beta}(x\mid c_1,c_2)$, the distribution function of a beta random variable with shape parameters $c_1>0,c_2>0$. Note that $p(z;b_1,b_2)=0$ for $z \leq b_1$ and $p(z;b_1,b_2)=1$ for $z \geq b_2$ and so the lower- and upper-tails of $G_{b}$ are completely determined by $G_{G}$ and $G$, respectively. For continuity of $G_{b}$, we require $G_{G}(b_1)=G(b_1)$ and $G_{G}(b_2)=G(b_2)$; for this we set $b_1=G^{-1}(p_{b_1})$ and $b_2=G^{-1}(p_{b_2})$ for small $0 < p_{b_1},p_{b_2} <1$ and let $\tilde{q}_\alpha=b_1-(b_2-b_1)(l_\alpha-l_{p_{b_1}})/(l_{p_{b_{1}}}-l_{p_{b_{2}}})$
and $\tilde{s}_\beta=(b_2-b_1)(l_{\beta/2}-l_{1-\beta/2})/(l_{p_{b_1}}-l_{p_{b_2}})$.
 To ensure the log-density of $G_{b}$ is continuous, the parameters $c_1$ and $c_2$ are restricted to $c_1>3, c_2>3$. Values of the hyper-parameters must be chosen a priori; we follow \cite{castro2021practical} and \cite{vandeskog2021modelling} and set $\alpha=\beta=0.5,p_{b_1}=0.05,p_{b_2}=0.2,$ and $c_1=c_2=5$. {Note that the latter four hyper-parameters determine the shape of the bGEV distribution function below the $p_b$-quantile, and so their choice does not impact the upper-tail; see \cite{castro2021practical}.}\par
 As the $G_{b}$ distribution has the same lower-endpoint as $G_{G}$, i.e., negative infinity, the bGEV distribution can be used for modelling without the NN training issues associated with the GEV distribution. To utilise the bGEV model in a PP framework, we simply replace the mean measure $\Lambda_{s,t}$ in \eqref{PP_nll} with $\Lambda_{s,t}([z,\infty)):= -(1/n_y)\log G_{b}(z)$, which is a valid measure by construction; we denote this new model as the bGEV-PP approach and write $Y(s,t)\mid\mathbf{X}(s,t)\sim\mbox{bGEV-PP}(\boldsymbol{\theta}(s,t); u(s,t))$ where $\boldsymbol{\theta}(s,t) = (q_\alpha(s,t),s_\beta(s,t),\xi(s,t))^T$ when used within a spatiotemporal framework.
Notice that maxima $M_n=\max(Z_1,\dots,Z_n)$ conditional on $n$, where $Z_1,\dots,Z_n\sim \mbox{bGEV-PP}$, follow the bGEV distribution, i.e., $M_n\sim \mbox{bGEV}$. This can be verified as \begin{align*}
\Pr(M_n \leq z)&=\Pr(\mbox{No point in }[z,\infty))=\exp(-\Lambda_{s,t}(A_z))=G_b(z),
\end{align*} which makes it clear that the bGEV-PP model extends the classical extreme value PP model for threshold exceedances in the same way as the bGEV distribution extends the GEV distribution for maxima. \par
 We choose to use the proposed bGEV-PP model over the classic GPD. Our reasoning is twofold: firstly, the focus of our analysis is to provide extreme quantile estimation whilst simultaneously identifying the drivers of extreme wildfire risk. As described in Section~\ref{intro_evmodel_sec}, it is difficult to do identify drivers with the GPD as $\sigma_u$ is dependent on $u(s,t)$, which is itself dependent on the predictors. {For the bGEV-PP model, the threshold $u(s,t)$ defines observations deemed as ``extreme'' for inference, but, given that it is sufficiently large, the parameters $q_\alpha(s,t),s_\beta(s,t),$ and $\xi(s,t)$, are independent of $u(s,t)$.}  Secondly, application of the GPD requires training of three networks; one for each of the $u(s,t)$, GPD, and $\Pr\{Y(s,t) \leq u(s,t)\}$ models \citep[see, e.g.,][]{opitz2018inla}, whilst the bGEV-PP approach requires only one for $u(s,t)$ and one for the bGEV-PP. The added computational demand and the potential for increased model uncertainty leads us to favour the bGEV-PP approach. We find in Section~\ref{app_sec} that the bGEV-PP approach provides fantastic model fits to the data. {For details on inference with the bGEV-PP PINN model, see Appendix~\ref{NN_train_sec} \citep{papersup}.}
  \subsection{Simulation study}
 \label{sim_mainpaper_sec}
{We illustrate the efficacy of the bGEV-PP approach by applying it to simulated data; the full details of our study are provided in Appendix~\ref{sim_study} of the Supplementary Material \citep{papersup}, but we summarise the salient findings here. Two simulation studies are conducted; the first investigates estimation of the interpretable function $m_\mathcal{I}(\cdot)$. Appendix~\ref{par_est_sec} illustrates that our approach can accurately estimate linear and spline-based representations for $m_\mathcal{I}(\cdot)$, when the interpretable predictor set $\mathbf{x}_\mathcal{I}(s,t)$ is correctly specified; this is the case even when $m_\mathcal{I}(\cdot)$ is zero everywhere, i.e., the interpretable covariates have no effect on the response. The second study in Appendix~\ref{misspec_sec} considers the scenario where the PINN structure in \eqref{full_model} or the bGEV-PP response distribution is misspecified. In both cases, we show that our approach is still able to accurately estimate extreme quantiles.  }
\section{Data application}
\label{app_sec}
\subsection{Pre-processing and setup}
\label{sec:pre_process}
Exploratory analysis reveals that the response variable is very heavy-tailed, with a number of fitted stationary GPD models giving estimates $\hat{\xi} > 1$, and so we use the square root of the response for inference and rescale the predicted distributions accordingly; the response $\sqrt{Y}$ can be interpreted as the ``diameter" of an affected region. A $\log$ transformation was also considered instead of the square root but this led to negative shape parameter estimates, and hence the finite bound problem discussed in Section~\ref{bgev_PP_sec}. As we believe that the driving factors that lead to occurrence and spread of wildfires differ between the two components of the wildfire distribution, we propose separate PINN models for $p_0(s,t):=\Pr\{Y(s,t)>0\mid\mathbf{X}(s,t)=\mathbf{x}(s,t)\}$ and $\sqrt{Y}(s,t)\mid\{Y(s,t)>0,\mathbf{X}(s,t)\}\sim \mbox{bGEV-PP}(\boldsymbol{\theta}(s,t); u(s,t))$; there are $216713$ observations of $Y(s,t)\mid Y(s,t)>0$. Models for $p_0(s,t)$ are fitted by specifying $h$ in \eqref{full_model} as the logistic/sigmoid link function.  To fit the bGEV-PP model we require an exceedance threshold $u(s,t)$, which we estimate as the $\tau$-quantile of strictly positive $\sqrt{Y}(s,t) \mid \mathbf{X}(s,t)$ for some $\tau \in (0,1)$. {This is represented by a neural network of the form $\exp[m_\mathcal{N}\{\mathbf{x}(s,t)\}]$,  i.e., a PINN model with no interpretable aspects (a regular NN) that provides strictly positive quantile estimates, which is trained via minimisation of the tilted quantile loss function \citep{Koenker_2005}; see Appendix~\ref{NN_train_sec} for details. Note that the quantile-level $\tau$ is a tunable hyper-parameter; see Section~\ref{sec:arch}. }\par
{
Climate and wildfire dynamics vary spatially across the contiguous U.S. \citep{gedalof2010climate} and so, although a common approach in many environmental applications, we choose not to fix the shape parameter $\xi(s,t)$ across $(s,t)$. However, estimation of $\xi(s,t)$ in a regression setting can be difficult, particularly in the presence of many predictor variables. Thus, we represent $q_\alpha(s,t)$ and $s_\beta(s,t)$ using the flexible PINN formulation given in \eqref{full_model}, but adopt a simpler model for $\xi(s,t)$: a MLP applied to the latitude and longitude coordinates at $s$, and with $h$ taken to be the sigmoid link function to ensure that $\xi(s,t)\in(0,1)$ for all $(s,t)$. Hence, $\xi(s,t)$ varies spatially but not temporally, and is not dependent on the $d=42$ predictors described in Section~\ref{Wildfire_Data_sec}; we hereafter drop the time $t$ from the notation (i.e., $\xi(s):=\xi(s,t)$). We find that this specification facilitates fantastic model fits and has the added benefit of allowing interpretation of spatial patterns in estimates of $\xi(s)$; see Section~\ref{sec:hazard_maps}. 
}

We assess model and parameter uncertainty by using a stationary bootstrap \citep{politis1994stationary} \citep[see Appendix~\ref{boot_sm_sec} of][]{papersup} with $200$ samples and expected block size of 2 months. The distribution of quantities of interest, e.g., quantiles and models parameters, are estimated by taking the empirical median and pointwise quantiles across all bootstrap samples. {To reduce computation time, we first train a single $p_0(s,t)$, $u(s,t)$, and bGEV-PP PINN models on one bootstrap sample, for 1000 epochs each. Their trained weights and parameter estimates are then used as initial values for training of all subsequent models (for each bootstrap sample), with these networks trained for another 1000 epochs.} 

{To derive more accurate out-of-sample predictions, we use a training and validation scheme. The original data are partitioned into two complementary subsets with $20\%$ of observations used for validation and the remainder used for training.} As the goal is to predict spatiotemporal quantiles, we remove observations for validation that are slightly clustered in space and time. To this end, for each seven month block of observed space-time locations, we simulate a standard Gaussian process $\{Z(s,t)\}$ with separable correlation function $\rho((s_i,t_i),(s_j,t_j))={\exp\{-\|s_i-s_j\|^*/100\}\exp\{-| t_i-t_j|/5\}}$; here $\|\cdot\|^*$ denotes the (geodesic) great-Earth distance in miles. Observation $y(s,t)$ is then assigned to the validation set if the realisation $z(s,t)$ falls below the $0.2$-quantile of the simulated $\{z(s,t):s \in \mathcal{S}, t\in\mathcal{T}\}$. {Models are fitted by applying stochastic gradient descent over a finite number of epochs, via the adaptive moment estimation (Adam) algorithm \citep{kingma2014adam}. As we have few observations of the space-time process, we use all training data to estimate the gradient of the loss at each epoch (maximal mini-batch size), rather than a subset (as is common practice with stochastic gradient descent). Whilst the model is optimised using the training data only, we evaluate the model loss on the validation data and save the current state of the model (i.e., the current parameter estimates) at each epoch. In order to obtain the model fit that provides the best out-of-sample predictions, we use only the ``best'' model fit, which is the state that minimises the validation loss over all epochs. Note that, whilst a different partitioning of training data is used for each bootstrap sample, the same partition is used within a single bootstrap sample for the three model components: $u(s,t)$, $p_0(s,t)$, and the bGEV-PP model.}
 
{We consider two formulations of our PINN models. Recall that the three parameters $p_0(s,t)$, $q_\alpha(s,t),$ and $s_\beta(s,t),$ are represented using the PINN formulation, given by \eqref{full_model}, with $I=2$ interpreted predictors comprising $\mathbf{x}_{\mathcal{I}}(s,t)$: 2m air temperature and SPI. The other $d-I=40$ predictors feed the neural network used to estimate $m_{\mathcal{N}}(\cdot)$. We consider two models for the interpretable function $m_I(\cdot)$ in \eqref{full_model}: a \textit{global} and \textit{local} effect model. For the first model, we use splines to model the global impact of $\mathbf{x}_{\mathcal{I}}(s,t)$ on $p_0(s,t),q_\alpha(s,t),$ and $s_\beta(s,t)$. This is global in the sense that the splines' basis coefficients do not vary with $(s,t)$, and so estimates of $m_I(\cdot)$ model the aggregated effect of $\mathbf{x}_{\mathcal{I}}(s,t)$ on the wildfire distribution. To explore spatial heterogeneity in the impact of the interpreted predictors, we also consider a local model: a spatially-varying coefficient model, whereby $m_I\{\mathbf{x}_{\mathcal{I}}(s,t)\}$ is represented as a linear function with coefficients varying by the state in which location $s$ lies\footnote{Due to their relatively small sizes, we merge the states of Rhode Island and Connecticut with Massachusetts, as well as Maryland with Delaware.}.}

{For the global PINN models, the interpretable function $m_\mathcal{I}$ is modelled using thin-plate splines \citep{wood2003thin}, represented via radial basis functions that are evaluated at ten fixed knots; see Appendix~\ref{GAM_sec} of the Supplementary Material \citep{papersup}. Knots are taken to be marginal quantiles of the predictors $\{\mathbf{X}(s, t) : (s, t) \in \mathcal{S} \times \mathcal{T}\}$ with equally spaced probabilities.} For all models, we improve the numerical stability during model training by standardising all inputs; both linear and NN predictors, and the knot-wise radial basis function evaluations are standardised by subtracting and dividing by their marginal means and standard deviations.  To use convolutional layers, the spatial locations must lie on a rectangular grid \citep[see Appendix~\ref{cnn_sec},][]{papersup}. We map the observation locations to a rectangular grid and we then treat $Y(s,t)$ at locations sampled over water as missing by removing their influence on the loss function. {Note that the meteorological predictors that we consider in our analyses are available over sea and on a sufficiently large domain (encompassing the contiguous U.S.) as to avoid any edge effects related to the use of CNNs.}
{\subsection{Architecture optimisation and measures of performance}
\label{sec:arch}
In order to determine the optimal model architecture (the NN depth, width, and layer type) and quantile-level $\tau$, we optimise the following measures of model performance. For the occurrence probability model $p_0(s,t)$, we measure the models' predictive capabilities using the out-of-sample (evaluated on the validation data) area under the receiver operating characteristic curve (AUC). For the bGEV-PP model, we propose two measures.} The first, a measure of goodness-of-fit, was proposed by \cite{richards2021joint}. We use the predicted distributions to transform all data onto standard exponential margins. Then for $\{0 < p_1 < \dots < p_m < 1\}$, i.e., a grid of $m\geq 2$ equally spaced values with $p_m=1-(p_2-p_1)$, we calculate the standardised mean absolute deviation (sMAD) $(1/m)\sum^m_{j=1}\mid\tilde{q}(p_j)-F_{E}^{-1}(p_j)\mid$, where $\tilde{q}(p_j)$ denotes the empirical $p_j$-quantile of the standardised data and $F_{E}$ denotes the standard exponential distribution function. That is, we estimate the expected deviation in the Q-Q plot for the standardised data against exponential quantiles from the line $y=x$, but only above the $p_1$-quantile. As our interest lies in the fit for extreme values only, we set $p_1=0.95$ and $m=10835$, i.e., $100(1-p_1)\%$ of the number of observations. We evaluate this metric on the original data, not the training nor validation bootstrap data.\par
To evaluate the out-of-sample predictive performance of the bGEV-PP model, we apply the threshold-weighted continuous ranked probability score (twCRPS) \citep{Gneiting2011} which is a proper scoring rule. Define $\mathcal{P}\subset \mathcal{S} \times \mathcal{T}$ as the set of out-of-sample space-time locations and let $\hat{p}(y; s,t)$ denote a model-based estimate of the probability $\Pr\{Y(s,t) < y\}$. Then for a sequence of increasing thresholds $\{v_1,\dots,v_{n_v}\}$ and weight function $w(x)=\tilde{w}(x)/\tilde{w}(v_{n_v}), \tilde{w}(x)=1-(1+(x+1)^2/1000)^{-1/4}$, we define
$\mbox{twCRPS}=\sum_{(s,t)\in\mathcal{P}}\sum^{n_v}_{i=1}w(v_i)[\mathbbm{1}\{y(s,t)\leq v_i\}-\hat{p}(v_i; s,t)]^2$ where the function $w(x)$ puts more weight on extreme predictions. We use 24 irregularly spaced thresholds ranging between $v_1=\sqrt{30}$ and $v_{24}=\sqrt{100000}$.

{The optimal architectures are found by minimising the twCRPS and sMAD, and maximising the AUC, using a coarse grid-search scheme. The optimal quantile level is found to be $\tau=0.8$. For $p_0(s,t)$, $u(s,t)$, $q_\alpha(s,t)$, and $s_\beta(s,t)$, the optimal choice of architecture for $m_\mathcal{N}$ in \eqref{full_model} uses three convolutional layers with $3\times3$ filters and widths of $(32,32,32), (12,12,12), (24,24,24)$, and $(10,10,10)$, respectively. The optimal model for $\xi(s)$ is a MLP with three dense layers, each with a width of eight. The optimal NN model for $u(s,t)$ comprises $7177$ estimable parameters; see Table~\ref{compareTab} for the number of estimable parameters in the $p_0(s,t)$ and bGEV-PP models. Note that the optimal $\tau$ and NN architecture for the $p_0(s,t)$ and bGEV-PP models is consistent across both the global and local model settings (see Section~\ref{sec:pre_process}). } 

\subsection{Model comparison}
\label{sec:model_compare}
{We now perform a study to compare the effect of the functional form of $\theta(s,t)$ in \eqref{full_model} on the fitted models for $p_0(s,t)$ and $\sqrt{Y}(s,t)\mid \{Y(s,t)>0,\mathbf{X}(s,t)\}\sim \mbox{bGEV-PP}(\boldsymbol{\theta}(s,t);u(s,t))$. Three candidate PINN models are considered: the local and global PINN models, described in Section~\ref{sec:arch}, and a simplified local PINN model with homogeneous shape parameter, $\xi(s)=\xi>0$ for all $s\in\mathcal{S}$. We compare these PINN models against two classical statistical methods: a generalised linear model and a generalised additive model (GAM). These models can be considered special cases of the PINN framework, with $I=d$ and $m_\mathcal{N}(\mathbf{x})=0$ for all $\mathbf{x}$. The GAM model uses the same spline specification as the global PINN formulation (see Section~\ref{sec:pre_process}). We also consider a PINN model which permits no interpretable structure (i.e., simply a NN), with $I=0$ and $m_\mathcal{I}(\mathbf{x})=0$ for all $\mathbf{x}$; this model adopts the same neural network architecture as the local and global PINN models (see Section~\ref{sec:arch}). All models are fitted using the same training scheme, validation data, and estimates of $u(s,t)$ (with $\tau=0.8$; as described in Section~\ref{sec:pre_process}).} 

{We fit each of the six candidate models across all bootstrap samples and calculate the performance metrics detailed in Section~\ref{sec:arch}. Table~\ref{compareTab} gives the average of the goodness-of-fit and prediction metrics, as well as the validation loss for the bGEV-PP model, across all bootstrap samples; note that the twCRPS and AUC are evaluated on the validation data, whilst the sMAD is evaluated on the original data. For both the $p_0(s,t)$ and bGEV-PP models, we observe vast improvements in performance and fit when including a neural network in the formulation for $\theta(s,t)$. Both the linear and GAM models give much higher (lower) values of the sMAD, twCRPS, and validation loss (AUC) compared to the NN and PINN models, suggesting that the flexibility afforded by a deep learning approach has drastically improved the model performance. We find that, for the bGEV-PP component, the local and global PINN models facilitate better model performance than the NN model, which may be because we did not optimise the architecture for the latter. However, the NN model does provide the lowest AUC for the $p_0(s,t)$ model component, suggesting that it provides the best prediction of wildfire occurrence probabilities. Interestingly, the local PINN model with homogeneous $\xi(s)$ provides the best predictive power (by minimising twCRPS) for quantiles of the non-zero burnt area distribution, but at the cost of a poorer model fit (in terms of the sMAD).}

{In Section~\ref{sec:interp_res}, we showcase estimates of the interpretable function, $m_\mathcal{I}(\cdot)$, from the local and global PINN models (with heterogeneous $\xi(s)$). In Sections~\ref{sec:hazard_maps} and \ref{sec:temp_trends}, we discuss spatial and temporal trends in estimates of $\theta(s,t)$ and extreme quantiles of the burnt area distribution. For brevity, we present there only model estimates and diagnostics for the local PINN model (with heterogeneous $\xi(s)$), as this model provides a good balance of interpretability, goodness-of-fit, and predictive performance. To further illustrate the goodness-of-fit and suitability of the local PINN model for the bGEV-PP component of our model, we illustrate a pooled {Q-Q} plot in Figure~\ref{fig_exp_fit} of Appendix~\ref{sup_figs} of the Supplementary Material \citep{papersup}, transforming all non-zero data onto standard Exponential margins using the bGEV-PP fits; this procedure is repeated for each bootstrap sample and we observe good fits in the upper-tails as the estimated tolerance bands include the diagonal.}
\begin{table}
\caption{Comparison of functional forms for $\theta(s,t)$. Performance metrics are the median over 200 bootstrap samples. The AUC is calculated for the occurrence probability, $p_0$, models. All other metrics are calculated for the bGEV-PP models and are given as the absolute difference to the lowest value across all models. The same $p_0$ model is used for both ``local'' model formulations; hence, these results are reported only one.}
 \begin{tabular}{c|c|c| c| c|c} 
\hline
$\theta(s,t)$ &\shortstack{Number of parameters \\
($\mbox{bGEV-PP}$/$p_0$)}  & \shortstack{Validation\\ loss}  &  \shortstack{ sMAD \\ ($\times 10^{-2}$)}  & \shortstack{twCRPS} & \shortstack{AUC \\ ($\times 10^{-2}$)} \\
\hline
\hline
linear & 263 / 43&3180.6&4.87&642.1&79.6\\
GAM & 1019 / 421&3043.6&5.78&630.5&81.1\\
NN &25335 / 30657&54.4&9.87&77.1&91.4\\
global PINN & 24763 / 30101&32.9&0&51.1&91.2\\
local PINN &24903 / 30171&0&2.03&66.3&91.1\\
local PINN (homogeneous $\xi$) & 24727 / - &15.0&25.5&0&-\\
\hline
  \end{tabular}
\label{compareTab}
  \end{table}

\subsection{Identifiable drivers of U.S. wildfires}
\label{sec:interp_res}
\begin{figure}[t!]
\centering
\begin{minipage}{0.32\linewidth}
\centering
\includegraphics[width=\linewidth]{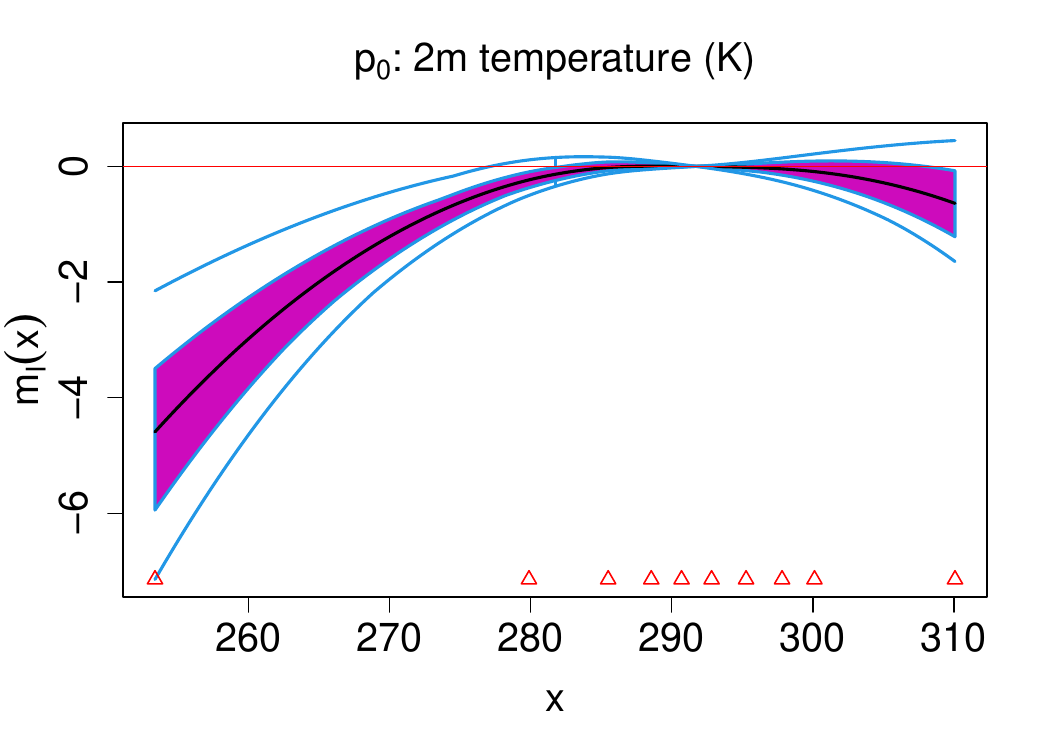} 
\end{minipage}
\begin{minipage}{0.32\linewidth}
\centering
\includegraphics[width=\linewidth]{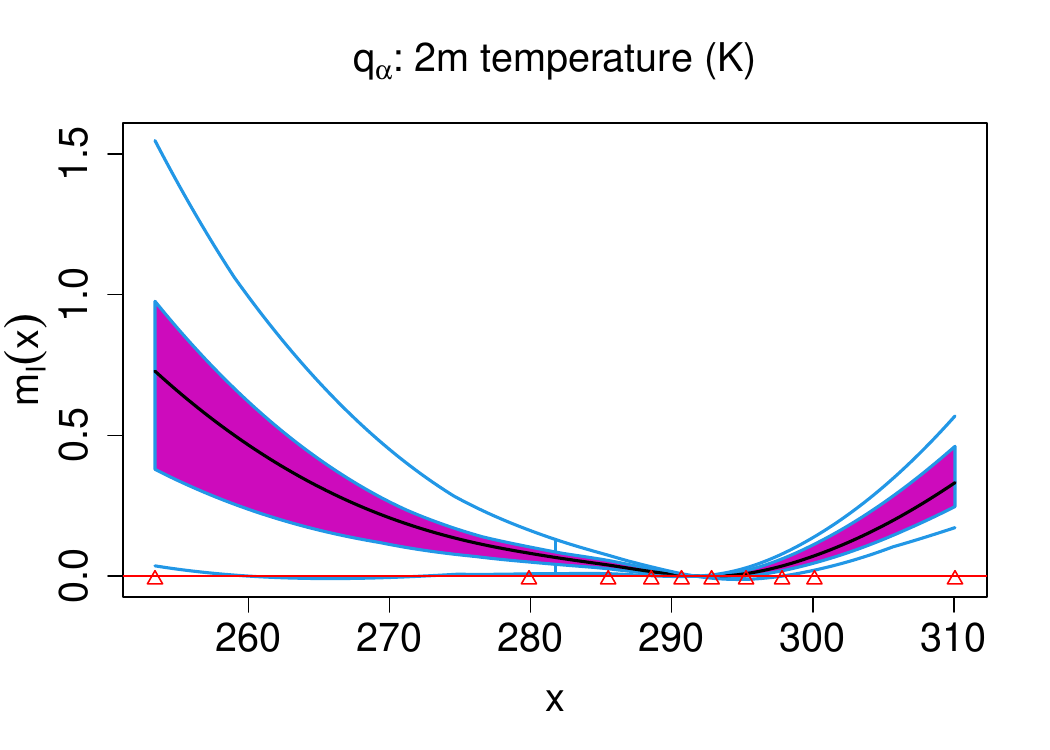} 
\end{minipage}
\begin{minipage}{0.32\linewidth}
\centering
\includegraphics[width=\linewidth]{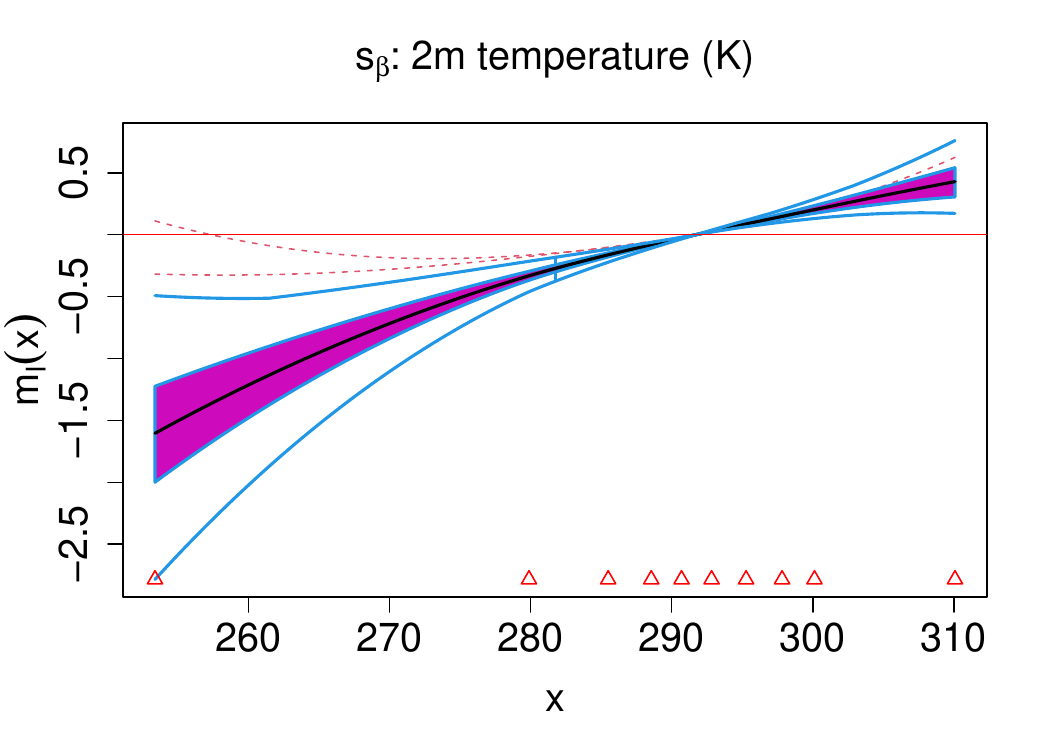} 
\end{minipage}
\begin{minipage}{0.32\linewidth}
\centering
\includegraphics[width=\linewidth]{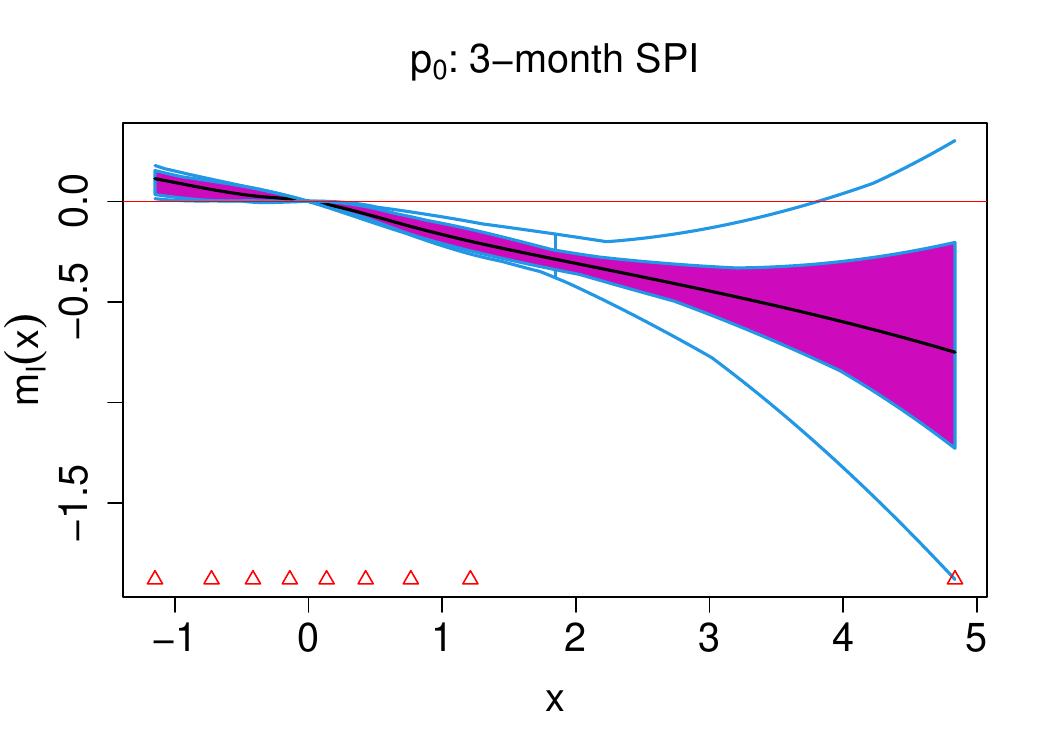} 
\end{minipage}
\begin{minipage}{0.32\linewidth}
\centering
\includegraphics[width=\linewidth]{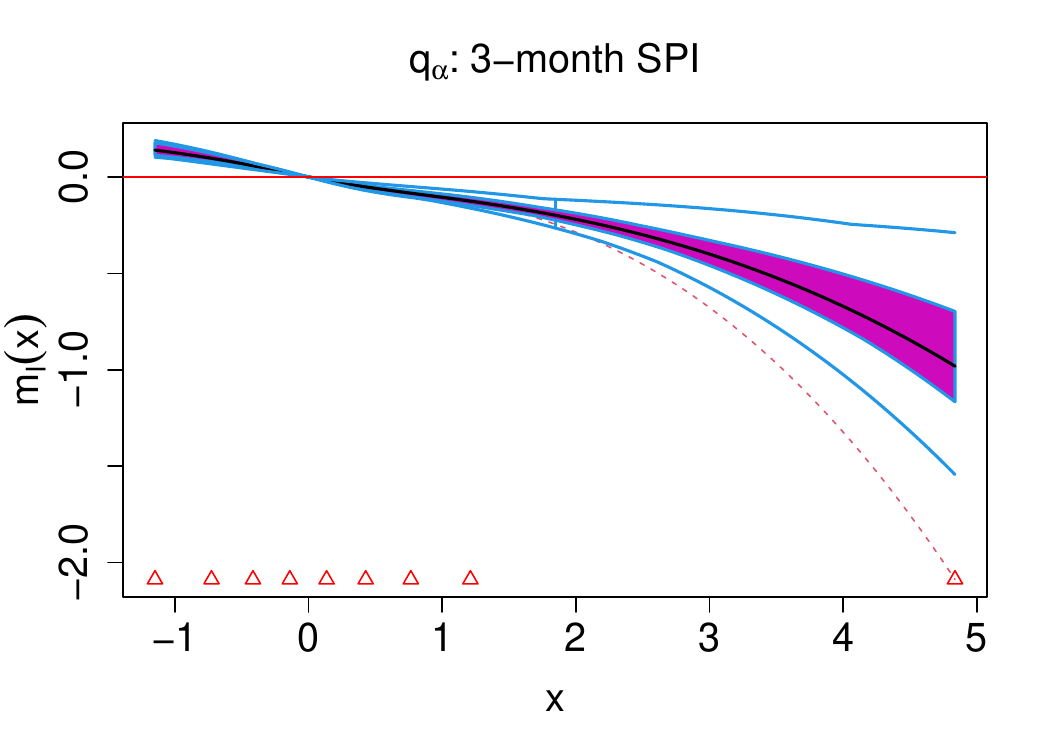} 
\end{minipage}
\begin{minipage}{0.32\linewidth}
\centering
\includegraphics[width=\linewidth]{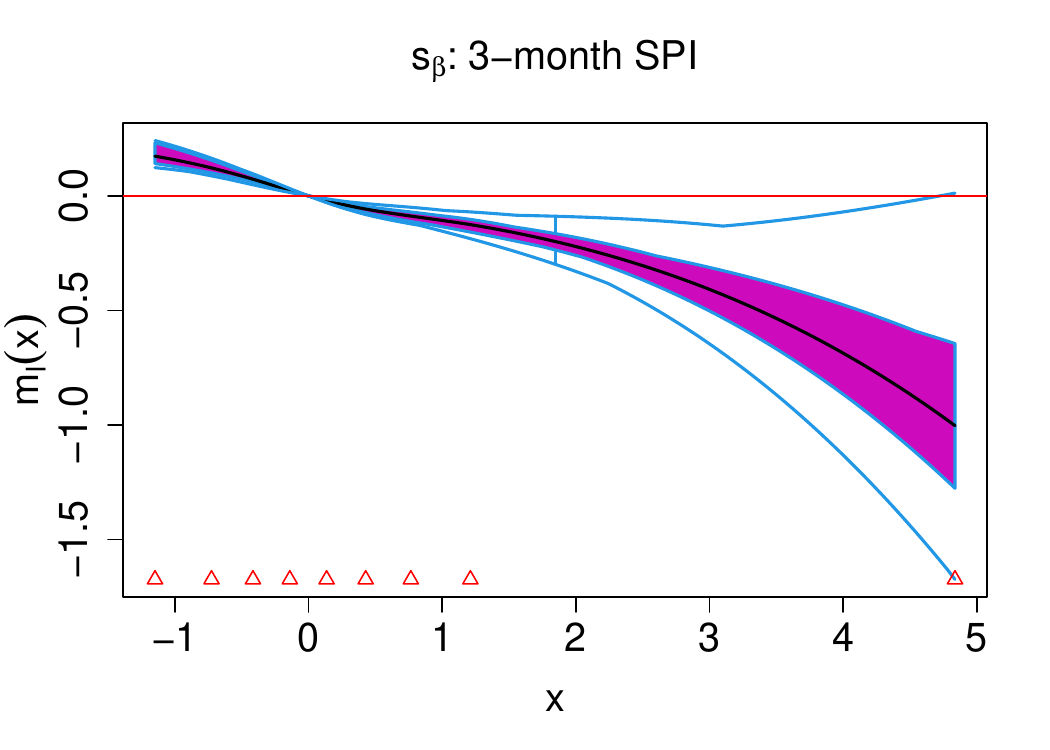} 
\end{minipage}
\caption{Functional boxplots of estimates of the additive function contributions of 2m air temperature (K; top row) and 3-month SPI (unitless; bottom row) to the log-odds of occurrence probability $p_0(s,t)$ (unitless; left column), log-location $\log \{q_\alpha(s,t)\}$ (log-$\sqrt{\mbox{acres}}$; centre column), and log-scale $\log \{s_\beta(s,t)\}$ (log-$\sqrt{\mbox{acres}}$; right column) of the {global} PINN model. Black curves gives the median function over all bootstrap samples, and are enclosed within the 50$\%$ central regions (magenta). Blue curves denote the maximum envelopes, and the red dashed curves represent outlier candidates that fall outside these envelopes. Red horizontal lines denote a zero effect, whilst the red triangles denote the placement of the splines' knots.}
\label{occur_GAM}
\end{figure}

{Here, we present estimates of the interpretable function $m_\mathcal{I}(\cdot)$ for the local and global PINN models. For the global PINN model, we consider the individual contribution of each interpreted predictor to $m_\mathcal{I}$, denoted here $\hat{m}_{\mathcal{I},j}(x_j)$ for $j=1,2$, where $x_1$ and $x_2$ correspond to 2m air temperature and SPI, respectively. We centre estimates of each spline $\hat{m}_{\mathcal{I},j}(x_j)$ for each bootstrap sample, by subtracting evaluations of the spline at the median of the predictor values. We illustrate, in Figure~\ref{occur_GAM}, their estimates and uncertainty using functional box plots \citep{sun2011functional}; note that we determine if a predictor has a ``significant'' effect on a model parameter if the magenta 50\% central region does not encompass the horizontal line (corresponding to zero effect). We observe that increases in air temperature lead to an increase in both the log-odds of wildfire occurrence probability, $p_0$, and the log-scale, $\log(s_\beta)$, of extreme wildfire spread; the latter can be interpreted as a measure of severity for extreme wildfires, given an occurrence. Interestingly, there is no further significant increase in the effect of 2m air temperature on $p_0$ when the temperature exceeds roughly its median value of 290K; the same does not hold true for $s_\beta$. This suggests that, above a certain threshold, increasing temperatures exacerbate only the severity, not the frequency, of U.S. wildfires. Increasing temperatures above the median lead to an increase in the log-location parameter, $\log(q_\alpha)$, of extreme wildfire spread, but the converse holds true for temperatures below the median. This may be due to interactions between low temperature and fuel abundance or type, which have a significant impact on the spread of wildfires. For the drought index, we consistently observe across all parameters ($p_0$, $q_\alpha$, and $s_\beta$) that larger values (corresponding to relatively more monthly rainfall) lead to a reduction in the parameter value. }\par
\begin{figure}[t!]
\centering
\begin{minipage}{0.49\linewidth}
\flushleft
\includegraphics[width=0.8\linewidth]{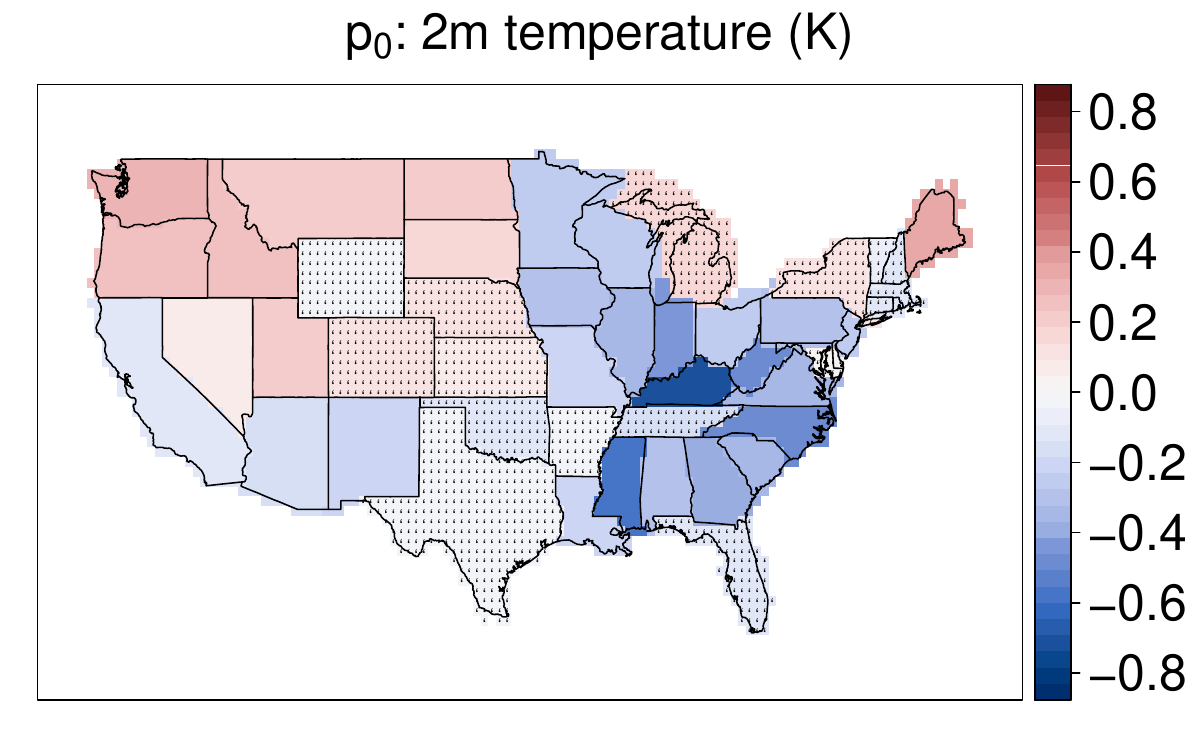} 
\end{minipage}
\vspace{-0.3cm}
\begin{minipage}{0.49\linewidth}
\flushleft
\includegraphics[width=0.8\linewidth]{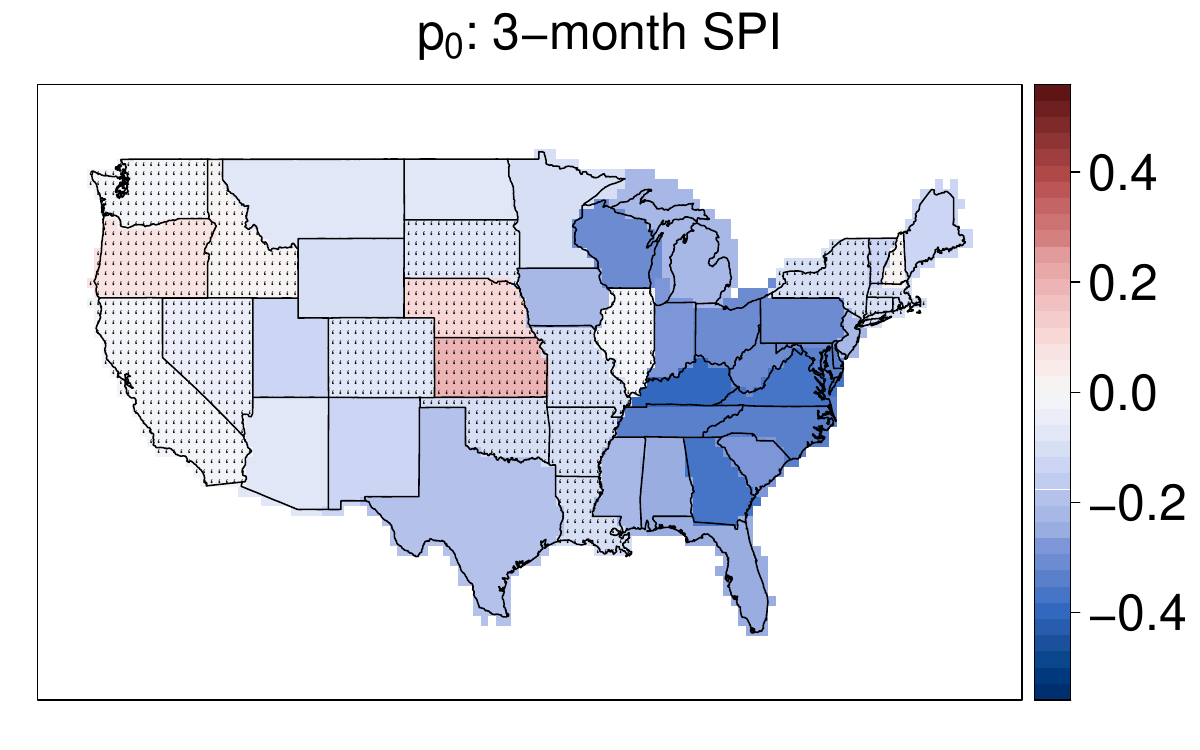} 
\end{minipage}
\begin{minipage}{0.49\linewidth}
\flushleft
\includegraphics[width=0.8\linewidth]{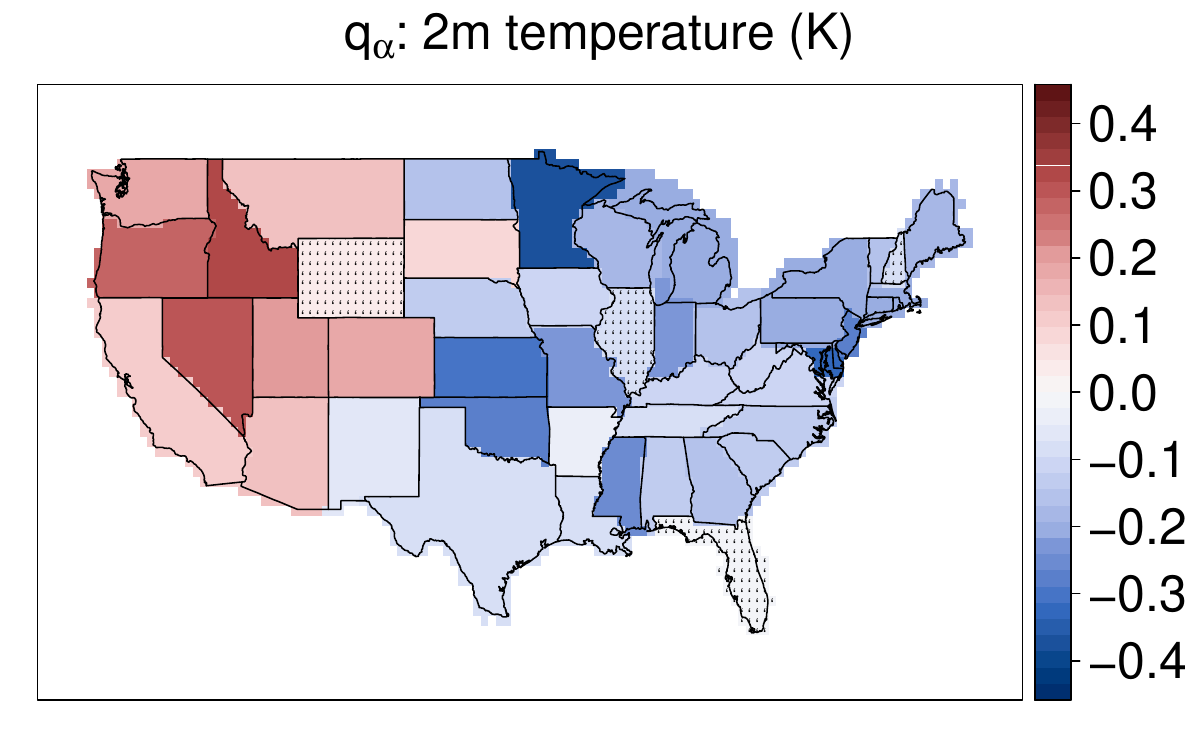} 
\end{minipage}
\vspace{-0.3cm}
\begin{minipage}{0.49\linewidth}
\flushleft
\includegraphics[width=0.8\linewidth]{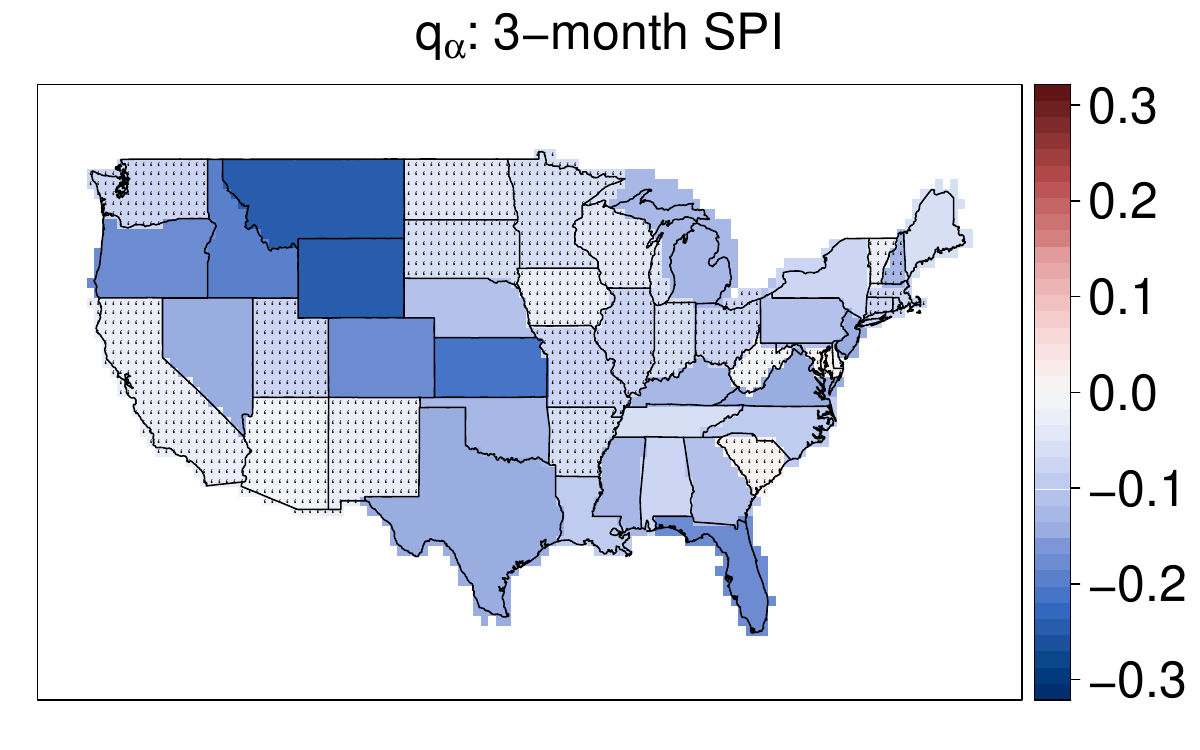} 
\end{minipage}
\begin{minipage}{0.49\linewidth}
\flushleft
\includegraphics[width=0.8\linewidth]{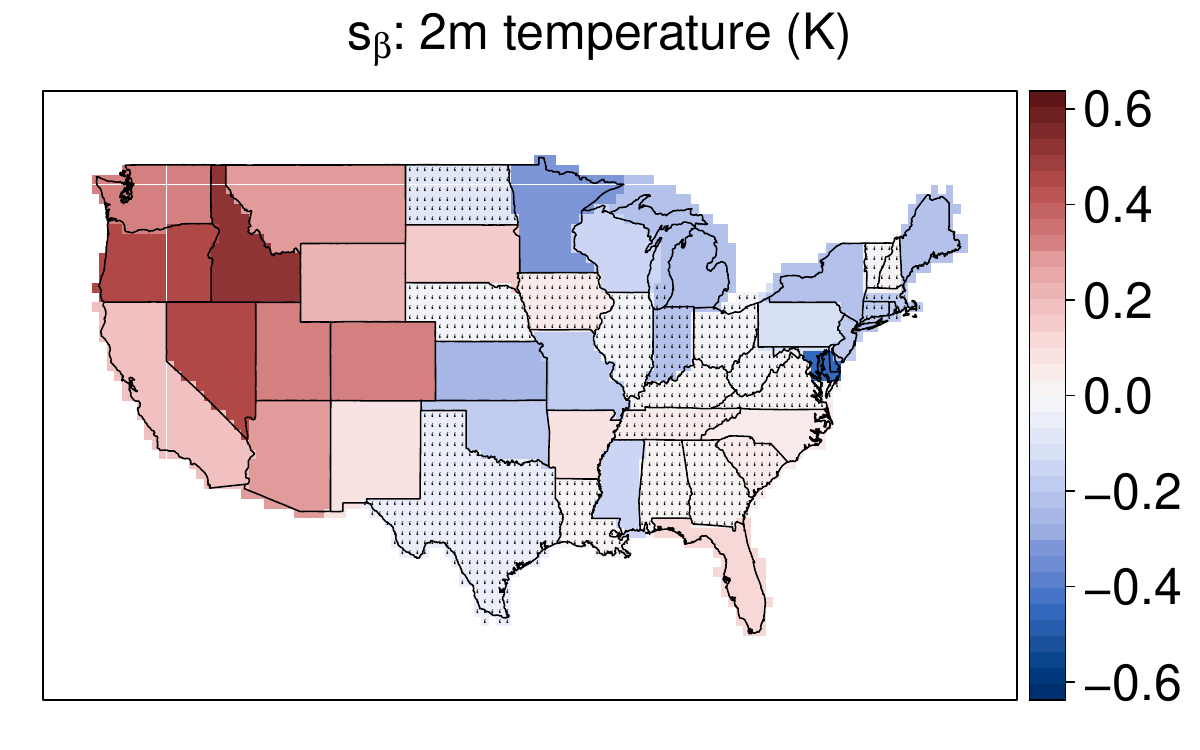} \end{minipage}
\begin{minipage}{0.49\linewidth}
\flushleft
\includegraphics[width=0.8\linewidth]{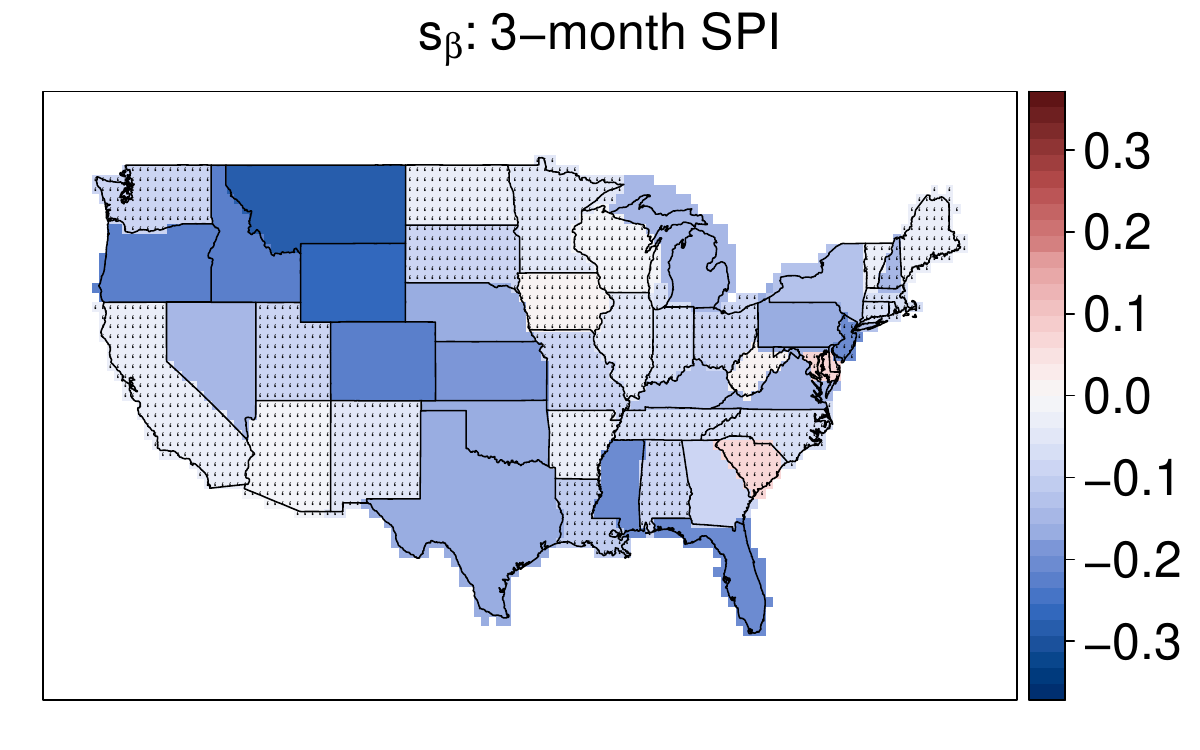} 
\end{minipage}
\vspace{-0.5cm}
\caption{Maps of median estimated $m_\mathcal{I}(\cdot)$ for the local PINN model, where $m_\mathcal{I}(\cdot)$ is a linear model with state-varying coefficients. Mapped values correspond to the contribution of 2m air temperature (K; left column) and 3-month SPI (unitless; right column) to the log-odds of occurrence probability $p_0(s,t)$ (unitless; top row), log-location $\log \{q_\alpha(s,t)\}$ (log-$\sqrt{\mbox{acres}}$; centre row), and log-scale $\log \{s_\beta(s,t)\}$ (log-$\sqrt{\mbox{acres}}$; bottom row). Shaded regions correspond to insignificant estimates, where the $95\%$ bootstrap confidence interval includes zero. }
\label{fig:SVC}
\end{figure}

{Figure~\ref{fig:SVC} presents maps of the median estimates of $m_\mathcal{I}$ for the local PINN model, with uncertainty estimates provided in Figure~\ref{fig:SVC_IQR} \citep{papersup}. We observe spatial variation in the state-varying effects of temperature and SPI on the model parameters. For 2m air temperature, we generally observe (for all three parameters) positive trends in the western U.S., but negative trends in the central and eastern regions. This could suggest that global warming will exacerbate wildfires in the western U.S., but lead to a reduction in extreme wildfire occurrences and severity in eastern regions of the U.S. We note that some states, e.g., Arizona, New Mexico, North and South Carolina, and Virginia, illustrate disagreement between the sign of the trends for the occurrence probability ($p_0$) and severity ($q_\alpha$/$s_
\beta$) components of the model; here the model estimates suggest that increasing temperatures will lead to fewer but more severe wildfires, corresponding to lower $p_0$ and higher $q_\alpha$ or $s_\beta$ values. For the 3-month SPI index, we generally observe negative trends, in all parameters, across the entirety of the contiguous U.S., suggesting that increased drought mostly leads to increased U.S. wildfire frequency and severity. Only one state, Nebraska, shows a positive relationship between SPI and $p_0$, which may be caused by increased fuel production \citep{scasta2016droughts}. Figure~\ref{fig:SVC_sensitivity} \citep{papersup} investigates the sensitivity of estimates for the local bGEV-PP PINN model to the choice of $\tau$; we find that taking $\tau=0.7$ or $\tau=0.9$ does not significantly alter the spatial patterns in estimates of $m_\mathcal{I}$.}

\subsection{Hazard map estimates}
\label{sec:hazard_maps}
\begin{figure*}[t]
\begin{minipage}{0.49\linewidth}
\flushleft
\hspace{-.22cm}
\includegraphics[width=0.8\linewidth]{Images/obs_t103.pdf} 
\end{minipage}
\vspace{-0.3cm}
\begin{minipage}{0.49\linewidth}
\flushleft
\hspace{.005cm}
\includegraphics[width=0.8\linewidth]{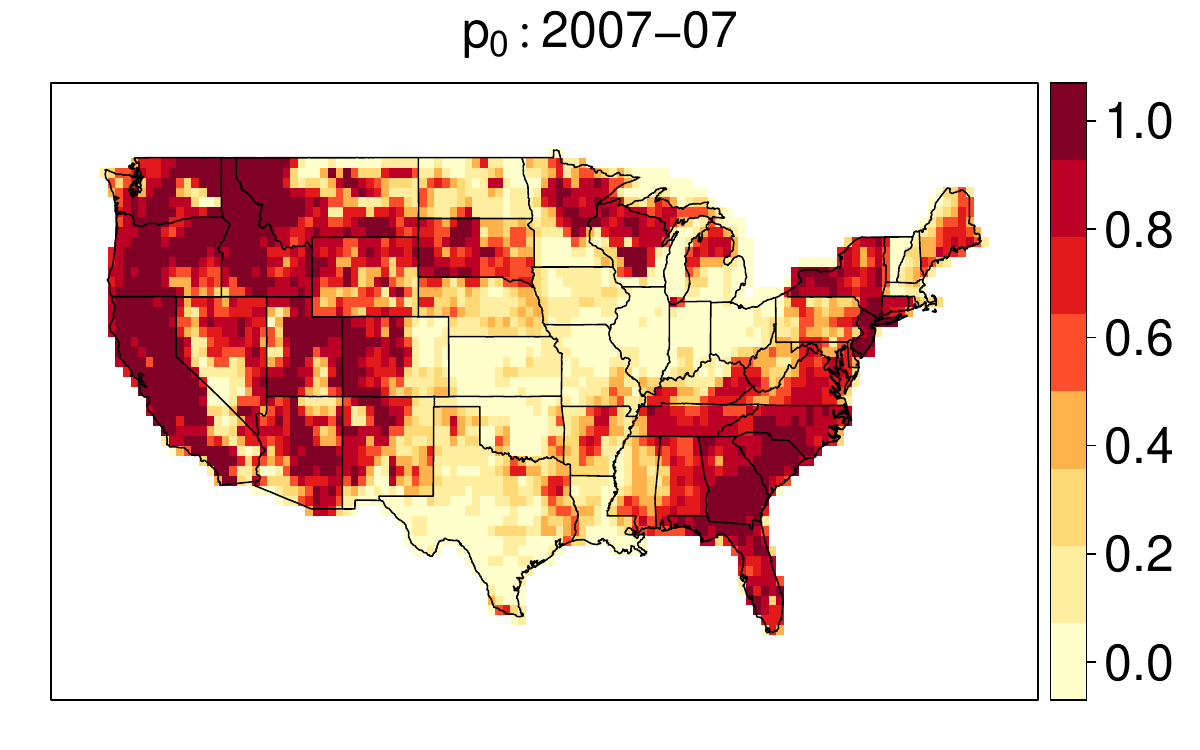} 
\end{minipage}
\begin{minipage}{0.49\linewidth}
\flushleft
\includegraphics[width=0.8\linewidth]{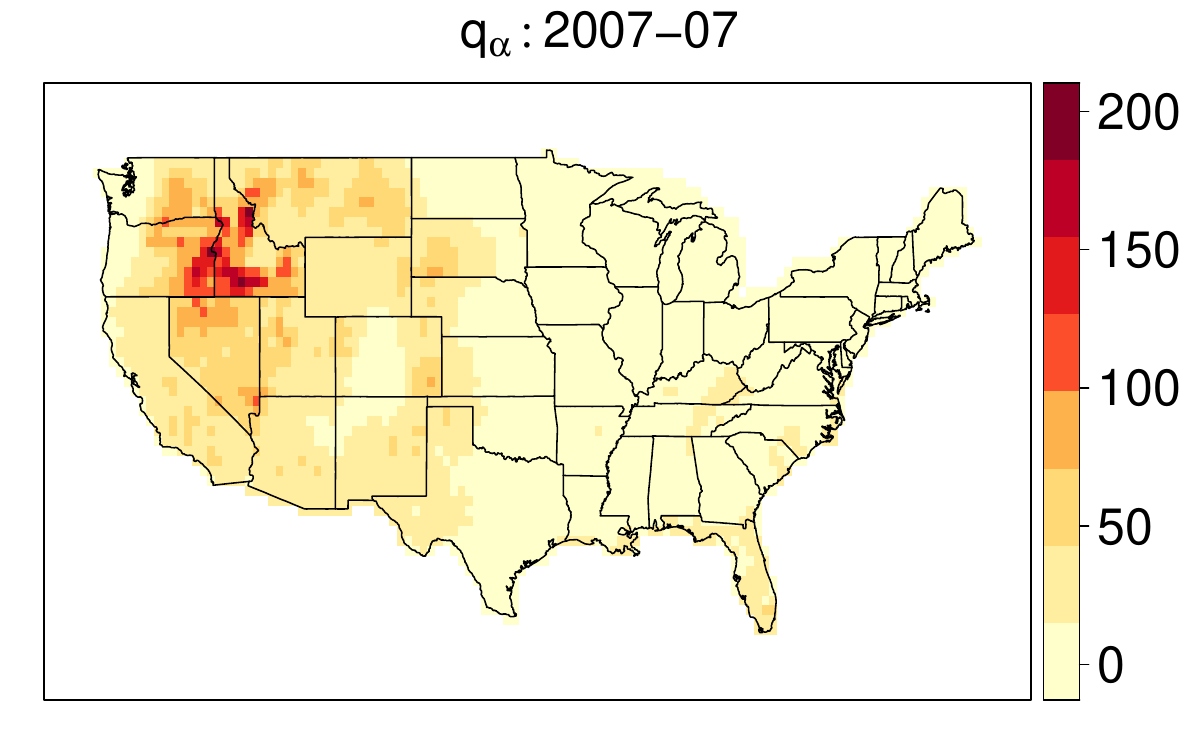} 
\end{minipage}
\vspace{-0.3cm}
\begin{minipage}{0.49\linewidth}
\flushleft
\hspace{-.2cm}
\includegraphics[width=0.8\linewidth]{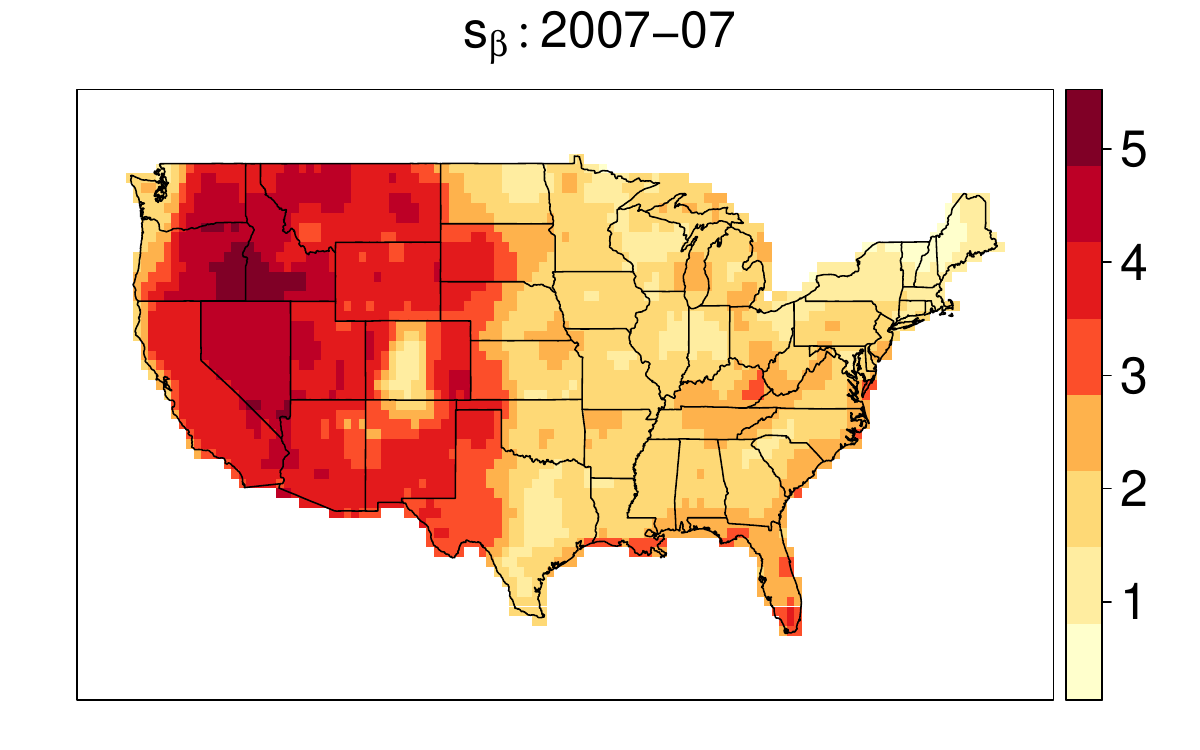} 
\end{minipage}
\vspace{-0.3cm}
\begin{minipage}{0.49\linewidth}
\flushleft
\includegraphics[width=0.8\linewidth]{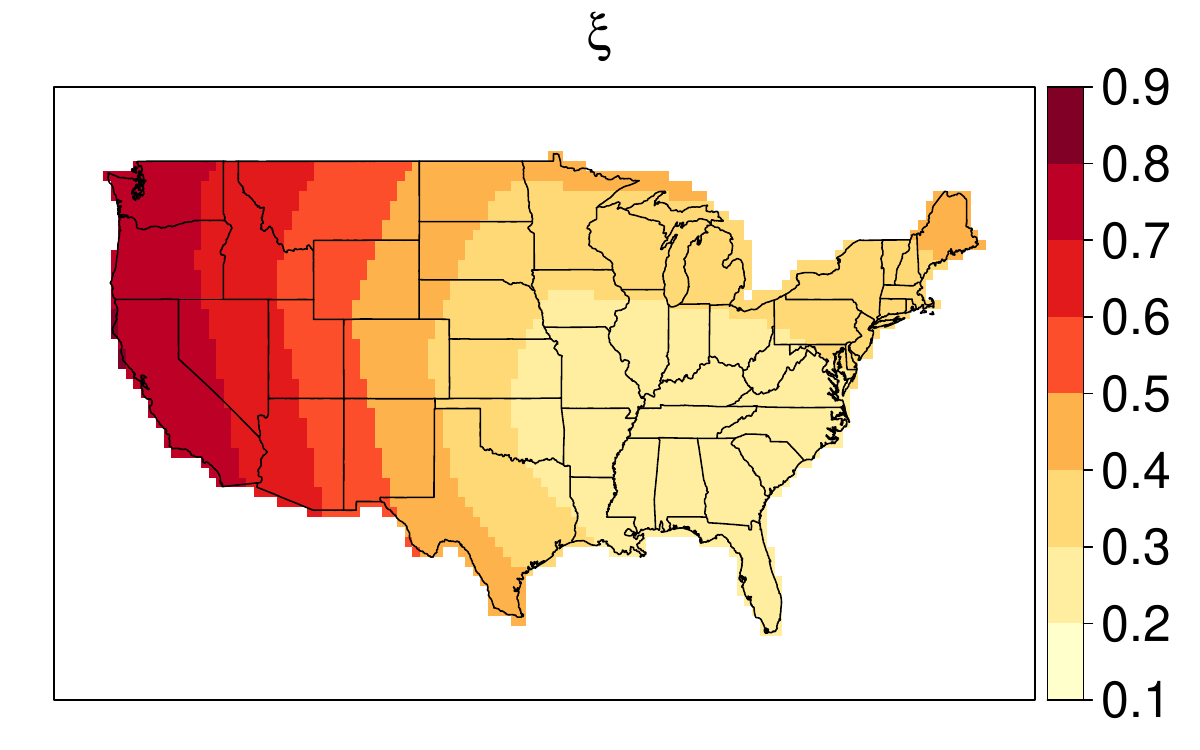} 
\end{minipage}
\begin{minipage}{0.49\linewidth}
\flushleft
\hspace{.04cm}
\includegraphics[width=0.8\linewidth]{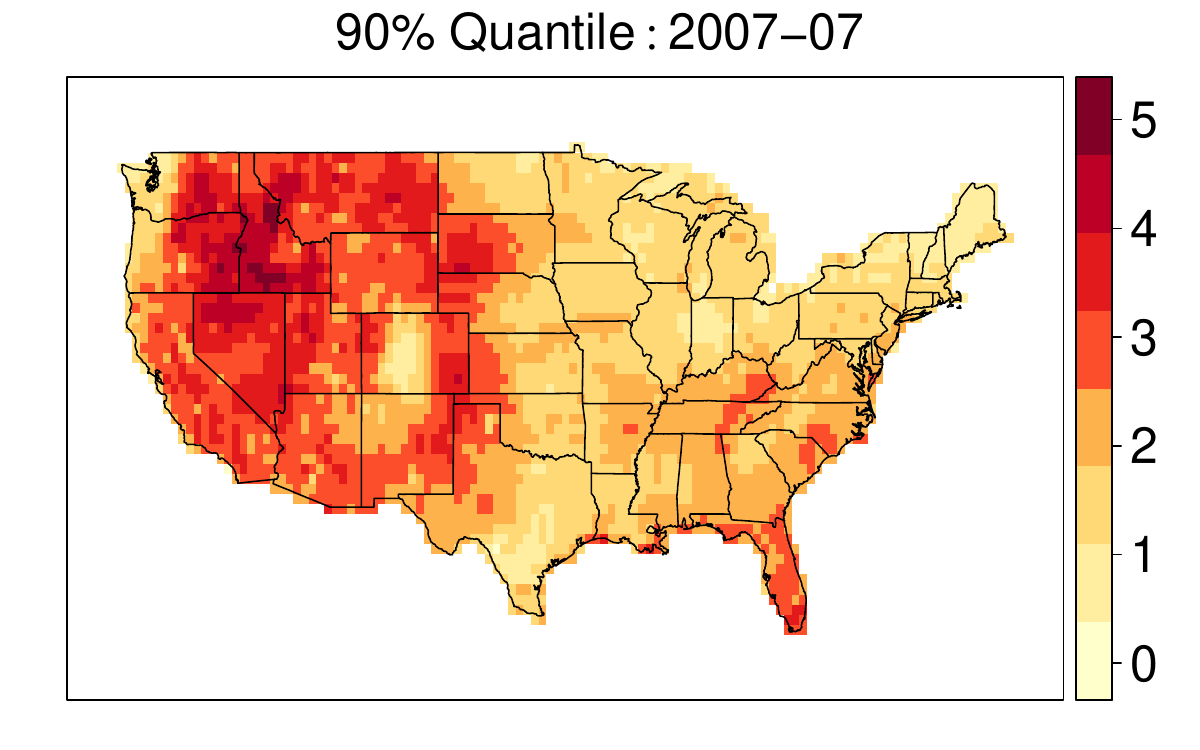} 
\end{minipage}
\vspace{-0.2cm}
\caption{Maps of observed $\log\{1+\sqrt{Y}(s,t)\}$ (log-$\sqrt{\mbox{acres}}$; top-left) and median estimated $p_0(s,t):=\Pr\{Y(s,t)>0\mid\mathbf{X}(s,t)\}$ (unitless; top-right), $q_\alpha(s,t)$ ($\sqrt{\mbox{acres}}$; centre-left), $\log \{1+s_\beta(s,t)\}$ (log-$\sqrt{\mbox{acres}}$; centre-right), $\xi(s)$ (unitless; bottom-left), and $90\%$ quantile of $\log\{1+ \sqrt{Y}(s,t)\}\mid\{Y(s,t)>0,\mathbf{X}(s,t)\}$ (log-$\sqrt{\mbox{acres}}$; bottom-right) for July, 2007.}
\label{fig:maps}
\end{figure*}

{Figure~\ref{fig:maps} provides maps of the median estimated $p_0(s,t)$, $q_\alpha(s,t)$, and $s_\beta(s,t)$, for a fixed time $t$; for brevity, we consider here only the month July, 2007, as this month was the most devastating in terms of burnt area across the entire U.S. Uncertainty estimates for Figure~\ref{fig:maps} are provided in Figure~\ref{spread_map_IQR}, and similar maps for other selected months are provided in Figures~\ref{occur_map_sup} and \ref{spread_map_sup} of the Supplementary Material \citep{papersup}.
The maps of predicted probabilities match well with the observed wildfires (see Figure~\ref{fig:maps}), suggesting that this component of the local PINN model fits well to the data. We observe relatively large values of $p_0(s,t)$ across the entirety of the spatial domain, but particularly in the north-east, south-east, and western regions. We generally observe that the magnitude and variability of wildfire spread are intrinsically linked; areas of large $q_\alpha$ typically occur with large $s_\beta$. For July 2007, large values of $q_\alpha$ tend to occur in the western areas of the U.S., particular Idaho and Oregon, whilst large values of $s_\beta$ occur over a larger region that stretches into the centre of the domain.}

{Alongside estimates of temporally-varying parameters, Figure~\ref{fig:maps} provides a map of the median estimated shape parameter, $\xi(s)$. Values vary between 0.2 and 0.8, suggesting that U.S. wildfire spread is particularly heavy-tailed; an observation also made by \cite{d2021flexible} and \cite{koh2021gradient} for these data. Note that this is an estimate of the tail index for the square-root of the burnt area, i.e., the burnt area ``diameter", and so the area itself, $Y(s,t)$, will have even heavier tails and (potentially) non-finite first and second-order moments. We observe interesting spatial variation in $\xi(s)$, with significantly larger values observed in the western U.S. Wildfire spread is particularly heavy-tailed in these regions, with California experiencing the most extreme events.}

{The bGEV-PP model is composed of a location, scale, and shape parameter; we can combine these into a single metric to access the overall hazard of wildfire spread, conditional on an occurrence. Figure~\ref{fig:maps} provides estimates of the (log-transformed) $90\%$ quantile\footnote{As the bGEV-PP distribution has real support, but the response takes only positive values, we set negative quantile estimates to zero. {Across all bootstrap samples and spatiotemporal locations $(s,t)$, we found only 1.6$\%$ of estimates of the $90\%$ quantile of $ \sqrt{Y}(s,t)\mid\{Y(s,t)>0,\mathbf{X}(s,t)\}$ are negative.}} of $\sqrt{Y}(s,t)\mid\{Y(s,t)>0,\mathbf{X}(s,t)\}$, for July 2007. Although we do not observe large values of $q_\alpha$ in the east, we still observe high estimates of the $90\%$ quantile there, due to the effect of $s_\beta$ and $\xi$. Note that the presented quantiles are for strictly positive spread only, and so do not necessarily characterise the overall or compound risk of wildfire burnt areas. Instead, these maps can be used to identify areas where it is pertinent to prevent wildfires; here wildfires are unexpected, but, should they occur, they are particularly devastating. For example, there are regions where there is a low probability of wildfire occurrence $p_0$, whilst high estimates of the (conditional) quantile in these regions illustrate that the spread distribution there is particularly extreme, e.g., parts of California, Nevada, and northern Texas.}
\par
\begin{figure*}[t!]
\centering
\begin{minipage}{0.32\linewidth}
\centering
\includegraphics[width=\linewidth]{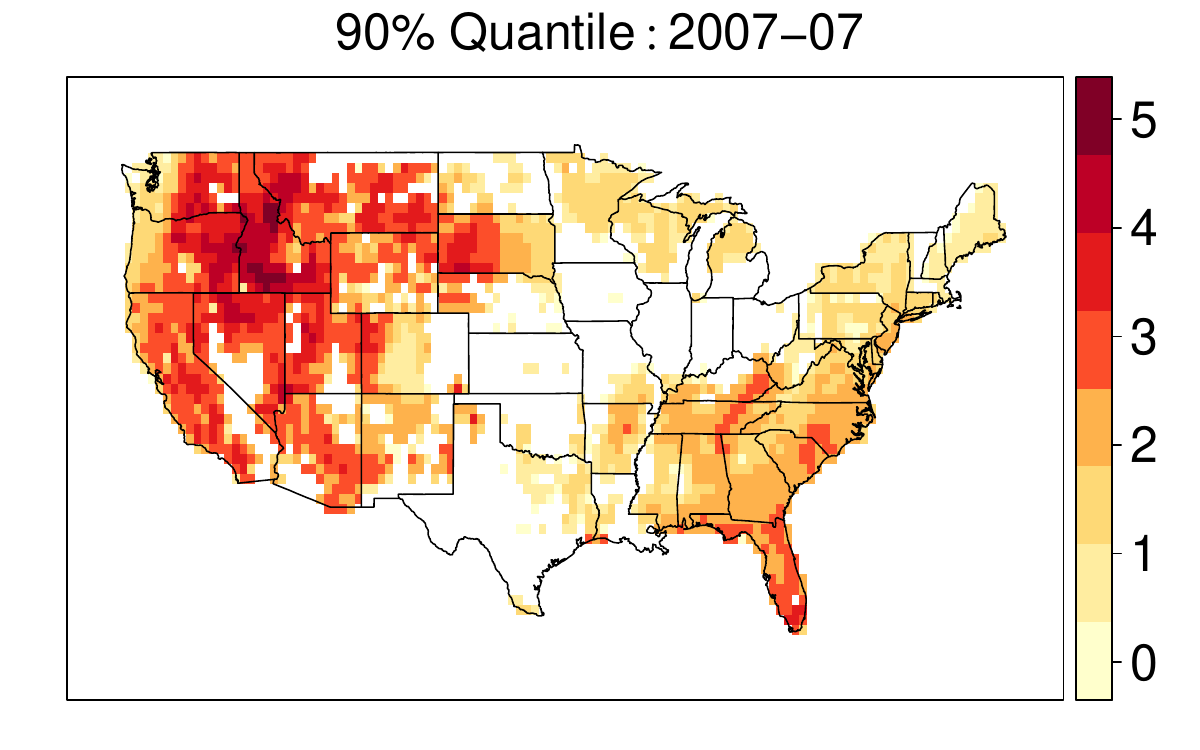} 
\end{minipage}
\begin{minipage}{0.32\linewidth}
\centering
\includegraphics[width=\linewidth]{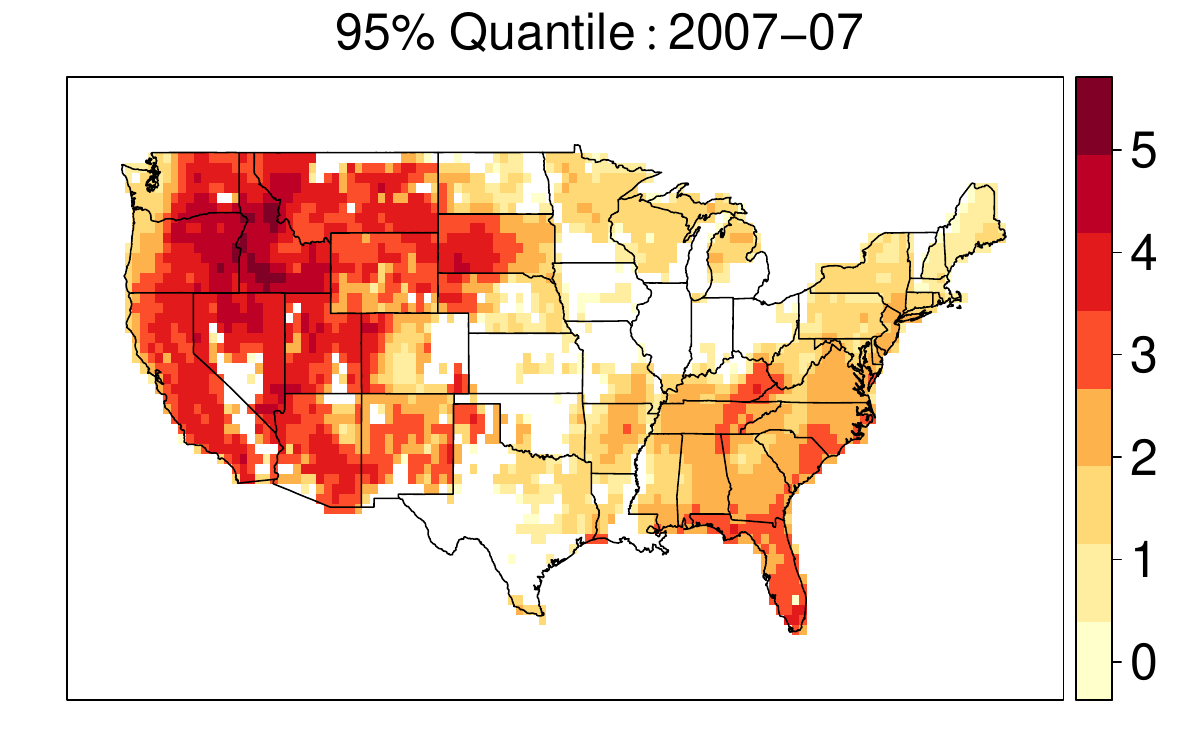} 
\end{minipage}
\begin{minipage}{0.32\linewidth}
\centering
\includegraphics[width=\linewidth]{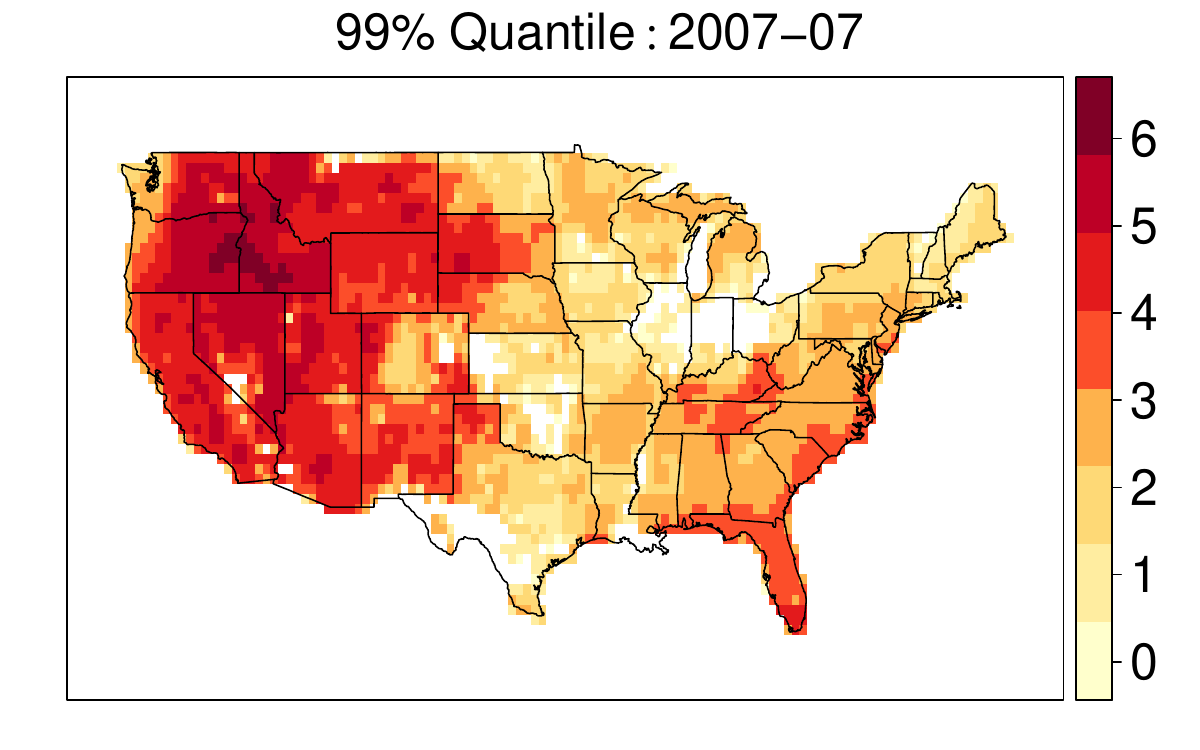} 
\end{minipage}
\vspace{-0.5cm}
\caption{Maps of median estimated quantiles for $\log\{1+\sqrt{Y}(s,t)\}\mid\mathbf{X}(s,t)$ (log-$\sqrt{\mbox{acres}}$) for July 2007. Quantiles, left to right: $90\%$, $95\%$, and $99\%$.}
\label{fig:cr_map}
\end{figure*}
{To assess the overall risk of U.S. wildfires, Figure~\ref{fig:cr_map} provides maps of estimated extreme quantiles for $\log\{1+\sqrt{Y}(s,t)\}\mid\mathbf{X}(s,t)$, i.e., the unconditional response, in July 2007 \citep[Figure~\ref{cr_map_sup},][shows similar maps for other months]{papersup}. This characterises the overall risk of burnt area, for this particular month, by combining estimates of $p_0$ with estimates of the bGEV-PP local PINN model. We note that areas of large values in Figure~\ref{fig:cr_map} do not necessarily correspond to areas of large values in estimates of $p_0$, $q_\alpha$, or $s_\beta$ (Figure~\ref{fig:maps}), and so we conclude that the spread and occurrence of wildfires are not always intrinsically linked; different factors lead to wildfire ignition as well as spread, and so our decision to model both processes separately is well-founded. We observe high values of the unconditional quantiles across large areas of the U.S., typically in the west of the U.S., including California, and the south-east, including Florida, with states in the north-east, and Texas, having relatively low compound risk. }
\subsection{Temporal trends}
\label{sec:temp_trends}
\begin{figure}[t!]
\flushleft
\begin{minipage}{0.495\textwidth}
\includegraphics[width=0.9\linewidth]{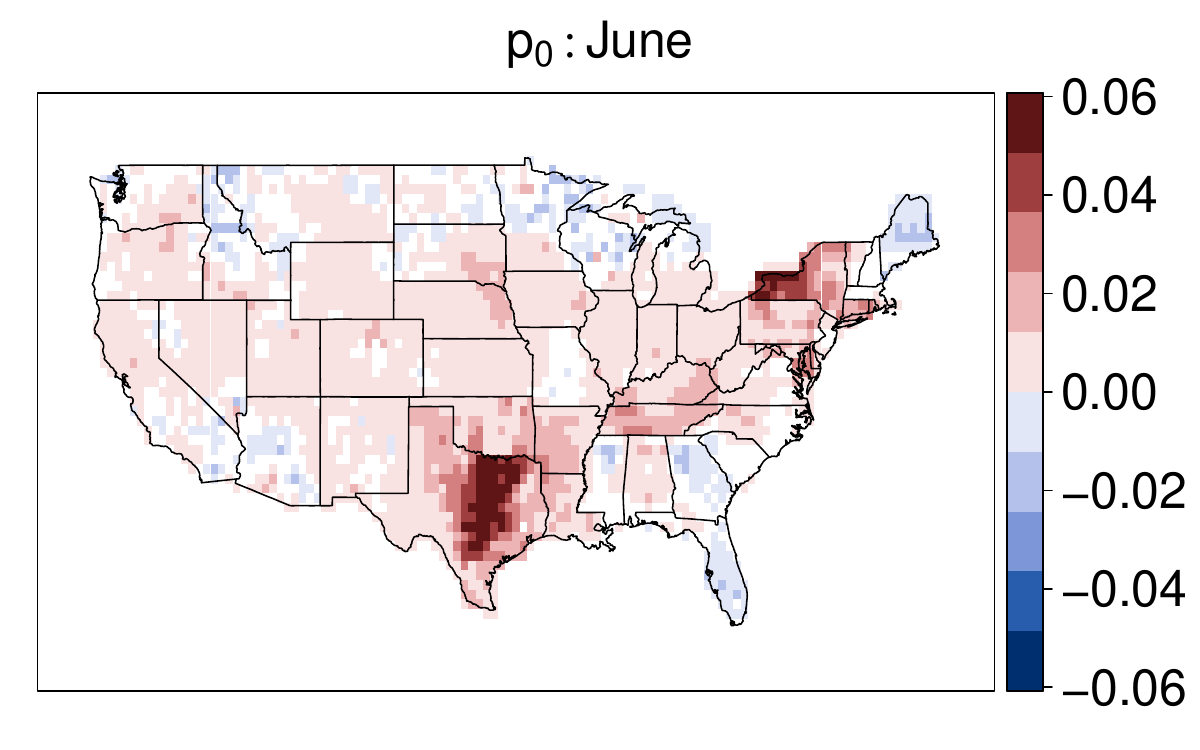} 
\end{minipage}
\begin{minipage}{0.495\textwidth}
\flushleft
\includegraphics[width=0.9\linewidth]{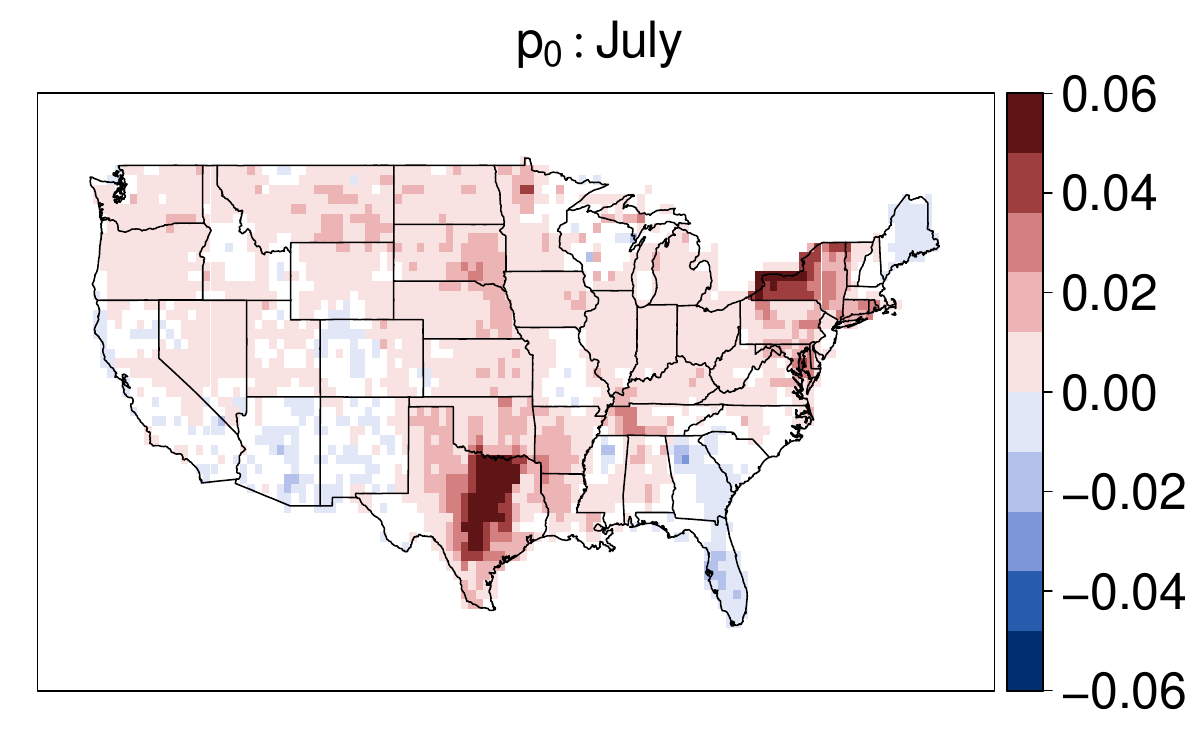} 
\end{minipage}
\vspace{-0.3cm}
\begin{minipage}{0.495\linewidth}
\flushleft
\includegraphics[width=0.9\linewidth]{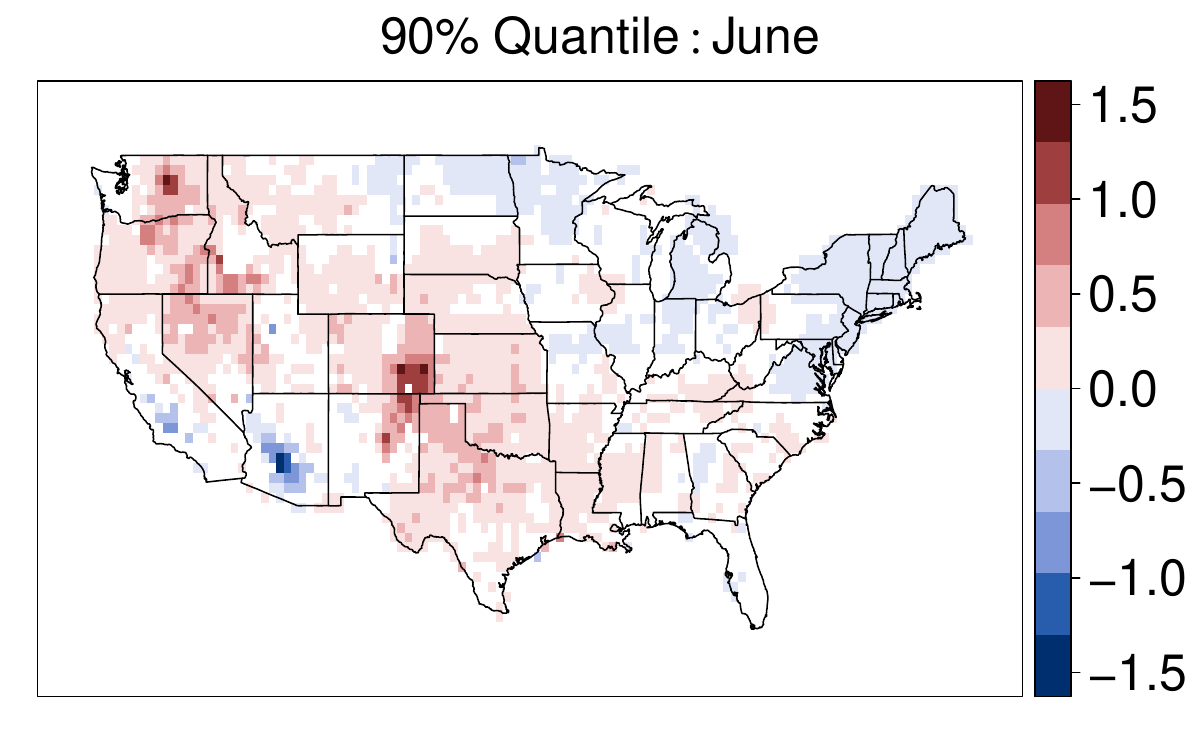} 
\end{minipage}
\begin{minipage}{0.495\linewidth}
\flushleft
\hspace{-0.18cm}
\includegraphics[width=0.9\linewidth]{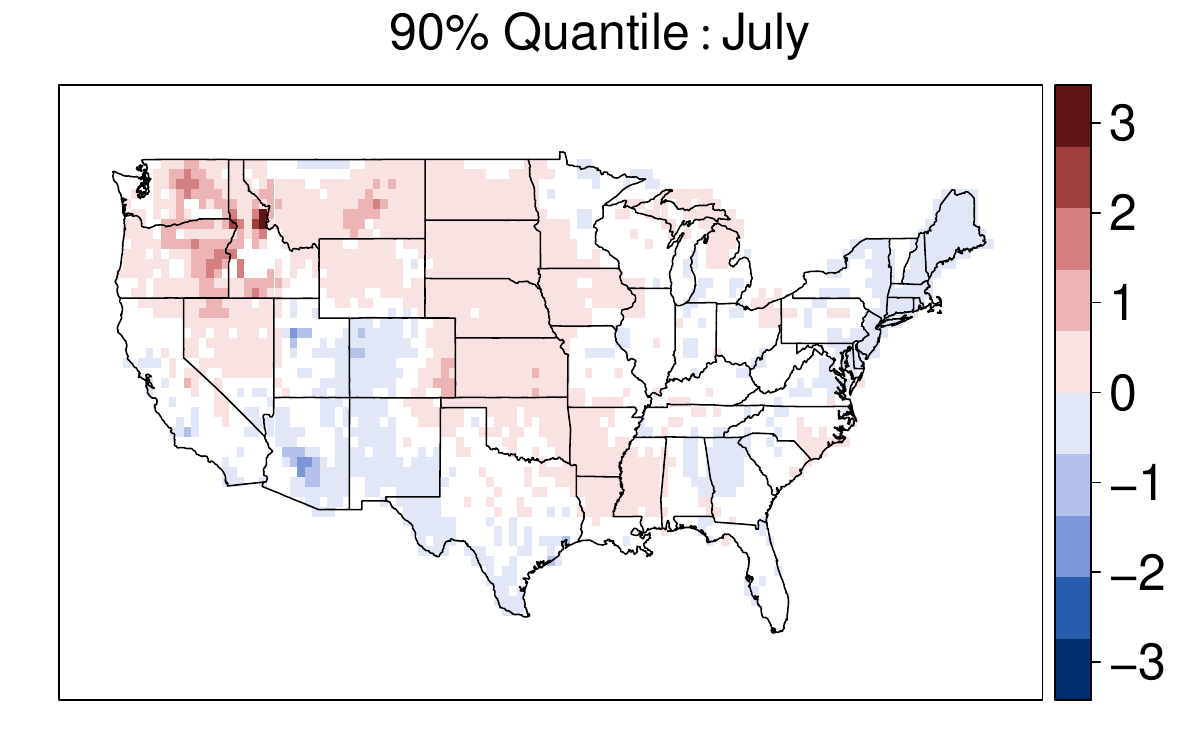} 
\end{minipage}
\vspace{-0.3cm}
\caption{Maps of median site-wise trends in estimated occurrence probability $p_0(s,t)$ (unitless; top) and $90\%$ quantile of $\sqrt{Y}(s,t)\mid\{Y(s,t)>0,\mathbf{X}(s,t)\}$ ($\sqrt{\mbox{acres}}$; bottom) across all bootstrap samples, stratified by month. Left; June. Right; July.}
\label{fig:temp_trends}
\end{figure}

To quantify temporal changes in the wildfire distribution, {we present maps of the site-wise trends in estimates of $p_0(s,t)$, stratified by month. Trends are estimated at each spatial location $s$ by fitting a linear regression to estimates of ${\{p_0(s,t):t \in \mathcal{T}^*\}}$, where $\mathcal{T}^*$ is the set of times corresponding to a specific month. A similar procedure is conducted to find temporal trends in the 90$\%$ quantile of $\sqrt{Y}(s,t)\mid\{Y(s,t)>0,\mathbf{X}(s,t)\}$, which we use here as a measure of wildfire severity. Results for two months, June and July, are presented in Figure~\ref{fig:temp_trends}, with others in Figures~\ref{occur_diff_map_sup} and \ref{spread_diff_map_sm} \citep{papersup}.
We generally observe increases in $p_0(s,t)$ during 1993--2015 for both June and July, and for the vast majority of locations, suggesting that climate change may be leading to an increase in the frequency of wildfires. Areas of particular concern include the north-eastern regions of the U.S. and Texas, which show large increases in $p_0(s,t)$. Interestingly, some parts of the U.S. show a decrease in $p_0(s,t)$, e.g., Georgia, Florida, and Maine, albeit very slightly. For the $90\%$ quantile of $\sqrt{Y}(s,t)\mid\{Y(s,t)>0,\mathbf{X}(s,t)\}$, we again observe positive trends across large parts of the U.S., particularly in the north-west and central regions. Negative trends can be observed in states closer to the boundaries of the contiguous U.S.,  particularly in the south-west and north-east. Again, we observe different spatial patterns in the behaviour of $p_0$ and the wildfire spread distribution, which provides further evidence to support our choice to model these two processes separately. }

\section{Discussion}
\label{discuss_sec}
We have proposed a framework for performing semi-parametric extreme quantile regression using partially-interpretable neural networks. The methodology unifies the tail robustness and theoretical strength of parametric extreme-value models, the high predictive accuracy and computational efficiency of neural networks, and the interpretability of linear and spline-based regression models, allowing us to accurately estimate extreme quantiles whilst simultaneously performing statistical inference. To complement this flexible methodological framework, we propose a new point process model for extreme values that circumvents issues concerning finite lower-bounds on the process intensity function, which is a key issue when training neural networks. We use a large ensemble of neural networks to model the compound risk of extreme wildfires in the contiguous U.S., with separate models proposed for the occurrence and spread of extreme wildfires. {Our approach allows for the identification of high-risk areas of the U.S. and spatiotemporal trends in the wildfire distribution, as well as facilitating inference that reveals interesting insight into the spatial variation of the driving behaviour of extreme wildfires.} \par
{Our extreme quantile regression model is computationally-scalable and capable of extracting information from a complex, high-dimensional predictor set ($d>40$). This comprises mostly environmental variables, and so our model omits anthropogenic factors that may also drive the behaviour of  U.S. wildfires. Wildfire politics, i.e., suppression and mitigation policies, vary across the U.S., and they are likely to directly impact the occurrence and extent of wildfires \citep[see, e.g., ][]{kupfer2020climate, barnett2016using, dunn2020wildfire}. Quantifying the efficacy of wildfire policies and incorporating this information into our model is a future endeavour that would be interesting to pursue.   }\par 
Section~\ref{sec:model_compare} highlights the predictive power of convolutional networks, which are capable of capturing spatial patterns in predictors, when they are applied to our data. Further extensions of our model could leverage recurrent neural networks that capture temporal structure in data. For example,  long short-term memory neural networks \citep{hochreiter1997long} have been successfully used for prediction in a number of environmental applications, e.g., North Atlantic Oscillation index \citep{yuan2019prediction}, influenza trends \citep{liu2018lstm}, sea surface temperatures \citep{liu2018td} and basin water levels \citep{shuofeng2021lstm}, and these can be combined with CNNs to 
build neural networks that model spatiotemporal structure. A drawback of CNNs is that they are only applicable to gridded spatial data, which have been constructed as a sequence of images. Hence, future studies may benefit from using alternatives to the CNN. For example, \cite{scarselli2008graph} propose graphical neural networks (GNNs), which are applicable to data with a graphical structure. Recent developments \citep[see review by][]{zhou2020graph} of GNNs have seen the proposal of convolutional \citep{kipf2016semi} and recurrent \cite[e.g.,][]{li2015gated} extensions that can handle data which are irregularly-spaced throughout $\mathcal{S}$ and $\mathcal{T}$, respectively; hence they could be used in our framework to model the extremes of processes observed at point locations or on irregularly-spaced grids.

\par
We have here considered one type of distribution for modelling extremes of the response, namely the newly proposed bGEV-PP model. However, our framework extends far beyond the use of this specific model, which can be easily replaced by any parametric model. For example, it is trivial to see how the bGEV point process could be replaced with the standard bGEV\footnote{For the reasons outlined in Sections \ref{bgev_PP_sec} and \ref{bGEV_sec}, we would always favour the bGEV over the GEV when using neural networks.} distribution; we would model block-maxima of the response and appropriately aggregate the predictors. As is common with many extreme value approaches, fitting of the bGEV-PP and GPD models requires a prerequisite step whereby a sufficiently high threshold $u(s,t)$ is estimated; $u(s,t)$ is typically estimated to ensure a constant exceedance probability $\Pr\{Y(s,t)>u(s,t)\}$ across all $(s,t)\in\mathcal{S}\times\mathcal{T}$. When modelling the response below $u(s,t)$, the bGEV-PP model performs poorly; alternatives that can be used to model both the upper-tails of the response, as well as the bulk, are mixture models with GPD tails, see, e.g., \cite{carreau2007hybrid} and \cite{carreau2011stochastic}, but these still require specification of $u(s,t)$. Estimating $u(s,t)$ requires additional modelling choices and can contribute to model uncertainty; \cite{papastathopoulos2013extended}, \cite{naveau2016modeling}, and \cite{yadav2021spatial} propose extensions of the GPD which model the whole distribution and removes this requirement, and so may provide better candidates for modelling extremes of burnt area than both the GPD and bGEV-PP. \par
{{Our modelling framework does not account for dependence between the burnt area response $Y(s, t)$ across different space-time locations, and so we are unable to account for occurrences of wildfires that persist between spatial grid-cells or subsequent times. Such a limitation of our regression modelling approach would require fitting of a complex dependence model for both the bulk and the tail, which accounts for the significant non-stationarity exhibited by a process observed over a very large or complex topographical domain, e.g., the contiguous U.S.; this would not only be challenging to model, but also extremely computationally demanding to fit, and so is beyond the scope of our work.}}\par
Finally, to make our unifying partially-interpretable neural network (PINN)-based framework for extreme quantile regression available to the whole statistical community at large, and for reproducibility, we have created the $\texttt{R}$ package $\texttt{pinnEV}$ \citep{pinnEV}. {All models considered in our application, as well as the GPD and bGEV extensions detailed above, are currently available to use in this package.}


\section*{Acknowledgments}
The authors would like to thank Thomas Opitz for providing the data and Michael O'Malley for supportive discussions. 
The research reported in this publication was supported by funding from King Abdullah University of Science and Technology (KAUST) Office of Sponsored Research (OSR) under Award No. OSR-CRG2020-4394. Support from the KAUST Supercomputing Laboratory is gratefully acknowledged. The data that supports our findings are available in the accompanying $\texttt{R}$ package, $\texttt{pinnEV}$, which is available at \url{https://github.com/Jbrich95/pinnEV}.


\baselineskip=14pt
\begingroup
\setstretch{0.75}

\bibliographystyle{apalike}
{\small
\bibliography{ref}
}
\endgroup
\baselineskip 10pt
\renewcommand{\theequation}{S.\arabic{equation}}
\renewcommand{\thefigure}{S\arabic{figure}}
\renewcommand{\thetable}{S\arabic{table}}
\renewcommand{\thesection}{S\arabic{section}}
\setcounter{figure}{0}
\setcounter{table}{0}
\setcounter{equation}{0}
\pagebreak
\begin{appendix}
\section*{Appendices}
\section{Modelling covariate-dependent parameters}
\label{modelling_sec}
\subsection{Connection to main text}
This appendix supports the material in Section~\ref{intro-towards-sec} of the main paper. Appendix~\ref{GAM_sec} and Appendix~\ref{NN_sec} detail the generalised additive framework and neural networks used in model \eqref{full_model} of the main text.
\subsection{Generalised additive models}
\label{GAM_sec}
Following \cite{wood2006generalized}, a generalised additive model (GAM) allows parameters to be represented through a basis of splines, creating a smooth function of predictors; we apply this approach to model the function $m_\mathcal{I}$ in \eqref{full_model}. Given a set of $I$ predictors $x_1(s,t),\dots,x_I(s,t)$, we can define $k_i\in\mathbb{N}$ knots for each predictor by $\{x^{*i}_1,\dots,x^{*i}_{k_i}\}$ for $i=1,\dots, I$. We then let
\begin{equation}
m_{\mathcal{I}}(s,t)=\sum^I_{i=1}\sum^{k_i}_{j=1}w_{ij}\psi_{ij}(x_i(s,t),x_j^{*i}),
\label{GAMeq}
\end{equation}
where $w_{ij} \in \mathbb{R}$ for all $i=1,\dots,I,j=1,\dots,k_i$ and where $\psi_{ij}$ are arbitrary smoothing basis functions. We constrain our focus to the simple case where $\psi_{ij}(x,y)=\psi(x,y)=\|x-y\|^2\log (\| x-y\|)$ for all $i,j$ and where $\|\cdot\|$ denotes the Euclidean norm, i.e., each $\psi$ is a radial basis function and $m_{\mathcal{I}}$ is a thin-plate spline \citep{wood2003thin}, and fix $k_i = k$ for all $i=1,\dots,I$.\par
If required, a smoothing penalty can be accommodated into the loss, or negative log-likelihood, function to ensure that the splines are smooth enough \citep{silverman1985some}. For smoothness parameters $\boldsymbol{\lambda}=(\lambda_1,\dots,\lambda_I)>\mathbf{0}$ and with $\boldsymbol{\omega}=(\boldsymbol{\omega}^T_1,\dots,\boldsymbol{\omega}_I^T)^T$ where $\boldsymbol{\omega}_i=(\omega_{i1},\dots,\omega_{ik_i})^T$, we add $\boldsymbol{\omega}^TS_{\boldsymbol{\lambda}}\boldsymbol{\omega}/2$ to the loss, where $S_{\boldsymbol{\lambda}}=\sum^I_{i=1}\lambda_iS_{i}$ and $S_i$ is an $|\boldsymbol{\omega}|\times|\boldsymbol{\omega}|$ matrix with all zero entries except for a $k_i \times k_i$ block, denoted $S_i^{*}$, which begins at the $[\sum^{i-1}_{j=1} k_i+1, \sum^{i-1}_{j=1} k_i+1]$-th entry of $S_i$ if 
$i>1$ and the $[1,1]$-th entry, otherwise. The $[j,k]$-th entry of $S_i^{*}$ is given by $\psi(x_j^{*i},x_k^{*i})$. Whilst we do not use the smoothing penalty in the main text, it is implemented in the accompanying \texttt{R} package \citep{pinnEV}.
\subsection{Neural networks}
\label{NN_sec}
\subsubsection{Overview}
We describe a neural network with $J$ hidden layers, each with $n_j$ neurons for $j=1,\dots,J$, and with a set of $d$ predictors $\mathbf{x}(s,t)=(x_1(s,t),\dots,x_d(s,t))^T$ for $(s,t)\in\mathcal{S}\times\mathcal{T}$. The output for layer $j$ and neuron $i$ is described by the function $m_{j,i}(s,t)$, which differs for the two considered layer types (MLP and CNN). Defining the vector of outputs from layer $j$ by $\mathbf{m}_j(s,t)=(m_{j,1}(s,t),\dots,m_{j,n_j}(s,t))^T$, then the output layer, i.e., $m_{\mathcal{N}}$ in \eqref{full_model} of the main text, can always be described by
\[
m_{\mathcal{N}}(\mathbf{x}(s,t))=(\mathbf{w}^{(J+1)})^T\mathbf{m}_{J}(s,t),
\]
for weight vector $\mathbf{w}^{(J+1)}\in \mathbb{R}^{n_J}$. We now describe the possible choices for the output of the nodes in the $j$-th layer, which is determined by one of two classes; densely-connected or convolutional.
\subsubsection{Densely-connected layers}
\label{dense_sec}
If the $j$-th layer is densely-connected, i.e., from an MLP, then the output is simply a weighted sum of the outputs from the previous layer; the output at each node $i=1,\dots,n_j$ is
\begin{equation}
\label{dense_eq}
m_{j,i}(s,t)=h_j\left\{b^{(j,i)}+(\mathbf{w}^{(j,i)})^T\mathbf{m}_{j-1}(s,t)\right\},
\end{equation}
for bias $b^{(j,i)}\in\mathbb{R}$, weights $\mathbf{w}^{(j,i)}\in \mathbb{R}^{n_{j-1}}$ and fixed activation function $h_j$, with the convention that $m_{0,i}(s,t):=x_i(s,t)$ for $i=1,\dots,d$. The activation function is typically non-linear, and many choices can be considered \citep{ramachandran2017searching}. Throughout we use only the rectified linear unit (ReLU) activation function, defined by $h_j(x)=\max\{0,x\}$, as \cite{glorot2011deep} illustrates that they lead to better predictive performance than alternatives. Moreover, they can emulate the capabilities of biological neurons by ``switching off'' if the input value is too small, and they are computationally efficient.
\subsubsection{Convolutional layers}
\label{cnn_sec}
We note that convolutional layers are only applicable if $\mathcal{S}$ can be represented as a $D_1 \times D_2$ grid of locations. If this is the case, then a convolution filter is created, with pre-specified dimension $d_{1,j}\times d_{2,j}$ for $d_{1,j},d_{2,j}$ odd integers, $d_{1,j}\leq D_1, d_{2,j}\leq D_2$; this filter is then applied over the input map. We define a neighbourhood $\mathcal{N}_j(s)$ as the locations $\{s\in\mathcal{S}\}$ such that they form a $d_{1,j}\times d_{2,j}$ grid with site $s$ at the centre and let $M^{(j-1,i)}(s,t)$ be a $d_{1,j}\times d_{2,j}$ matrix created by evaluating $m_{j-1,i}(s,t)$ on the grid $\mathcal{N}_j(s)$. For a $d_{1,j}\times d_{2,j}$ weight matrix $W^{(j,i)}$ with real entries, the $i$-th convolutional filter for layer $j$ is then of the form $\mathcal{C}_{j,i}(m_{j-1,i}(s,t))=\sum_{k=1}^{d_{1,j}}\sum_{l=1}^{d_{2,j}}W^{(j,i)}[k,l]M^{(j-1,i)}(s,t)[k,l]$ with operations taken entry-wise; here $A[k,l]$ denotes the $(k,l)$-th entry of matrix $A$. Then if layer $m_{j,i}(s,t)$ is convolutional, it takes the form
\begin{equation}
\label{cnn_eq}
m_{j,i}(s,t)=h_j\left\{b^{(j,i)}+\sum^{n_{j-1}}_{k=1}\mathcal{C}_{j,k}(m_{j-1,k}(s,t))\right\},
\end{equation}
for $i=1,\dots,d$; equivalence with the densely-connected layers is achieved by setting $d_{1,j}=d_{2,j}=1$.\par
To ensure that the convolutional filter can be applied at the boundaries of $\mathcal{S}$, we pad the predictor maps by mapping $\mathcal{S}$ to a new domain $\mathcal{S}^*$,  a $(D_1+\max_{j=1,\dots,J}\{d_{1,j}\}-1) \times (D_2+\max_{j=1,\dots,J}\{d_{2,j}\}-1)$ grid with $\mathcal{S}$ at its centre. Then, for all $i=1,\dots,p$, we set $\mathbf{x}_i(s,t)=0$ for all $s$ within distance $(\max_{j=1,\dots,J}\{d_{1,j}\}-1)/2$ and $(\max_{j=1,\dots,J}\{d_{2,j}\}-1)/2$ of the first and second dimension, respectively, of the boundaries of $\mathcal{S}^*$. {As described in Section~\ref{Wildfire_Data_sec} of the main text, we observe the predictor variables on a much larger spatial domain than the response variable, in order to ensure that the choice of $\mathbf{x}_i(s,t)=0$ is arbitrary and the model fits do not suffer from edge effects.}

\subsubsection{Training}
\label{NN_train_sec}
Parameters for neural networks are estimated through minimisation of a loss function. When fitting regression models, i.e., conditional density estimation networks \citep{rothfuss2019conditional}, the loss is taken to be the negative log-likelihood associated with the considered statistical model. {For our model, the negative log-likelihood for the bGEV-PP model is derived by substituting   $\Lambda_{s,t}([z,\infty)):=-(1/n_y)\log G_b(z)$ into \eqref{PP_nll} of the main text. That is, we replace the mean measure of the classical extreme value point process with one based on the bGEV distribution function, $G_b(\cdot)$.  Note that $n_y$ corresponds to the number of observations in a year.} In more classical applications of neural networks, such as logistic regression/classification and non-parametric quantile regression, the loss functions are taken to be the binary cross-entropy/Bernoulli negative log-likelihood and the pinball/tilted loss \citep{koenker_2005}, respectively; both of these types of models are fitted in Section~\ref{app_sec} of the main text. {Let $\theta(s,t)$ denote the $\tau$-quantile of $Y(s,t)\mid(\mathbf{X}(s,t)=\mathbf{x}(s,t))$ for $\tau\in(0,1)$. Then, the tilted loss is defined by
\[
\ell(\theta(s,t))=\sum_{t\in\mathcal{T}}\sum_{s\in\mathcal{S}}\rho_\tau\{y(s,t)-\theta(s,t)\},
\]
where $\rho_\tau(z):=z(\tau-\mathbbm{1}\{z < 0\})$. In our application, we represent the strictly-positive conditional quantile by $\theta(s,t):=\exp[m_\mathcal{N}\{\mathbf{x}(s,t)\}]$, where $m_\mathcal{N}(\cdot)$ is a neural network.}

The loss functions are minimised using variants of stochastic gradient descent, which is performed over a finite number of iterations, referred to throughout as epochs. In our application, we use the Adaptive Moment Estimation (Adam) algorithm, see \cite{kingma2014adam} for details, which uses adaptive learning rates for updating parameters and loss minimisation.  
Adam is performed using the deep learning \texttt{R} package \texttt{keras} \citep{kerasforR} with computationally efficient and exact evaluation of the gradient of the loss achieved through backpropagation. When training NNs with \texttt{keras}, an issue occurs if parameters are updated to infeasible values, i.e., where the loss function cannot be evaluated; training stops with parameters that provide a sub-optimal value of the loss. Whilst this problem does not occur for the majority of loss functions (if the parameter/link functions are correctly specified), it can be an issue for conditional density estimation when the endpoints of the considered distribution are dependent on the model parameters, e.g., in the case of the classical PP and GEV models described in Section~\ref{intro_evmodel_sec} of the main text. \par
To avoid over-fitting with neural networks, it is common practice to perform out-of-sample validation. The data are partitioned into a validation set, with the remaining data used for training the network. Performance of the model is evaluated by computing various metrics for the validation set, e.g., the loss function or the goodness-of-fit/predictions scores described in Section~\ref{sec:arch} of the main text. Note that the validation set is not used in estimating the parameters of the neural network at all, and so minimising the loss for the validation set, rather than the training set, {can produce models that exhibit less over-fitting and better generalise to unseen data}. In our application, the partitioning of the data into validation and training sets is performed at random; although this can be done uniformly, we instead assign space-time clusters (see Section~\ref{sec:pre_process} of the main text) of values to the validation set. Model validation further exacerbates the issue with the finite distributional endpoint; it is not guaranteed that a feasible endpoint will be predicted for the validation data, particularly in high-dimensional settings \citep{castro2021practical}. Hence to overcome this issue, and the training issue described above, we propose a new flexible class of extreme value PP models, in Section~\ref{bgev_PP_sec} of the main text, where the distributional endpoints are infinite, and so can be feasibly fitted using NNs.
 \section{Simulation study}
\label{sim_study}
\subsection{Connection to main text}
This appendix supports the material in Section~\ref{sim_mainpaper_sec} of the main text. Here we detail a simulation study to illustrate the efficacy of the proposed model proposed. Appendix~\ref{par_est_sec} and \ref{misspec_sec} consider parameter estimation and robustness under model misspecification, respectively. Given the complexity and high-dimensionality of the dataset used in our application, we consider here a simplified version of the model adopted in Section~\ref{app_sec} of the main text.
\subsection{Parameter estimation}
\label{par_est_sec}
{We conduct a simulation study to identify the efficacy of our approach in estimating the interpretable function $m_\mathcal{I}(\cdot)$ in \eqref{full_model}. We simulate $n$ replications of $Y\mid (\mathbf{X}=\mathbf{x})$, where $\mathbf{X}\in\mathbb{R}^{10}$ is a $10$-dimensional vector of covariates, drawn from a multivariate Gaussian distribution with all pairwise correlations equal to $1/2$. The response-predictor relationship of six components of $\mathbf{X}$ are taken to be highly non-linear and we estimate their relationship using $m_\mathcal{N}(\cdot)$; the remaining four components are interpreted. \par
We simulate $Y\mid(\mathbf{X}=x)$ from the limiting Poisson process model described in Section~\ref{intro_evmodel_sec} of the main text, albeit with the same parameterisation described in Section~\ref{bGEV_sec} of the main text. We let $Y\mid(\mathbf{X}=x) \sim \mbox{PP}(q_\alpha(\mathbf{x}),s_\beta(\mathbf{x}),\xi; u)$ where the functions $q_\alpha$ and $s_\beta$ are represented as PINNs; see \eqref{full_model} of the main text. We set $\xi=0.2 > 0$ and so the PP for $Y\mid(\mathbf{X}=\mathbf{x})$ is defined on the set $A_\mathbf{x}=(z,\infty)$ for $z> u > z_-(\mathbf{x})$; here $z_-(\mathbf{x})$ denotes the lower-endpoint of the corresponding GEV for $Y\mid\mathbf{X}$ which is dependent on $\mathbf{x}$. As $u$ must be a sufficiently high threshold and we must ensure the same number of exceedances of $Y\mid(\mathbf{X}=\mathbf{x})$ above $u$ for any $\mathbf{x}$, we also let $u$ depend on $\mathbf{X}$ and fix it to the known $99\%$ quantile of $Y\mid(\mathbf{X}=\mathbf{x})$.\par
We let $q_\alpha(\mathbf{x})=\eta_0^{(1)}+m^{(1)}_\mathcal{L}(\mathbf{x})+m^{(1)}_\mathcal{A}(\mathbf{x})+m^{(1)}_\mathcal{N}(\mathbf{x})$ and $s_\beta(\mathbf{x})=\exp\{\eta_0^{(2)}+m^{(2)}_\mathcal{L}(\mathbf{x})+m^{(2)}_\mathcal{A}(\mathbf{x})+m^{(2)}_\mathcal{N}(\mathbf{x})\}$, where, for $i=1,2$,  $\eta_0^{(i)}\in\mathbb{R}$, $m^{(i)}_\mathcal{L}(\mathbf{x})$ is a linear function, $m^{(i)}_\mathcal{A}(\mathbf{x})$ is a smooth additive function, and $m^{(i)}_\mathcal{N}(\mathbf{x})$ is a highly nonlinear and non-additive function. For $q_\alpha$ and $s_\beta$, we set the intercepts as $\eta^{(1)}_0=1$ and $\eta^{(2)}_0=-0.5$, respectively. All of the following functions are defined for $\mathbf{x}=(x_1,\dots,x_{10})\in\mathbb{R}^{10}$, but do not necessarily act on all components: $m_\mathcal{N}$, $m_\mathcal{L}$, and $m_{\mathcal{A}}$ act on the first six components, $(x_7,x_8),$ and $(x_9,x_{10})$, respectively. For $m_\mathcal{N}$, we adopt the approach of \cite{zhong2021neural} and use the two highly non-linear and non-additive functions  given by
{\begin{align}
\label{sim_m_n}
m^{(1)}_\mathcal{N}(\mathbf{x})&=0.1\bigg(x_1x_2+x_2\left[1-\cos(\pi x_2 x_3)\right]+2\frac{\sin(x_3)}{\mid x_3-x_4\mid+2}+0.2(x_4+x_4x_5/2)^2\nonumber\\&-\sqrt{x_5^2+x_6^2+2}+\exp\left\{-12+\sum^6_{i=1}x_i/10\right\}\bigg),\nonumber\\
m^{(2)}_\mathcal{N}(\mathbf{x})&=0.1\bigg(0.7x_1x_2-5+x_2\left[1-\cos(\pi x_2 x_3)\right]+3\frac{\sin(x_3)}{\mid x_3-x_4\mid+2}\nonumber\\&+0.2(x_4+x_4x_5/2-1)^2+\exp\left\{-18+\sum^6_{i=1}x_i/10\right\}\bigg).
\end{align}}
For the linear and additive functions, we consider two cases: defined for $\mathbf{x}=(x_7,x_8)\in\mathbb{R}^2$, we first consider $m^{(1)}_\mathcal{L}(\mathbf{x})=0.8x_7+2x_8$ and $m^{(2)}_\mathcal{L}(\mathbf{x})=0.4x_7-0.2x_8$ with the additive functions taken to be cubic polynomials; we let $m^{(1)}_\mathcal{A}(\mathbf{x})=0.2(0.1x_9^3-x_9^2+x_9+0.4x_{10}^3-2x_{10})$ and $m^{(2)}_\mathcal{A}(\mathbf{x})=0.2(0.2x_9^3-0.3x_9^2+x_9-0.1x_{10}^3+0.2x_{10}^2-0.5x_{10})$. For the second case, we instead let $m^{(1)}_\mathcal{L}(\mathbf{x})=-0.5x_8$ and $m^{(2)}_\mathcal{L}(\mathbf{x})=0.3x_8$ with additive functions $m^{(1)}_\mathcal{A}(\mathbf{x})=x_{10}$ and $m^{(2)}_\mathcal{A}(\mathbf{x})=0.2(0.1x_{10}^3-0.3x_{10}^2-x_{10})$. That is, in the second case we remove the influence of $X_7$ and $X_9$ on $Y$ in order to determine if our approach can identify this behaviour. We further let $m^{(1)}_\mathcal{A}$ be linear to investigate how well the splines can approximate such functions. \par
To estimate the functions for $Y\mid(\mathbf{X}=\mathbf{x})$, we fit the bGEV-PP PINN model, described in Section~\ref{bgev_PP_sec} of the main text, using $10000$ epochs for training. Before fitting, the components of $\mathbf{X}$ are subset into their known effect class, i.e., ``linear'', ``additive'', or ``NN'', and we exploit this knowledge during modelling. That is, we model the impact of the ``linear'' predictors on $q_\alpha$ or $s_\beta$ using linear functions, and similarly with the ``additive'' predictors and splines. All of the additive functions are estimated using splines with 20 knots taken to be marginal quantiles of the predictors with equally spaced probabilities, and with no smoothing penalty used in the loss. The highly non-linear NN functions are estimated using a densely-connected MLP with ReLU activation functions; the architecture is the same for both $q_\alpha$ and $s_\beta,$ and has four layers with widths $(8,6,4,2)$.\par
\begin{table*}
 \centering
  \caption{Estimates of the MSE and MISE, defined in \eqref{ISE}, of the function estimates for the parameter estimation study described in Section~\ref{par_est_sec}.}
\label{simTab}
 \begin{tabular}{l| c| c| c| c| c} 
\nocell{2} &\multicolumn{2}{| c|}{MSE $(\times 10^{-3})$ $(\eta_1/\eta_2)$}&\multicolumn{2}{c}{MISE $(\times 10^{-3})$ ($m_{\mathcal{A},1}/m_{\mathcal{A},2})$}\\
\cline{2-6}
 & $n$ & $q_\alpha$&$s_\beta$ & $q_\alpha$&$s_\beta$\\
\hline
\multirow{4}{*}{Case 1}&1$\times10^4$&53.1/313&85.0/38.3&31.2/75.3&49.5/56.0\\
&1$\times10^5$&0.51/0.54&8.04/3.63&6.04/35.9&9.22/13.7\\
&5$\times10^5$&0.21/0.12&5.13/1.38&3.65/37.1&6.09/10.8\\
&1$\times10^6$&0.20/0.13&4.46/1.46&3.66/40.2&5.35/10.1\\
\hline
\hline
\multirow{4}{*}{Case 2}&1$\times10^4$&2.11/15.2&16.6/16.6&7.8/44.8&41.0/73.8\\
&1$\times10^5$&0.59/0.63&1.51/5.24&2.44/4.70&5.83/15.1\\
&5$\times10^5$&0.12/0.16&0.42/2.48&0.50/1.61&1.39/7.19\\
&1$\times10^6$&0.05/0.12&0.16/1.87&0.31/1.23&0.74/6.20\\
\hline
 \end{tabular}
  \end{table*}
To quantify the accuracy of the parameter estimation of the model, we simulate $100$ sets of replications of $Y\mid(\mathbf{X}=x)$ and estimate the functions for each set. The mean squared error (MSE) for each of the linear regression coefficients (denoted $\eta_1, \eta_2$) is reported in Table~\ref{simTab}. For the additive functions, we first decompose these into the respective effect of each predictor, i.e., we let $m_\mathcal{A}(x_9,x_{10})=m_{\mathcal{A},1}(x_9)+m_{\mathcal{A},2}(x_{10})$, where $m_{\mathcal{A},1}(x_9)$ and $m_{\mathcal{A},2}(x_{10})$ act only on the first and second components, respectively. Then for each component of $m_\mathcal{A}(x_9,x_{10})$, we evaluate the mean integrated squared error (MISE) between translations of the true and estimated functions, with the latter denoted by $\widehat{m}_{\mathcal{A},i}$ for $i=1,2$. That is, we report the mean of 
\begin{equation}
\label{ISE}
\int_{\mathbb{R}}\left\{[m_{\mathcal{A},i}(x_i)-m_{\mathcal{A},i}(0)]-[\widehat{m}_{\mathcal{A},i}(x_i)-\widehat{m}_{\mathcal{A},i}(0)]\right\}^2\mathrm{d}x_i
\end{equation} for $i=1,2$. The centering ensures that we evaluate the MISE between two functions that are equal at zero and is done to account for the possibility that estimates $\widehat{m}_{\mathcal{A},i}$ are translated from the true function, which may be caused by identifiability issues between the intercept $\eta_0$ and $m_\mathcal{N}$; our focus is on correct estimation of the shape of $m_\mathcal{A}$, rather than its exact values. We estimate \eqref{ISE} numerically; the integrand is evaluated for a sequence of $x_i$ taken to be $5000$ quantiles with equally spaced probabilities. Table~\ref{simTab} illustrates that our modelling approach is capable of estimating the true underlying regression model, with increasingly small values of the MSE and MISE given for increasing $n$ in both cases. Note that although $n$ appears large, because we take $u$ to be the $99\%$ theoretical quantile of $Y\mid(\mathbf{X}=\mathbf{x})$ only $0.01n$ replications are deemed to be ``extreme''; for the datasets where our approach is applicable, our choice of $n$ is reasonable. The results for case 2 in Table~\ref{simTab} illustrate that our approach is also able to identify well the absence of significant relationships between $Y$ and predictors, as well as the presence; this is an important property of our approach as it suggests that the statistical inferences we draw from our fitted models are well-founded.

\subsection{Misspecification}
\label{misspec_sec}
\subsubsection{Overview}
We now perform two studies to highlight the robustness of our modelling approach when the response distribution or the functional form of the PINN are misspecified, and illustrate the flexibility of the bGEV-PP model for predicting extreme quantiles.
\subsubsection{Response distribution}
\label{misspec_sec_1}
For this study, we assume that $Y\mid(\mathbf{X}=\mathbf{x})$ does not follow the limiting PP model, but we fit our bGEV-PP model regardless. Two cases are considered for $Y\mid(\mathbf{X}=\mathbf{x})$: a heavy-tailed distribution in the maximum-domain of attraction of a GEV distribution with positive tail index $\xi>0$, and a lighter-tailed distribution with tail index $\xi=0$. Predictors are drawn from a $10$-dimensional standard Gaussian distribution with all pairwise correlations equal to $0.3$. In the heavy-tailed case, we let $Y\mid(\mathbf{X}=\mathbf{x})\sim \mbox{GPD}(\exp\{0.5-3m^{(2)}_\mathcal{N}(\mathbf{x})\},0.1)$, with $m^{(2)}_\mathcal{N}(\mathbf{x})$ defined in \eqref{sim_m_n}. In the lighter-tailed case, we draw $Y\mid(\mathbf{X}=\mathbf{x})$ from a log-normal distribution with density function $f(y)=(\sqrt{2\pi}\sigma y)^{-1}\exp\{-(\log y - \mu)^2/(2\sigma^2)\}$ for $y>0, \sigma >0,$ and $\mu \in \mathbb{R}$. The parameters $\mu$ and $\sigma$ are dependent on $\mathbf{x},$ and we set $\mu(\mathbf{x})=6+0.2m^{(1)}_\mathcal{N}(\mathbf{x})$ and $\sigma(\mathbf{x})=1/2,$ with $m^{(1)}_\mathcal{N}(\mathbf{x})$ defined in \eqref{sim_m_n}; hence only six components of $\mathbf{X}$ act on $Y$.  \par
In both cases, we fit a bGEV-PP PINN model with no intepretable functions (i.e., a fully-NN bGEV-PP model) using the same architecture as described in Section~\ref{par_est_sec} and trained for 10000 epochs; note that, for computational ease, we force $\xi\in(0,1)$ when fitting. We quantify the ability of our model to capture the tails of a misspecified response distribution by simulating 100 sets of $Y\mid(\mathbf{X}=\mathbf{x})$, training a model for each and then using the bGEV-PP parameter estimates to estimate the distribution function, denoted $\hat{F}_i$, for each observation of the predictor set $\mathbf{x}_i$ and for all $i=1,\dots,n$. We then define the set tail-weighted log-survival score (stLS) for $p_{-}\in[0,1)$ by $\mbox{stLS}(p_-)$ equals 
\begin{align}
\label{stLS}
\frac{1}{n}\sum^n_{i=1}\int^1_{p_-}(\log(1-\hat{F}_i\{\tilde{F}_i^{-1}(p)\})-\log(1-p))^2\mathrm{d}p,
\end{align}
where $\tilde{F}^{-1}_i$ denotes the true quantile function for the observed $\mathbf{x}_i$. As our interest lies in scoring the upper-tails of the predicted distribution, we set $p_-=0.99$.\par
\begin{table}
\caption{Estimates of the mean stLS$(0.99)$ (s.d.), defined in \eqref{stLS}, for the response distribution misspecification study described in Section~\ref{misspec_sec_1}.}
\label{misspecTab}
 \centering
 \begin{tabular}{c|c| c} 
n&Case 1&Case 2\\
\hline
2.5$\times10^5$ &7.79 (25.3)&1.62 (7.11)\\
5$\times10^5$&1.46 (3.38)&0.22 (0.14)\\
1$\times10^6$&0.56 (0.21)&0.10 (0.19)\\
2$\times10^6$&0.40 (0.05)&0.04 (0.01)\\
\hline
 \end{tabular}

  \end{table}
 We tabulate the mean stLS$(0.99)$ over all $100$ sets of $Y\mid(\mathbf{X}=\mathbf{x})$ in Table~\ref{misspecTab}. Estimates are derived numerically using $200$ equally-spaced values of $p\in[0.99,0.9999]$. We observe increasingly small values of mean stLS$(0.99)$ as $n$ increases for both cases, suggesting that the bGEV-PP model is able to approximate well extreme quantiles of heavy-tailed and lighter-tailed distributions, even when the response distribution is misspecified. Results for Case 2 highlight the robustness of our approach, as the log-normal distribution is in the maximum-domain of attraction of a Gumbel distribution; this is not a sub-class of our fitted bGEV-PP model as, for computational purposes, we force $\xi>0$ when training. Moreover, we fit the limiting bGEV-PP model which is itself misspecified as it lies on the boundary of the parameter space, yet in both cases it provides accurate estimates of extreme quantiles.
  \subsubsection{Functional form of $\theta$}
\label{misspec_sec_2}
We now simulate 100 sets of $n=4\times10^6$ replications of $Y\mid(\mathbf{X}=\mathbf{x})$ from the re-parameterised limiting point process model \[\mbox{PP}(q_\alpha(\mathbf{x}),s_\beta(\mathbf{x}),\xi=0.25; u)\] (see Section~\ref{par_est_sec}), with $u$ fixed as the $95\%$ quantile of $Y\mid(\mathbf{X}=\mathbf{x})$, and fit the bGEV-PP model to each set. Predictors are drawn from a $12$-dimensional standard Gaussian distribution with all pairwise correlations equal to $0.3$. Three cases are considered for the true functions $\theta(\mathbf{x})=(q_\alpha(\mathbf{x}),s_\beta(\mathbf{x}))$: (i) linearity,  (ii) additivity, and (iii) high non-linearity, and all functions are defined for $\mathbf{x}=(x_1,\dots,x_{12})\in\mathbb{R}^{12}$. In case $(i)$, we let $q_\alpha(\mathbf{x})=1+\mathbf{x}^T\boldsymbol{\eta}_1$ and $s_\beta(\mathbf{x}) = \exp(0.5+\mathbf{x}^T\boldsymbol{\eta}_2)$, where entries to the coefficient vectors $\boldsymbol{\eta}_1\in[-1,1]^{12}$ and $\boldsymbol{\eta}_2\in[-0.5,0.5]^{12}$ are drawn uniformly at random from $\{-1,-0.9,\dots,0.9,1\}$ and $\{-0.5,-0.4,\dots,0.4,0.5\}$, respectively. For the second case, we let $q_\alpha(\mathbf{x})=15+\mathbf{x}_*^T\boldsymbol{\eta}_3$ and $s_\beta(\mathbf{x})=\exp(1-0.05\mathbf{x}_*^T\boldsymbol{\eta}_4)$ where $\mathbf{x}_*= (x_1^3,x_1^2,x_1,\dots,x_{12}^3,x_{12}^2,x_{12})$; entries to $\boldsymbol{\eta}_3\in[-1,1]^{36}$ and $\boldsymbol{\eta}_4\in[-0.5,0.5]^{36}$ are drawn similarly to $\boldsymbol{\eta}_1$ and $\boldsymbol{\eta}_2$, respectively. We note that in the first two cases, the coefficients are drawn randomly but then fixed across all sets. For the final case, we let $q_\alpha(\mathbf{x})=20+25\{m_\mathcal{N}^{(1)}(x_1,\dots,x_6,0,\dots,0)+|m_\mathcal{N}^{(1)}(x_7,\dots,x_{12},0,\dots,0)|\}$ and \begin{align*}
s_\beta(\mathbf{x})=\allowbreak\exp[0.5-&\{m_\mathcal{N}^{(2)}(x_1,\dots,x_6,0,\dots,0)-m_\mathcal{N}^{(2)}(x_7,\dots,x_{12},0,\dots,0)\}]
\end{align*} for $m_\mathcal{N}^{(1)}$ and $m_\mathcal{N}^{(2)}$ defined in \eqref{sim_m_n}.
       \par 
       In each of the cases for the true functions, we fit four models: a linear model, a GAM, a PINN, and a fully-NN model (see, also, Section~\ref{sec:model_compare} of the main text). All additive functions are estimated using splines with 10 knots of equally-spaced marginal quantiles. For the PINN model, we represent $m_\mathcal{I}$ as the sum of linear and additive contributions (as in Section~\ref{par_est_sec}); we assume that the ``linear'' and ``additive'' predictors are $\{x_1,x_2\}$ and $\{x_3,x_4\}$, respectively. For the neural network architecture of the PINN and fully-NN models, we use four hidden layers with widths $(12,8,4,2)$. To mitigate the risk of overfitting, replications are randomly partitioned into $80\%$ training and $20\%$ validation data; networks are then trained on the training data for 15000 epochs with batch size $n/4$, with the best network taken to be that which provides the minimum validation loss. The mean validation stLS$(0.99)$, defined in \eqref{stLS}, for each combination of true and fitted distributions and over all $100$ sets are provided in Table~\ref{misspecTab2}, alongside the number of parameters for each fitted model.  \par
  Unsurprisingly, we observe that the best performing model in each case is that which is well-specified, e.g., the fully-linear model when the true $\theta$ is linear, which we deduce from the lower stLS(0.99) estimates. Whilst the fully-linear model performs very well when the true $\theta$ is linear, it performs particularly poorly in the more realistic scenarios, i.e, when $\theta$ is additive or highly non-linear. The fully-NN model performs well across the three cases and is the best performing model in the realistic highly non-linear case. We observe that the PINN model performs well when specified correctly, i.e., when the true $\theta$ is linear, but its performance can suffer when misspecified, such as when the true $\theta$ is additive. However, we observe that in the highly non-linear case where all models are misspecified, the PINN model outperforms both the baseline fully-linear and GAM models. This study suggests that we must take particular care to avoid misspecification of the linear and GAM components of our PINN model. 
\begin{table*}
\caption{Estimates of the mean validation stLS$(0.99)$  (s.d.), defined in \eqref{stLS}, for the $\theta$ misspecification study described in Section~\ref{misspec_sec_2}.}
\label{misspecTab2}
 \centering
  \begin{tabular}{c| c| c| c| c}
  \nocell{2}&\multicolumn{3}{| c}{True $\theta$ functional form}\\
  \hline
Fitted $\theta$ model &No. parameters & Linear&Additive &Highly non-linear \\
\hline
linear&27&0.002 (2$\times 10^{-4}$)&2.819 (0.048)& 1.502 (0.211)\\
GAM&243&0.528 (0.540)&0.275 (0.005)&0.800 (0.005)\\\
PINN &567&0.133 (0.095)&1.316 (0.027) &0.654 (0.009)\\
NN&619&0.014 (0.001)&0.522 (0.010) &0.514 (0.010)\\
\hline
 \end{tabular}
 
  \end{table*}
  }

\section{Stationary bootstrap}
\label{boot_sm_sec}
  \subsection{Connection to main text}
This appendix supports the material in Section~\ref{sec:pre_process} of the main paper.  Appendix~\ref{appendx-boot-details} provides details of the stationary bootstrap procedure used to assess parameter uncertainty in the final analysis.
\subsection{Details}
\label{appendx-boot-details}
In Section~\ref{sec:pre_process} of the main text, we assess model uncertainty by using a stationary bootstrap \citep{politis1994stationary} with expected block size $m_K$. To create a single bootstrap sample, we repeat the following until obtaining a sample of length greater than or equal to $\mid\mathcal{T}\mid$; draw a starting time $t^*\in\mathcal{T}$ uniformly at random and a block size $K$ from a geometric distribution with expectation $m_K$, then add the block of observations $\{y(s,t):s \in \mathcal{S}, t\in \{t^*,\dots,t^*+K-1\}\}$ to the bootstrap sample. In cases where $t^*$ is generated with $t^*+K-1 > \mid\mathcal{T}\mid$, we instead add $\{y(s,t):s\in\mathcal{S}, t\in \{1,\dots,t^*+K-\mid\mathcal{T}\mid-1\}\cup\{t^*,\dots,\mid\mathcal{T}\mid\}\}$. The sample is then truncated to have length $\mid\mathcal{T}\mid$.

\newpage
\section{Supplementary figures}
\label{sup_figs}

\begin{figure}[h!]
\centering
\includegraphics[width=0.45\linewidth]{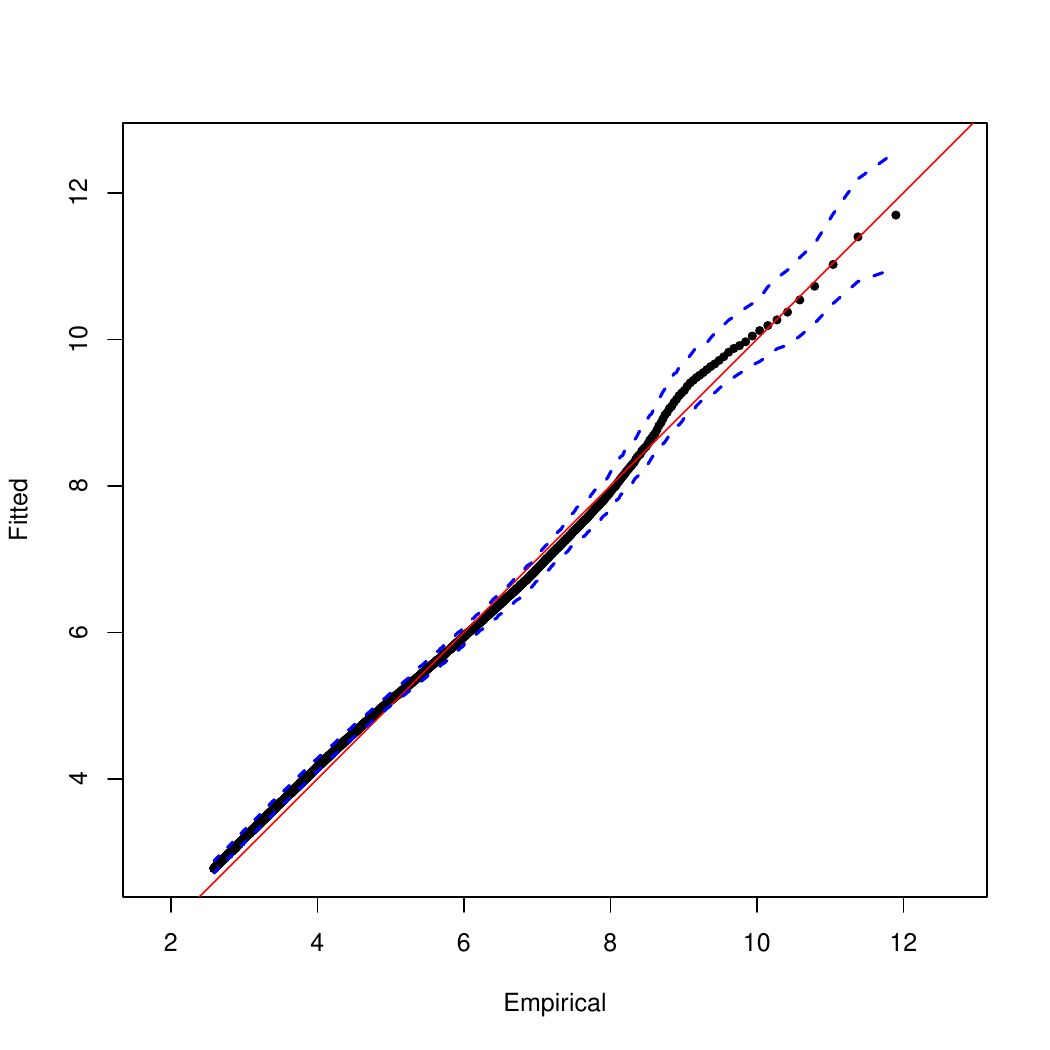} 
\caption{Pooled Q-Q plot for the local PINN bGEV-PP model fit on standard Exponential margins, averaged across all bootstrap samples. $95\%$ tolerance bounds are given by the blue dashed lines. Black points give the median quantiles across all samples, with the quantile levels ranging from $0.925$ to a value corresponding to the maximum observed value.}
\label{fig_exp_fit}
\end{figure}

\begin{figure}[t!]
\centering
\begin{minipage}{0.49\linewidth}
\flushleft
\includegraphics[width=0.9\linewidth]{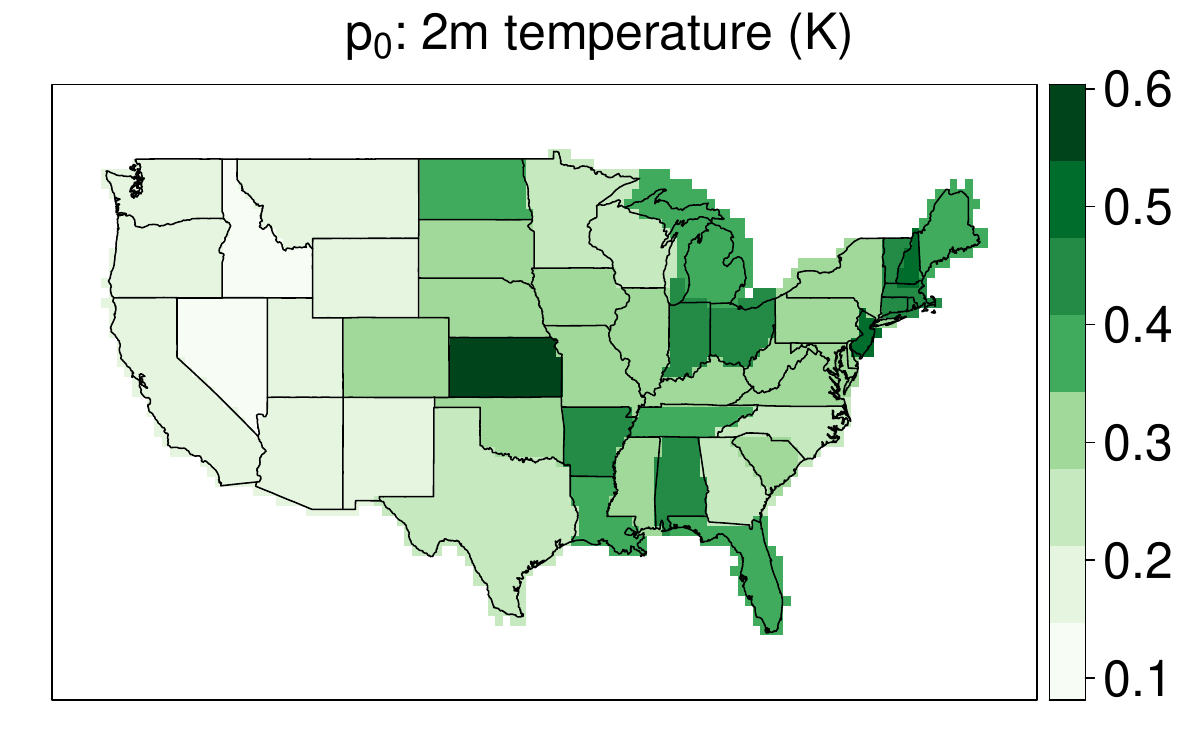} 
\end{minipage}
\begin{minipage}{0.49\linewidth}
\flushleft
\includegraphics[width=0.9\linewidth]{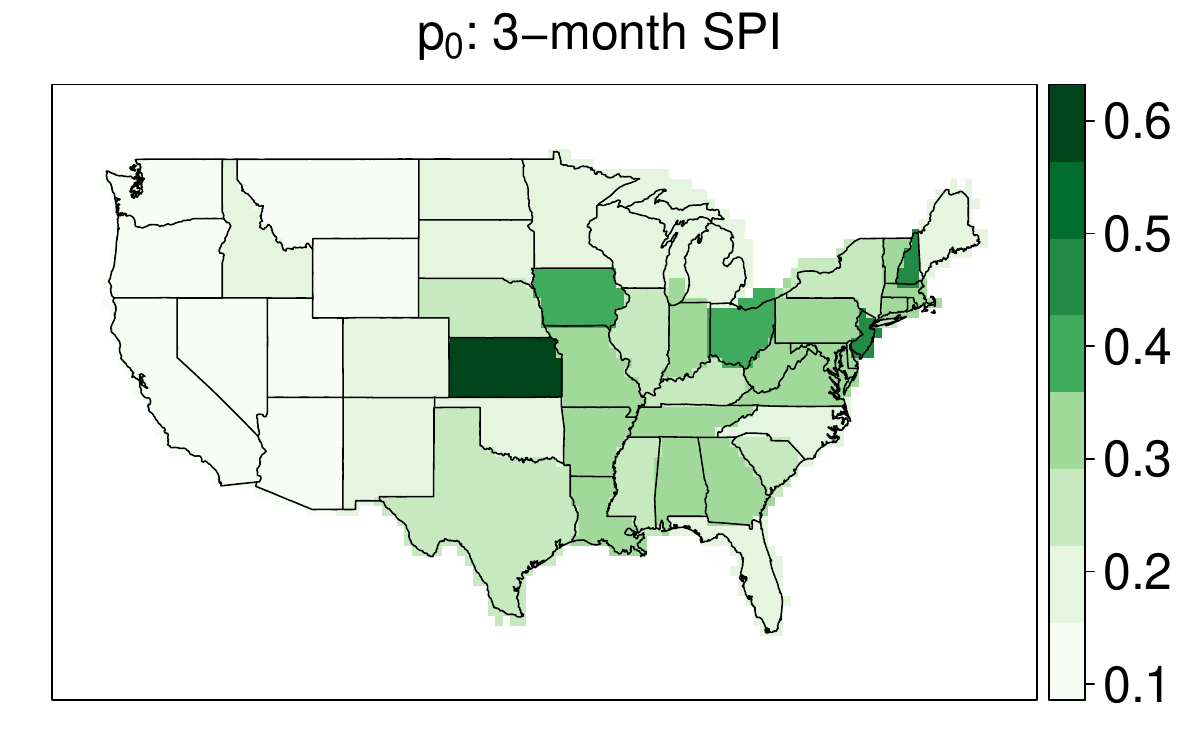} 
\end{minipage}
\begin{minipage}{0.49\linewidth}
\flushleft
\hspace{0.02cm}
\includegraphics[width=0.9\linewidth]{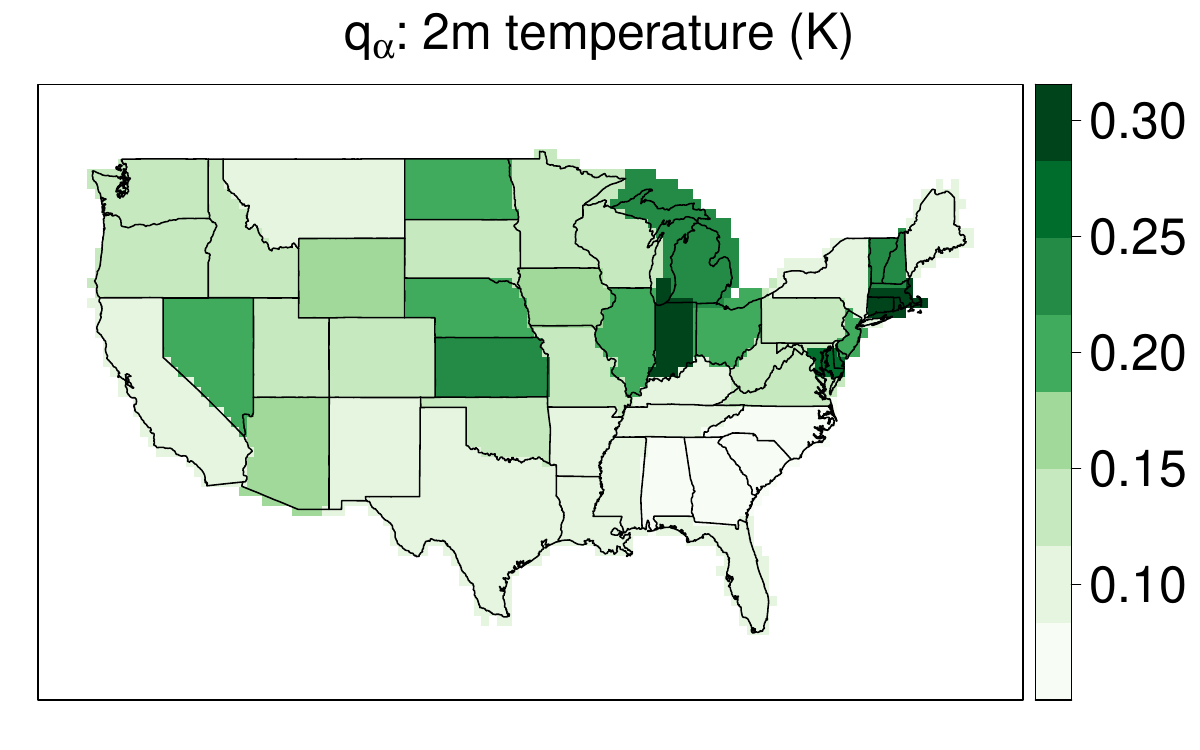} 
\end{minipage}
\vspace{-0.3cm}
\begin{minipage}{0.49\linewidth}
\flushleft
\hspace{0.025cm}
\includegraphics[width=0.9\linewidth]{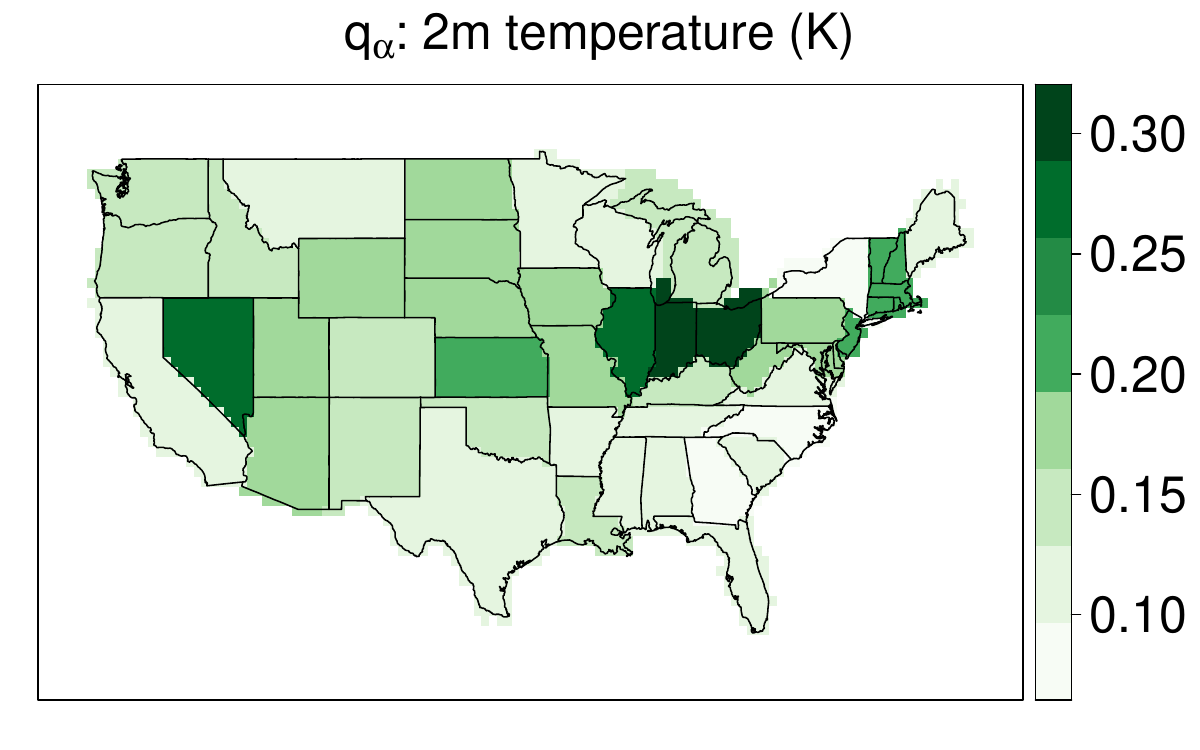} 
\end{minipage}
\begin{minipage}{0.49\linewidth}
\flushleft
\includegraphics[width=0.9\linewidth]{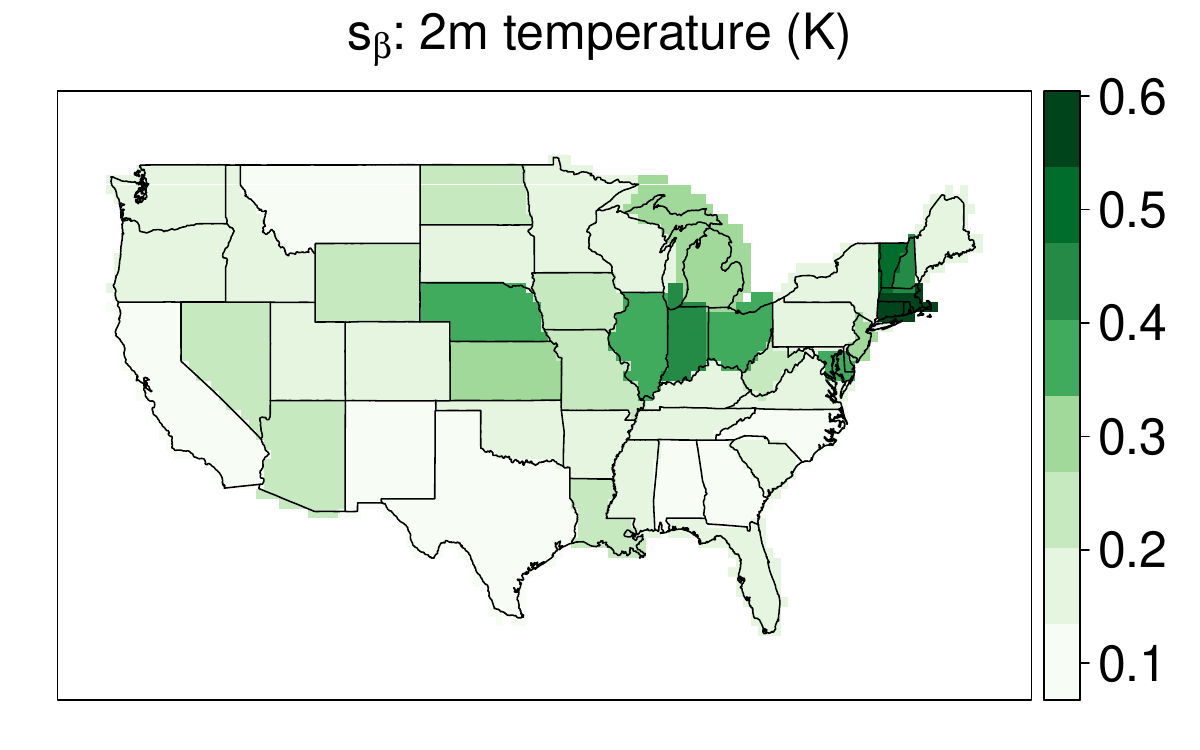} \end{minipage}
\begin{minipage}{0.49\linewidth}
\flushleft
\includegraphics[width=0.9\linewidth]{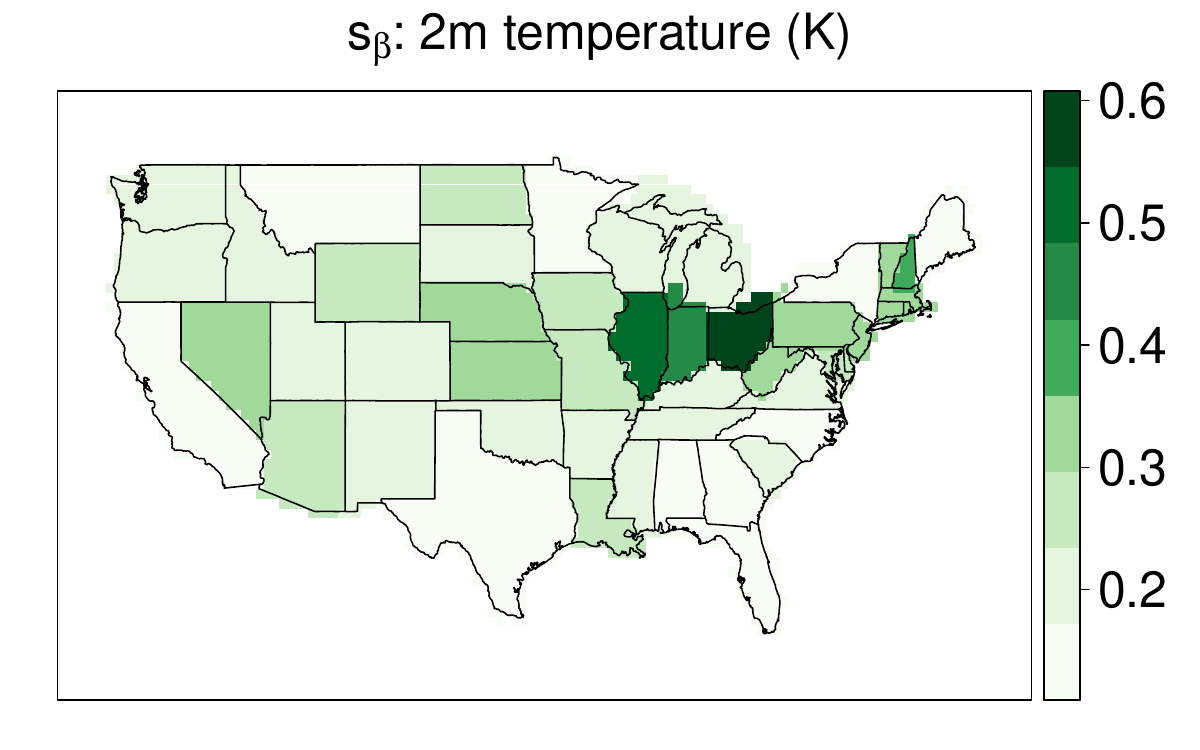} 
\end{minipage}
\vspace{-0.3cm}
\caption{Maps of the inter-quartile range for the estimated $m_\mathcal{I}$ for the local PINN model. Here $m_\mathcal{I}$ is a linear model with state-varying coefficients. Mapped values correspond to the contribution of 2m air temperature (K; left column) and 3-month SPI (unitless; right column) to the log-odds of occurrence probability $p_0(s,t)$ (unitless; top row), log-location $\log \{q_\alpha(s,t)\}$ (log-$\sqrt{\mbox{acres}}$; centre row), and log-scale $\log \{s_\beta(s,t)\}$ (log-$\sqrt{\mbox{acres}}$; bottom row). }
\label{fig:SVC_IQR}
\end{figure}

\begin{figure}[t!]
\centering
\begin{minipage}{0.49\linewidth}
\flushleft
\hspace{0.02cm}
\includegraphics[width=0.9\linewidth]{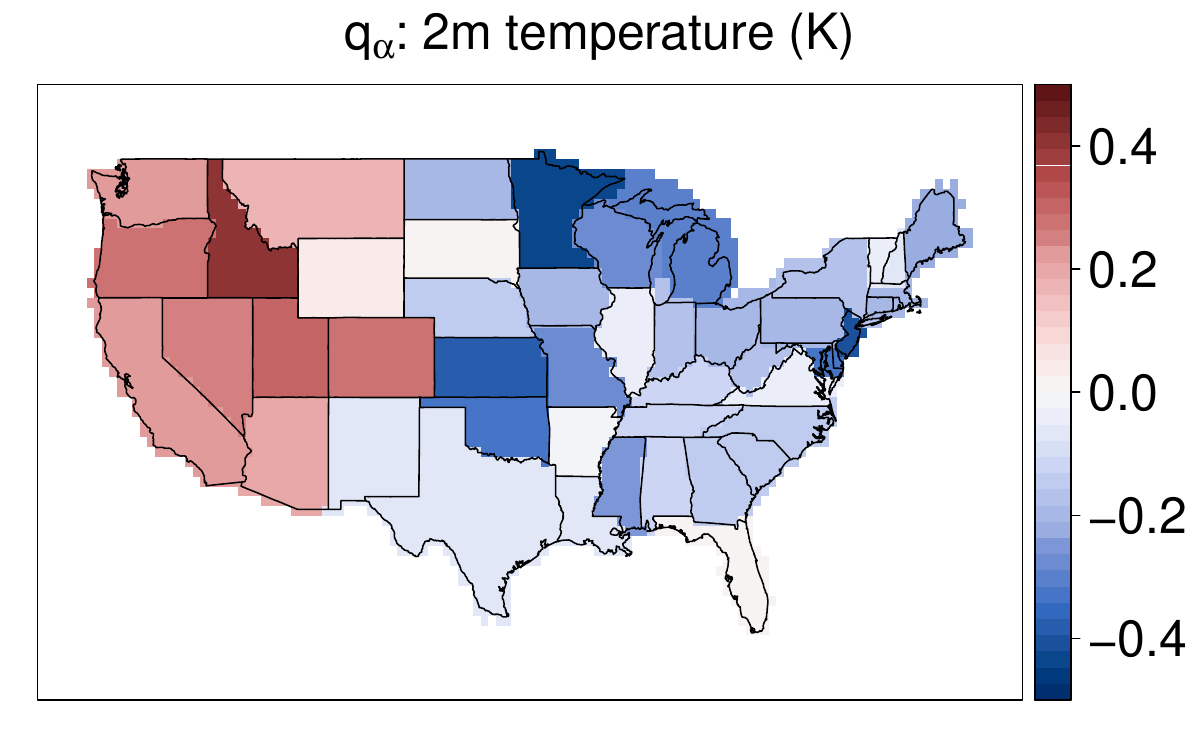} 
\end{minipage}
\vspace{-0.3cm}
\begin{minipage}{0.49\linewidth}
\flushleft
\hspace{0.025cm}
\includegraphics[width=0.9\linewidth]{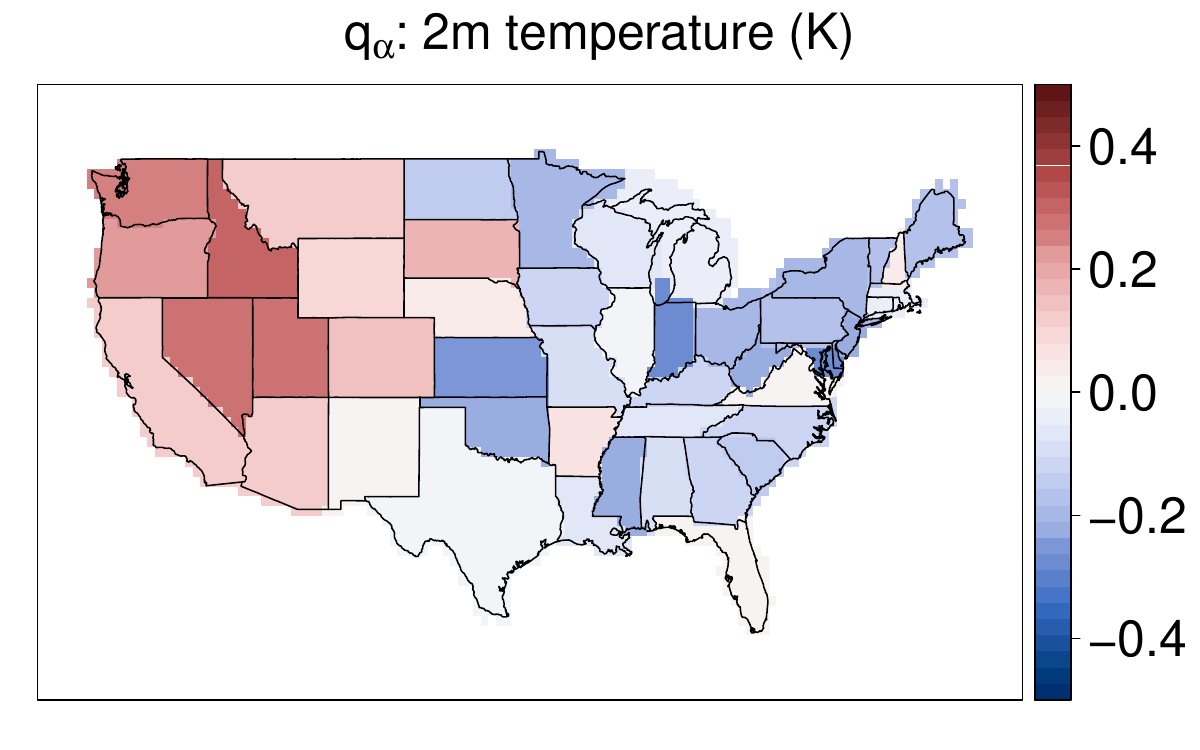} 
\end{minipage}
\begin{minipage}{0.49\linewidth}
\flushleft
\hspace{0.02cm}
\includegraphics[width=0.9\linewidth]{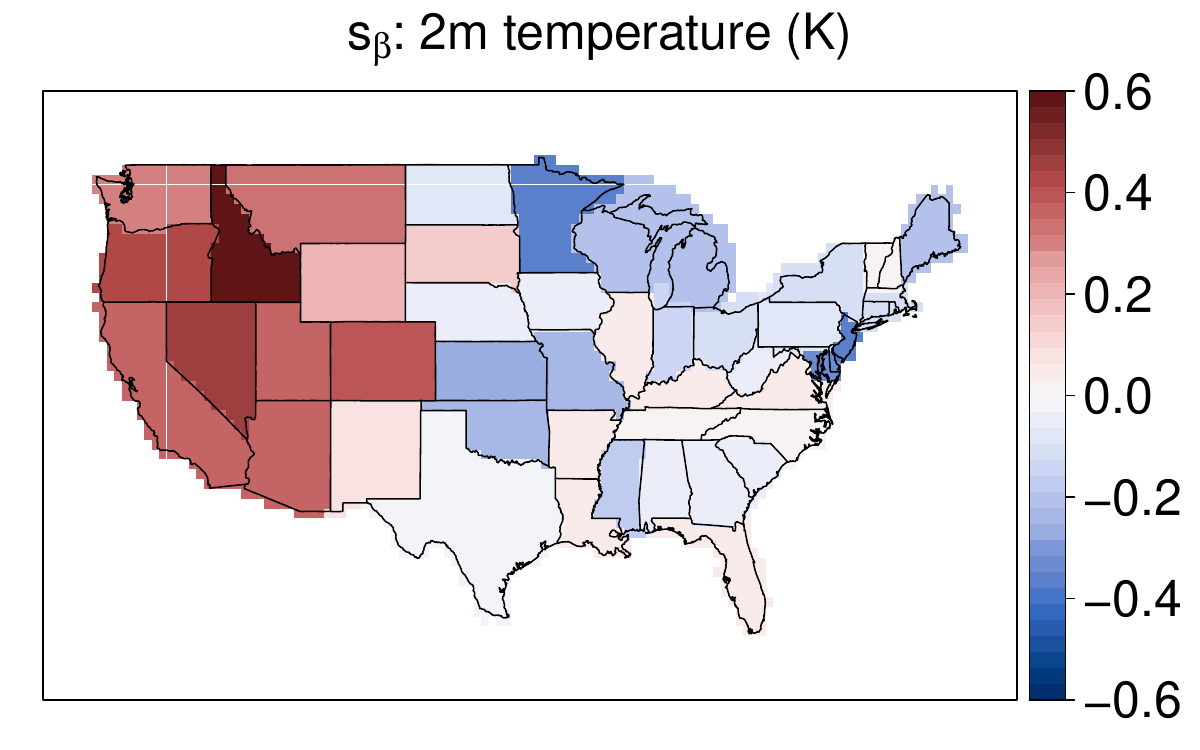} 
\end{minipage}
\vspace{-0.3cm}
\begin{minipage}{0.49\linewidth}
\flushleft
\hspace{0.025cm}
\includegraphics[width=0.9\linewidth]{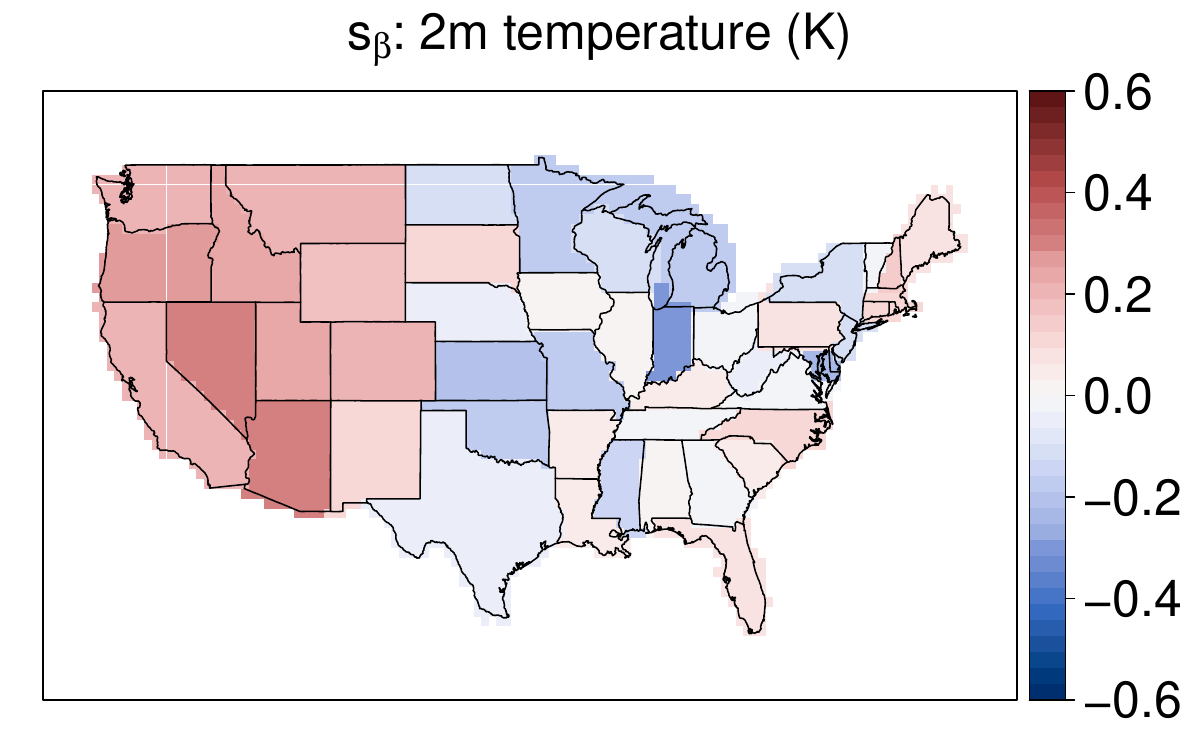} 
\end{minipage}
\begin{minipage}{0.49\linewidth}
\flushleft
\hspace{0.02cm}
\includegraphics[width=0.9\linewidth]{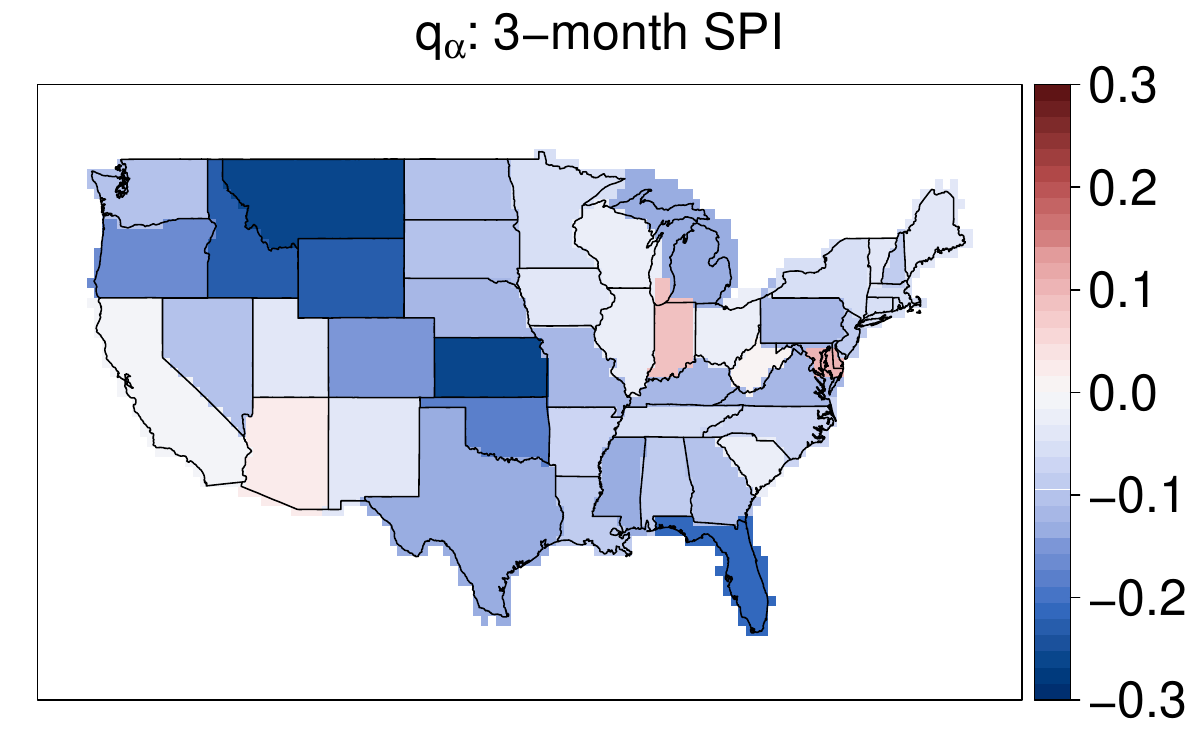} 
\end{minipage}
\vspace{-0.3cm}
\begin{minipage}{0.49\linewidth}
\flushleft
\hspace{0.025cm}
\includegraphics[width=0.9\linewidth]{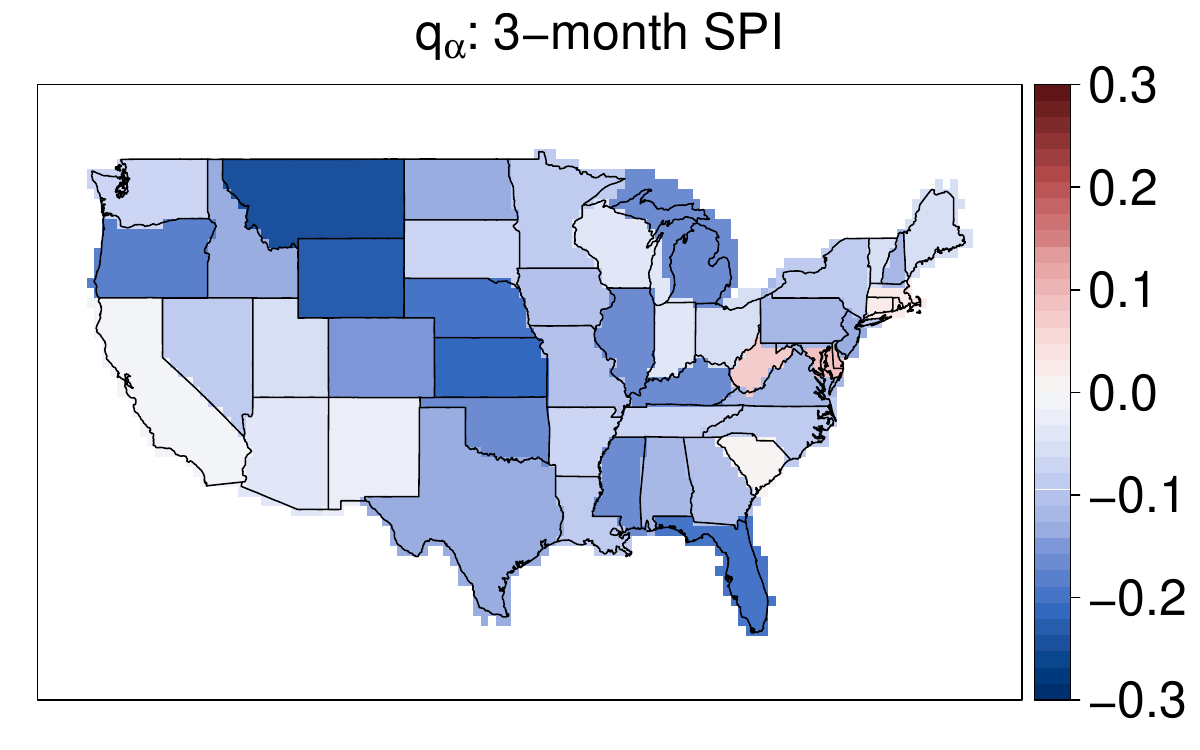} 
\end{minipage}
\begin{minipage}{0.49\linewidth}
\flushleft
\hspace{0.02cm}
\includegraphics[width=0.9\linewidth]{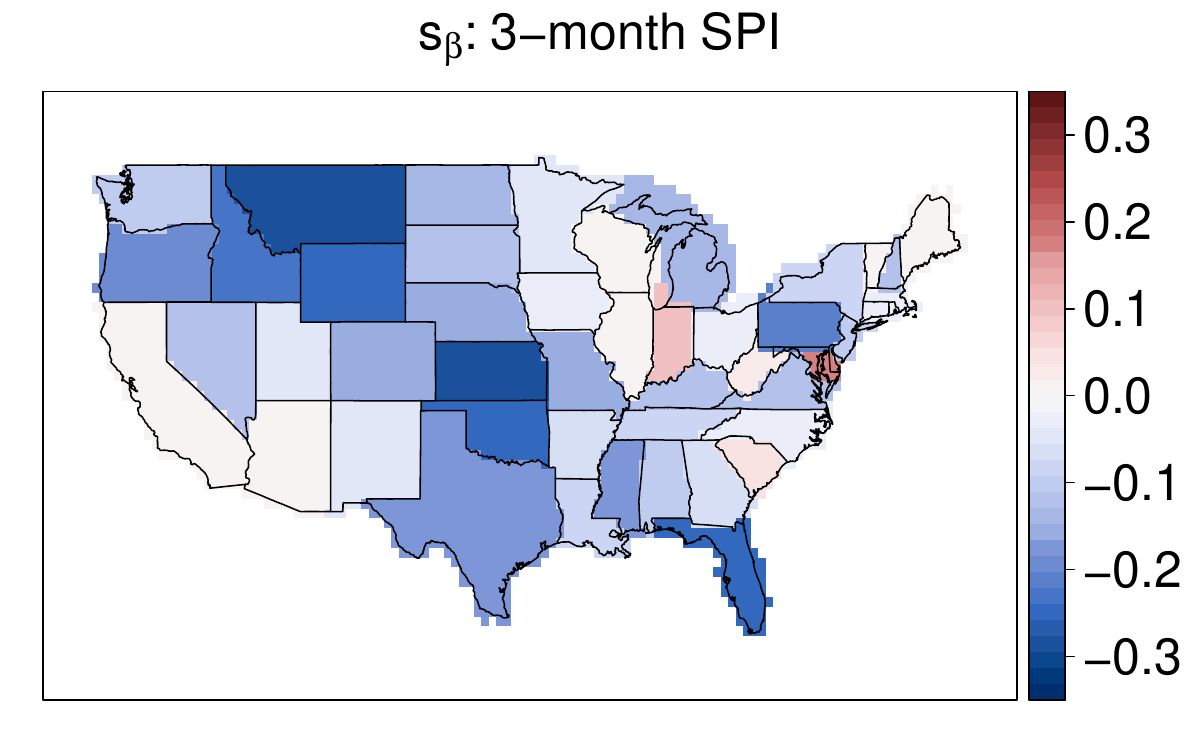} 
\end{minipage}
\vspace{-0.3cm}
\begin{minipage}{0.49\linewidth}
\flushleft
\hspace{0.025cm}
\includegraphics[width=0.9\linewidth]{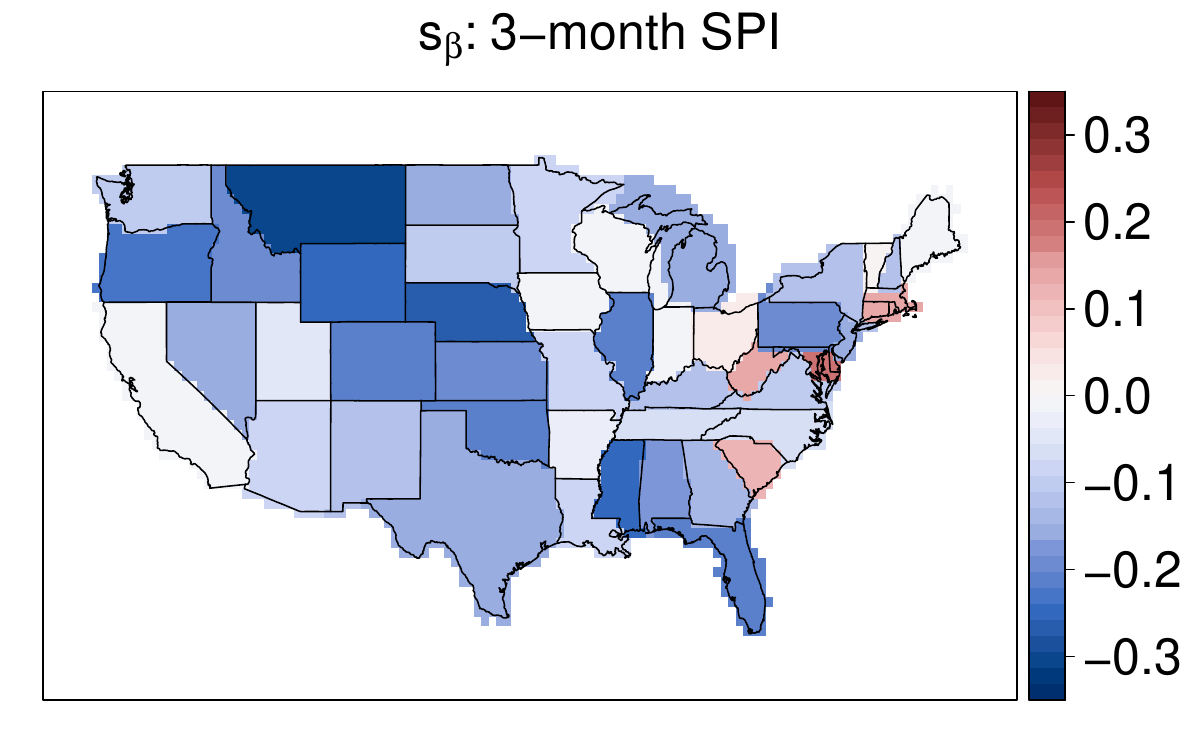} 
\end{minipage}
\vspace{-0.2cm}
\caption{Investigation of the sensitivity of $m_\mathcal{I}(\cdot)$ estimates to the choice of $\tau$ for the local bGEV-PP PINN model. Maps of the estimated $m_\mathcal{I}(\cdot)$ for the local bGEV-PP PINN model, where $m_\mathcal{I}(\cdot)$ is a linear model with state-varying coefficients. Mapped values correspond to the contribution of 2m air temperature (first two rows) and 3-month SPI (bottom two rows) to log-location $\log \{q_\alpha(s,t)\}$ (log-$\sqrt{\mbox{acres}}$; first and third rows) and log-scale $\log \{s_\beta(s,t)\}$ (log-$\sqrt{\mbox{acres}}$; second and fourth rows). Estimates in the left and right columns are derived using $\tau=0.7$ and $\tau=0.9$, respectively. Note that estimates are derived without bootstrap, i.e., using the original data only, and using the same model formulation and neural network architecture as described in Section~\ref{sec:arch} of the main text. We observe broadly similar spatial patterns in estimates, suggesting that our results are robust to the choice of $\tau$.}
\label{fig:SVC_sensitivity}
\end{figure}
\begin{figure*}[t]
\begin{minipage}{\linewidth}
\centering
\includegraphics[width=0.49\linewidth]{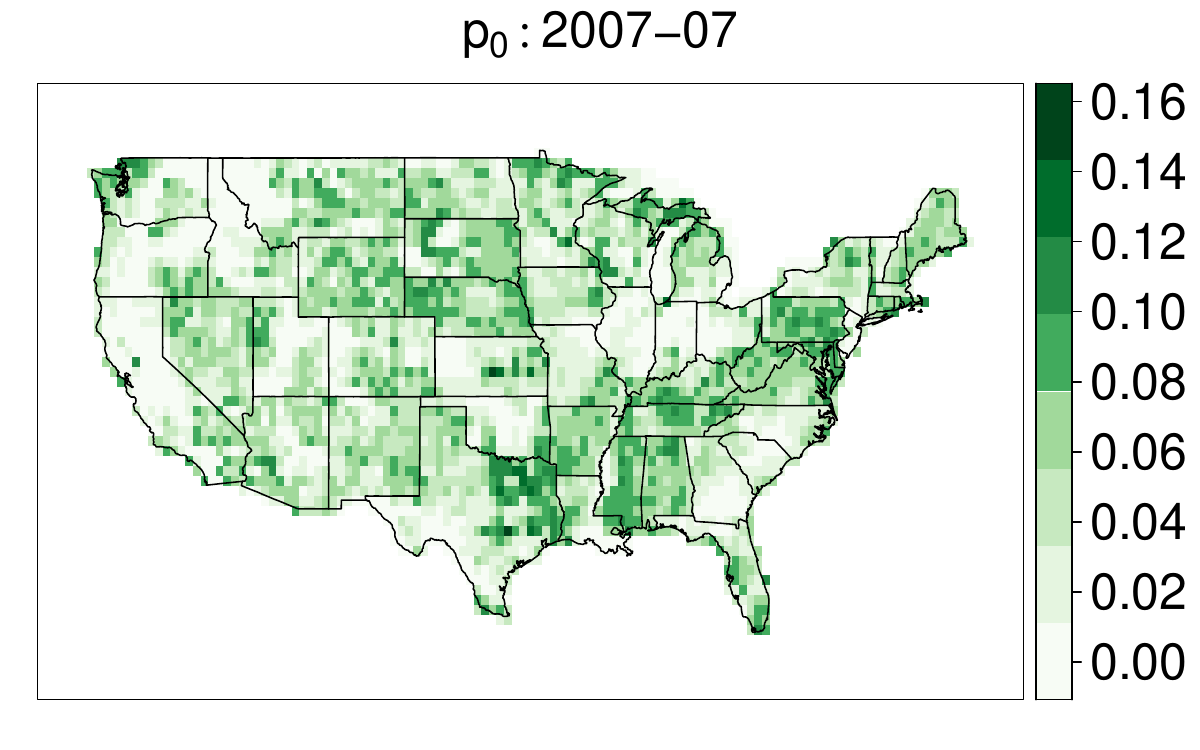} 
\vspace{-0.2cm}
\end{minipage}
\begin{minipage}{0.49\linewidth}
\flushleft
\includegraphics[width=0.9\linewidth]{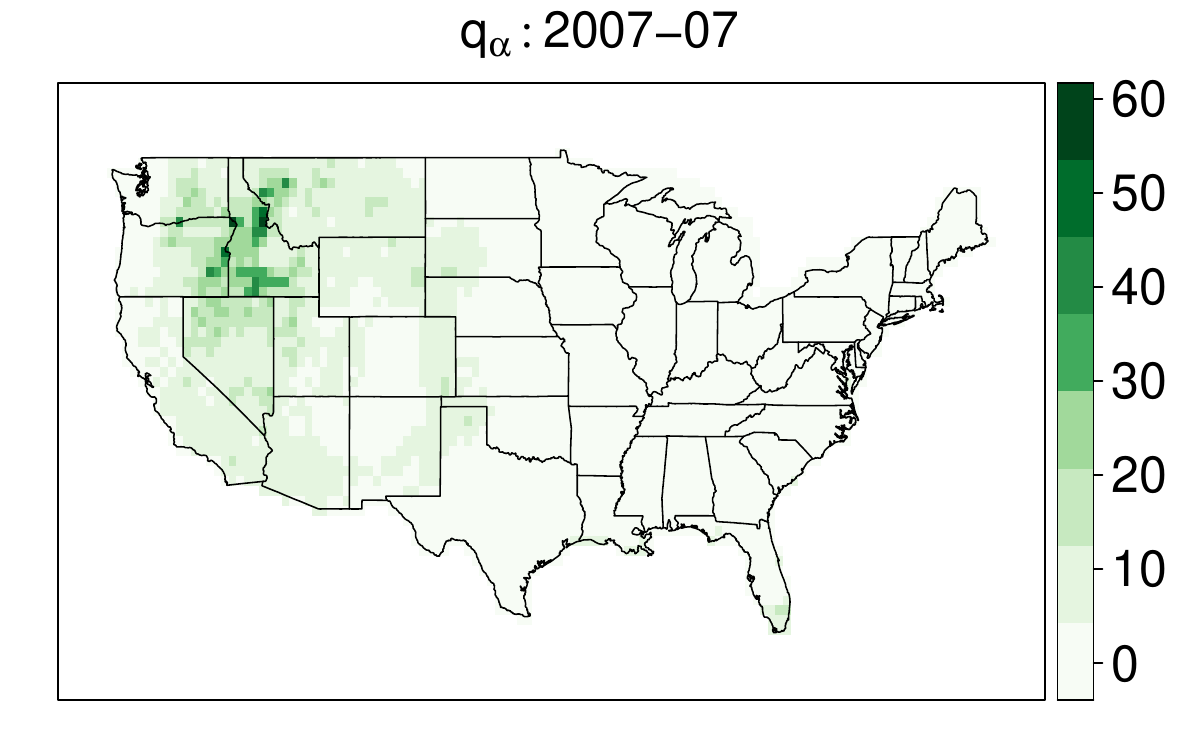} 
\end{minipage}
\vspace{-0.2cm}
\begin{minipage}{0.49\linewidth}
\flushleft
\includegraphics[width=0.9\linewidth]{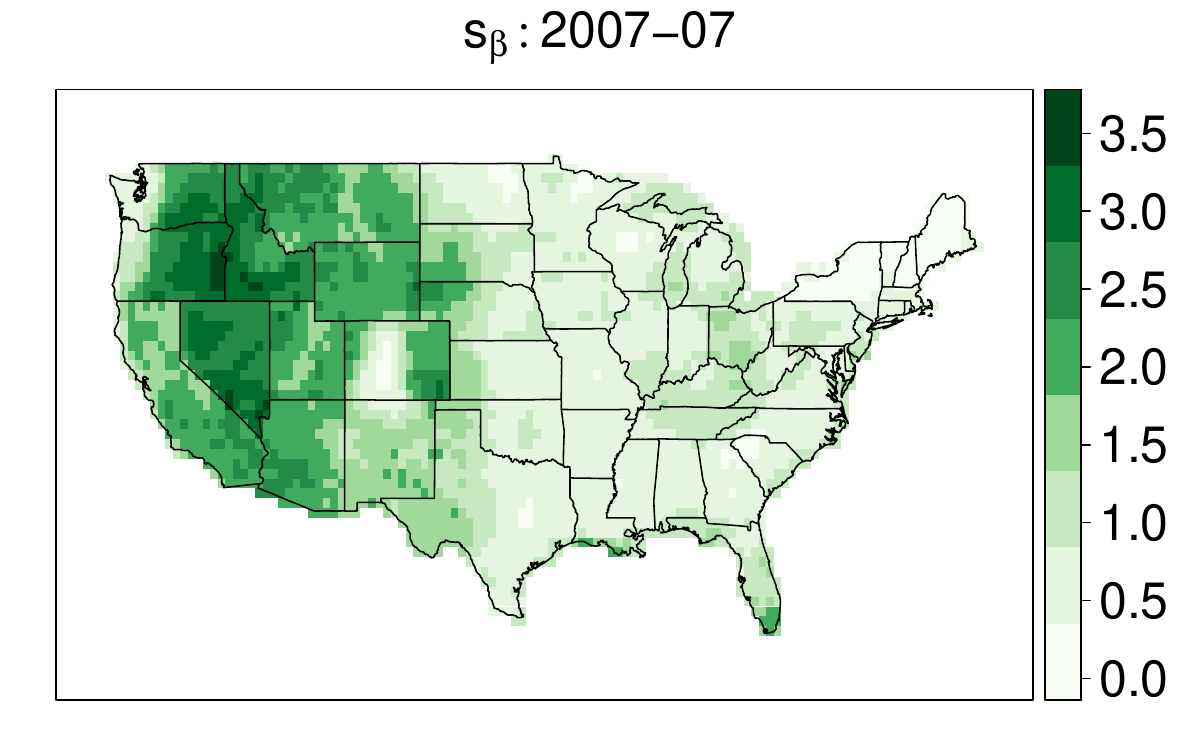} 
\end{minipage}
\begin{minipage}{0.49\linewidth}
\flushleft
\hspace{0.04cm}
\includegraphics[width=0.9\linewidth]{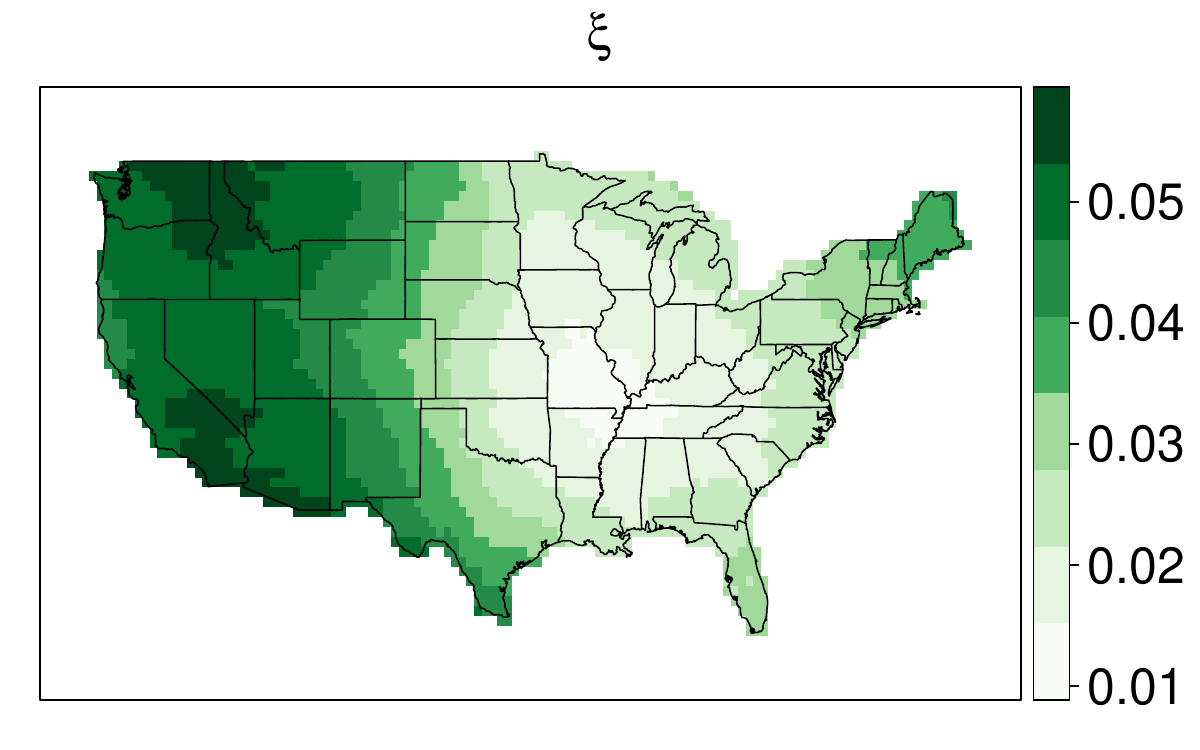} 
\end{minipage}
\begin{minipage}{0.49\linewidth}
\flushleft
\hspace{0.04cm}
\includegraphics[width=0.9\linewidth]{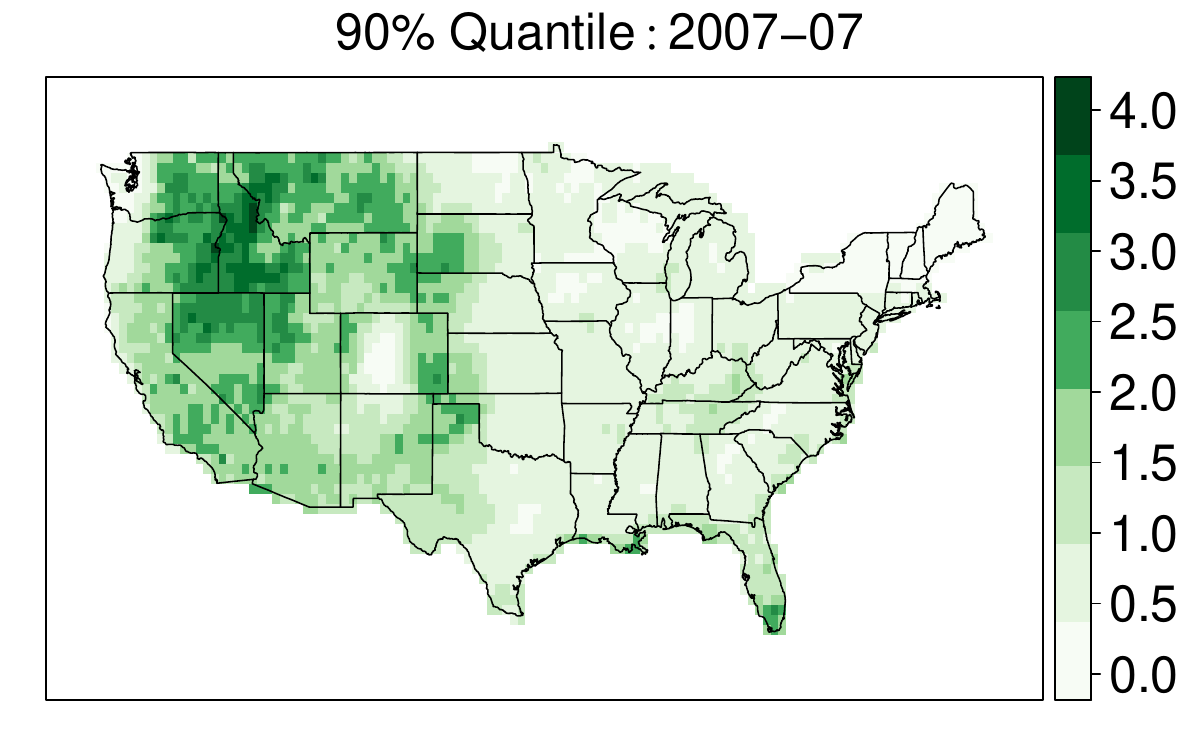} 
\end{minipage}
\vspace{-.3cm}
\caption{Maps of the inter-quartile range for estimated $p_0(s,t):=\Pr\{Y(s,t)>0\mid\mathbf{X}(s,t)\}$ (unitless; top), $q_\alpha(s,t)$ ($\sqrt{\mbox{acres}}$; centre-left), $\log \{1+s_\beta(s,t)\}$ (log-$\sqrt{\mbox{acres}}$; centre-right), $\xi(s)$ (unitless; bottom-left), and $90\%$ quantile of $\log\{1+ \sqrt{Y}(s,t)\}\mid\{Y(s,t)>0,\mathbf{X}(s,t)\}$ (log-$\sqrt{\mbox{acres}}$; bottom-right) for July, 2007. }
\label{spread_map_IQR}
\end{figure*}

\begin{figure}[t!]
\centering
\begin{minipage}{0.32\linewidth}
\flushleft
\includegraphics[width=\linewidth]{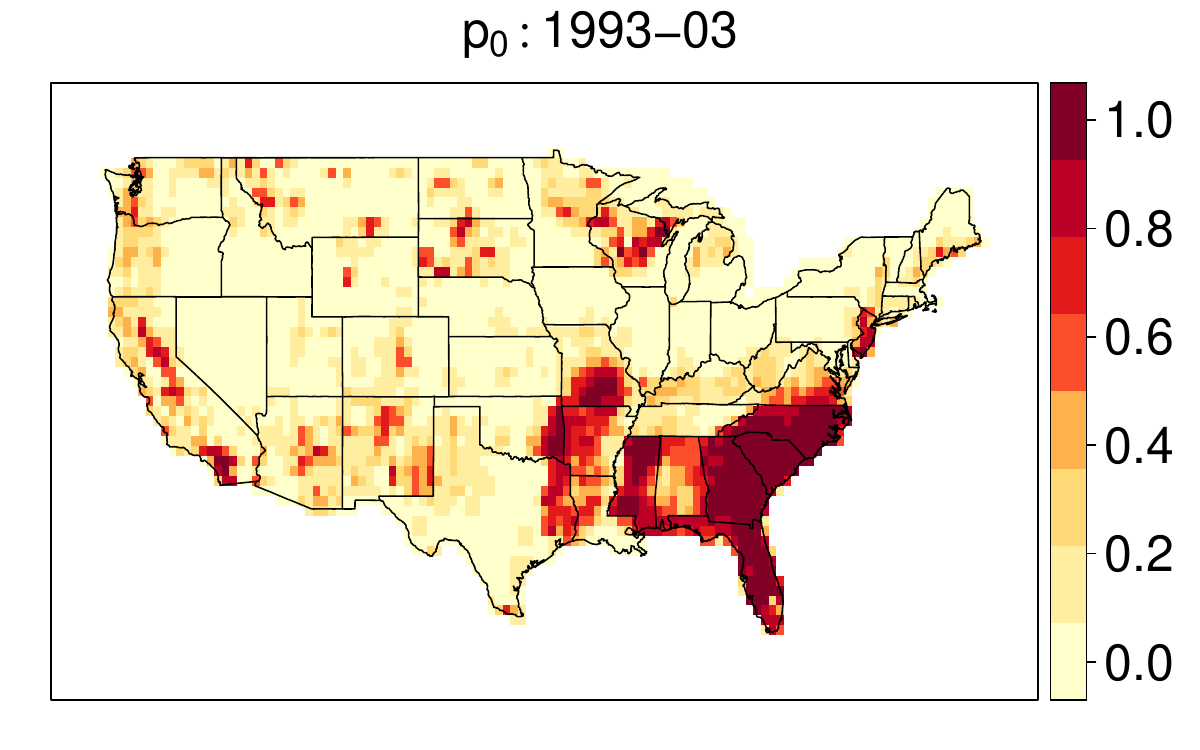} 
\end{minipage}
\begin{minipage}{0.32\linewidth}
\flushleft
\includegraphics[width=\linewidth]{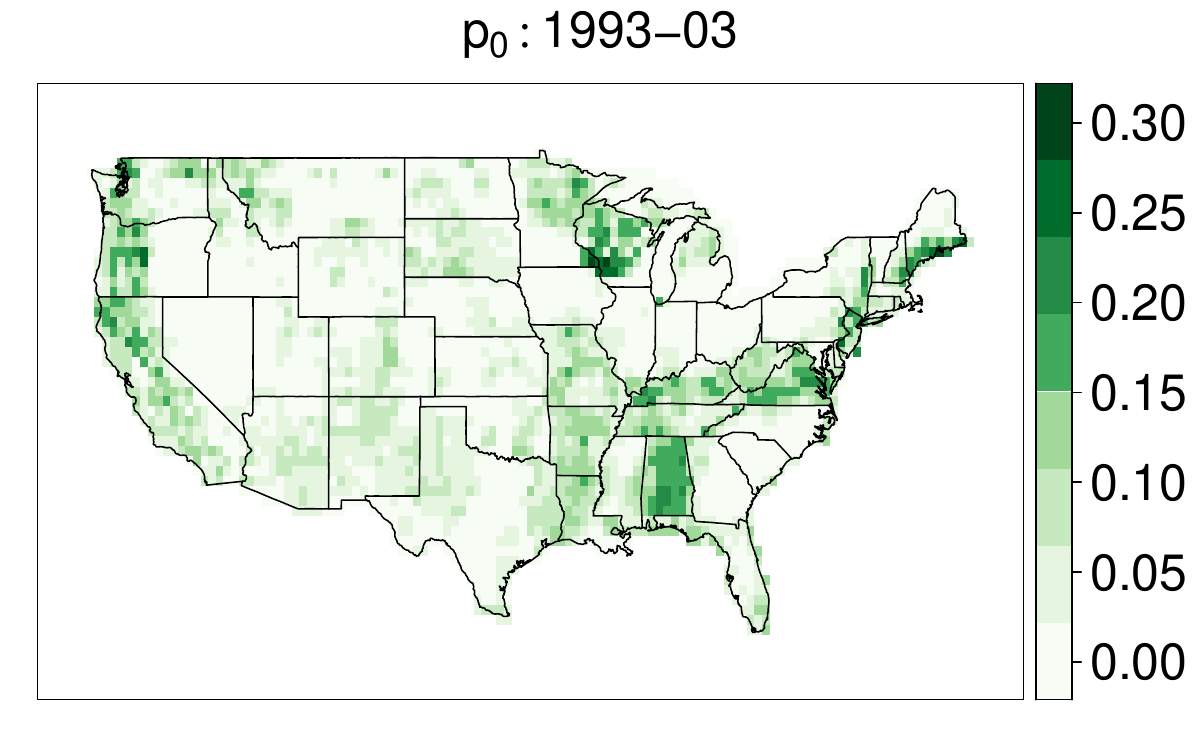} 
\end{minipage}
\vspace{-.2cm}
\begin{minipage}{0.32\linewidth}
\flushleft
\includegraphics[width=\linewidth]{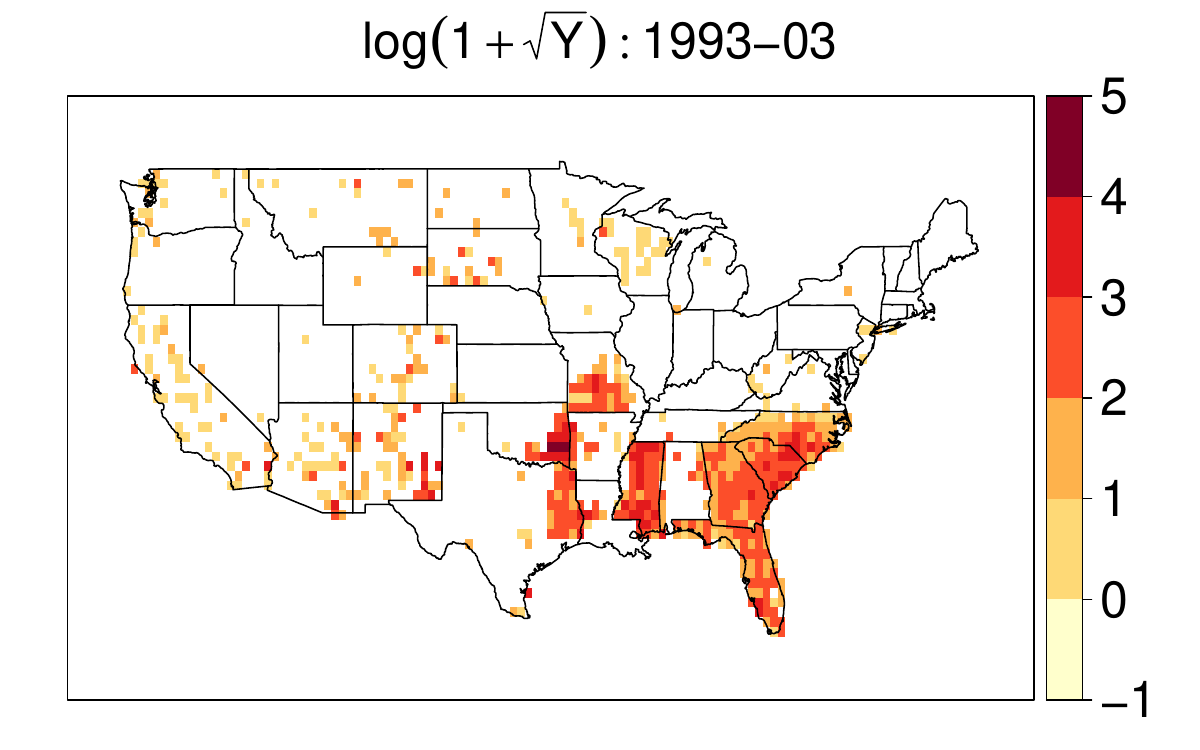} 
\end{minipage}
\begin{minipage}{0.32\linewidth}
\flushleft
\includegraphics[width=\linewidth]{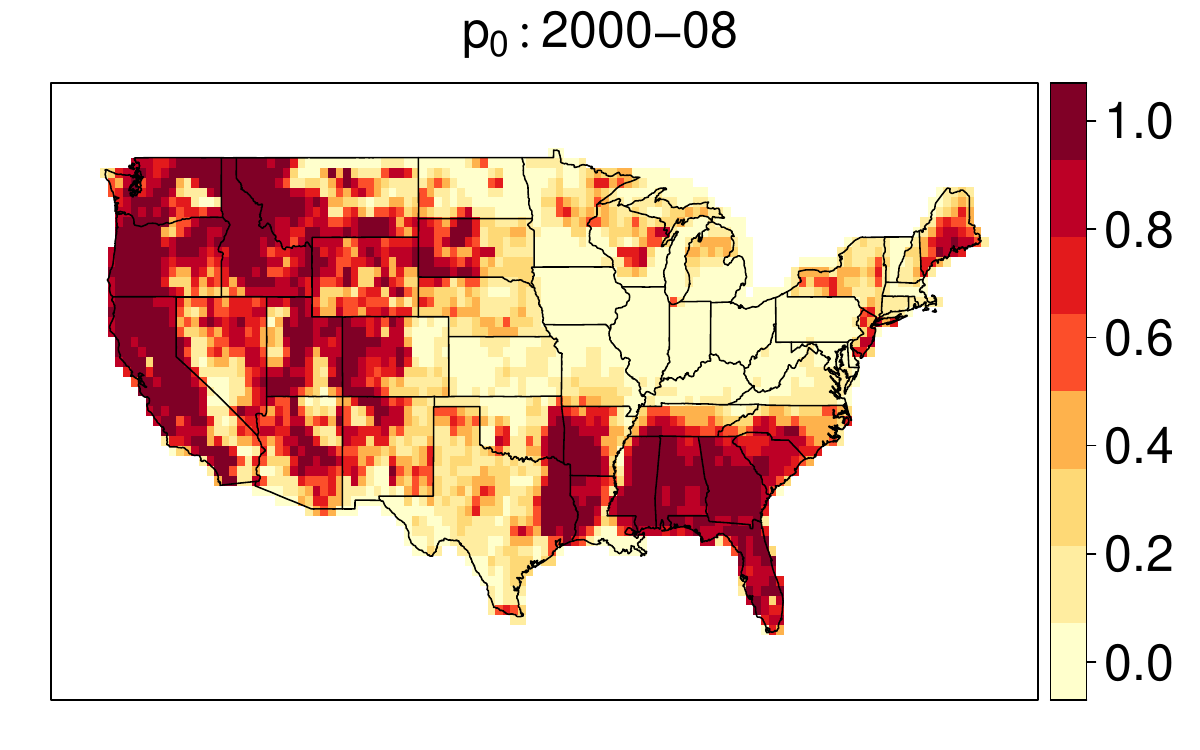} 
\end{minipage}
\begin{minipage}{0.32\linewidth}
\flushleft
\includegraphics[width=\linewidth]{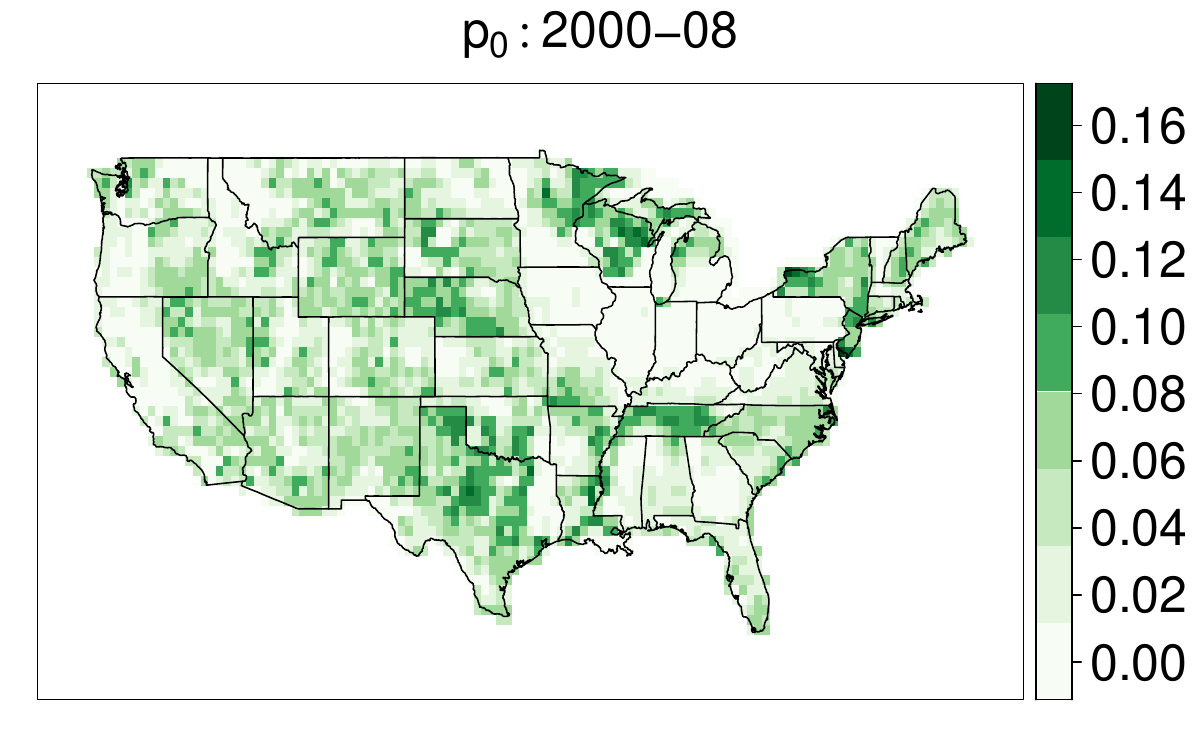} 
\end{minipage}
\vspace{-.2cm}\begin{minipage}{0.32\linewidth}
\flushleft
\includegraphics[width=\linewidth]{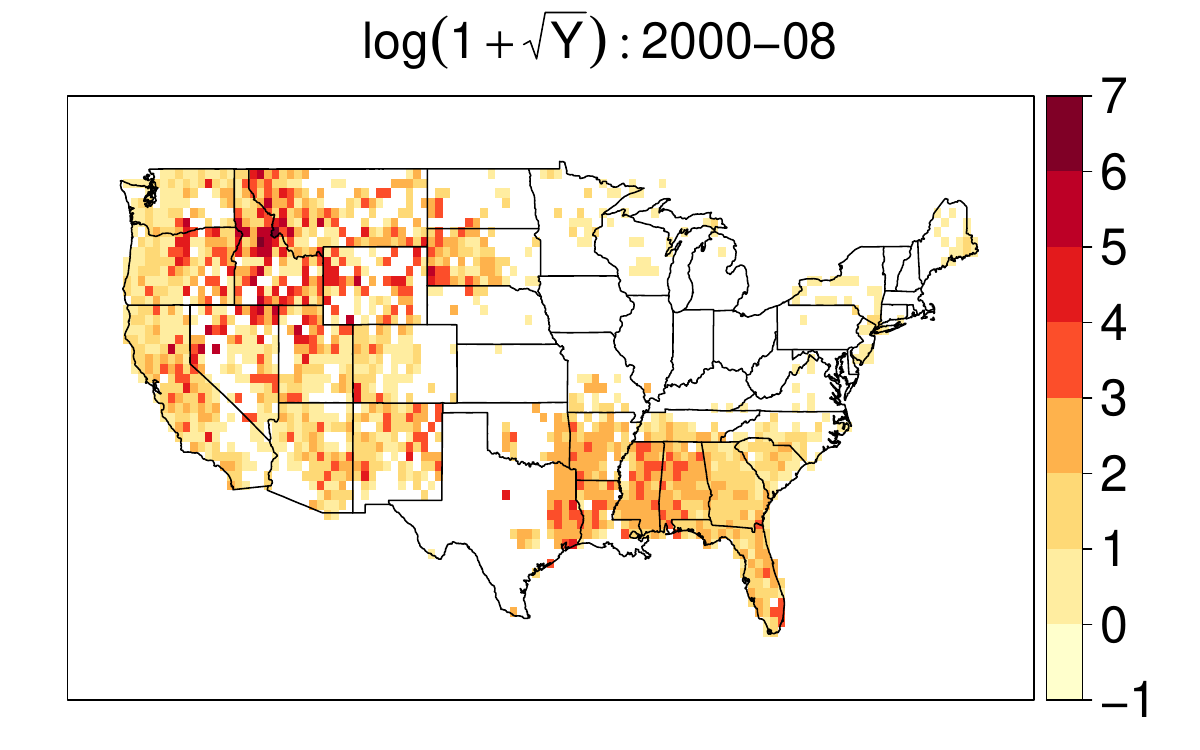} 
\end{minipage}
\begin{minipage}{0.32\linewidth}
\flushleft
\includegraphics[width=\linewidth]{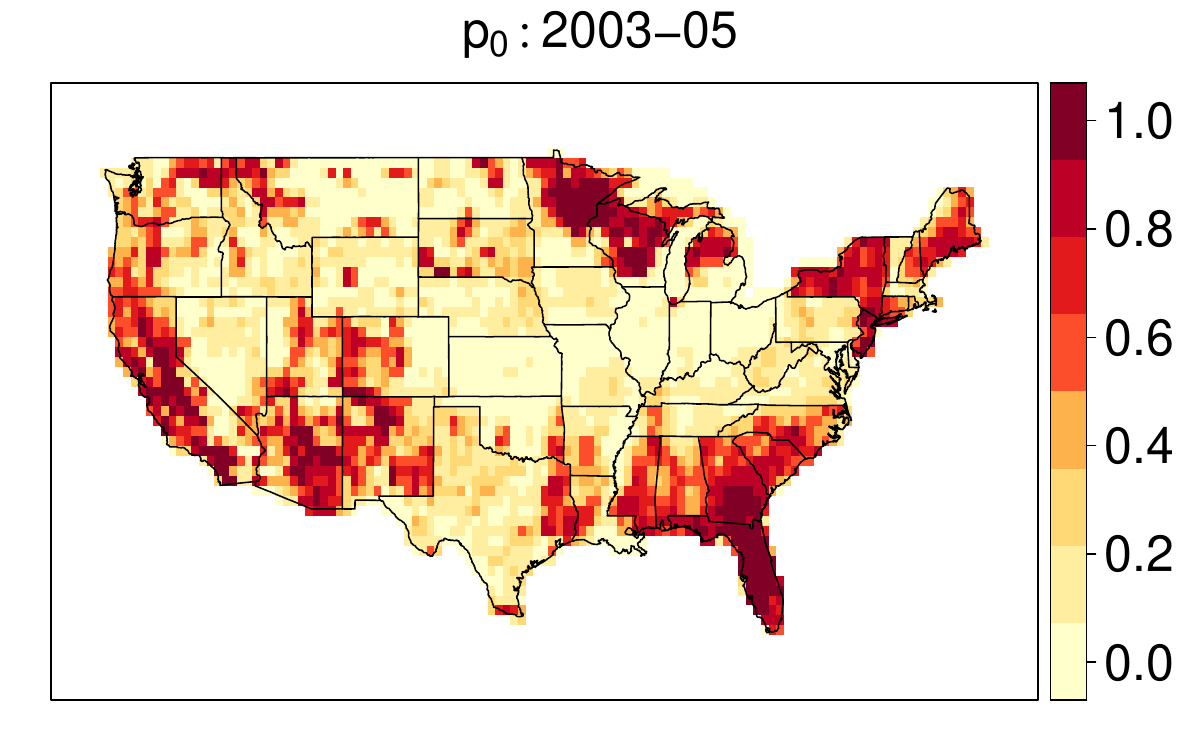} 
\end{minipage}
\begin{minipage}{0.32\linewidth}
\flushleft
\includegraphics[width=\linewidth]{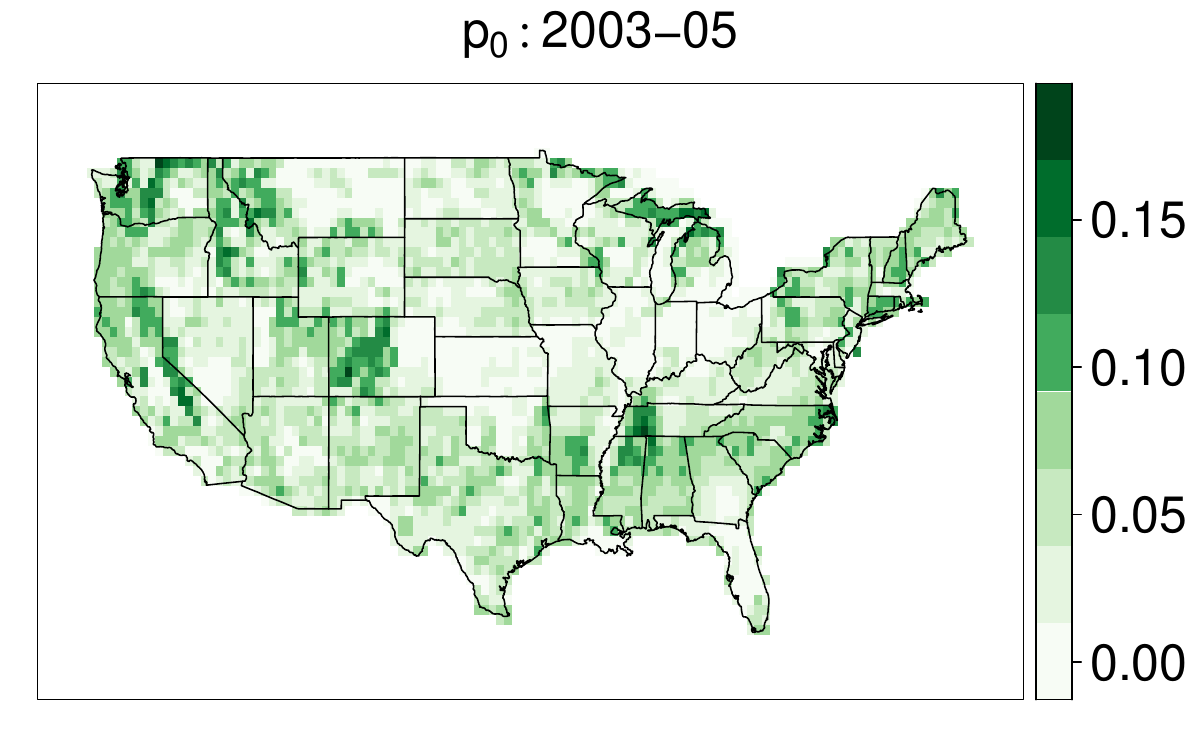} 
\end{minipage}
\vspace{-.2cm}
\begin{minipage}{0.32\linewidth}
\centering
\includegraphics[width=\linewidth]{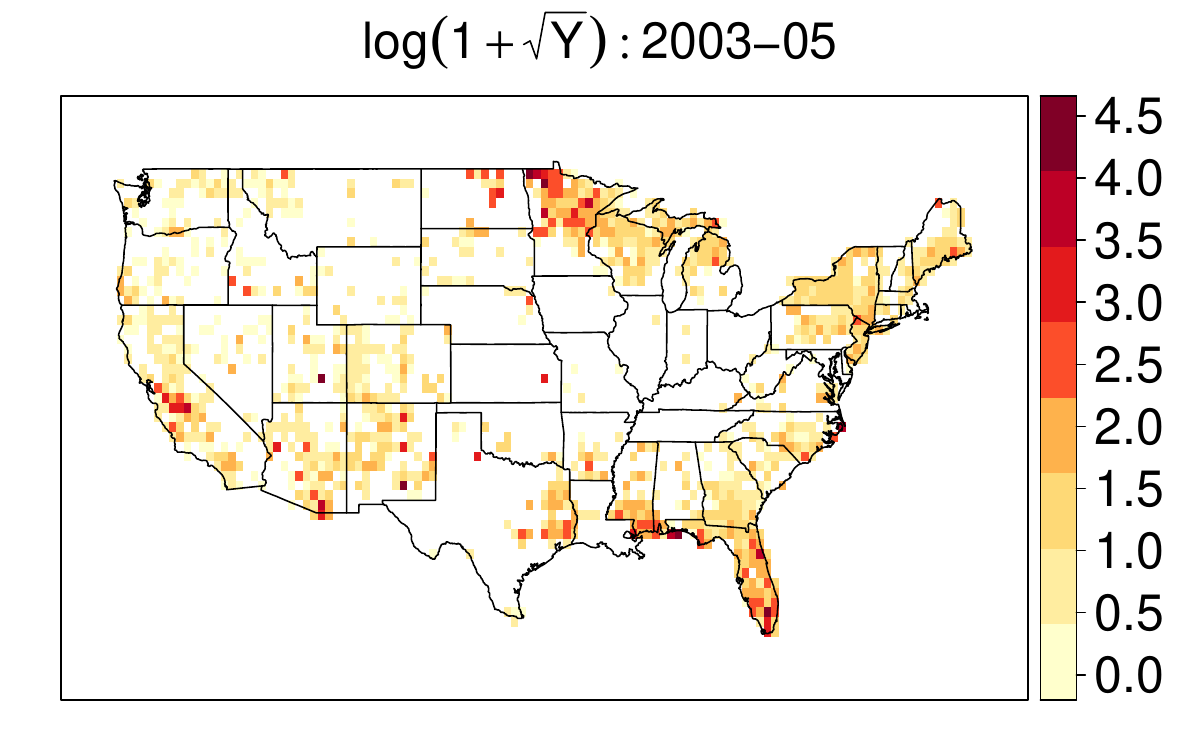} 
\end{minipage}
\vspace{-.2cm}
\begin{minipage}{0.32\linewidth}
\flushleft
\includegraphics[width=\linewidth]{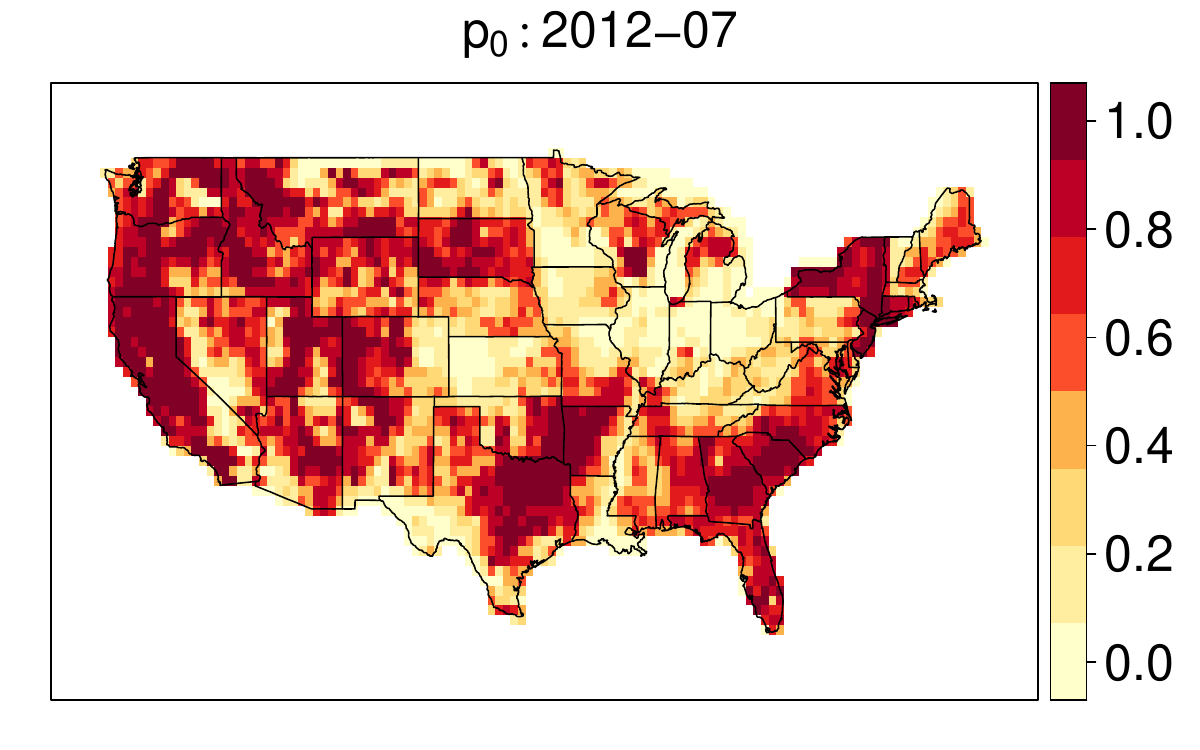} 
\end{minipage}
\begin{minipage}{0.32\linewidth}
\flushleft
\includegraphics[width=\linewidth]{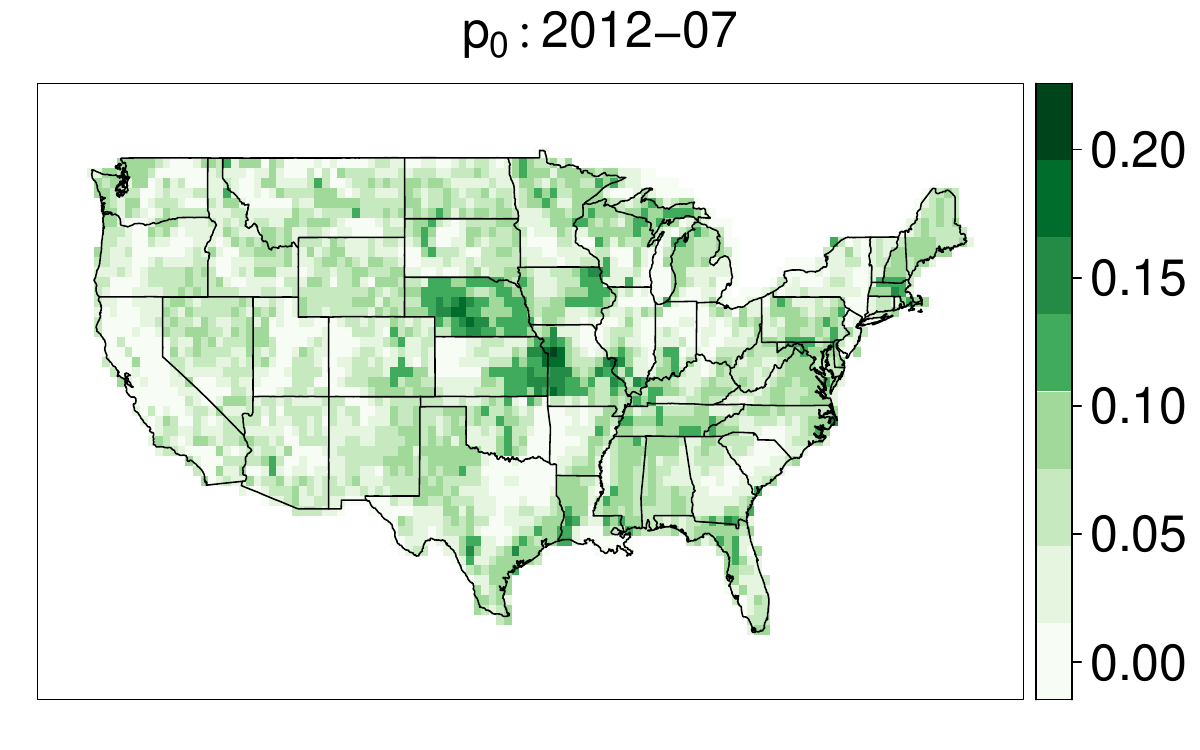} 
\end{minipage}
\begin{minipage}{0.32\linewidth}
\flushleft
\includegraphics[width=\linewidth]{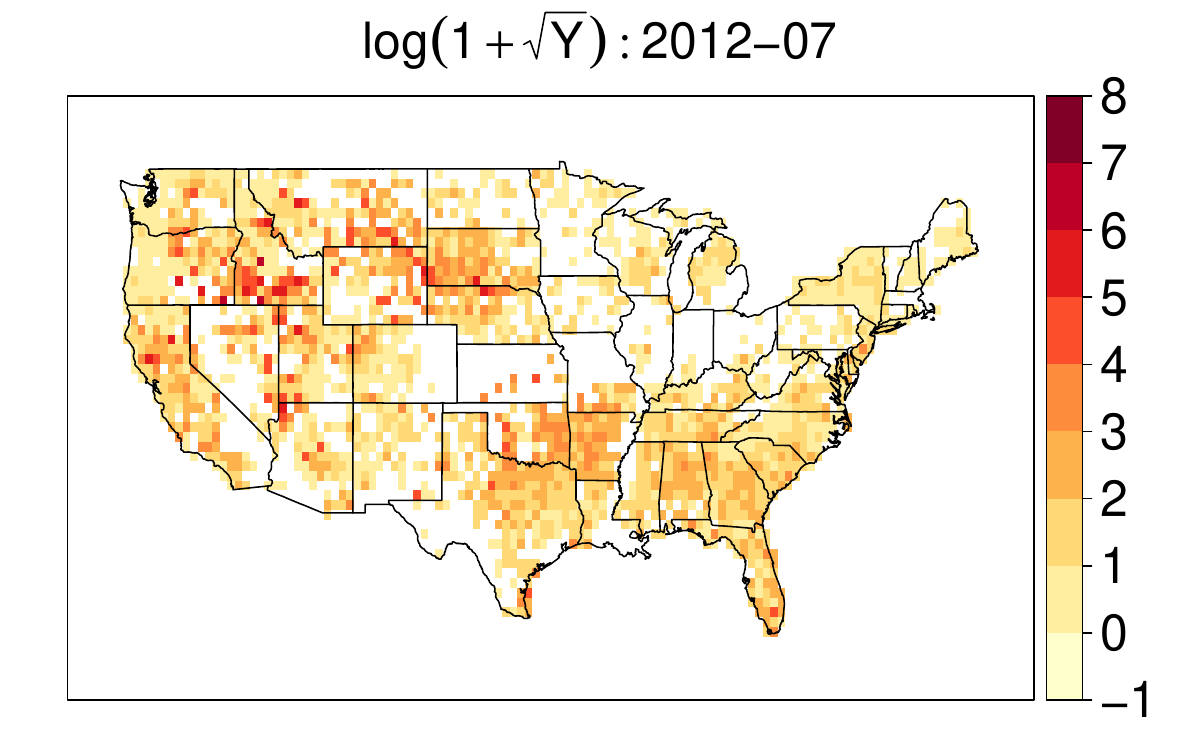} 
\end{minipage}
\begin{minipage}{0.32\linewidth}
\flushleft
\includegraphics[width=\linewidth]{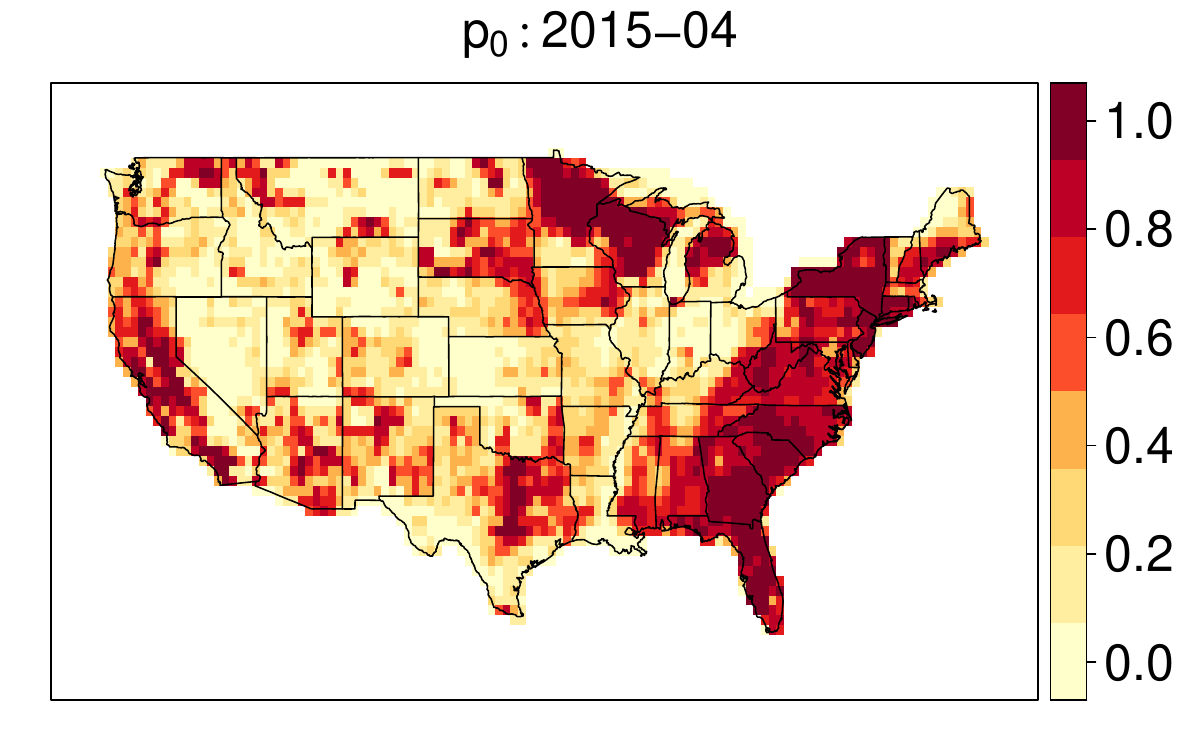} 
\end{minipage}
\begin{minipage}{0.32\linewidth}
\flushleft
\includegraphics[width=\linewidth]{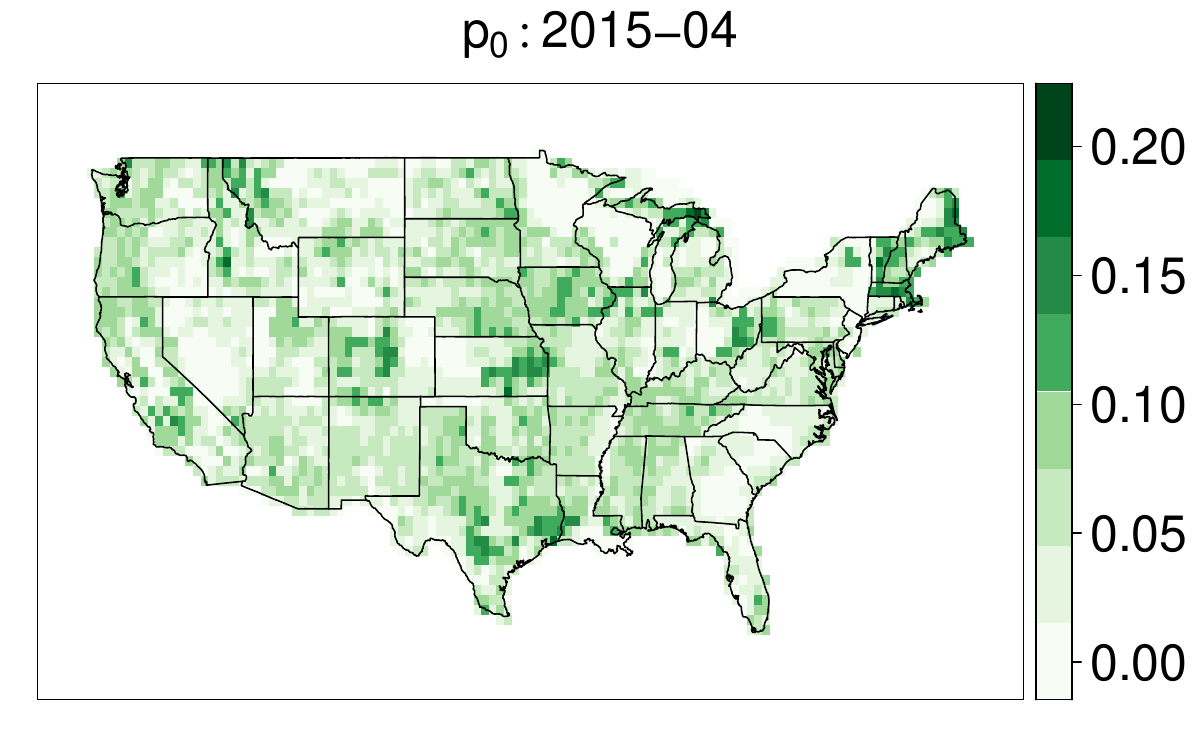} 
\end{minipage}
\begin{minipage}{0.32\linewidth}
\flushleft
\includegraphics[width=\linewidth]{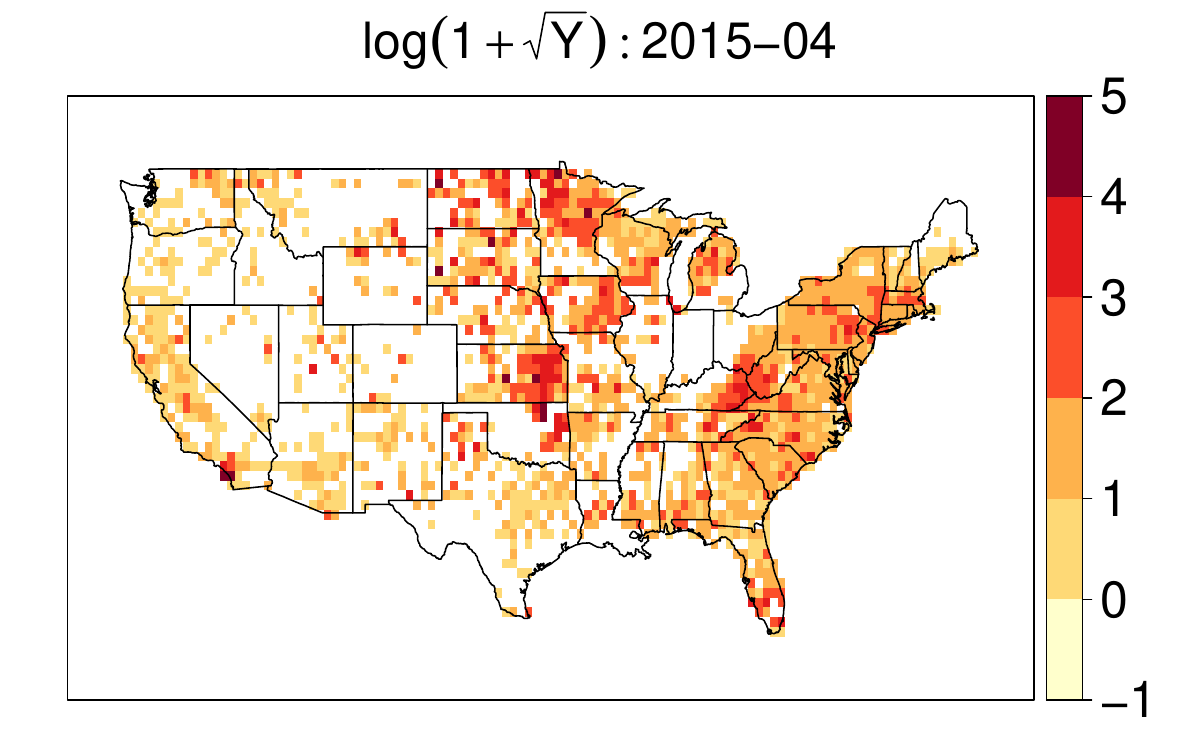} 
\end{minipage}
\vspace{-.3cm}
\caption{Maps of $p_0(s,t):=\Pr\{Y(s,t)>0\mid\mathbf{X}(s,t)\}$ for different times $t$ (unitless; by row). Left and middle columns: estimated median and inter-quartile range, respectively,  across all bootstrap samples. Right-column: observed $\log\{1+\sqrt{Y}(s,t)\}$ (log-$\sqrt{\mbox{acres}}$). Times, top to bottom row: March 1993, August 2000,  May 2003, July 2012, April 2015.}
\label{occur_map_sup}
\end{figure}

\begin{figure}[h!]
\centering
\begin{minipage}{0.32\linewidth}
\flushleft
\includegraphics[width=\linewidth]{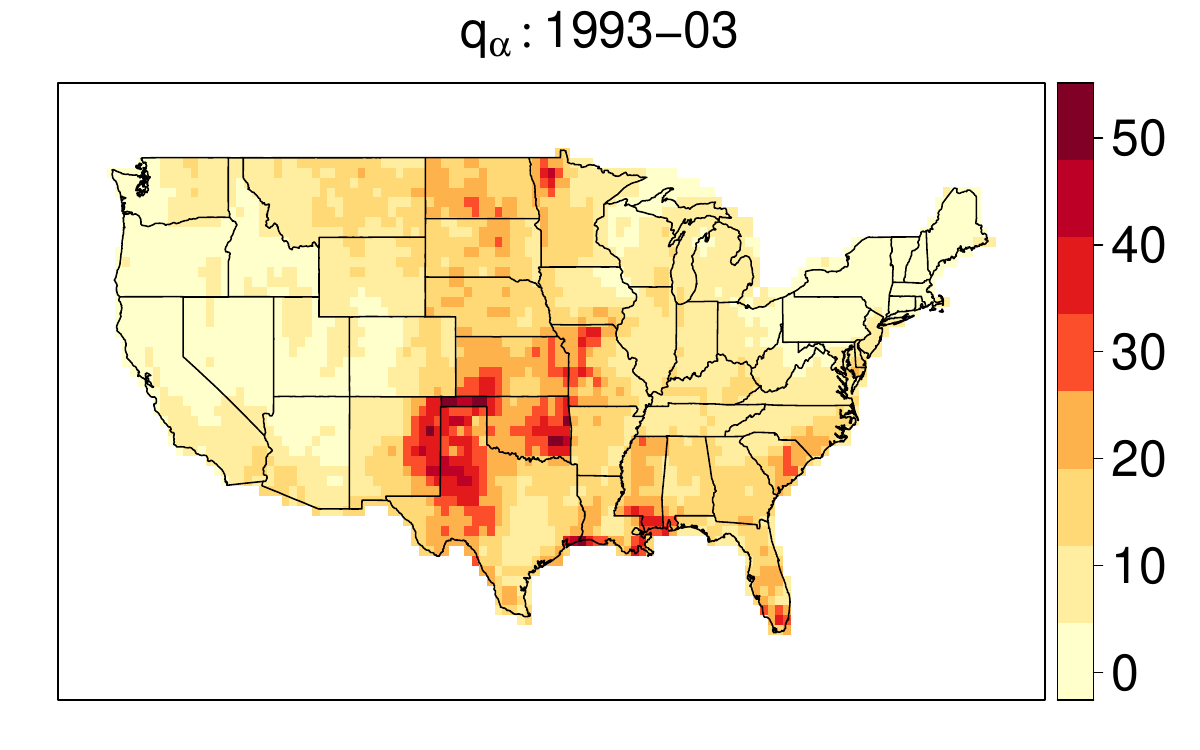} 
\end{minipage}
\begin{minipage}{0.32\linewidth}
\flushleft
\includegraphics[width=\linewidth]{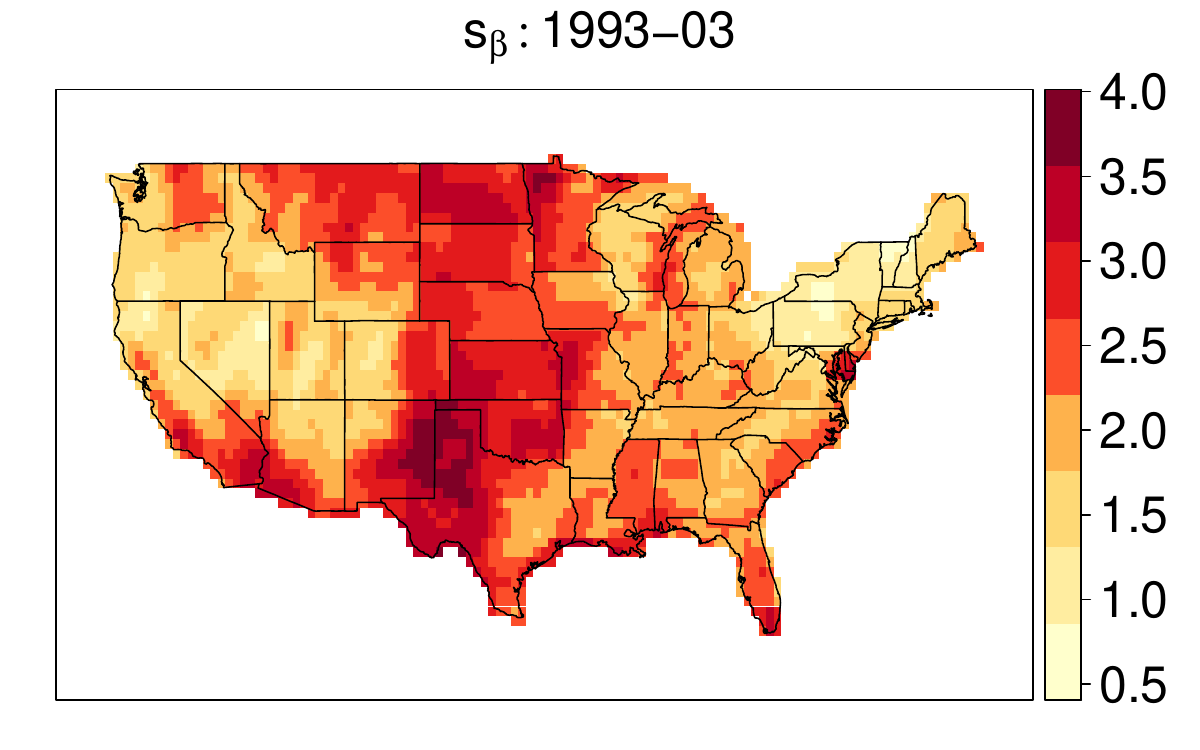} 
\end{minipage}
\vspace{-.2cm}
\begin{minipage}{0.32\linewidth}
\flushleft
\includegraphics[width=\linewidth]{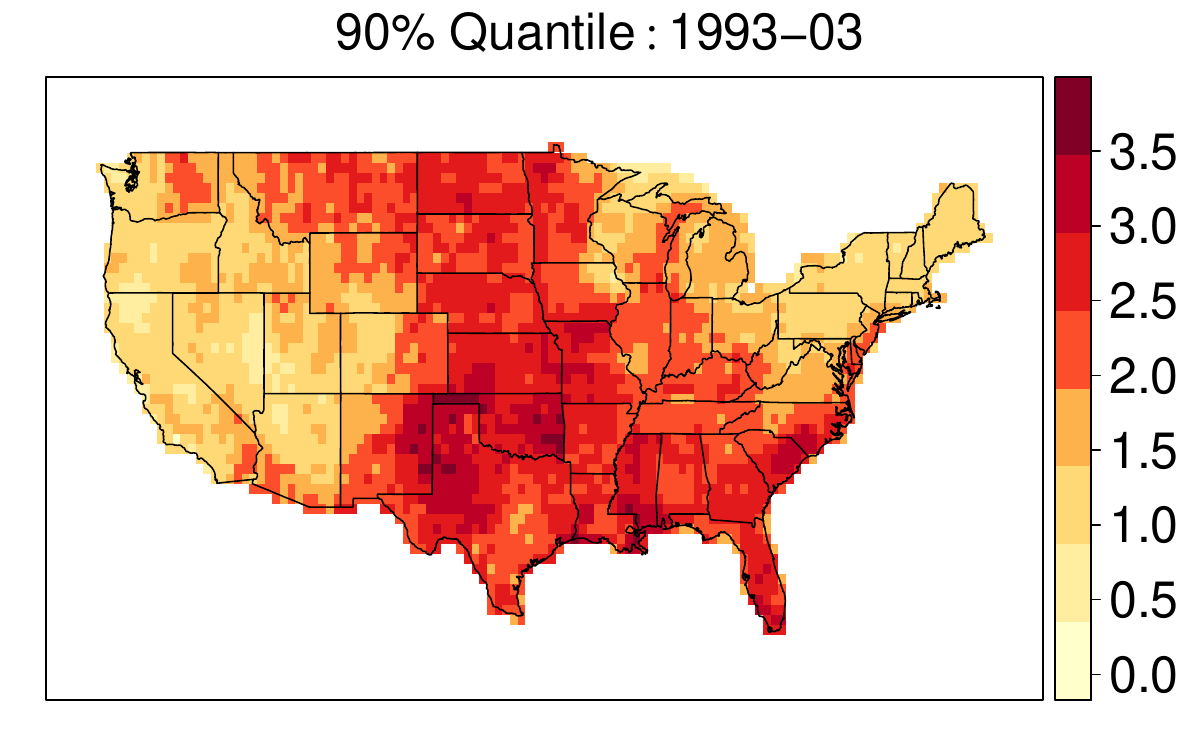} 
\end{minipage}
\begin{minipage}{0.32\linewidth}
\flushleft
\includegraphics[width=\linewidth]{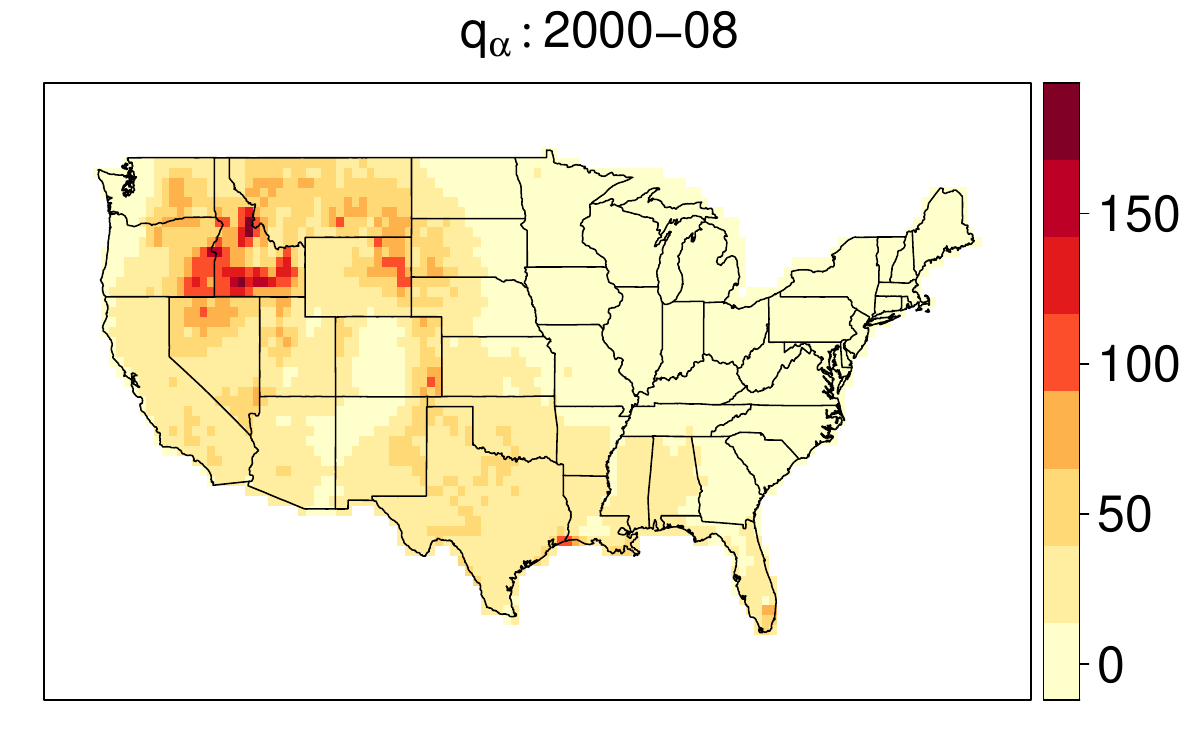} 
\end{minipage}
\begin{minipage}{0.32\linewidth}
\flushleft
\includegraphics[width=\linewidth]{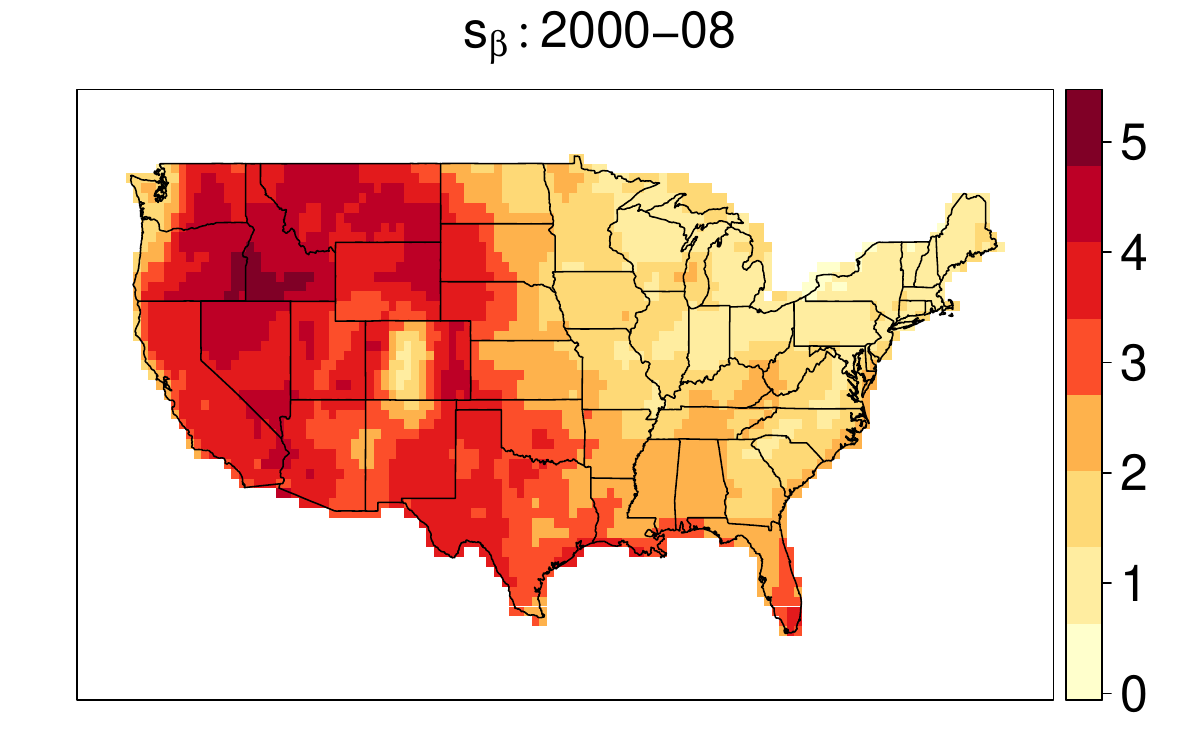} 
\end{minipage}
\vspace{-.2cm}
\begin{minipage}{0.32\linewidth}
\flushleft
\includegraphics[width=\linewidth]{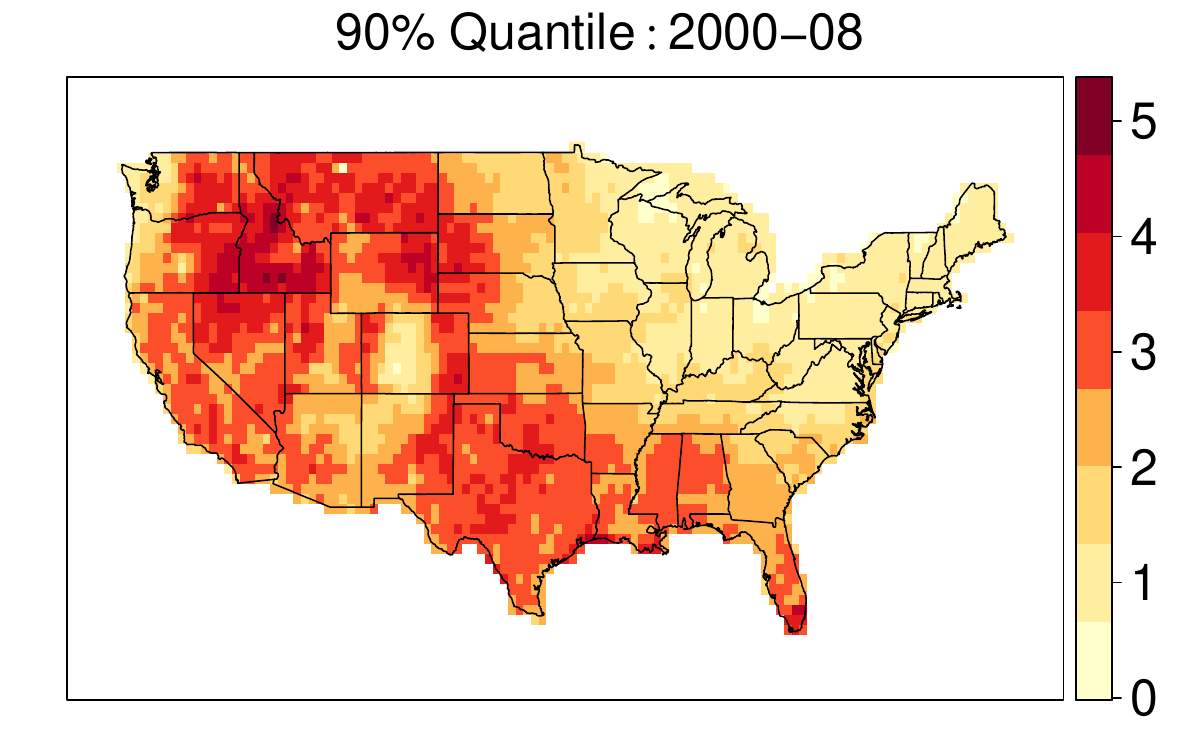} 
\end{minipage}
\vspace{-.2cm}
\begin{minipage}{0.32\linewidth}
\flushleft
\includegraphics[width=\linewidth]{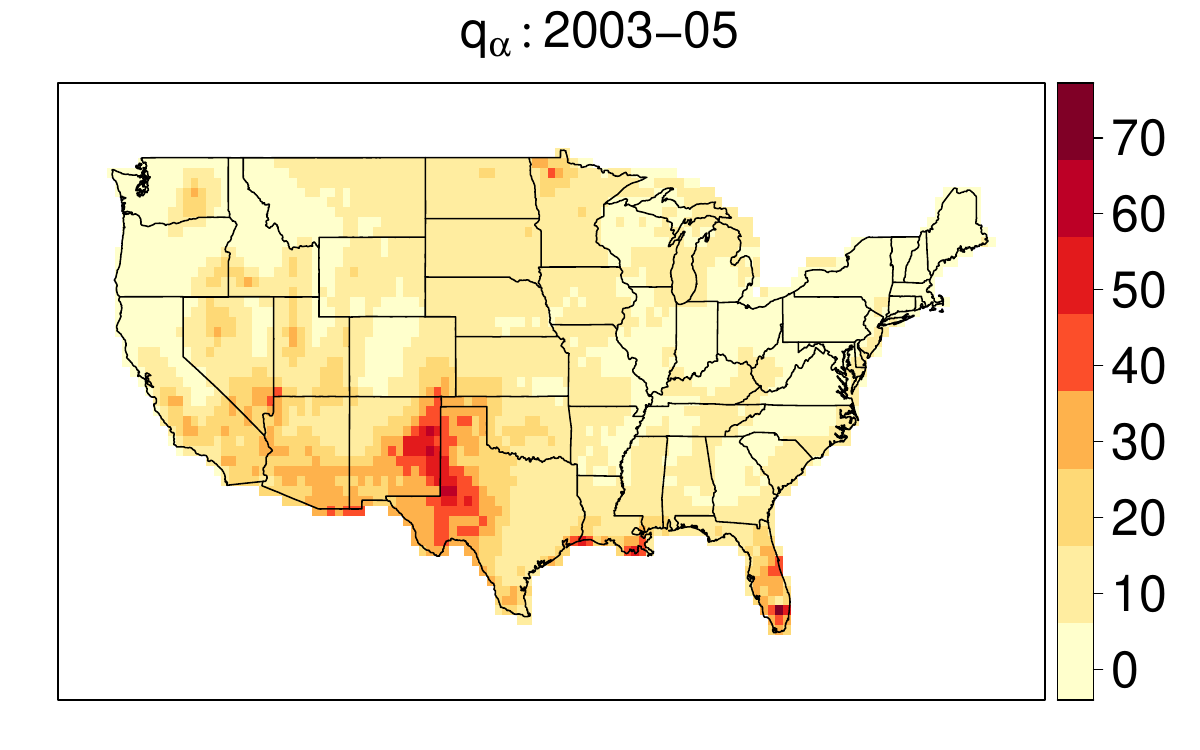} 
\end{minipage}
\begin{minipage}{0.32\linewidth}
\flushleft
\includegraphics[width=\linewidth]{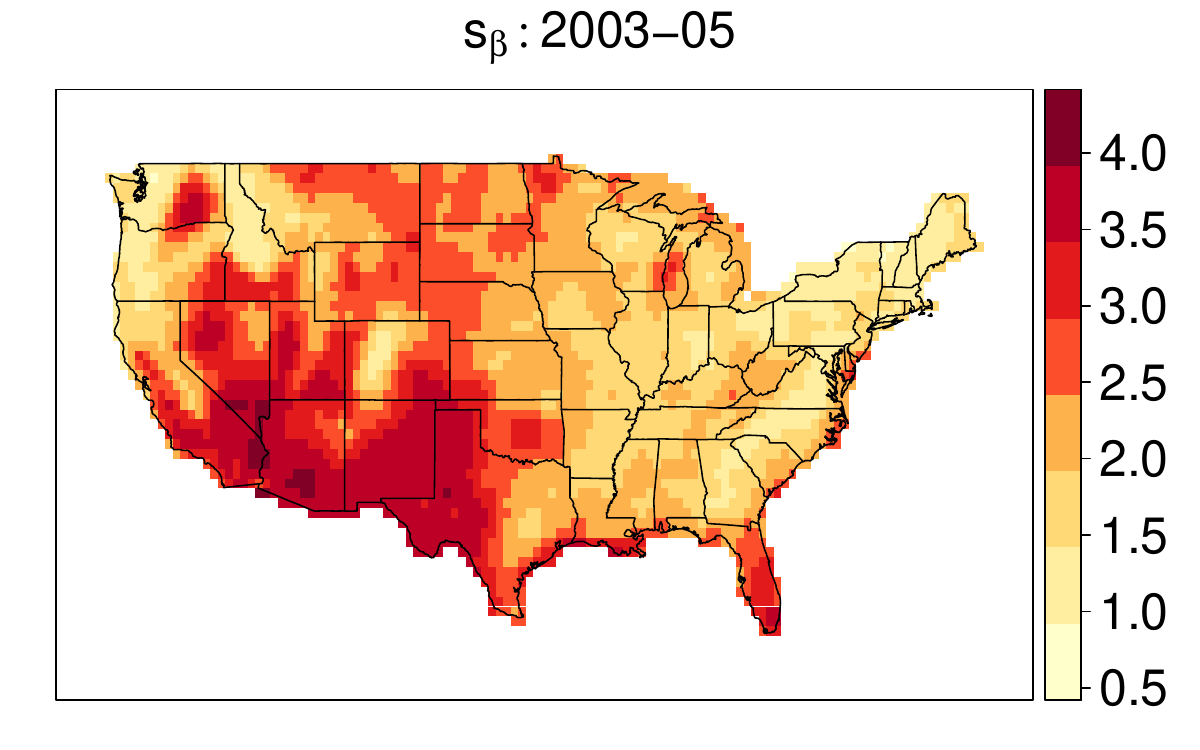} 
\end{minipage}
\begin{minipage}{0.32\linewidth}
\flushleft
\includegraphics[width=\linewidth]{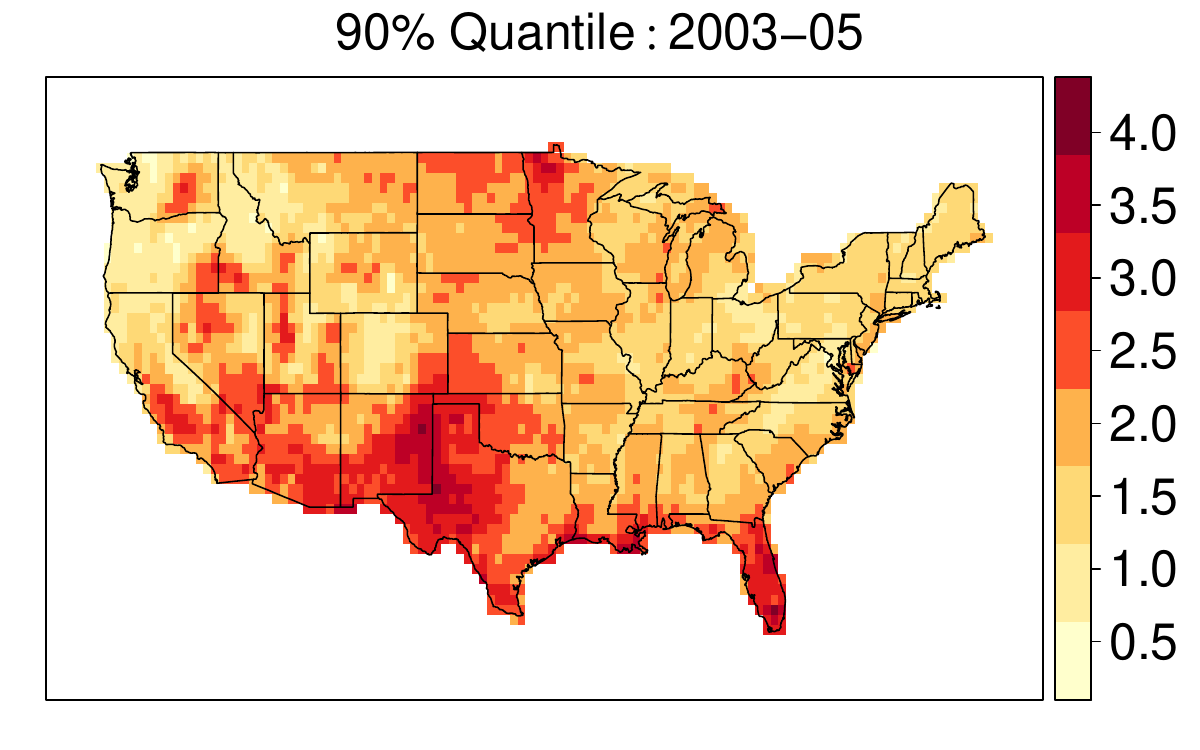} 
\end{minipage}
\vspace{-.2cm}
\begin{minipage}{0.32\linewidth}
\flushleft
\includegraphics[width=\linewidth]{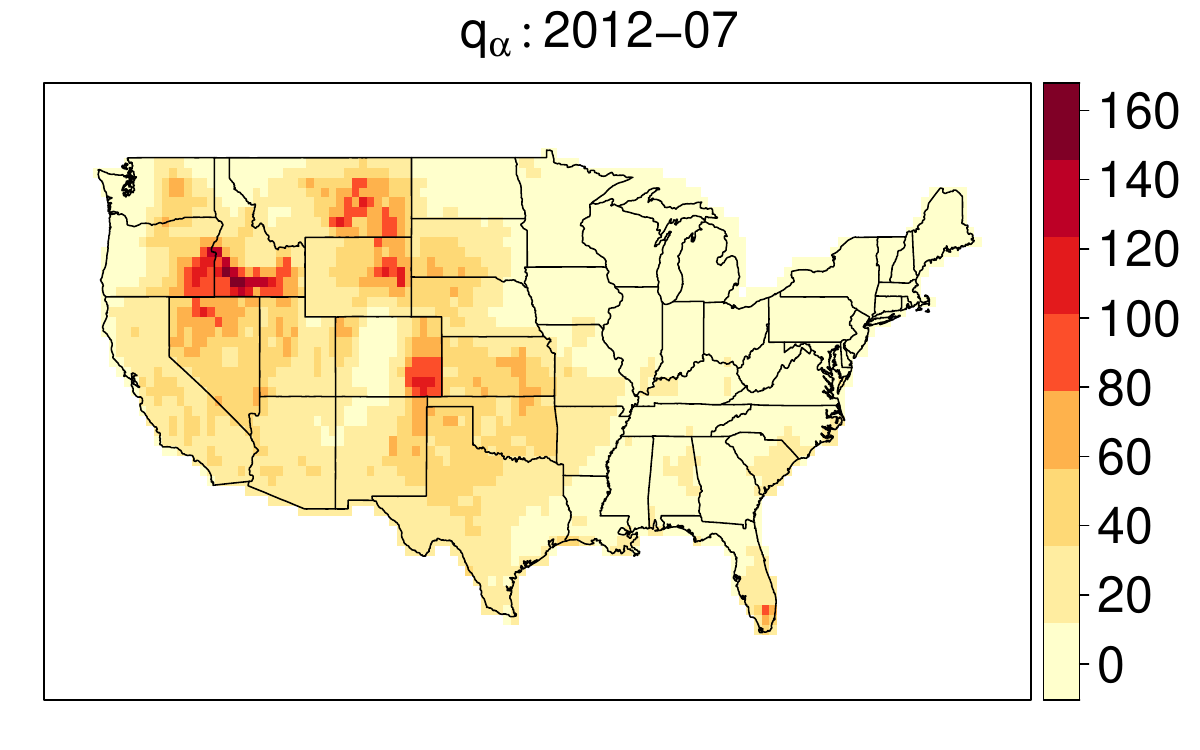} 
\end{minipage}
\begin{minipage}{0.32\linewidth}
\flushleft
\includegraphics[width=\linewidth]{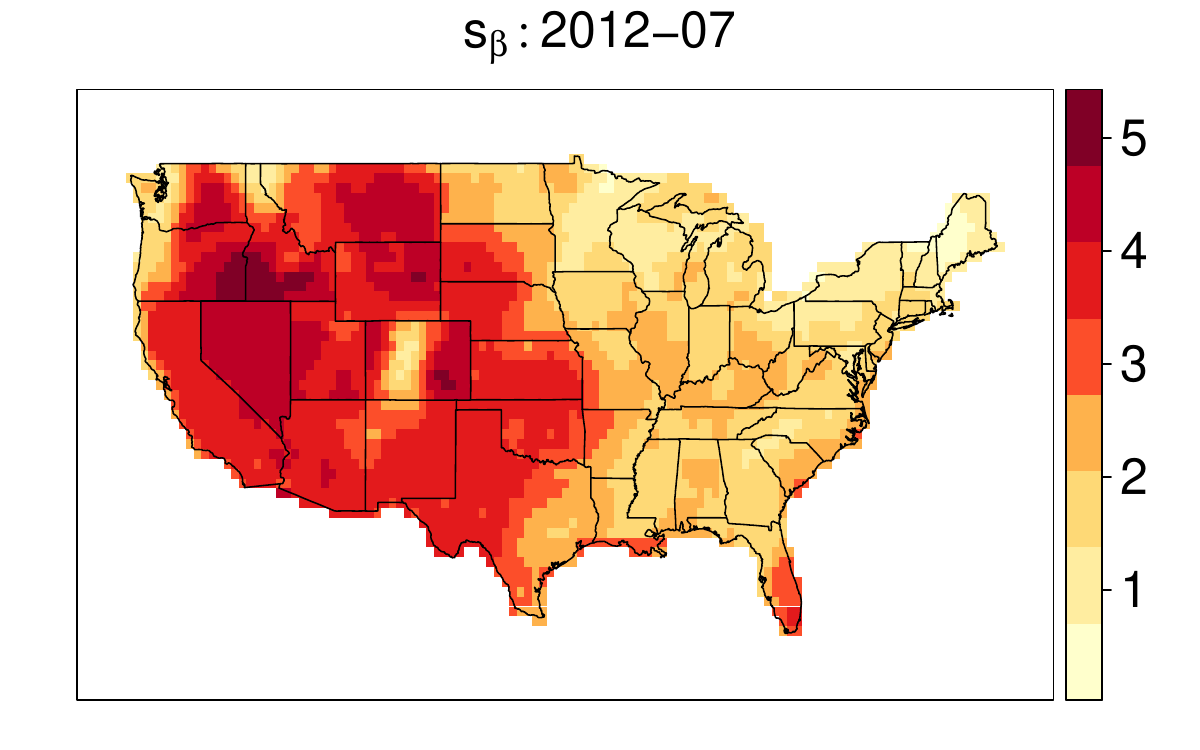} 
\end{minipage}
\begin{minipage}{0.32\linewidth}
\flushleft
\includegraphics[width=\linewidth]{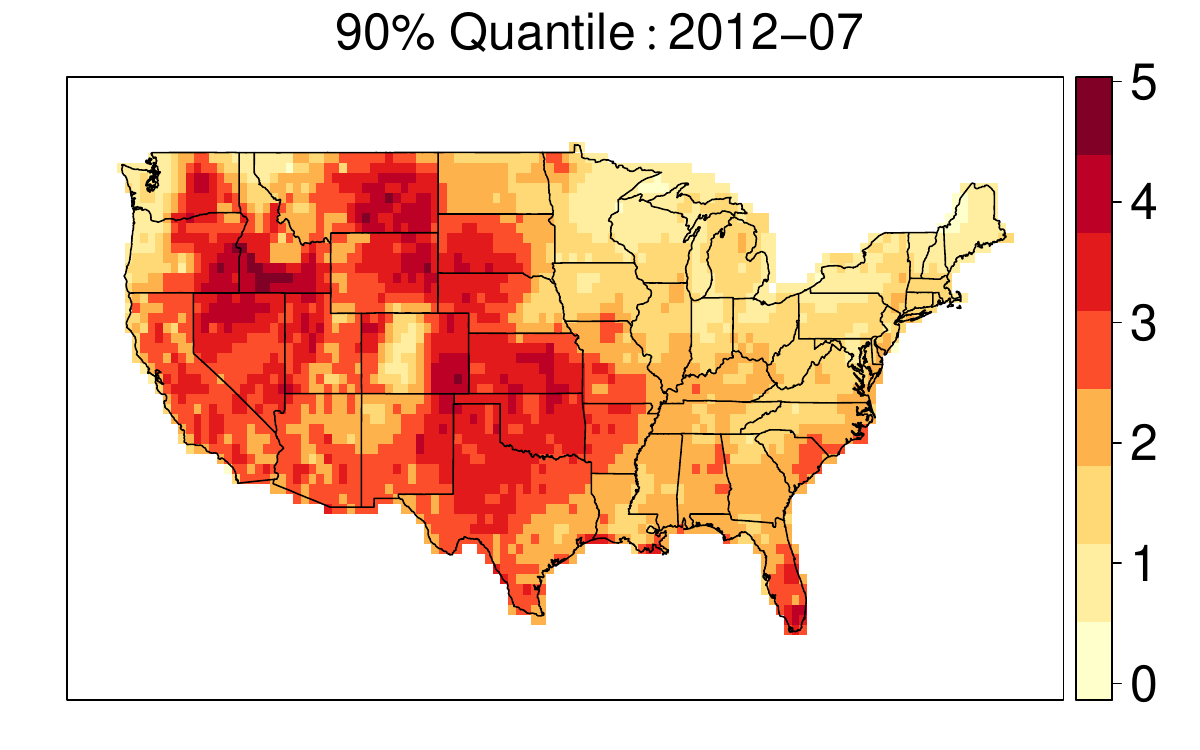} 
\end{minipage}
\begin{minipage}{0.32\linewidth}
\flushleft
\includegraphics[width=\linewidth]{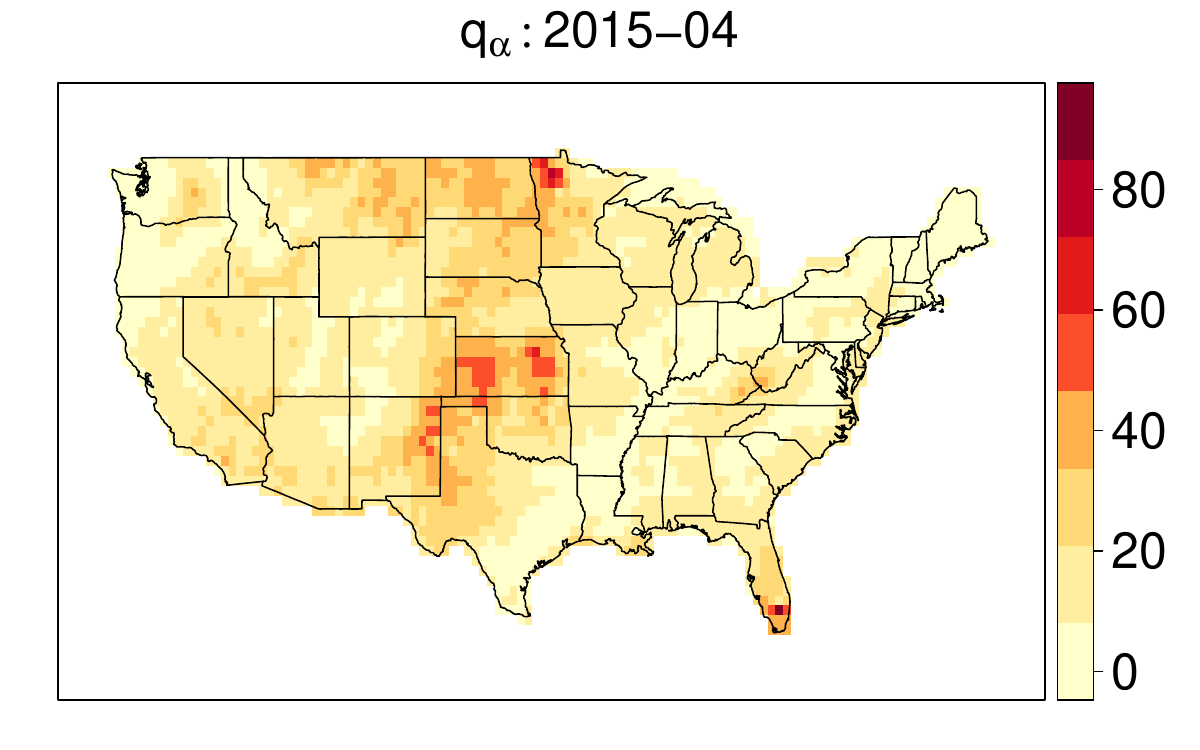} 
\end{minipage}
\begin{minipage}{0.32\linewidth}
\flushleft
\includegraphics[width=\linewidth]{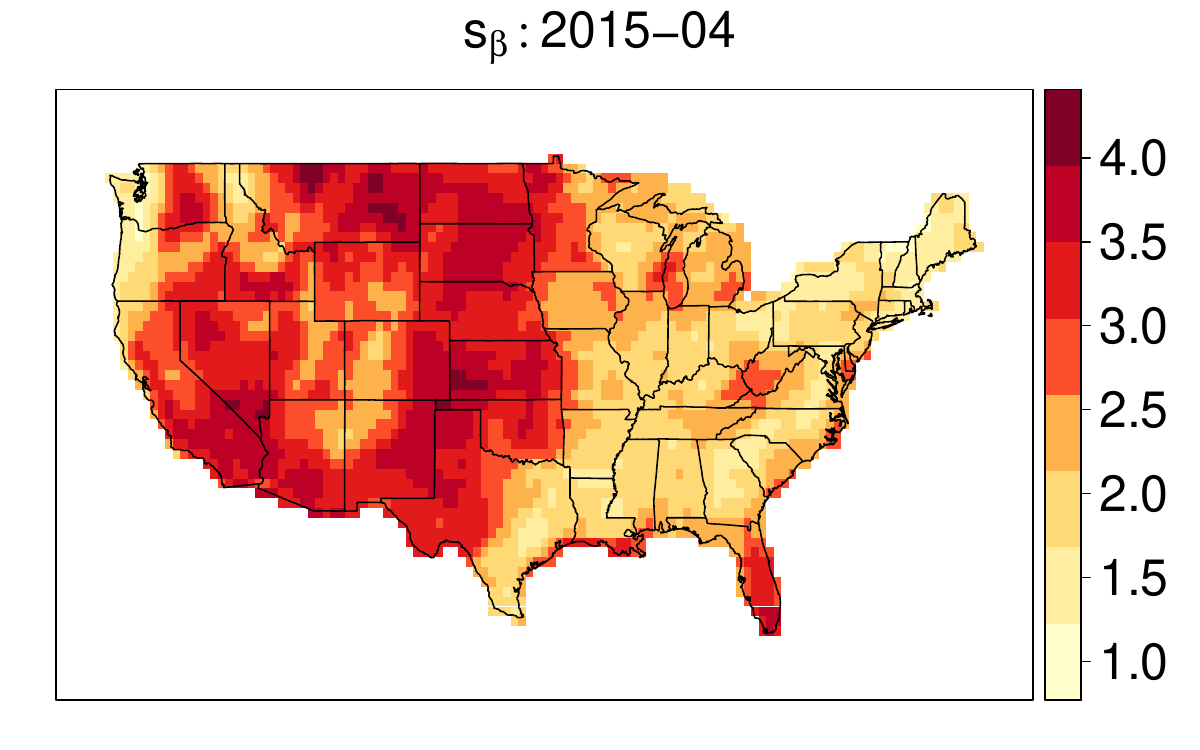} 
\end{minipage}
\begin{minipage}{0.32\linewidth}
\flushleft
\includegraphics[width=\linewidth]{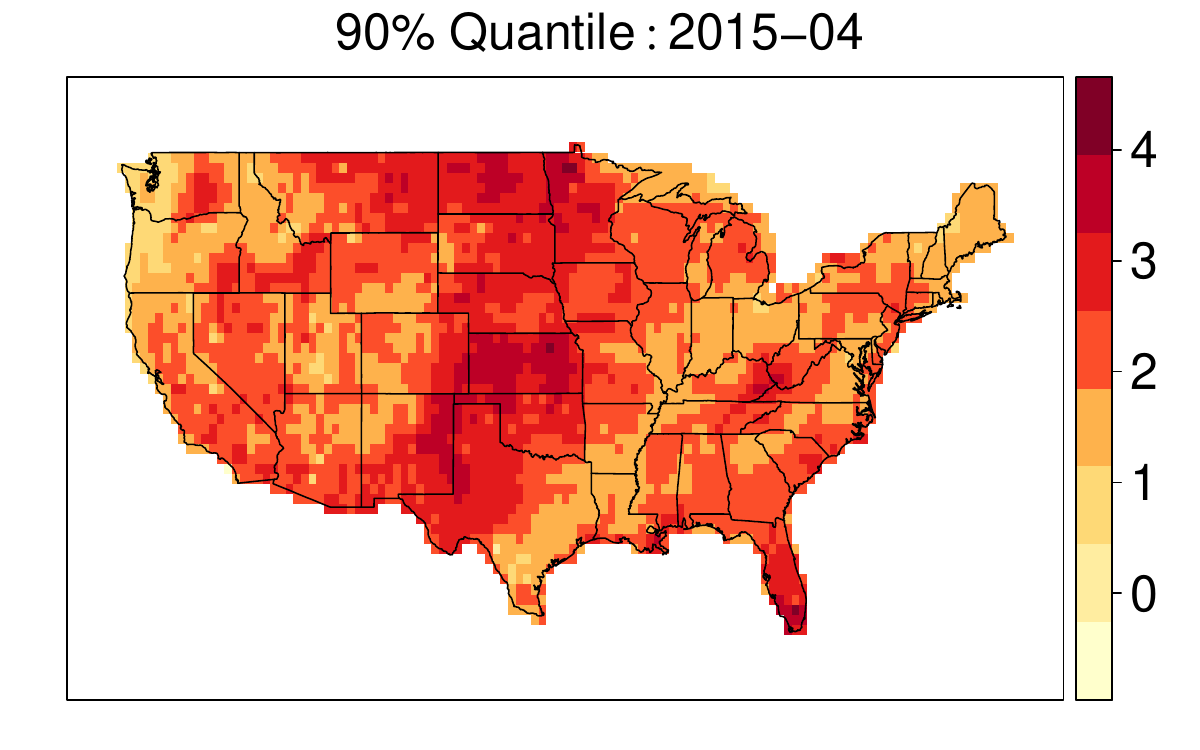} 
\end{minipage}
\vspace{-0.3cm}
\caption{Maps of median estimated $q_\alpha(s,t)$ ($\sqrt{\mbox{acres}}$; left), $\log \{1+s_\beta(s,t)\}$ (log-$\sqrt{\mbox{acres}}$; middle) and $90\%$ quantile of $\log\{1+ Y(s,t)\}\mid\{Y(s,t)>0,\mathbf{X}(s,t)\}$ (log-$\sqrt{\mbox{acres}}$; right) for different times $t$ (by row). Times, top to bottom row: March 1993, August 2000,  May 2003, July 2012, April 2015. }
\label{spread_map_sup}
\end{figure}

\begin{figure}[h!]
\centering
\begin{minipage}{0.32\linewidth}
\flushleft
\includegraphics[width=\linewidth]{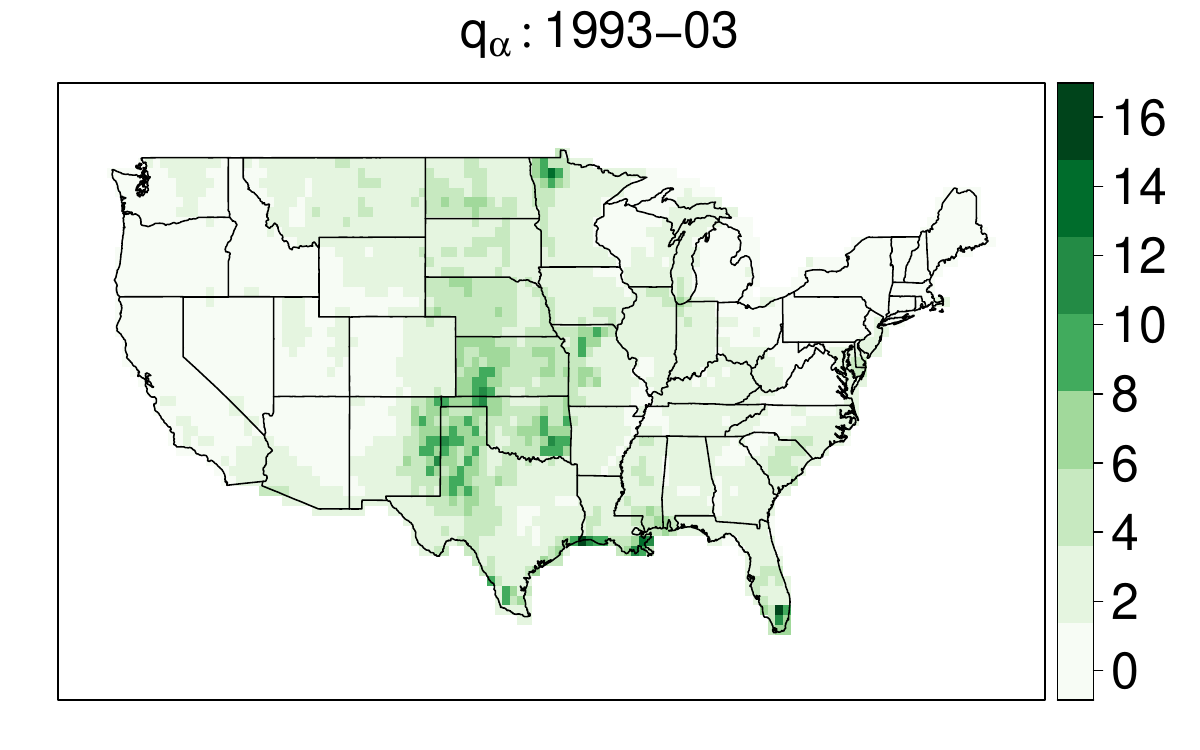} 
\end{minipage}
\begin{minipage}{0.32\linewidth}
\flushleft
\includegraphics[width=\linewidth]{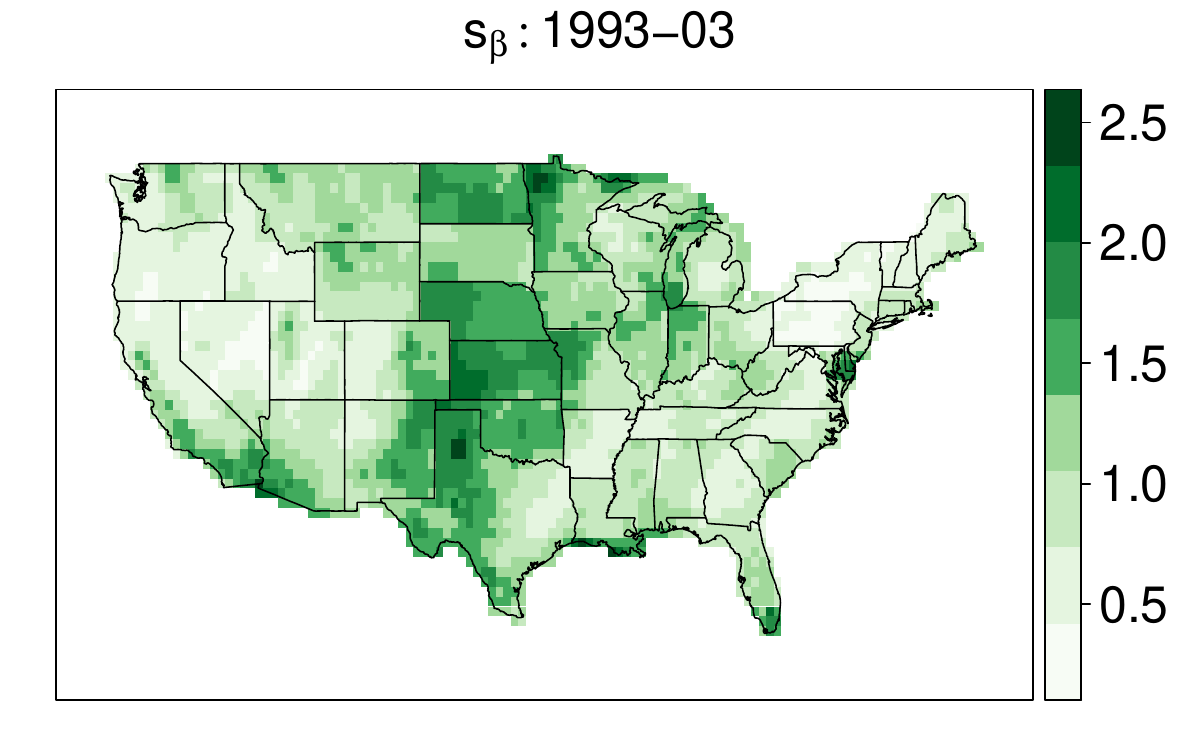} 
\end{minipage}
\vspace{-.2cm}
\begin{minipage}{0.32\linewidth}
\flushleft
\includegraphics[width=\linewidth]{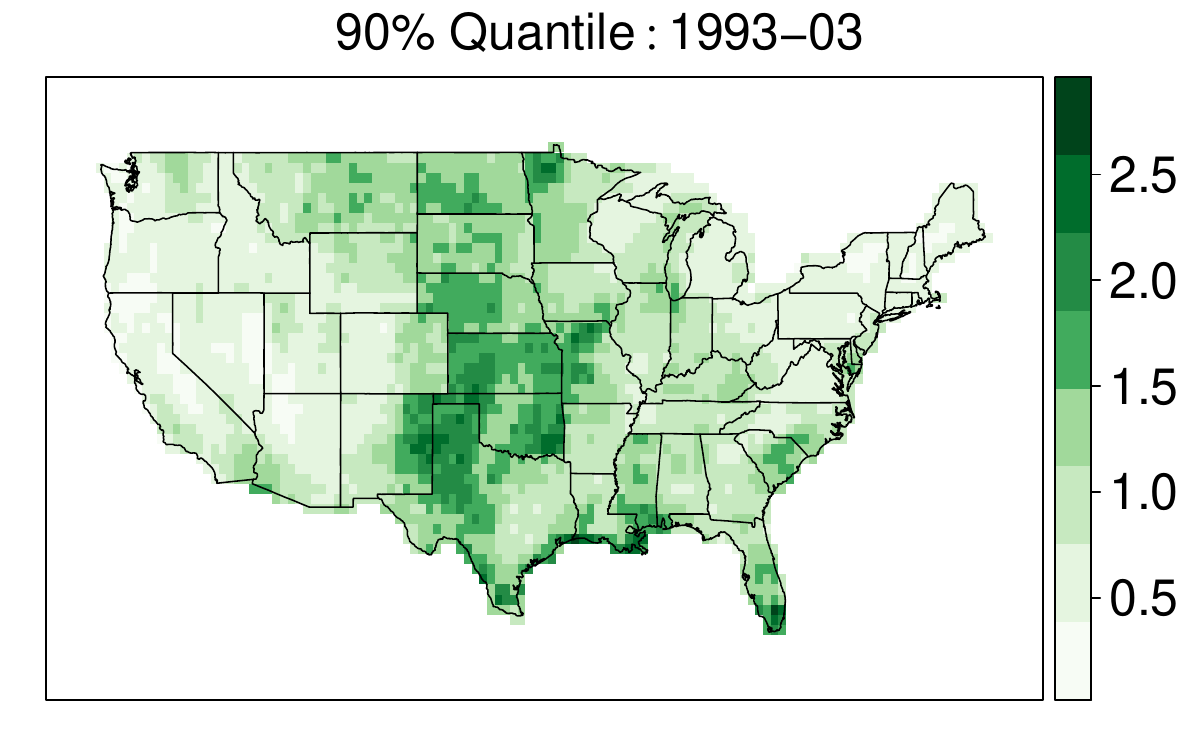} 
\end{minipage}
\begin{minipage}{0.32\linewidth}
\flushleft
\includegraphics[width=\linewidth]{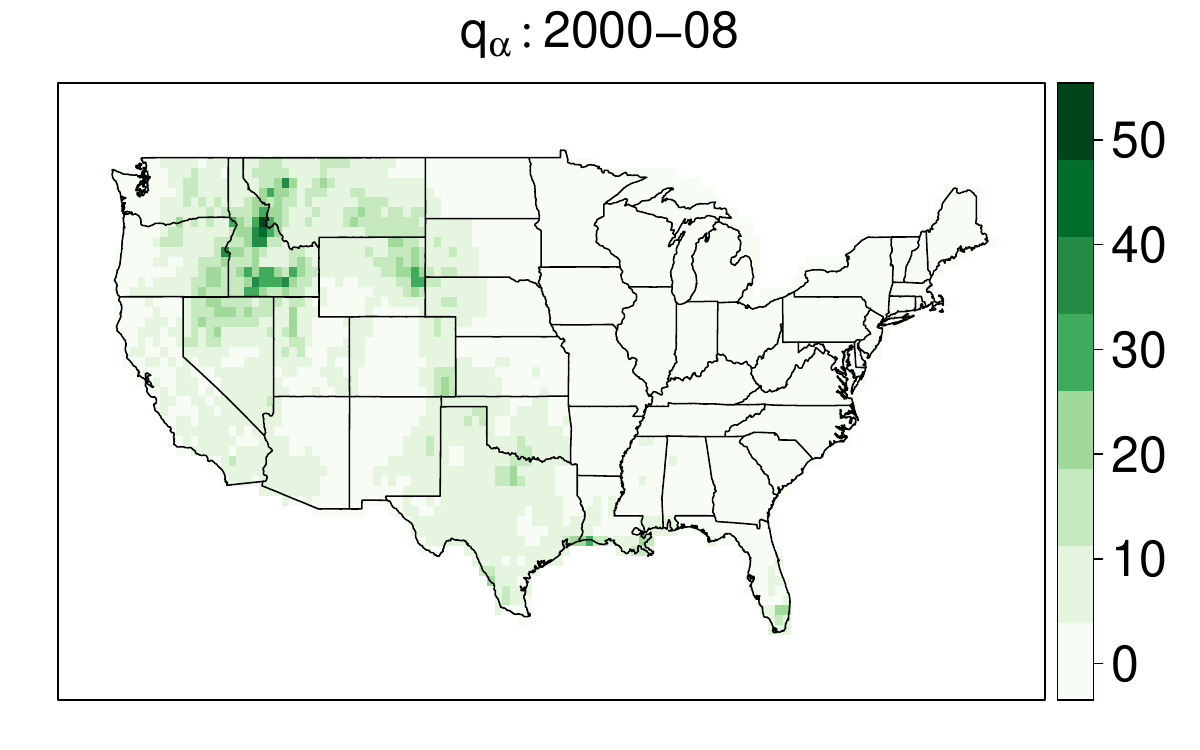} 
\end{minipage}
\begin{minipage}{0.32\linewidth}
\flushleft
\includegraphics[width=\linewidth]{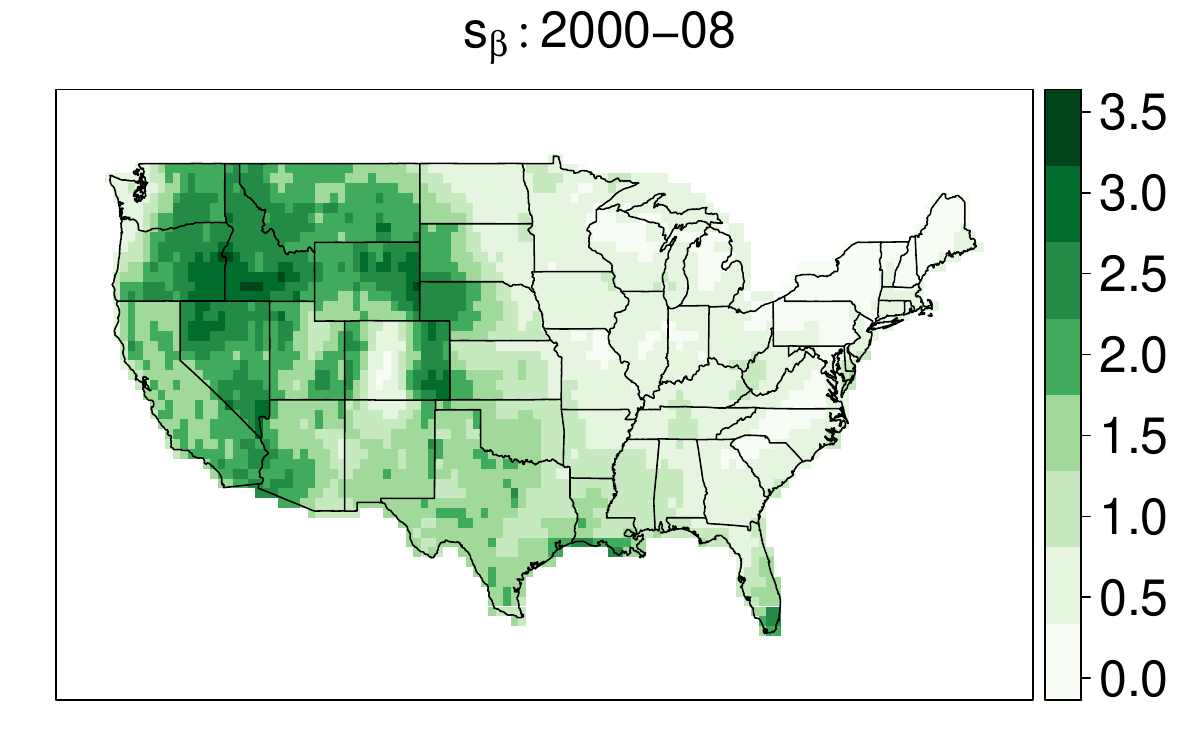} 
\end{minipage}
\vspace{-.2cm}
\begin{minipage}{0.32\linewidth}
\flushleft
\includegraphics[width=\linewidth]{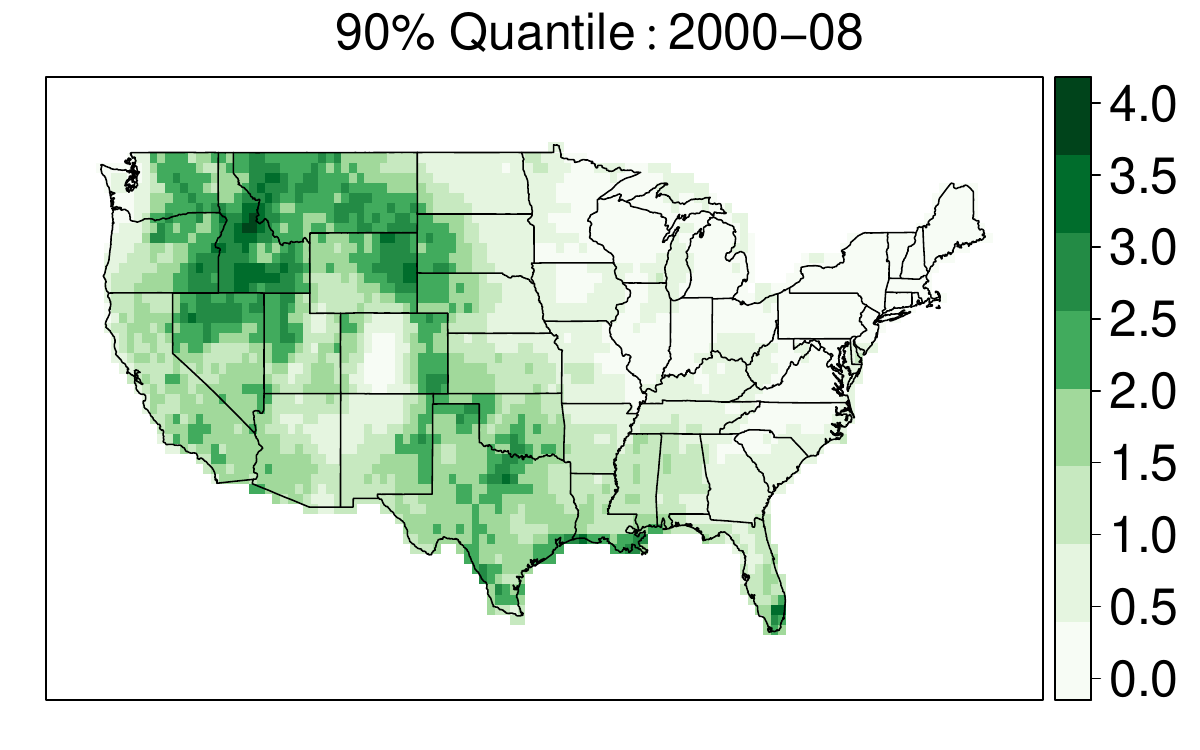} 
\end{minipage}
\vspace{-.2cm}
\begin{minipage}{0.32\linewidth}
\flushleft
\includegraphics[width=\linewidth]{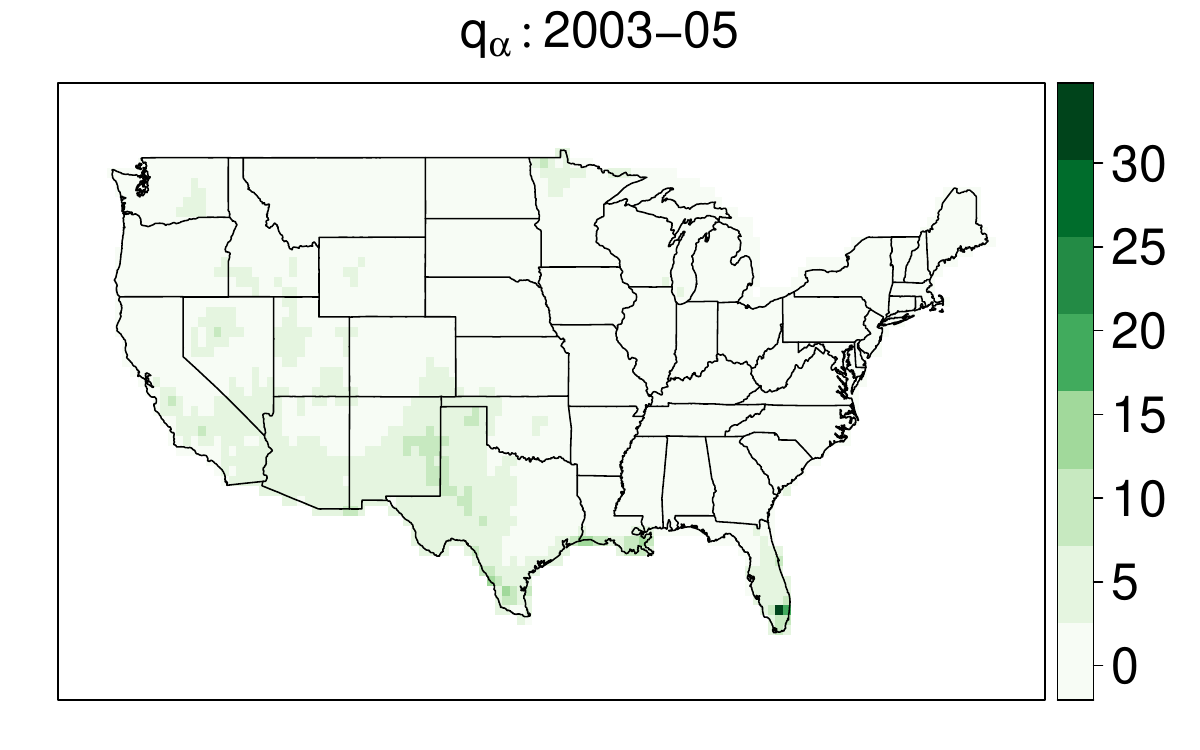} 
\end{minipage}
\begin{minipage}{0.32\linewidth}
\flushleft
\includegraphics[width=\linewidth]{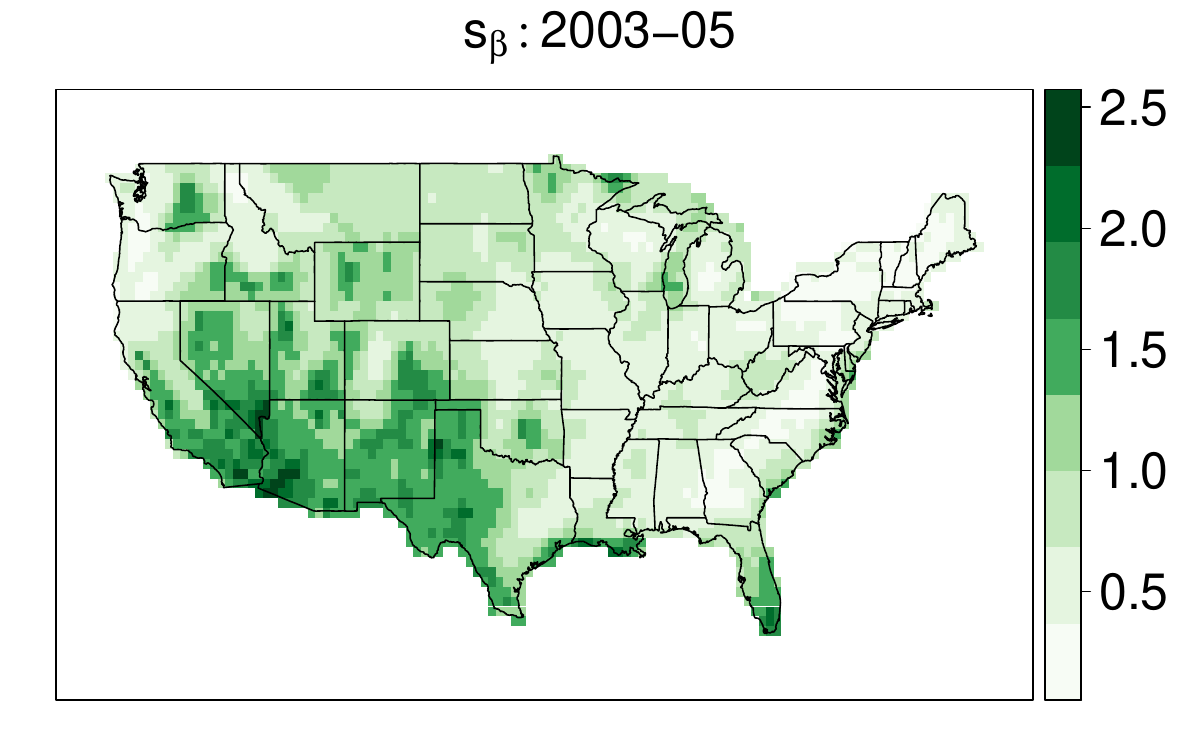} 
\end{minipage}
\begin{minipage}{0.32\linewidth}
\flushleft
\includegraphics[width=\linewidth]{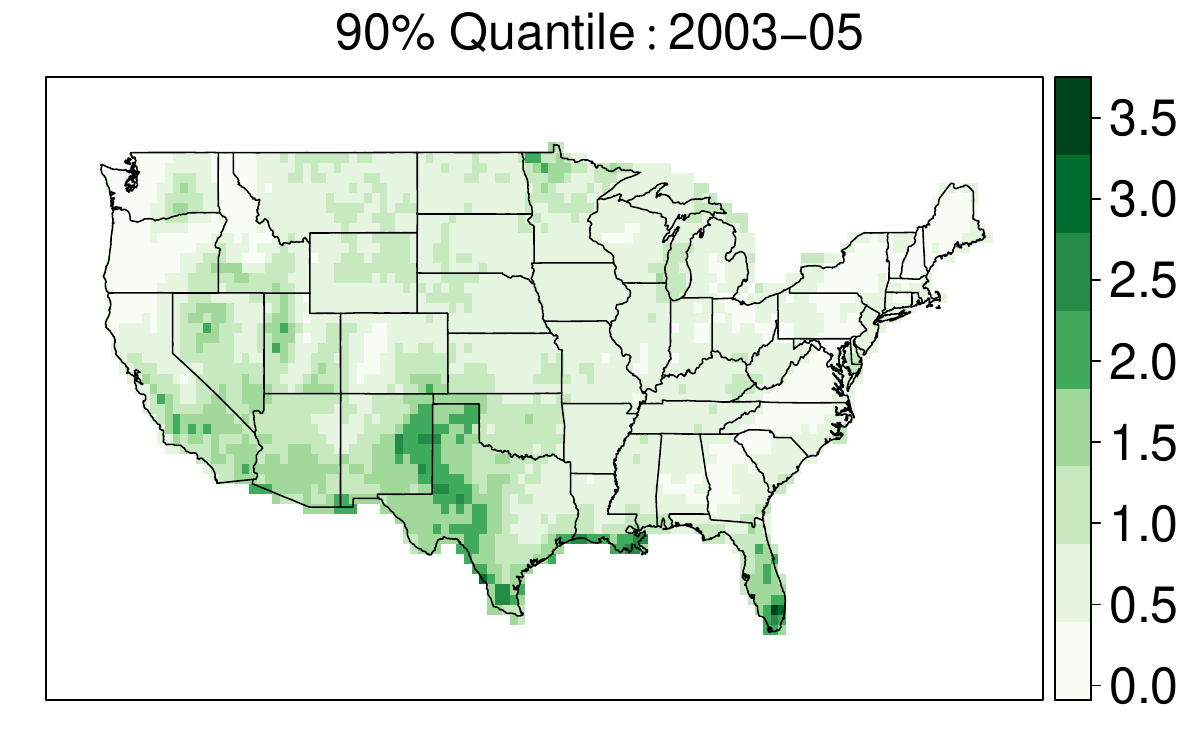} 
\end{minipage}
\vspace{-.2cm}
\begin{minipage}{0.32\linewidth}
\flushleft
\includegraphics[width=\linewidth]{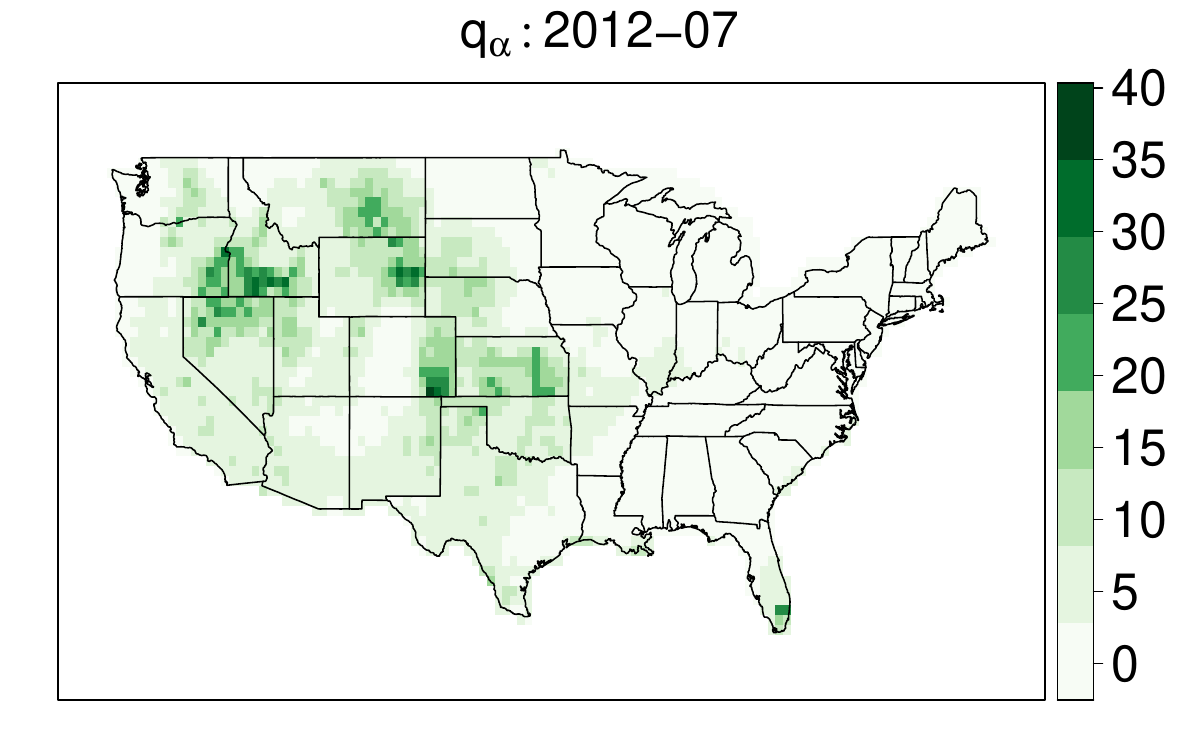} 
\end{minipage}
\begin{minipage}{0.32\linewidth}
\flushleft
\includegraphics[width=\linewidth]{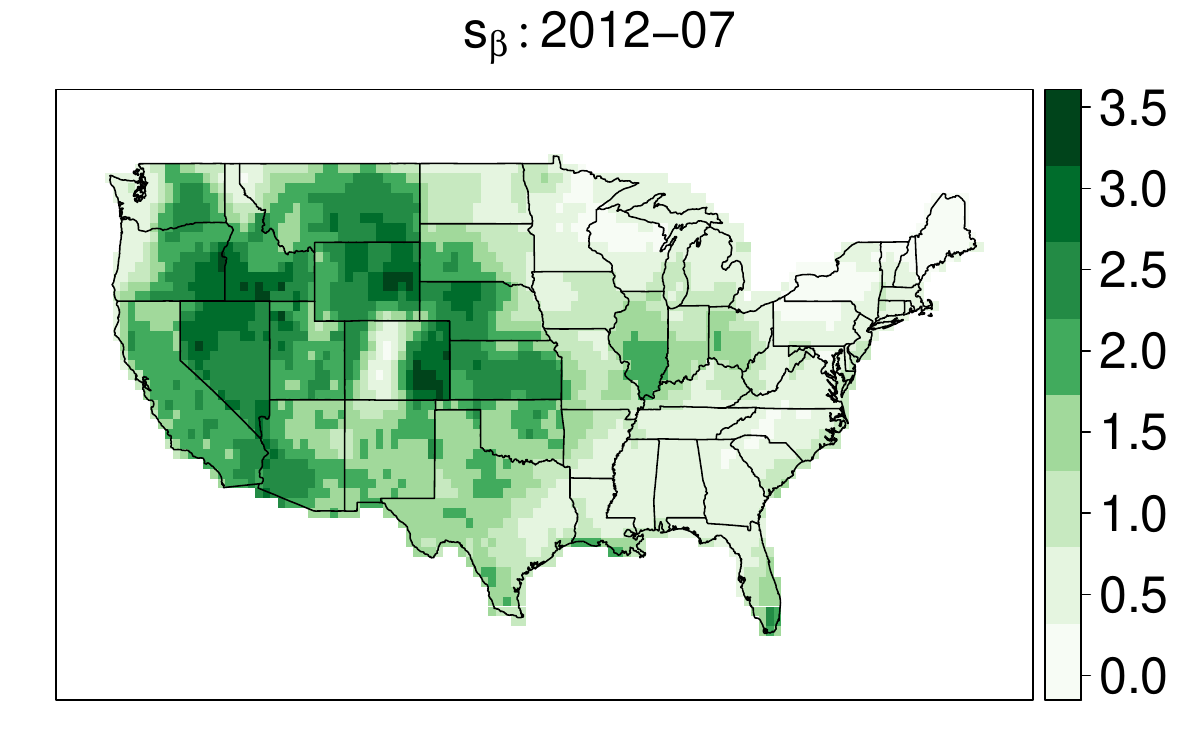} 
\end{minipage}
\begin{minipage}{0.32\linewidth}
\flushleft
\includegraphics[width=\linewidth]{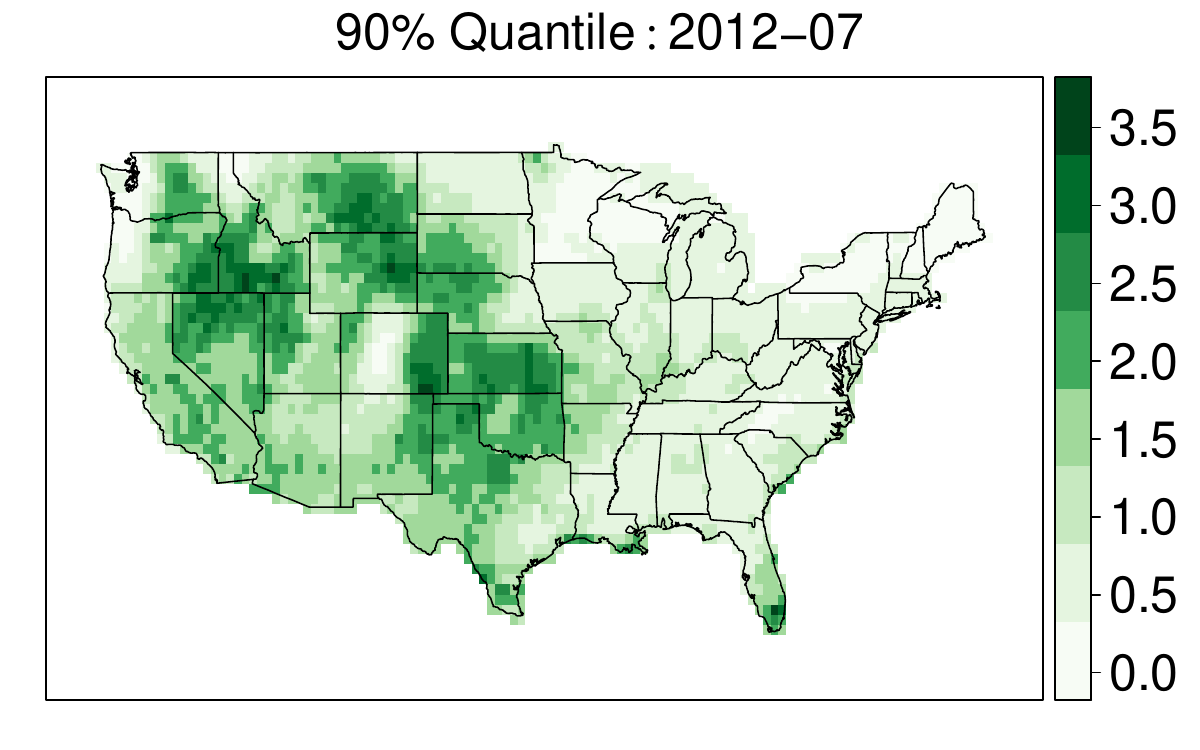} 
\end{minipage}
\begin{minipage}{0.32\linewidth}
\flushleft
\includegraphics[width=\linewidth]{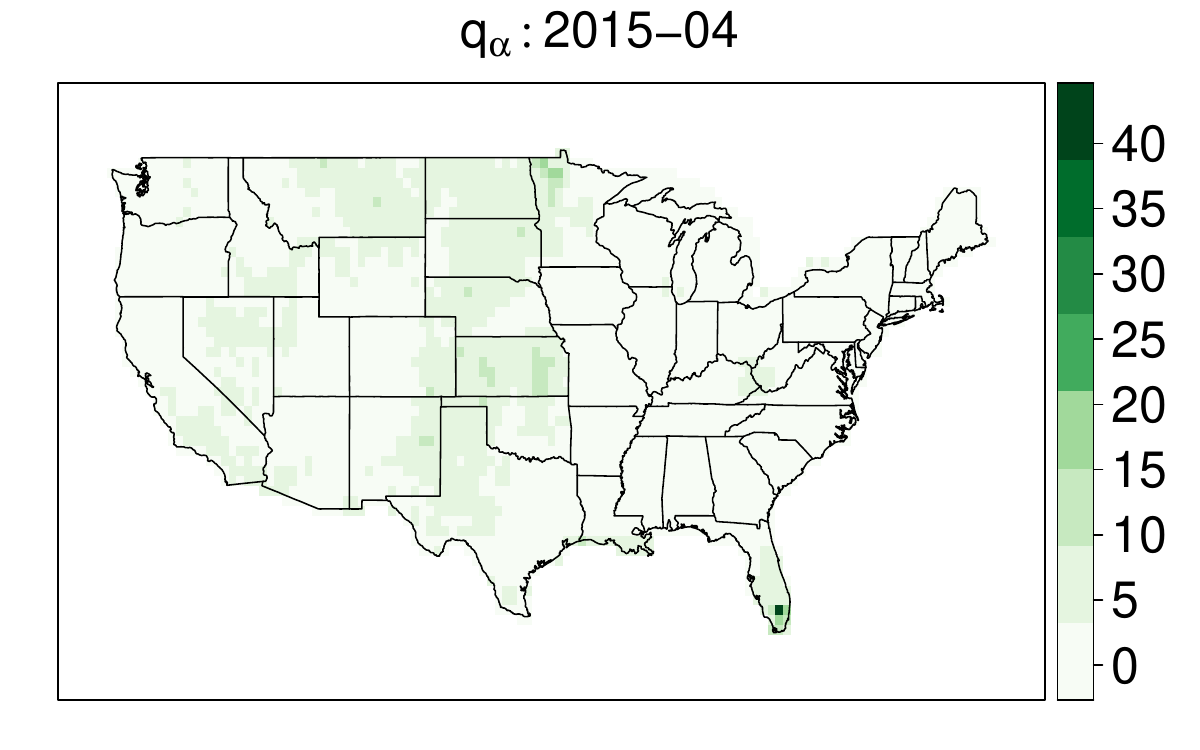} 
\end{minipage}
\begin{minipage}{0.32\linewidth}
\flushleft
\includegraphics[width=\linewidth]{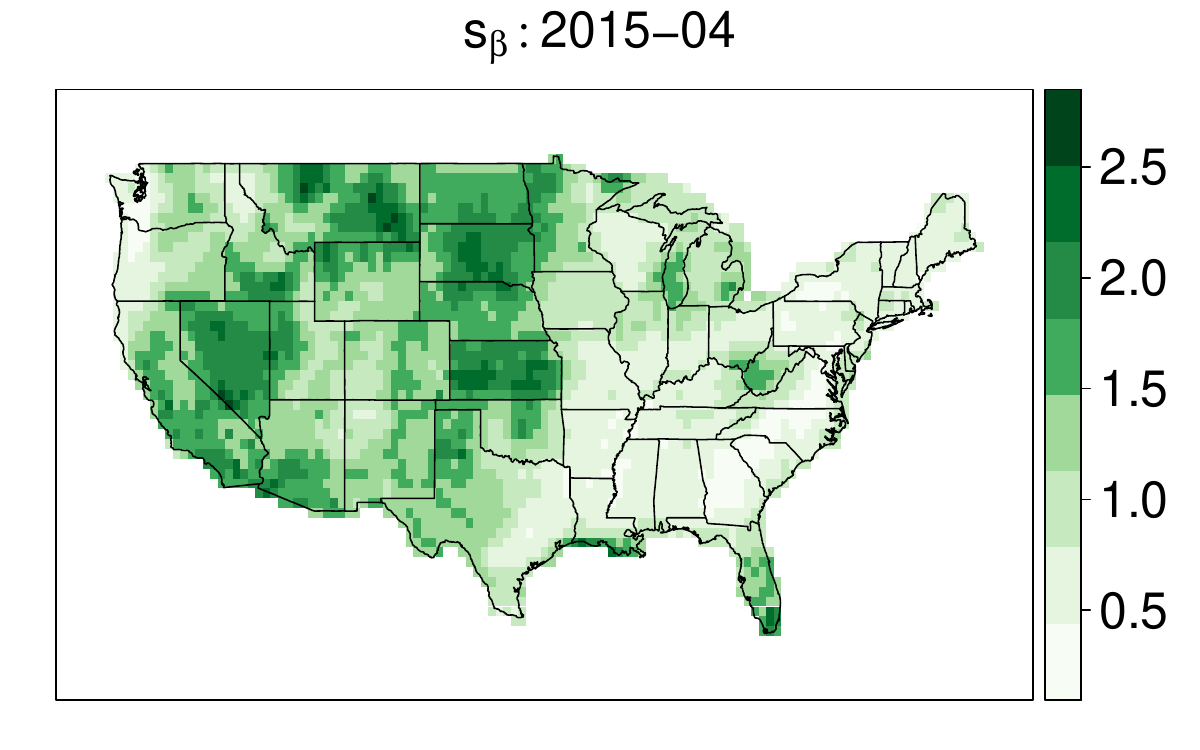} 
\end{minipage}
\begin{minipage}{0.32\linewidth}
\flushleft
\includegraphics[width=\linewidth]{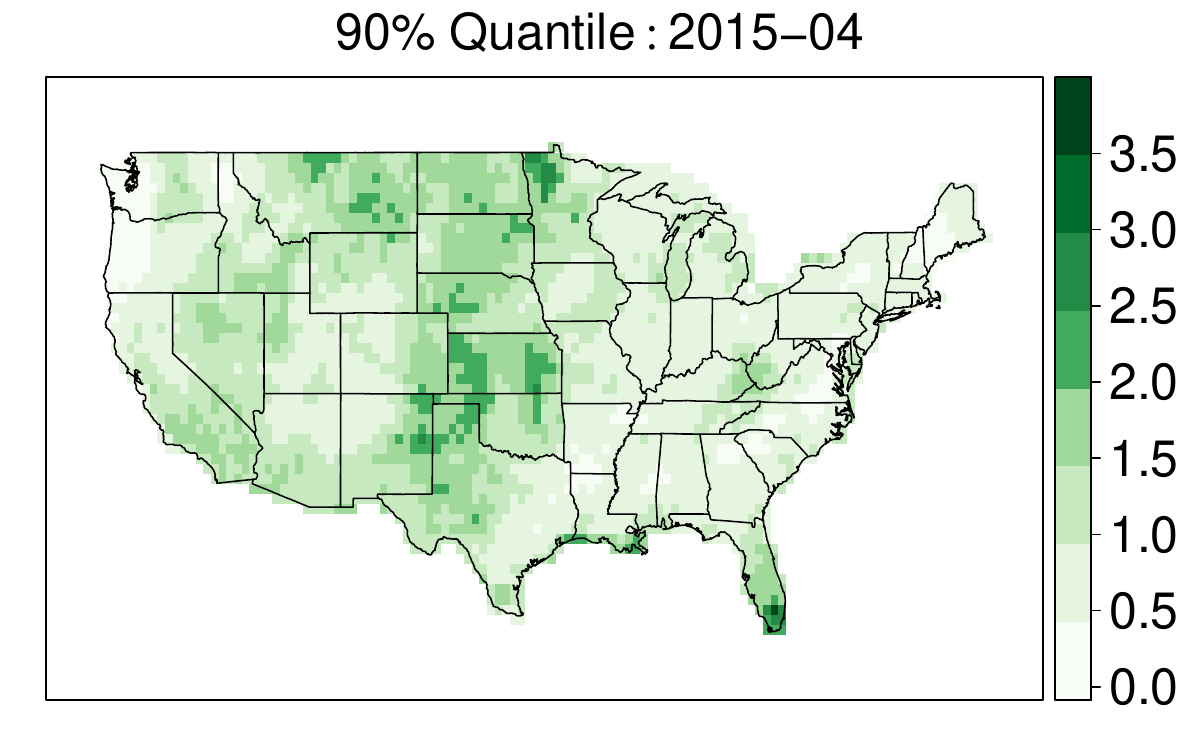} 
\end{minipage}
\vspace{-0.3cm}
\caption{Maps of the inter-quartile range for estimated $q_\alpha(s,t)$ ($\sqrt{\mbox{acres}}$; left), $\log \{1+s_\beta(s,t)\}$ (log-$\sqrt{\mbox{acres}}$; middle), and $90\%$ quantile of $\log\{1+ Y(s,t)\}\mid\{Y(s,t)>0,\mathbf{X}(s,t)\}$ (log-$\sqrt{\mbox{acres}}$; right) for different times $t$ (by row). Times, top to bottom row: March 1993, August 2000,  May 2003, July 2012, April 2015. }
\label{spread_IQR_sup}
\end{figure}

\begin{figure}[t!]
\centering
\begin{minipage}{0.32\linewidth}
\flushleft
\includegraphics[width=\linewidth]{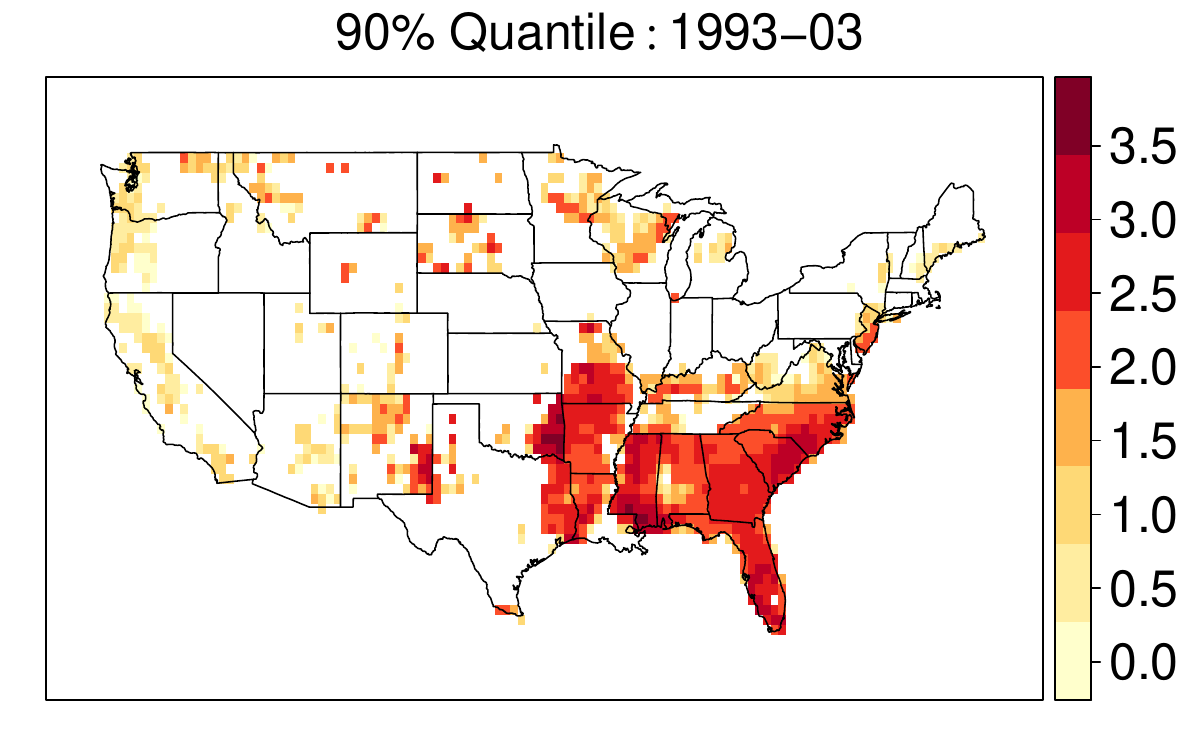} 
\end{minipage}
\vspace{-.2cm}
\begin{minipage}{0.32\linewidth}
\flushleft
\includegraphics[width=\linewidth]{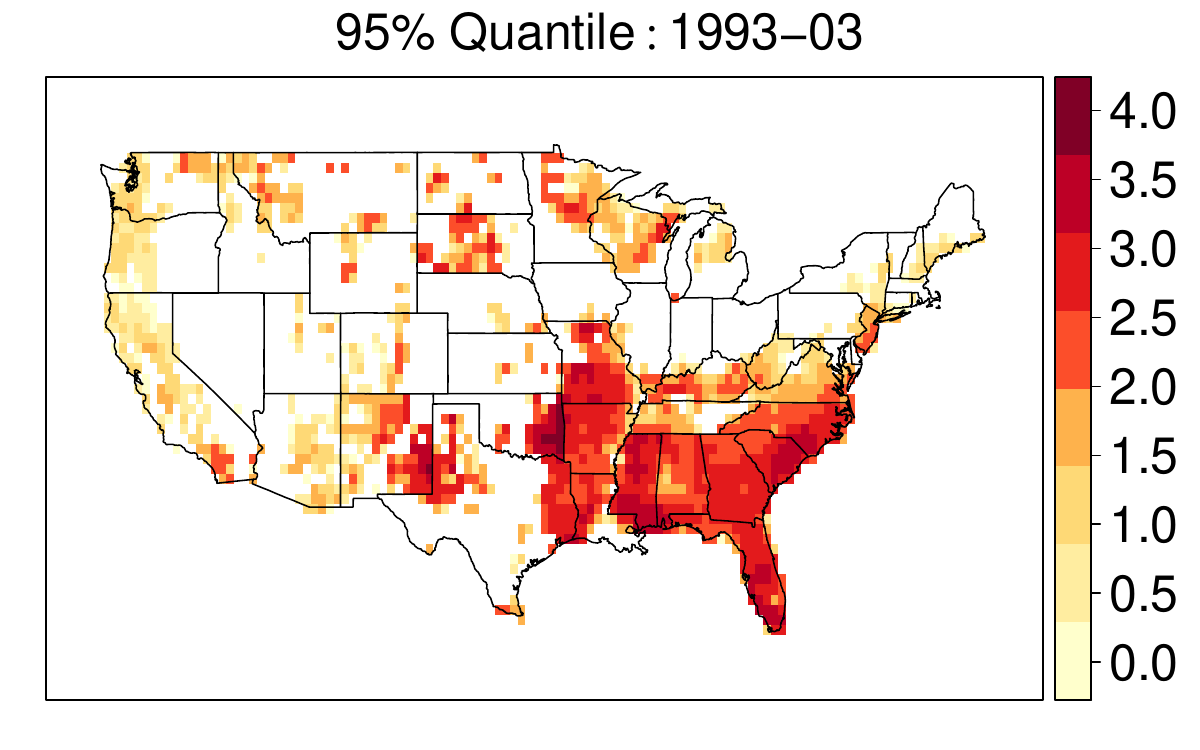} 
\end{minipage}
\begin{minipage}{0.32\linewidth}
\flushleft
\includegraphics[width=\linewidth]{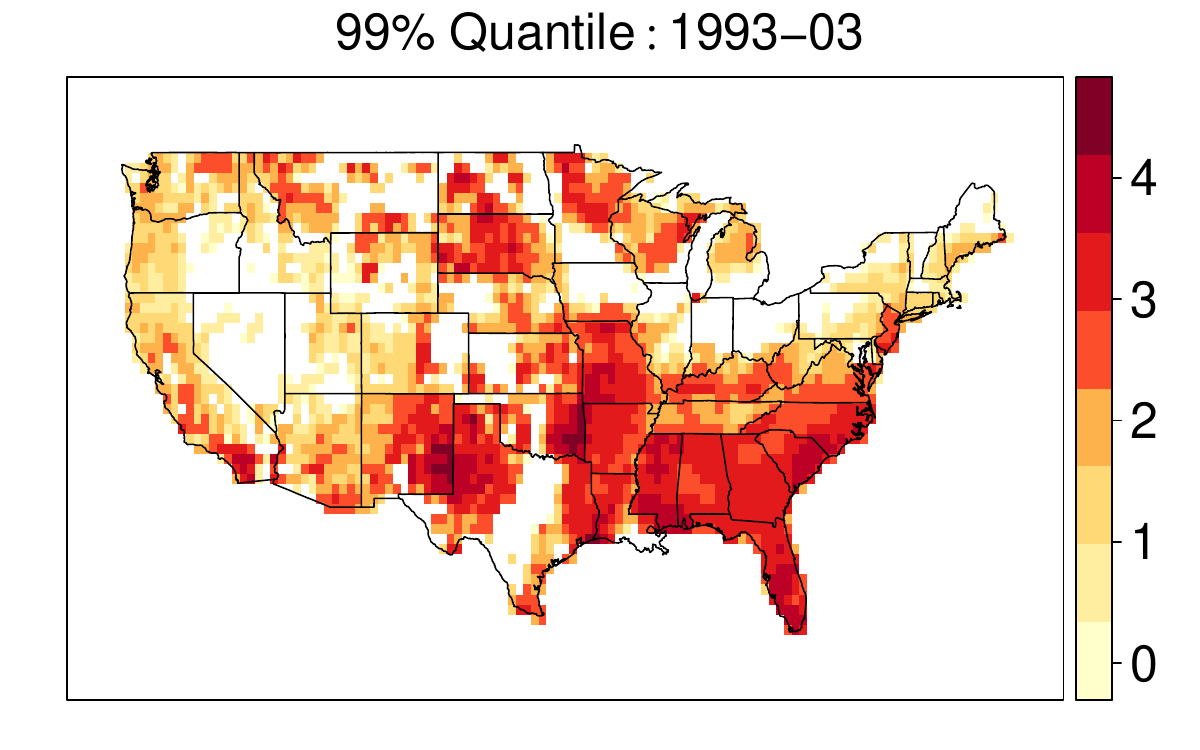} 
\end{minipage}
\begin{minipage}{0.32\linewidth}
\flushleft
\includegraphics[width=\linewidth]{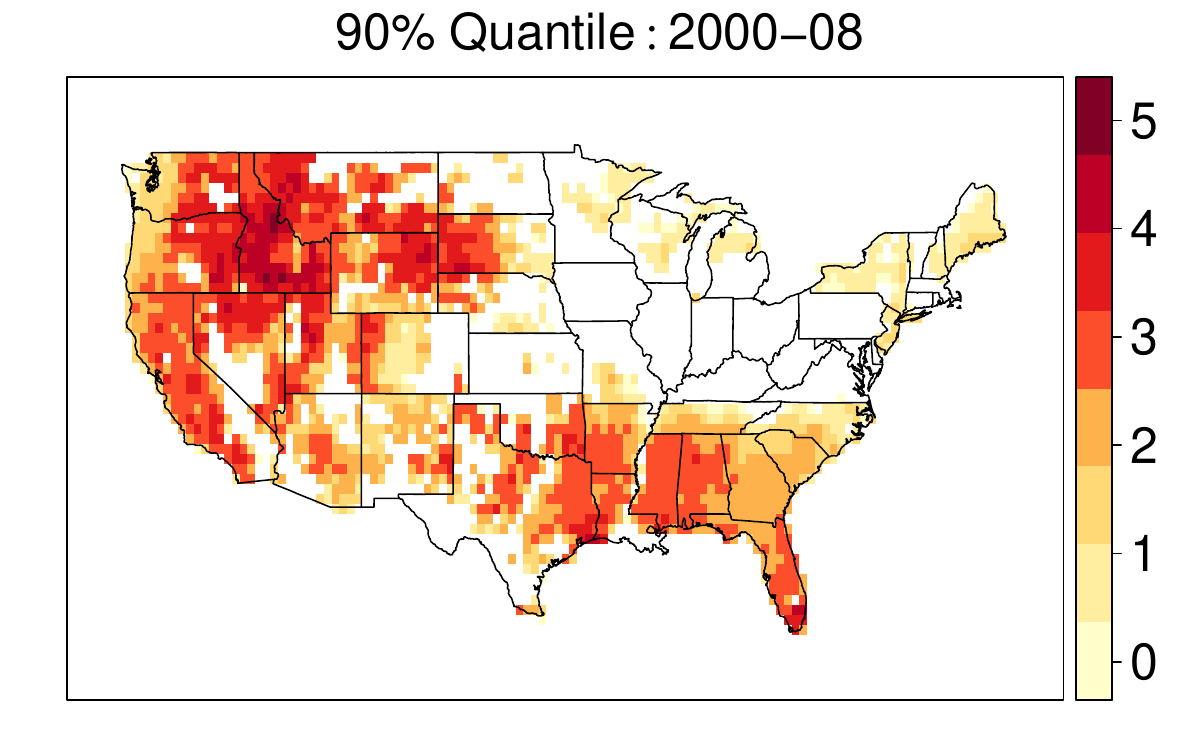} 
\end{minipage}
\begin{minipage}{0.32\linewidth}
\flushleft
\includegraphics[width=\linewidth]{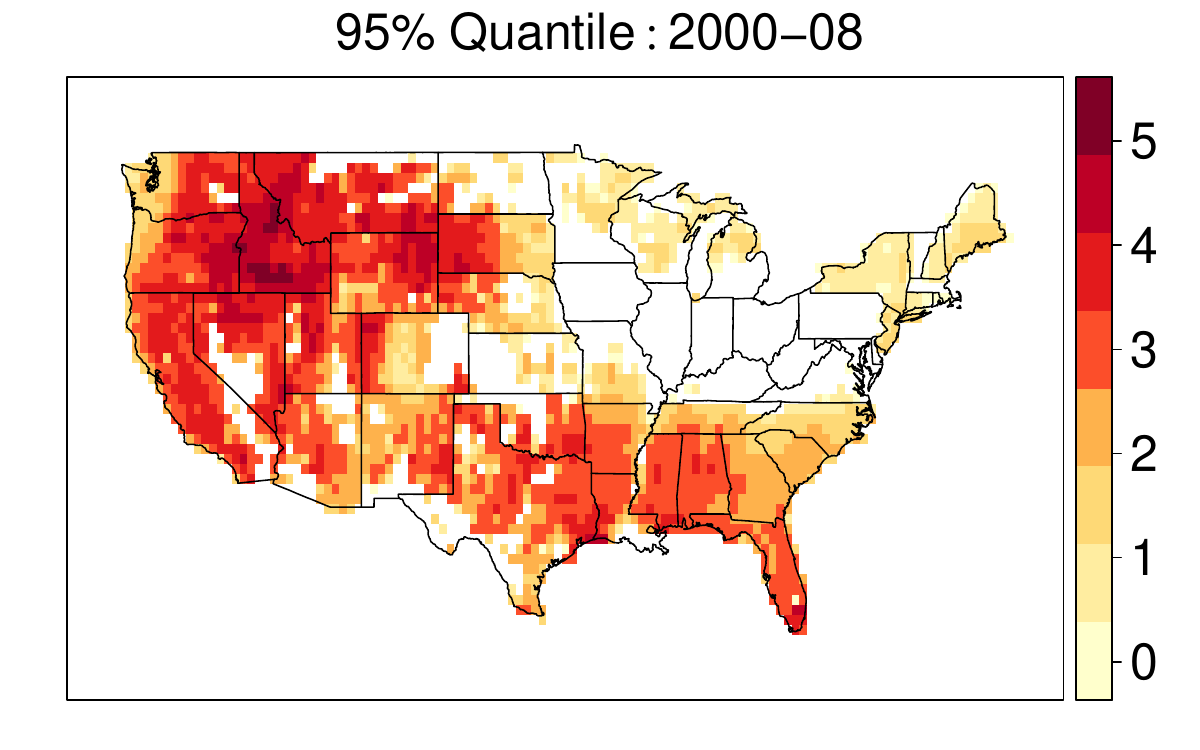} 
\end{minipage}
\vspace{-.2cm}
\begin{minipage}{0.32\linewidth}
\flushleft
\includegraphics[width=\linewidth]{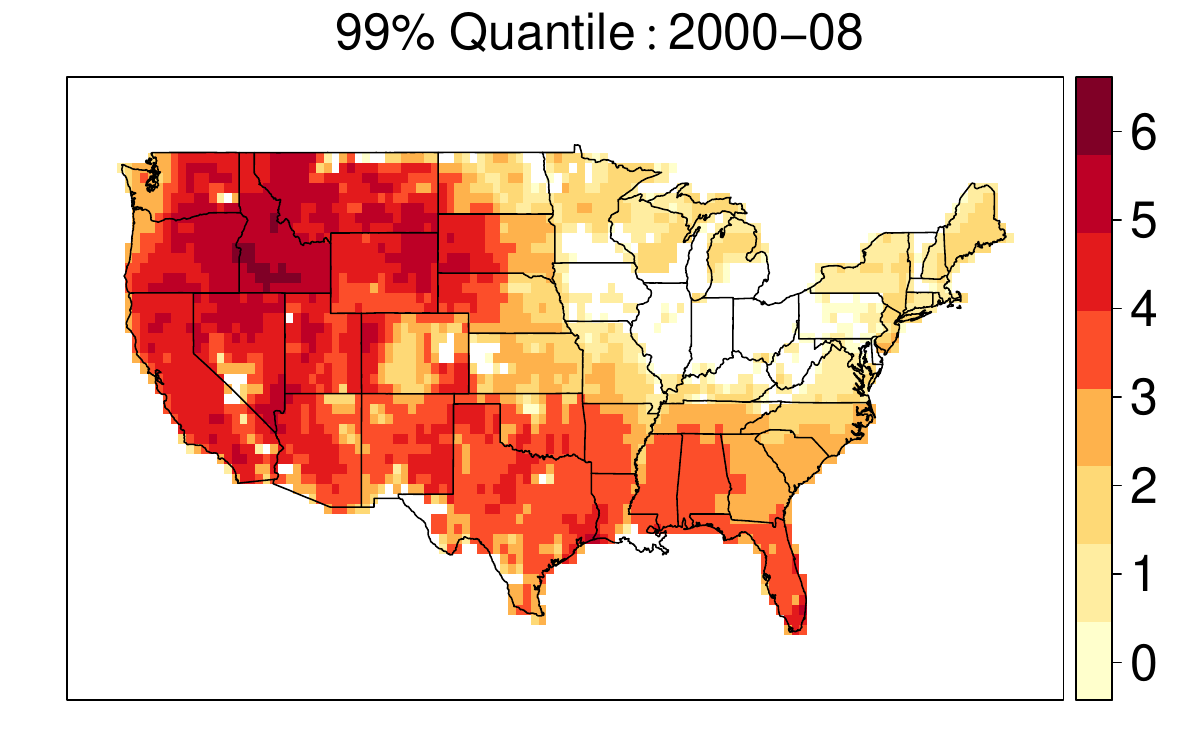} 
\end{minipage}
\begin{minipage}{0.32\linewidth}
\flushleft
\includegraphics[width=\linewidth]{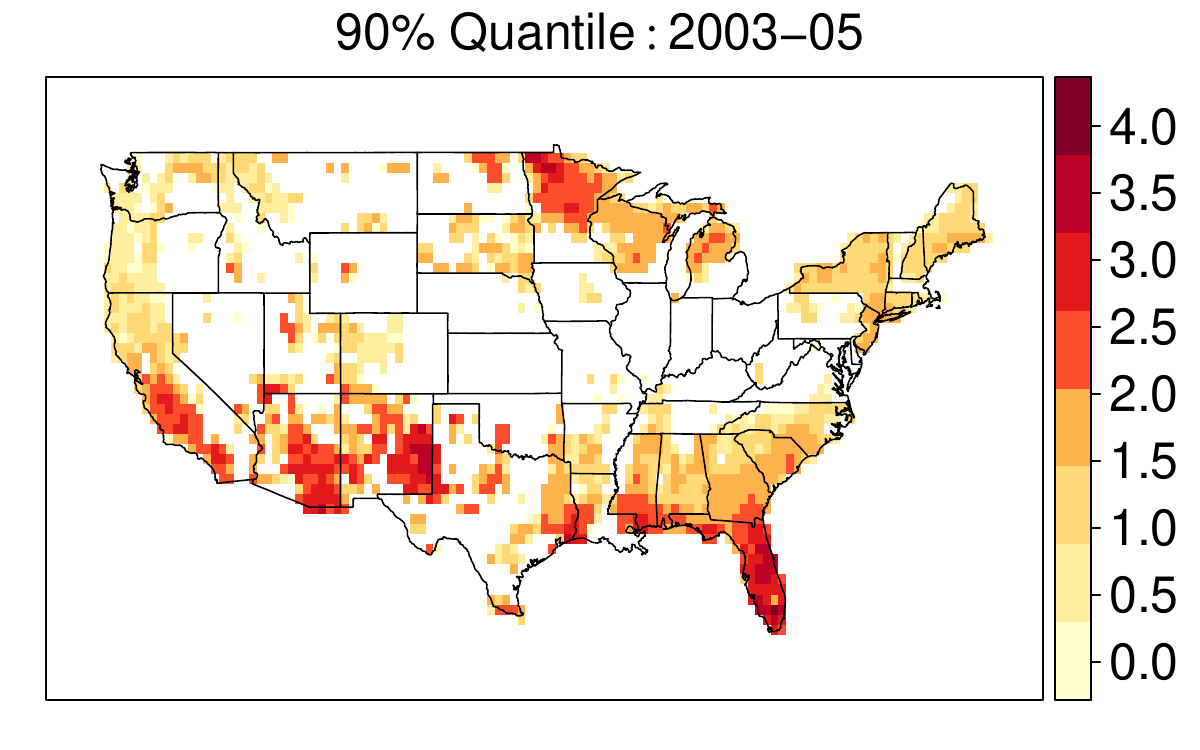} 
\end{minipage}
\begin{minipage}{0.32\linewidth}
\flushleft
\includegraphics[width=\linewidth]{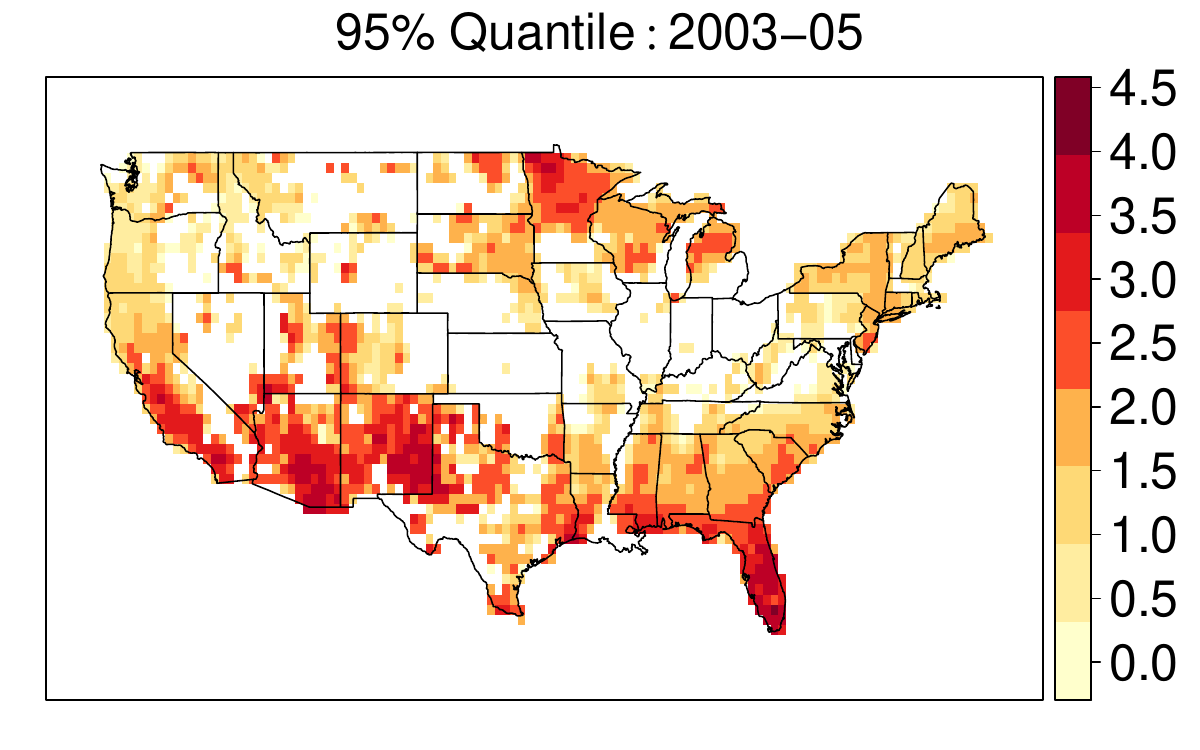} 
\end{minipage}
\vspace{-.2cm}
\begin{minipage}{0.32\linewidth}
\flushleft
\includegraphics[width=\linewidth]{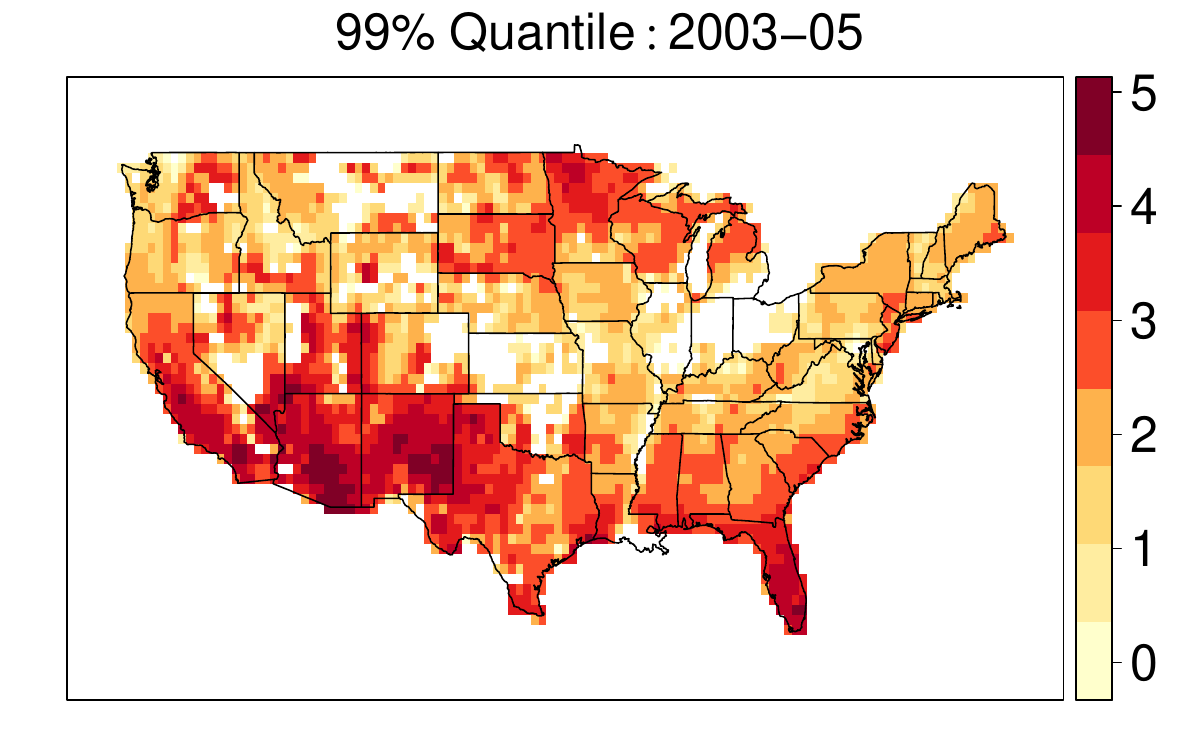} 
\end{minipage}
\begin{minipage}{0.32\linewidth}
\flushleft
\includegraphics[width=\linewidth]{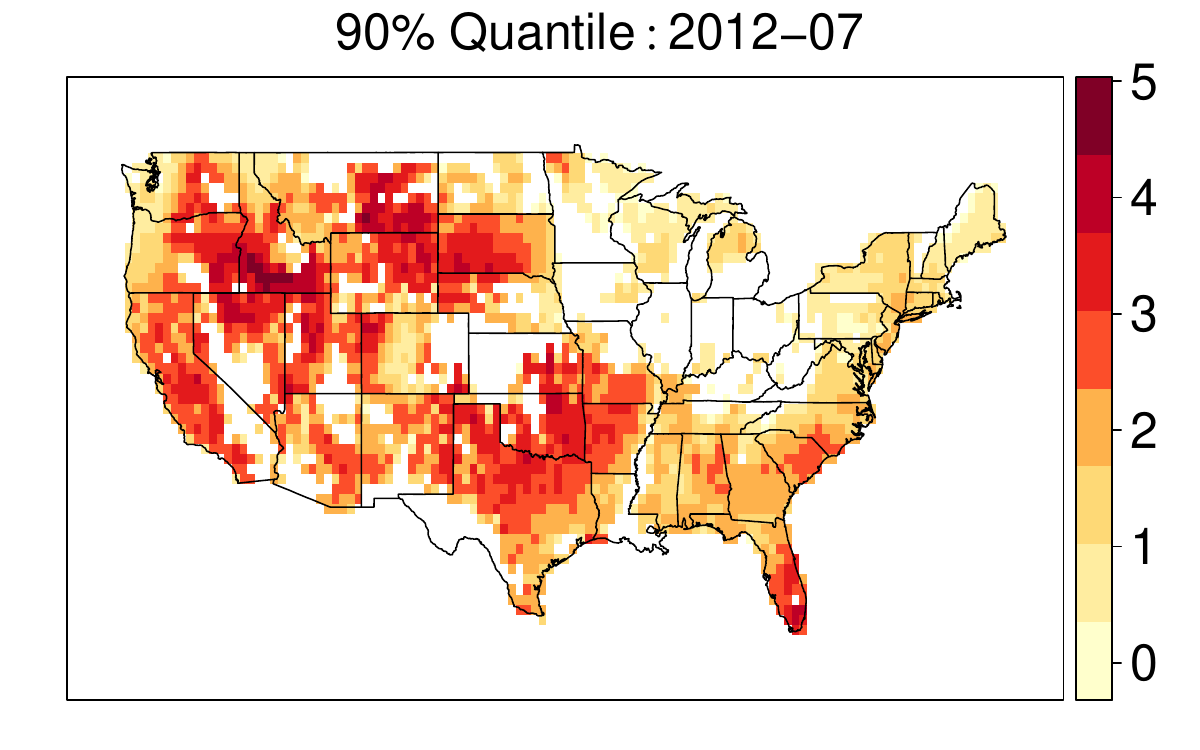} 
\end{minipage}
\begin{minipage}{0.32\linewidth}
\flushleft
\includegraphics[width=\linewidth]{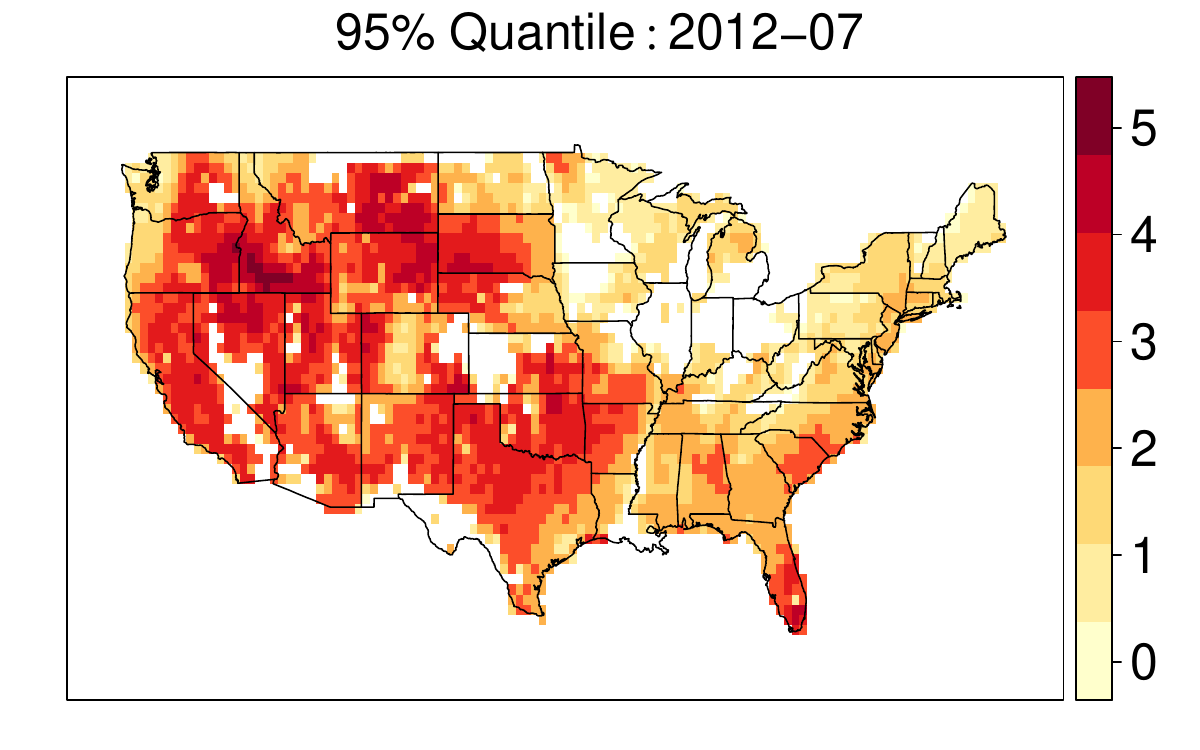} 
\end{minipage}
\vspace{-.2cm}
\begin{minipage}{0.32\linewidth}
\flushleft
\includegraphics[width=\linewidth]{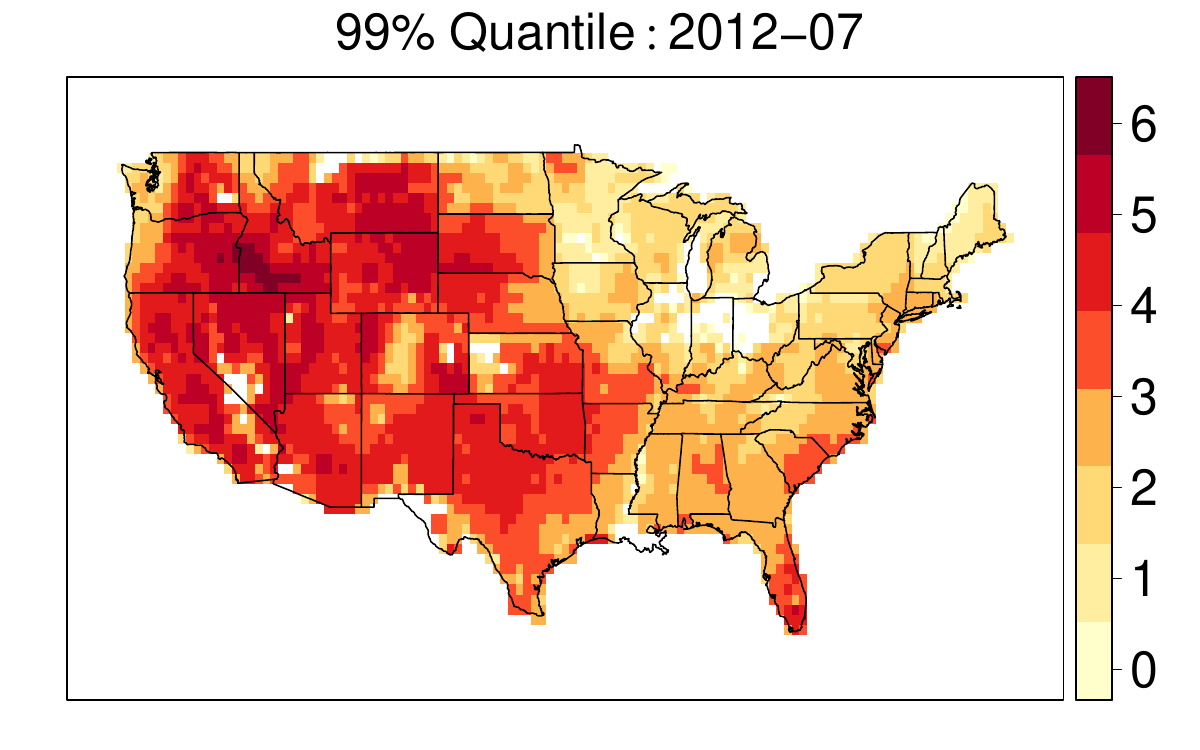} 
\end{minipage}
\begin{minipage}{0.32\linewidth}
\flushleft
\includegraphics[width=\linewidth]{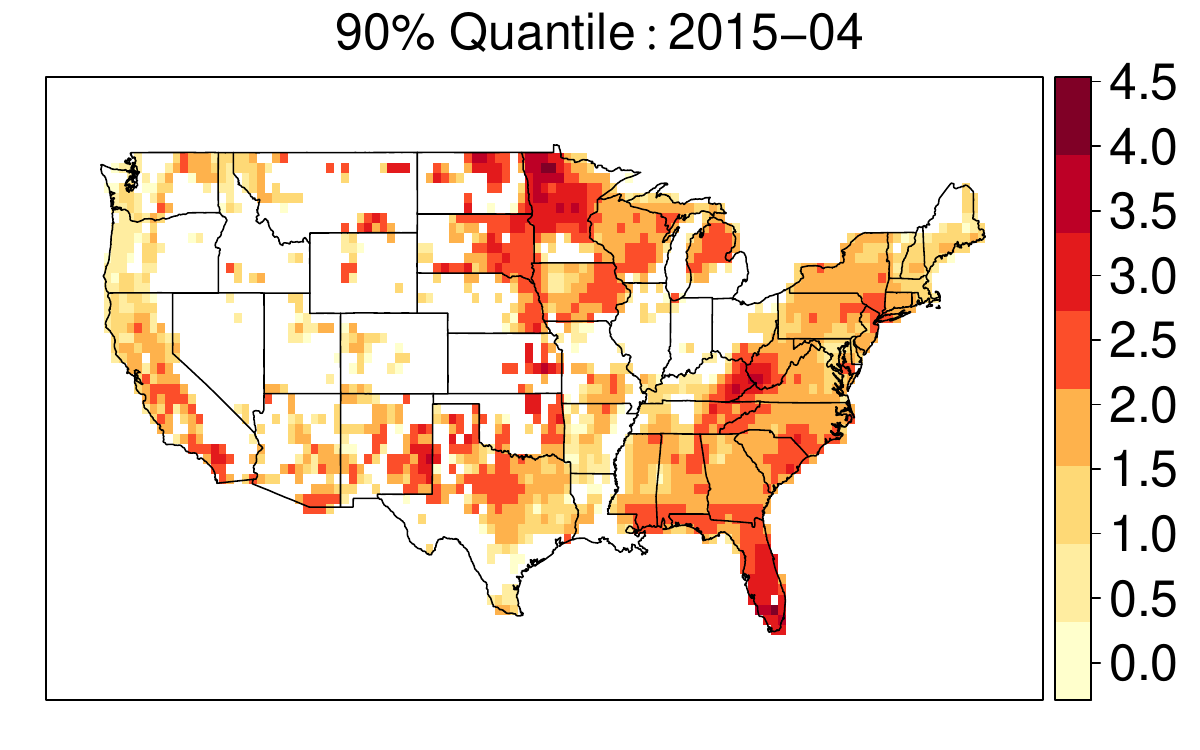} 
\end{minipage}
\begin{minipage}{0.32\linewidth}
\flushleft
\includegraphics[width=\linewidth]{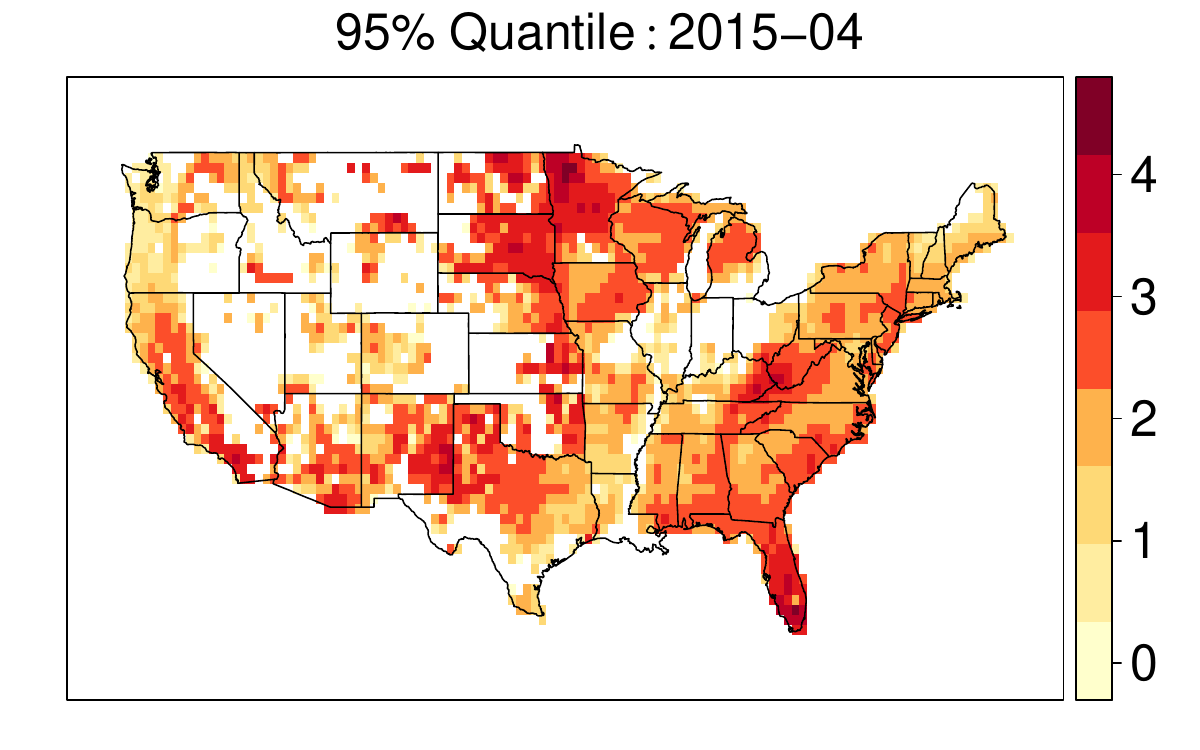} 
\end{minipage}
\begin{minipage}{0.32\linewidth}
\flushleft
\includegraphics[width=\linewidth]{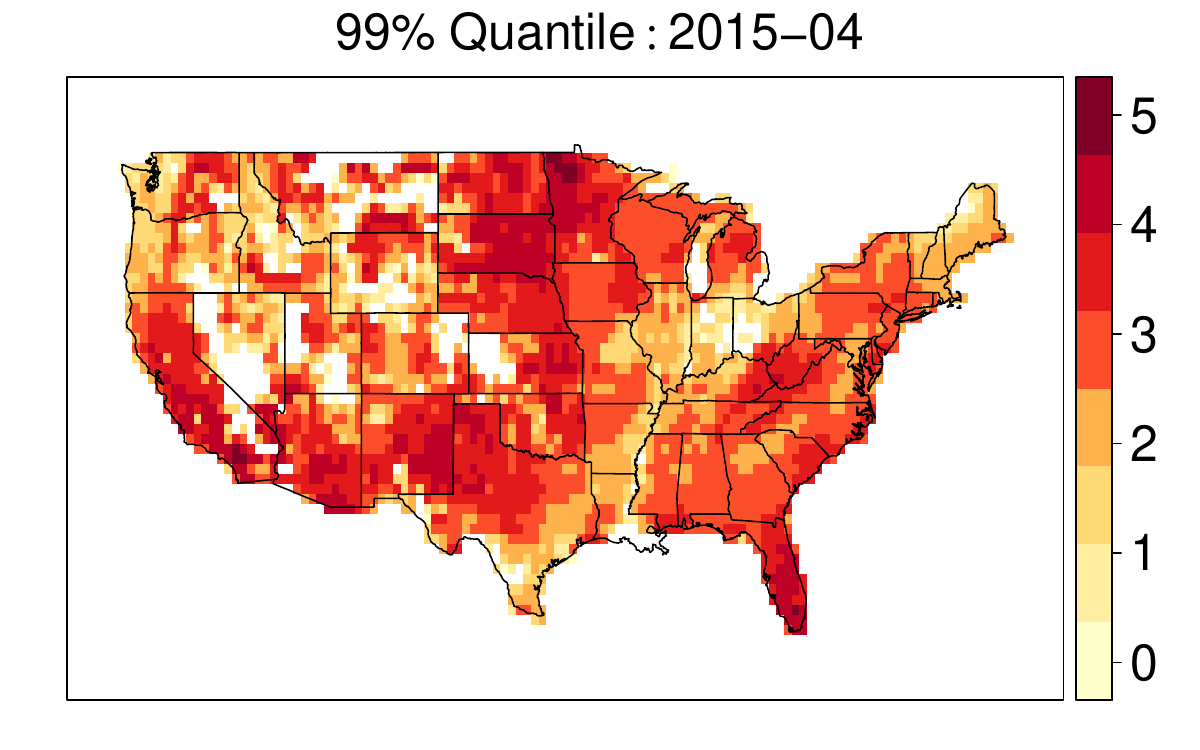} 
\end{minipage}
\vspace{-0.3cm}
\caption{Maps of median estimated quantiles for $\log\{1+\sqrt{Y}(s,t)\}\mid\mathbf{X}(s,t)$ (log-$\sqrt{\mbox{acres}}$) for different times $t$ (by row). Times, top to bottom row: March 1993, August 2000,  May 2003, July 2012, April 2015. Quantiles, left to right: $90\%$, $95\%$ and $99\%$.}
\label{cr_map_sup}
\end{figure}

\begin{figure}[t!]
\centering
\begin{minipage}{0.32\textwidth}
\flushleft
\includegraphics[width=\linewidth]{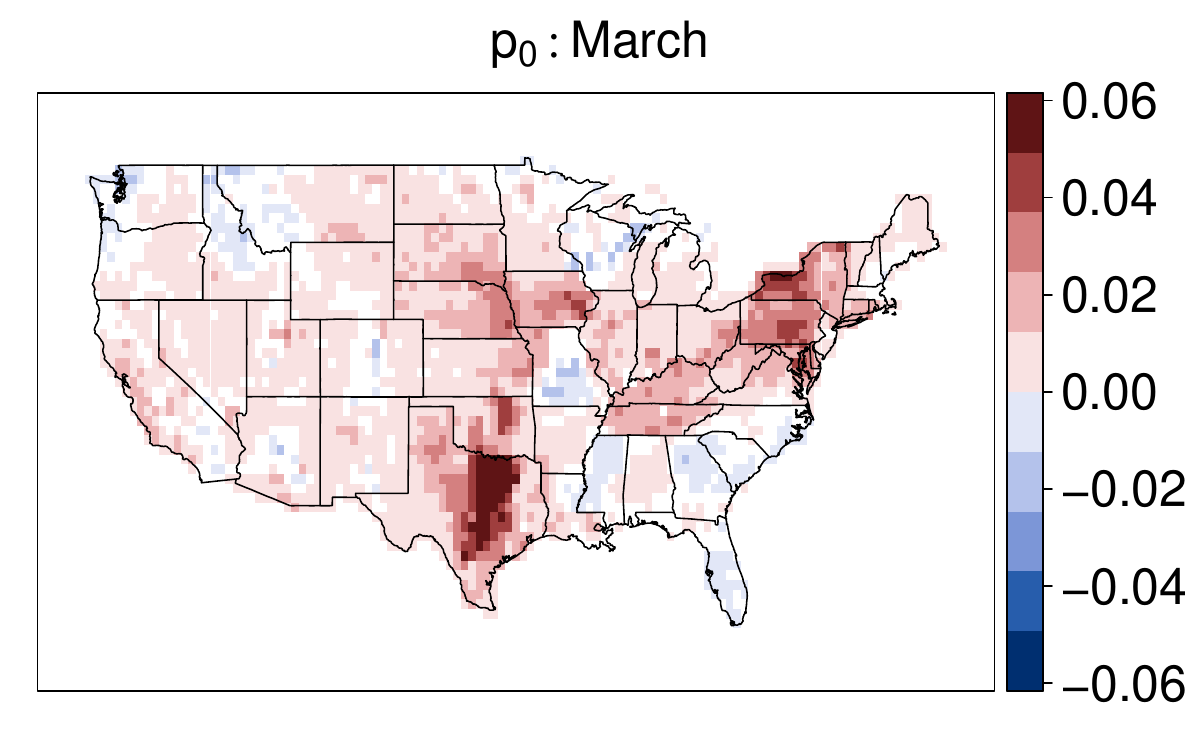} 
\end{minipage}
\vspace{-.3cm}
\begin{minipage}{0.32\textwidth}
\flushleft
\includegraphics[width=\linewidth]{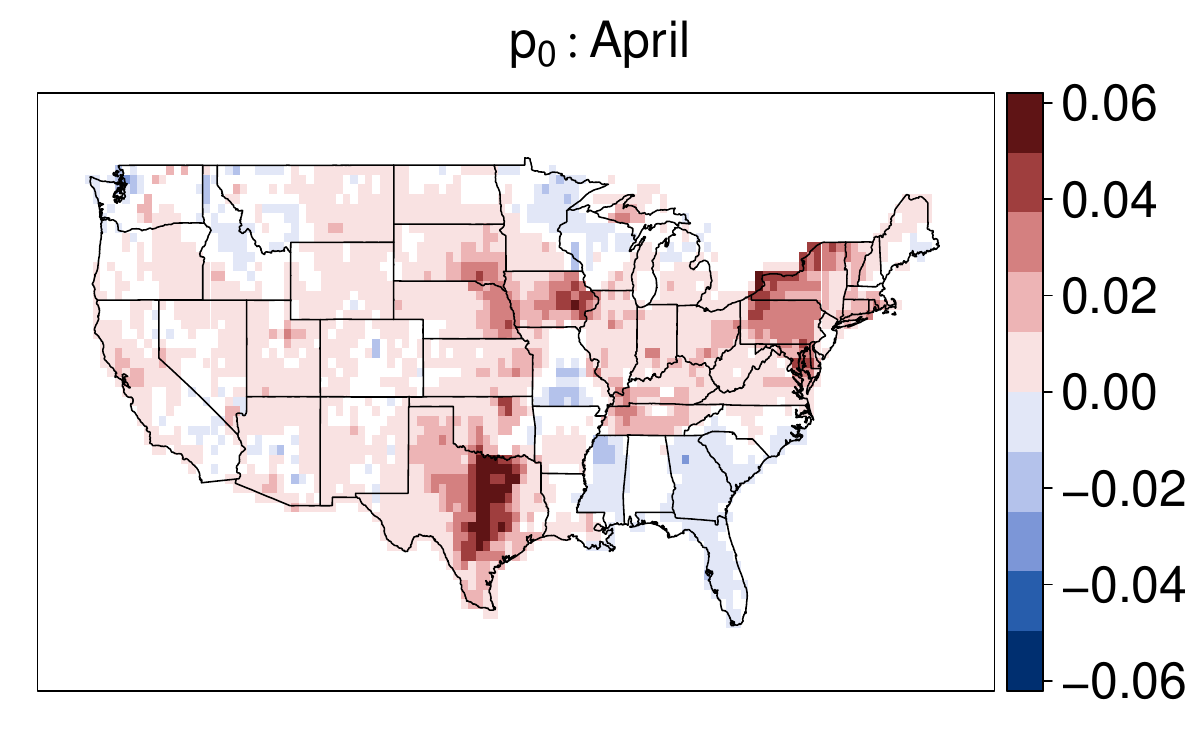} 
\end{minipage}
\begin{minipage}{0.32\textwidth}
\flushleft
\includegraphics[width=\linewidth]{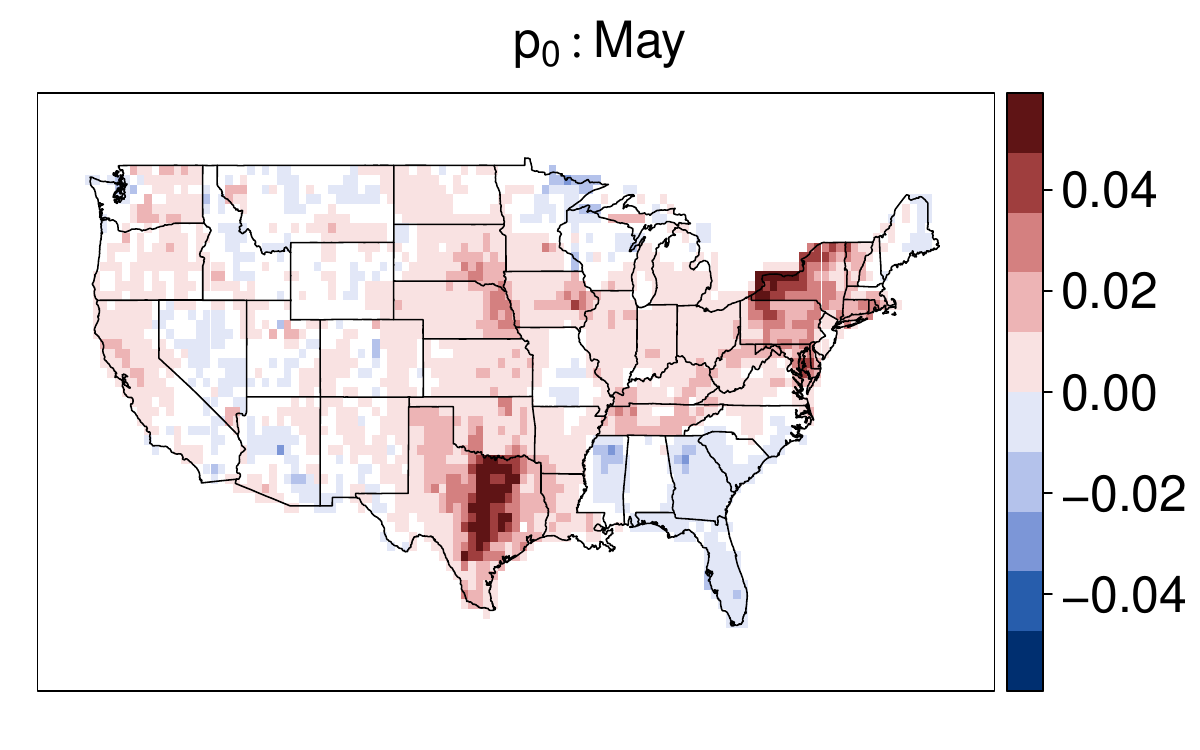} 
\end{minipage}
\begin{minipage}{0.32\textwidth}
\flushleft
\includegraphics[width=\linewidth]{Images/pZeroMap_diff_m7.pdf} 
\end{minipage}
\begin{minipage}{0.32\textwidth}
\flushleft
\includegraphics[width=\linewidth]{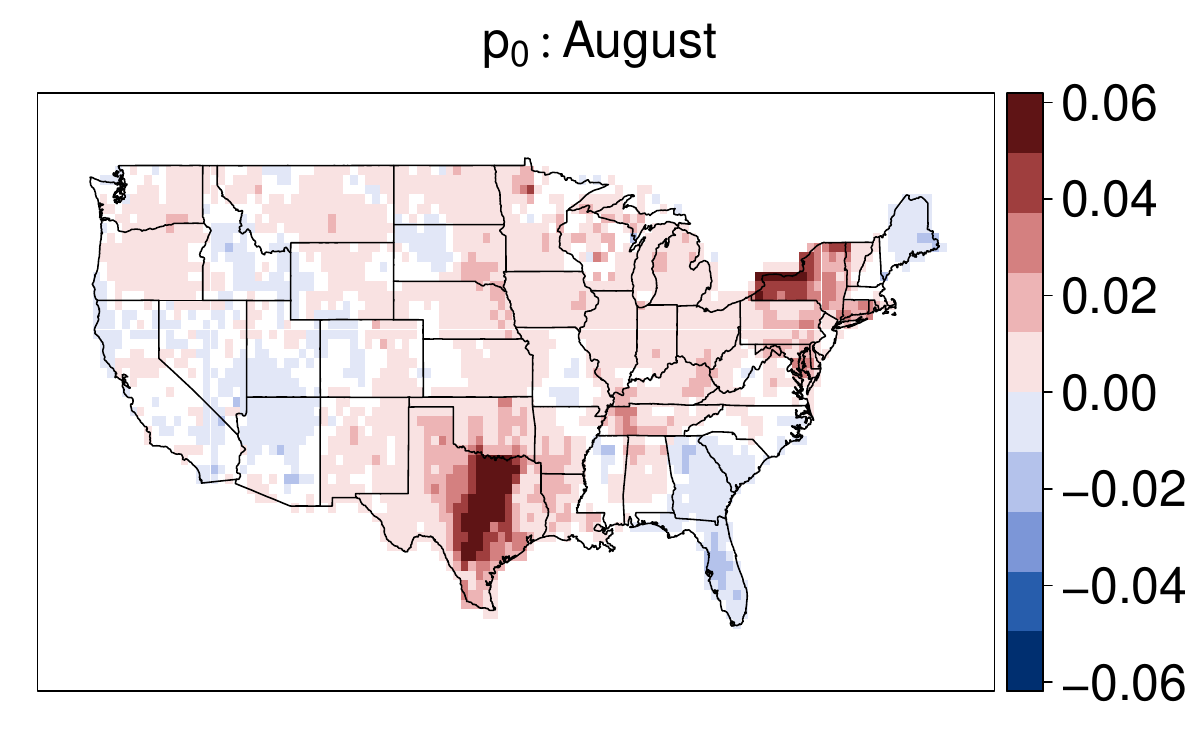} 
\end{minipage}
\begin{minipage}{0.32\textwidth}
\flushleft
\includegraphics[width=\linewidth]{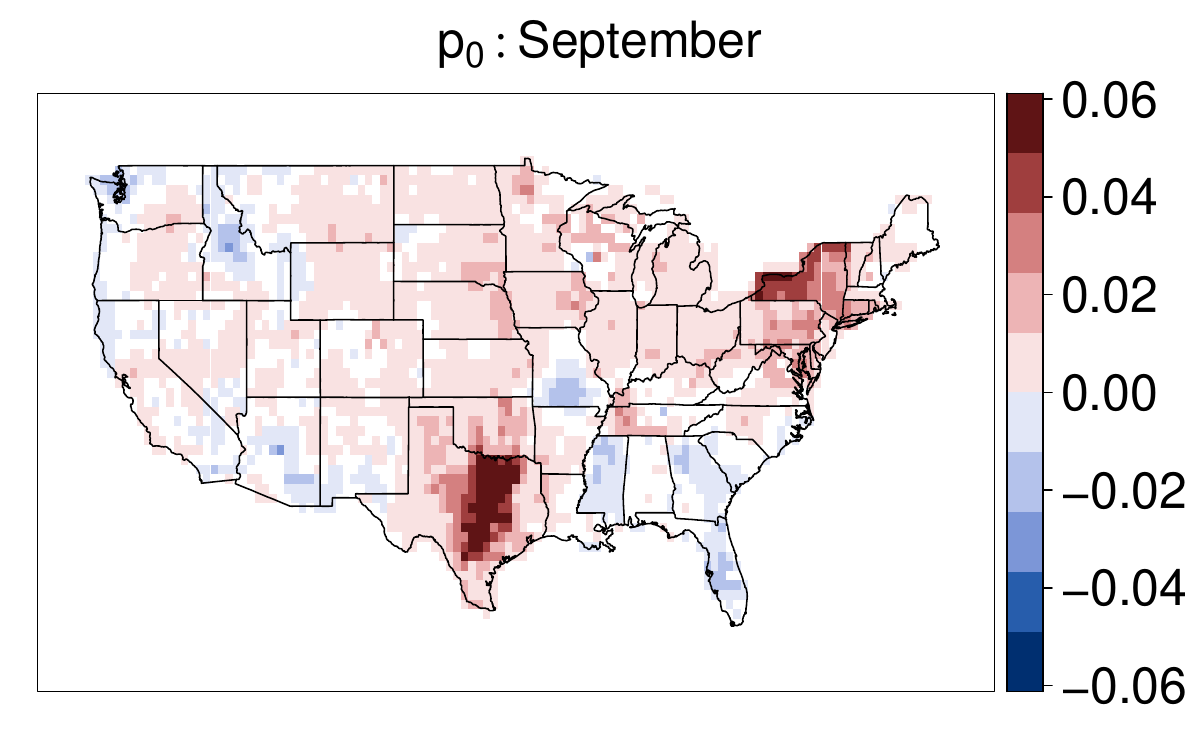} 
\end{minipage}
\vspace{-0.5cm}
\caption{Maps of median site-wise trends in estimated occurrence probability $p_0(s,t)$ (unitless) across all bootstrap samples, stratified by month. Top row: March--May. Bottom row: July--September.  }
\label{occur_diff_map_sup}
\end{figure}

\begin{figure}[t!]
\centering
\begin{minipage}{0.32\linewidth}
\flushleft
\includegraphics[width=\linewidth]{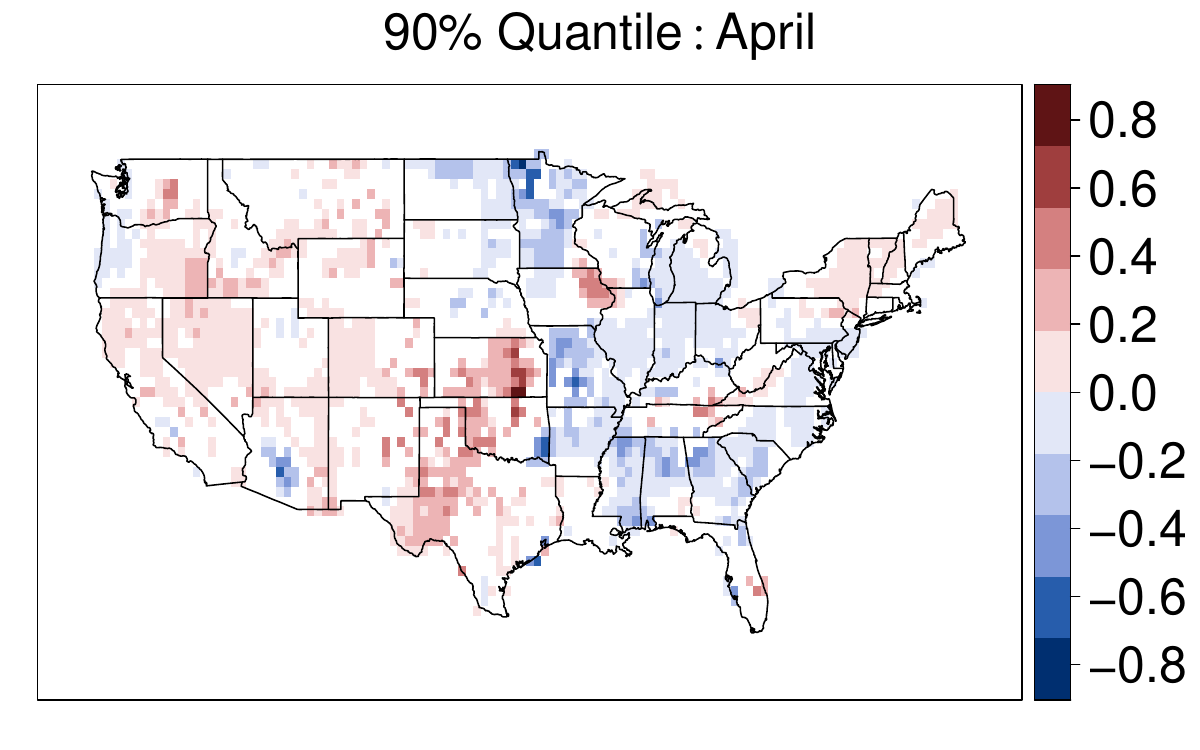} 
\end{minipage}
\vspace{-.2cm}
\begin{minipage}{0.32\linewidth}
\flushleft
\includegraphics[width=\linewidth]{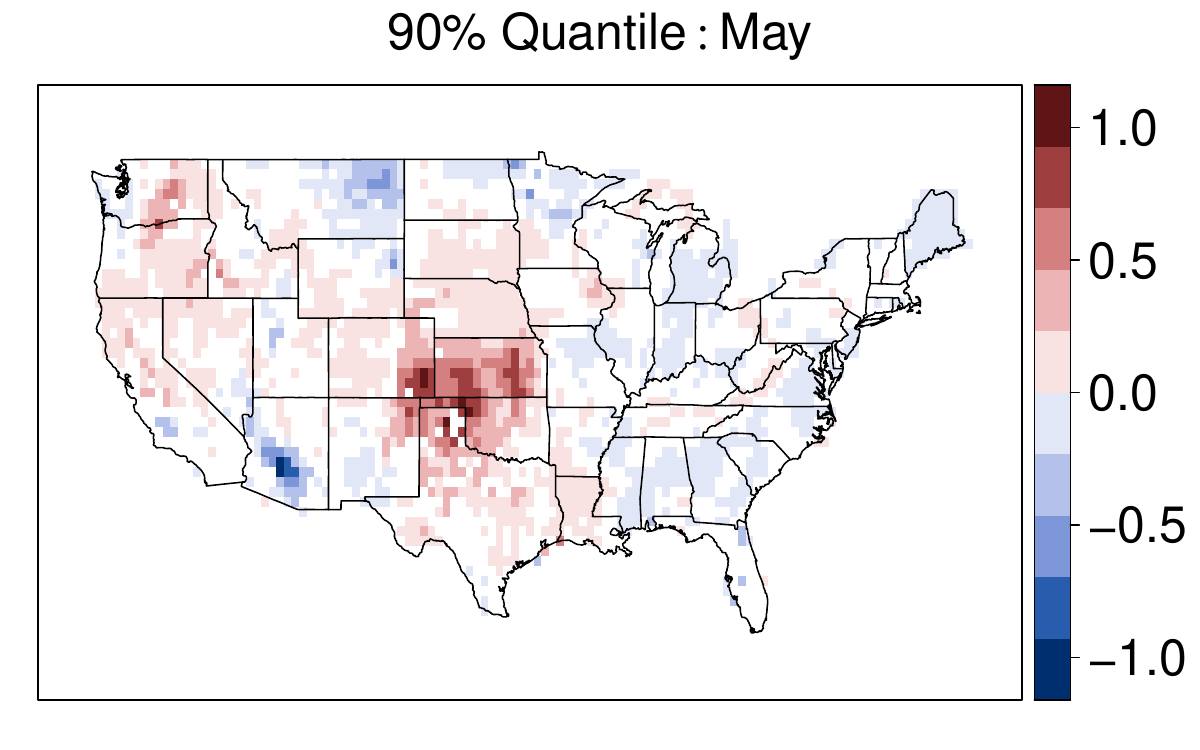} 
\end{minipage}
\begin{minipage}{0.32\linewidth}
\flushleft
\includegraphics[width=\linewidth]{Images/q90Map_diff_m6.pdf} 
\end{minipage}
\begin{minipage}{0.32\linewidth}
\flushleft
\includegraphics[width=\linewidth]{Images/q90Map_diff_m7.pdf} 
\end{minipage}
\begin{minipage}{0.32\linewidth}
\flushleft
\includegraphics[width=\linewidth]{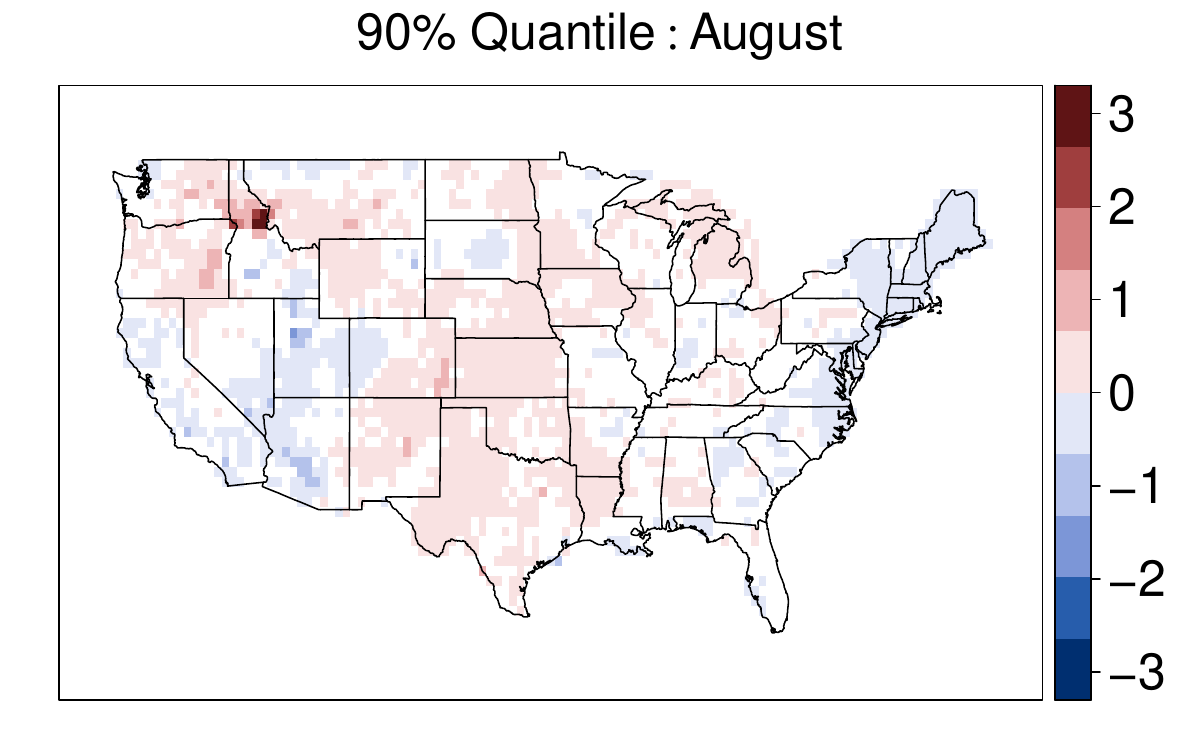} 
\end{minipage}
\begin{minipage}{0.32\linewidth}
\flushleft
\includegraphics[width=\linewidth]{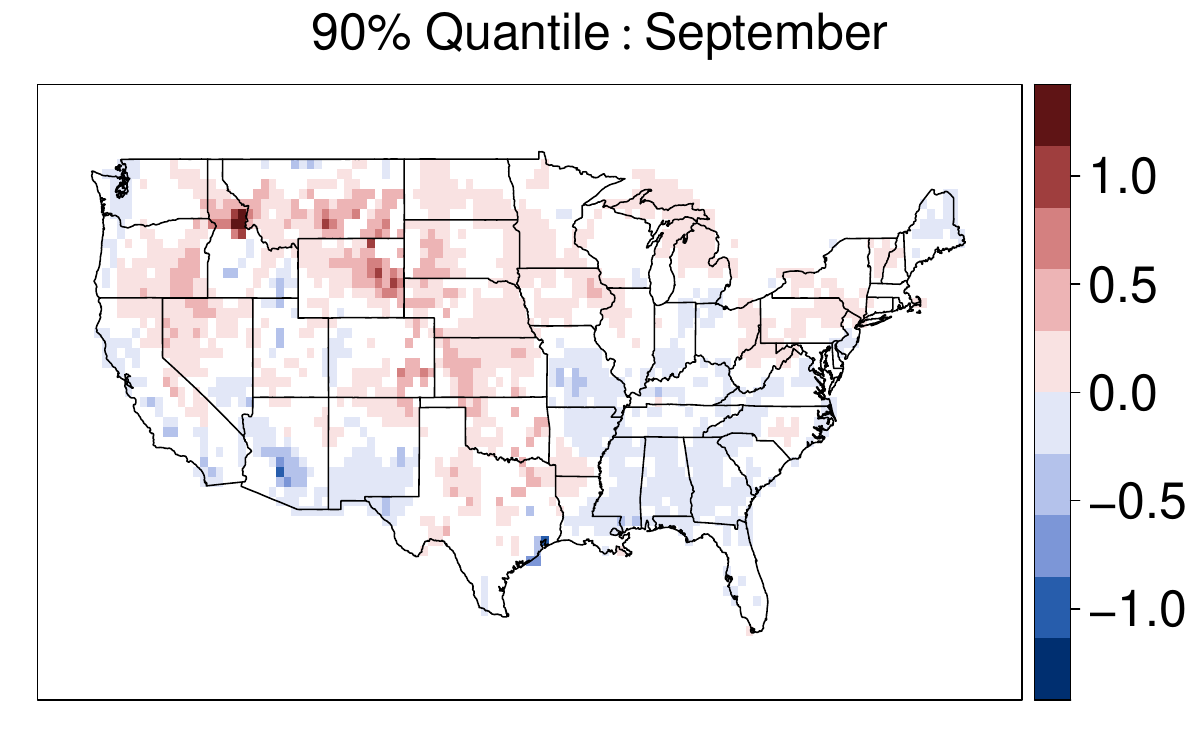} 
\end{minipage}
\vspace{-.5cm}
\caption{Maps of median site-wise trends in estimated estimated $90\%$ quantile of $\sqrt{Y}(s,t)\mid\{{Y(s,t)>0}, \mathbf{X}(s,t)\}$ ($\sqrt{\mbox{acres}}$) across all bootstrap samples, stratified by month. Top row: March--May. Bottom row: July--September. }
\label{spread_diff_map_sm}
\end{figure}
\end{appendix}

\end{document}